\documentclass[]{book}
\usepackage{times}
\usepackage{makeidx}
\usepackage{epsfig}
\usepackage{graphicx}
\usepackage{amsmath}
\usepackage{amssymb}
\usepackage{amsfonts,mathrsfs}
\usepackage{color}
\definecolor{orange}{rgb}{.5,.1,.02} 
\definecolor{purple}{rgb}{.5,0,.5} 
\definecolor{pea}{rgb}{.02,.4,.01} 
\definecolor{pink}{rgb}{.8,0,.6} 
\usepackage[pagebackref=true,breaklinks=true,letterpaper=true,colorlinks,bookmarks=false]{hyperref}
\newcommand{\be}{\begin{equation}}
\newcommand{\ee}{\end{equation}}
\newcommand{\ba}{\left[ \begin{array}}
\newcommand{\ea}{\end{array} \right]}
\newcommand{\bea}{\begin{eqnarray}}
\newcommand{\eea}{\end{eqnarray}}

\def\real{\mathbb{R}}
\DeclareRobustCommand\onedot{\futurelet\@let@token\@onedot}
\def\@onedot{\ifx\@let@token.\else.\null\fi\xspace}
\def\eg{\emph{e.g.} } 
\def\ie{\emph{i.e.} }

\newtheorem{defn}{Definition}
\newtheorem{rem}{Remark}
\newtheorem{example}{Example}
\newtheorem{lemma}{Lemma}
\newtheorem{theorem}{Theorem}
\newtheorem{corollary}{Corollary}

\newenvironment{proof}{{\bf Proof: \ }}{ \hfill \QED}
\def\cut#1{{}}
\def\cutOne#1{{}}
\def\cutTwo#1{{#1}}
\def\cutThree#1{{#1}}

\makeindex

\begin{document}
\title{Steps Toward a Theory of Visual Information:\\~\\ {\large Active Perception, Signal-to-Symbol Conversion \\ and the Interplay Between Sensing and Control}}
\date{Stefano Soatto \\~\\ \copyright ~ {\small 2008-2013, All Rights Reserved} \\~\\ Technical Report UCLA-CSD100028 \\ May 29, 2008 -- Last Updated January 2, 2013}
\author{\includegraphics[height=8cm]{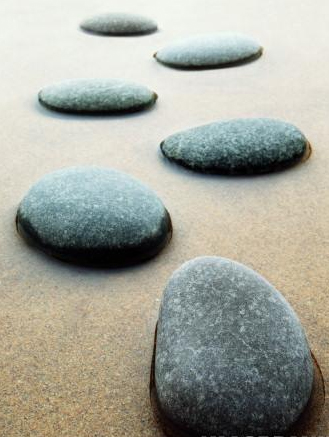}}
\titlepage
\maketitle

\tableofcontents

\newpage

\section*{Preface}

This manuscript has been developed starting from notes for a summer course at the First International Computer Vision Summer School (ICVSS) in Scicli, Italy, in July of 2008. They were later expanded and amended for subsequent lectures in the same School in July 2009. Starting on November 1, 2009, they were further expanded for a special topics course, CS269, taught at UCLA in the Spring term of 2010. 

I acknowledge contributions, in the form of discussions, suggestions, criticisms and ideas, from all my students and postdocs, especially Andrea Vedaldi, Paolo Favaro, Ganesh Sundaramoorthi, Jason Meltzer, Taehee Lee, Michalis Raptis, Alper Ayvaci, Yifei Lou, Brian Fulkerson, Teresa Ko, Daniel O'Connor, Zhao Yi, Byung-Woo Hong, Luca Valente. 

 In addition to my students, I am grateful to my colleagues Ying-Nian Wu, Andrea Mennucci, Anthony Yezzi, Veeravalli Varadarajan, Peter Petersen, and Alessandro Chiuso for discussions leading to many of the insights described in this manuscript. I also wish to acknowledge discussions with Joseph O'Sullivan, Richard Wesel, Alan Yuille, Serge Belongie, Pietro Perona, Jitendra Malik, Ruzena Bajcsy, Yiannis Aloimonos, Sanjoy Mitter, Roger Brockett, Peter Falb, Alan Willsky, Michael Jordan, Judea Pearl, Deva Ramanan, Charless Fowlkes, Giorgio Picci, Sandro Zampieri, Lance Williams, George Pappas, Tyler Burge, James Clark, Jan Koenderink, Yi Ma, Olivier Faugeras, John Tsotsos, Tomaso Poggio, Allen Tannenbaum, Alberto Pretto. I wish to thank Andreas Krause, Daniel Golovin, Andrea Censi, Lorenzo Rosasco, Max Welling and Taco Cohen for many useful comments and corrections. Finally, I wish to thank Roberto Cipolla, Giovanni Farinella and Sebastiano Battiato for organizing ICVSS that has initiated this project.

 The work leading to this manuscript was made possible by guidance, encouragement, and generous support of three federal research agencies, under the leadership of their Program Managers: Behzad Kamgar-Parsi of ONR, Liyi Dai of ARO, and Fariba Fahroo of AFOSR. Their help in making this project happen is gratefully acknowledged. I also wish to acknowledge discussion and feedback from Tristan Nguyen, Alan Van Nevel, Gary Hewer, William McEneany, Sharon Heise, and Belinda King.

 {\color{red} Note from 2017: Much of this work is superseded and distilled in the 2017 paper ``Emergence of Invariance and Disentanglement in Deep Representations'', \cite{achille2017emergence}.}

\chapter{Preamble}

{\quotation \em ``Intelligent Behavior'' is often assumed to involve some sort of ``internal representation'' made of discrete ``symbols.'' However, the Data Processing Inequality suggests that such a signal-to-symbol conversion is detrimental to  the optimality of decision and control actions downstream. This opens a number of questions that motivate this manuscript. Readers uninterested in such philosophical issues can skip this section.}~\\

Perceptual agents, from plants to humans, measure samples of physical processes (``signals'') that are {\em essentially continuous}.\footnote{The continuum is an abstraction; a continuous entity is to be understood as one existing at a level of granularity significantly finer than the resolution of the measurement devices. For instance, the radiance of an object is sampled by retinal photoreceptors that are finite in number; the more receptors (or the closer the viewer), the more details are being revealed, so one never has a ``sufficient number of pixels'' and therefore the radiance can be thought of as a continuous function.\cut{ Even when the sensory signals and the actions are discrete (due to digital encoders or transducers), the ``analog-to-digital'' conversion usually occurs in a manner that is independent of the signal being sampled (\eg fixed-rate sampling), or dependent only on coarse phenomenological aspects of the signal (\eg adaptive sampling based on frequency characteristics or sparsity constraints).}}  They also perform actions in the continuum of physical space. And yet, cognitive science, epistemology, and in general modern philosophy associate ``intelligent behavior'' with some kind of ``internal representation'' built upon {\em discrete symbols} (\eg  ``concepts'', ``ideas'', ``objects'', ``categories'') that can be manipulated and inferred with logic and probabilistic inference. But {\em why} should such a ``signal-to-symbol'' conversion occur? How does it yield an evolutionary advantage? What principles should guide it?

Information Theory suggests that such a signal-to-symbol conversion may be counter-productive: If we consider biological systems as machines that perform actions or make decisions in response to stimuli in a way that maximizes some decision or control objective,\cut{ for instance reproductive success,} then the Data Processing Inequality\footnote{See Section \ref{sect-dpi} or page 88 of \cite{shao98}.} indicates that the best possible agents would avoid {\em data analysis},\footnote{\label{foot-1} {\color{pea}Note that I refer to {\em data analysis} as the process of ``breaking down the data into pieces'' (cfr. gr. {\em analyein}), \ie the generally lossy conversion of data into ``local'' discrete entities (symbols). These do not include Fourier Analysis, Principal Component Analysis (PCA) and other global transforms.}} \marginpar{\tiny \sc {\color{pea} green for signal processing and information theory}}{\em i.e.,} the process of breaking down signals into discrete entities or symbols.\footnote{Discretization is often advocated on complexity grounds, but complexity calls for data {\em compression}, not necessarily for data {\em analysis}. Any complexity cost could be added to the decision or control functional in Section \ref{sect-dpi}, and the best decision would still avoid data analysis.\cut{ For instance, to simplify a segment of a radio signal one could represent it as a linear combination of a small number of (high-dimensional) bases, so few numbers (the coefficients) are sufficient to represent it in a parsimonious manner. This is different than breaking down the signal into pieces, \eg partitioning its domain into subsets, as implicit in the process of encoding a visual signal through a population of neurons each with a finite receptive field.} Is there an evolutionary advantage in data analysis, beyond it being just a way to perform data compression?\cut{ Another example, with motivations other than compression, is provided by the {\em packet} encoding of data for transmission in large networks such as the Internet.} } These considerations apply regardless of the specific control or decision task, from the simplest binary decision (\eg ``is a specific object present in the scene?'') to the most complex (\eg ``survival'').

So, why would we need, or even benefit from, an internal representation? Is ``intelligence'' not possible in an ``analog'' setting? Or is {\em data analysis} necessary for cognition? If so, what would be the mathematical and computational principles that guide it? What if the task is not known; is a notion of representation still meaningful in the absence of a task?

In the fields of Signal Processing, Image Processing, and Computer Vision, we routinely torture the data {\color{pea} (filtering, sampling, anti-aliasing, edge detection, feature selection, segmentation, etc.)}, seemingly against the basic tenets of Information and Decision Theory. The Data Processing Inequality would instead suggest an approach whereby data are fed directly into a ``black-box'' decision or control machine designed to optimize a (possibly very complex) cost functional.\cut{ Such an approach would outperform one based on a generic, task-independent ``internal representation.'' Should we then discard decades of work in attempting to infer such internal representations?}

One could argue that data analysis in biological systems is not guided by any optimality principle, but an accident due to the constraints imposed by biological hardware. In \cite{turingMorphogenesis}, Turing showed that (continuous) reaction-diffusion partial differential equations (PDEs) that govern ion concentrations in neurons exhibit discrete/discontinuous solutions. This may explain ``spikes'' in neuronal signals and perhaps from there symbols. But if we want to build machines that interact intelligently with their surroundings and are not bound by the constraints of biological hardware, should we draw inspiration from biology, or can we do better by following the principles of Information Theory? \\

The question of existence of an ``internal representation'' is best framed within the scope of a task, \marginpar{\tiny \sc task} \index{Task} which provides a falsifiability mechanism.\footnote{Of course one could construe the inference of the internal representation as the task itself, but this would be self-referential.} In the context of visual analysis I distinguish four broad classes of tasks, which I call \marginpar{\tiny \sc the 4 r's of vision} {\em the four ``R's'' of vision:} \index{Four R's} Reconstruction (building models of the {\em geometry}, or shape, of the scene), Rendering (building models of the {\em photometry}, or material properties,  of the scene), Recognition and other vision-based {\em decisions} such as detection, localization, categorization and more in general scene semantics, and Regulation or, more in general, vision-based {\em control} such as tracking, navigation, obstacle avoidance, manipulation etc.. 
\cut{For Reconstruction and Rendering, I am not aware of any principle that suggests an advantage in data analysis. It is not surprising that the current best approaches to infer the geometry and photometry of a scene from collections of images recover (piecewise) continuous surfaces and radiance functions directly from the data \cite{jinSY05IJCV}, unlike the traditional multi-step pipeline\footnote{A sequence of ``steps'' including point-feature selection, wide-baseline matching, epipolar geometry estimation, motion estimation, triangulation, epipolar rectification, dense re-matching, surface triangulation, mesh polishing, texture mapping.} that was long favored on complexity grounds (or pedagogical motivations, see \cite{maSKS} and references therein).}

In this manuscript, we will explore the issue of {\em representation} from visual data for decision and control tasks. To avoid philosophical entanglements, we will not attempt to define ``intelligent behavior'' or even ``knowledge,'' other than to postulate that knowledge -- whatever it is -- {\em comes from data}, but it is {\em not} {\em data}. This leads to the notion of the ``useful portion'' of the data, \marginpar{\tiny \sc information} which one might call ``information.'' So, our first step will be a definition of what ``information'' means in the context of performing a decision or action based on sensory data.

\begin{figure}[htb]
\begin{center}
\includegraphics[width=.5\textwidth]{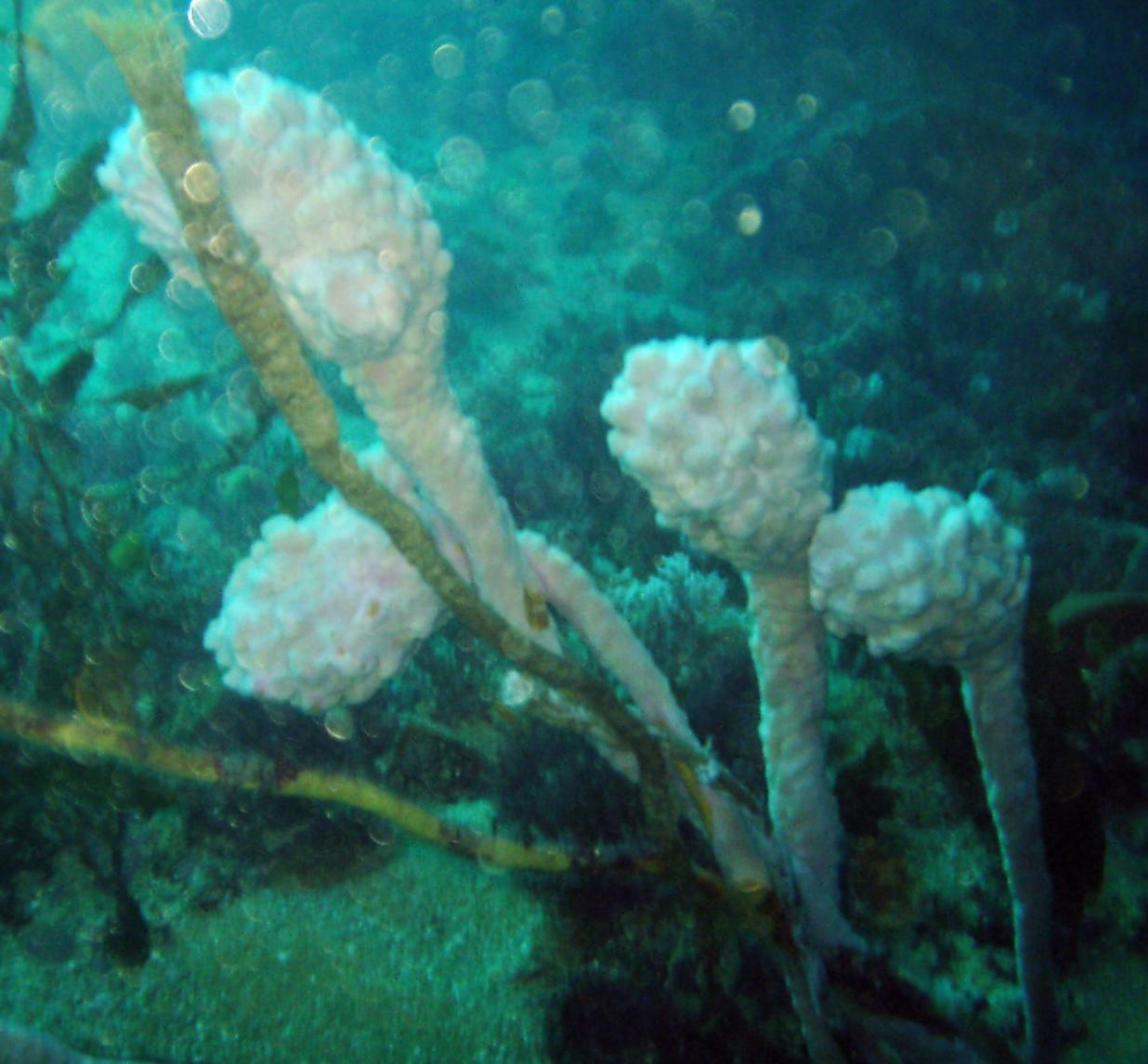}
\end{center}
\caption{\sl The {\bf Sea Squirt}, or Tunicate, is an organism capable of mobility until it finds a suitable rock to cement itself in place. Once it becomes stationary, it digests its own cerebral ganglion cells, or ``eats its own brain,'' and develops a thick covering, a ``tunic'' for self defense.${}^6$}
\label{fig-tunicate}
\end{figure}

As we will see, {\em visual perception} plays a key role in the signal-to-symbol barrier. As a result, much of this manuscript is about {\em vision}. Specifically, the need to perform {\em decision and control tasks} in a manner that is independent of nuisance factors \marginpar{\tiny \sc nuisance factors} \index{Nuisance} including {\em scaling} and {\em occlusion} phenomena require the perceptual agent (or, more in general, its evolved species) to exercise {\em control} \marginpar{\tiny \sc control} over certain aspects of the sensing process. This inextricably ties sensing, information and control. A case-in-point is provided by Sea Squirts, or Tunicates, shown in Figure~\ref{fig-tunicate}. These are organisms that possess a nervous system (ganglion cells) and the ability to move. They spend part of their lives as predators, but eventually settle on a rock, become stationary and thence swallow their own brain.\footnote{This is sometimes used as a metaphor of tenure in academic institutions.} Scaling and occlusion play a critical role: The first makes the continuum limit relevant, the second makes control a critical element in the analysis. These are present in a number of remote sensing modalities, including optical, infrared, multi-spectral imaging, as well as active ranging such as radar, lidar, time-of-flight, etc. 

\section{How to read this manuscript}

This manuscript is designed to allow different levels of reading. Some of the material requires some background beyond calculus and linear algebra. To make the manuscript self-contained, basic elements of {\color{orange} topology}, {\color{purple} variational methods and optimization}, {\color{blue} image processing,} {\color{pink} radiometry,} etc. are provided in a series of appendices. These are color coded. The parts of the main text that require background in the corresponding discipline are coded with the same color. The reader can then either read through the colored text if he or she is familiar with that subject, disregarding the appendices, or use the appendix as a reference in case he or she is not familiar with the subject, or skip the colored text altogether. The manuscript is structured to allow getting the ``big picture'' without any mathematical formalism by just reading the black text.

\cut{\marginpar{\tiny \sc mumble-jumble}
The next sections in this chapter are motivational mumble-jumble, that can be skipped without loss off continuity if the reader is already committed. If not, they elaborate the discussion above and provide a summary of the logic of the argument developed in the following chapters.}

\section*{Summary for the experts (to be skipped by others)} 

This section summarizes the content of the manuscript in a succinct manner. It can be used as a summary, or as a reference to the broader picture while reading the rest of the manuscript. For most readers, this summary will be cryptic or confusing at a first reading. If it was otherwise, there would be no need for a manuscript to follow. 

{\scriptsize \cut{The long-term motivation for this study is three-fold. First, a desire to see a theory of information emerge in support of {\em decision and control tasks}, as opposed to transmission and storage of data. Then, a desire to enable the computation of {\em performance bounds}  in a vision-based decision or control task, for instance bounds on the probability of error in visual recognition. Finally, the desire to give grounding to low-level vision operations such as segmentation, edge detection, feature selection etc. It goes without saying that these problems cannot be settled by one person in one manuscript, so this work aims to make a few steps in the direction of these long-term goals.}

What makes vision difficult, and might explain the fact that almost half of the primate brain is devoted to it \cite{fellemanV91,poggio11}, is the fact that {\em nuisance factors} in the data formation process {\em account for almost all the complexity of the data} \cite{sundaramoorthiPVS09}. Such factors include {\em invertible nuisances} such as contrast and viewpoint (away from occlusions), and {\em non-invertible} ones such as occlusions, quantization, noise, and general illumination changes. After {\em discounting the effects of the nuisances in the data} (invariance), even if one had started with infinite-resolution data, what is left is ``thin'' (supported on a {\em zero-measure} subset of the image domain). The complexity of the data after the effects of invertible nuisances has been remove is called {\em Actionable Information}. The fact that Actionable Information can be {\em thin in the data} is relevant to the {signal-to-symbol barrier} problem.

How can we deal with nuisances? At {\em decision time} one can marginalize them (Bayes) or search for the ones that best explain the data (max-out, or maximum-likelihood). Some, however, may be eliminated in a process called {\em canonization}. While marginalization and max-out require solving complex integration or optimization problems at decision time, canonization can be pre-computed, and hence it enables straight comparison of statistics at decision time. It is preferable if time-complexity is factored in. However, this benefit comes with a predicament, in that canonization cannot decrease the expected risk, but at best leave it unchanged. Among the statistics that leave the risk unchanged ({\em sufficient statistics}), the ones that are also {\em invariant} to the nuisances would be the ideal candidates for a {\em representation}: They would contain all and only the functions of the data that matter to the task.

Unfortunately, while for invertible nuisances one can construct {\em complete features} (invariant sufficient statistics), that act as a lossless representation, occlusion and quantization are {\em not invertible.} Thus, there is a {\em gap} between the maximal invariant and the minimal sufficient statistics. This gap cannot, in general, be filled by processing {\em passively gathered} data.

However, when one can exercise {\em control on the sensing process}, then {\em some non-invertible nuisances can become invertible.} Occlusions can be inverted by moving around the occluder. Scaling/quantization can be inverted by moving closer. Even the effects of noise can be countered by increasing temporal sampling and performing suitable averaging operations. Therefore, in an {\em active sensing} scenario one can construct {\em representations} that are (asymptotically) lossless for decision and control tasks, and yet have low complexity relative to the volume of the raw data. This inextricably ties {\em sensing and control}. It also may enable achieving {\em provable bounds}, by generalizing Rate-Distortion theory to {\em Perception-Control} tradeoffs, whereby the ``amount of control authority'' over the sensing process (to be properly defined) trades off the expected error in a visual decision task.

\marginpar{\tiny \sc how the theory unfolds}
In this manuscript, we characterize {\em representations} as complete invariant statistics, and call {\em hallucination}\footnote{The characterization of images as ``controlled hallucinations'' was introduced by J. Koenderink.} the simulation of the data formation process starting from a representation (as opposed to the actual {\em scene}). We define {\em Actionable Information} as the complexity of the maximal invariant statistics of the data, and {\em Complete Information} as the complexity of the (minimal sufficient statistic of a) complete representation. We define {\em co-variant detectors}, that enable the process of {\em canonization}, and their associated {\em invariant descriptors}. We define {\em canonizability}, and address the following  questions: (i) When is a classifier based on an invariant descriptor optimal? (in the sense of minimizing the expected risk) (ii) what is the best possible descriptor? (iii) what nuisances are canonizable? (and therefore can be dealt with in pre-processing, as opposed to having to be marginalized or max-outed at decision time).

\marginpar{\sc \tiny 4 key concepts}
Four concepts introduced in this study are key to the analysis and design of practical systems for performing visual decision and control tasks: {\em Canonizability, Commutativity, Structural Stability, and Proper Sampling.}
\marginpar{\sc \tiny canonizability}\index{Canonizability}
\marginpar{\sc \tiny commutativity}\index{Commutativity}
\marginpar{\sc \tiny structural stability}\index{Structural stability}
\marginpar{\sc \tiny proper sampling}\index{Proper Sampling}

Canonization is not sufficient to infer a complete representation, unless nuisances {\em commute} with one another. We show that the only nuisance that is canonizable and commutative is the isometric group of the plane. Affine transformations in general, and the scale group in particular, should {\em not} be canonized, but should instead be {\em sampled} and marginalized. 

Canonizing functionals, designed to select an element of the canonizable nuisance group, should be {\em stable} with respect to variations of the non-canonizable nuisances. We introduce the notion of {\em Structural Stability}, that is related to catastrophe theory and persistent topology. Selection by maximum structural stability margins gives rise to a novel feature selection scheme \cite{leeS10}. 

Whether the structure detected by a canonizing functional is ``real'' (\ie it arises from phenomena in the scene) or an ``alias'' (\ie it originates from artifacts of the image formation process, for instance quantization) depends on whether the signal is {\em properly sampled}. We introduce a notion of proper sampling that, unlike traditional (Nyquist-Shannon) sampling, cannot be decided based on a single datum (one image snapshot), but instead requires multiple images. This notion gives rise to a novel feature tracking scheme \cite{leeS10}.

\cut{This also brings the issues of {\em mobility} and {\em time} front and center in both the theory, and in the implementation of effective recognition systems. It gives rise to a novel set of descriptors, such as the {\em Best Template Descriptor} and {\em Time HOG}, and to visual recognition schemes that are based on video \cite{leeS10}, as opposed to collections of isolated snapshots. }

Intra-class variability can be captured by endowing the space of representations (which are discrete entities) with a probabilistic structure and learning distributions of individual objects or parts, clustered by labels. Objects are not necessarily rigid/static, but can also include ``actions'' or ``events'' that unravel in time. Time can be treated as yet another nuisance variable, which unfortunately is not invertible and therefore cannot be canonized without a loss. Time is, therefore, best dealt with by marginalization or max-out at decision time \cite{raptisS10}.

Along the way in our investigation we also discuss the role of ``textures'' and their dual (``structures''),  and characterize them as the complement of canonizable regions \cite{boltzNS10}.

\cut{The elements of this theory are presented following the trace below:
\begin{enumerate}
\item The starting point is the notion of {\em visual decision} leading to a definition of ``visual information'' as the part of the data that matters for the task. ``Knowledge,'' ``cognition,'' ``understanding,'' ``semantics,'' ``meaning,'' are higher-order concepts that will will remain undefined and not tackled in this manuscript (Section 2.1).
\item Visual decision problems (including  detection, localization, recognition, categorization) are {\em classification} tasks based on {\em data} measured from images or video $\{I\}$. A class is identified by a label $c$, and specified by means of examples (\eg a training set, Section 2.2).
\item Optimal classification is based on a {\em risk} functional $R$. 
To compute a risk functional we need a forward (image-formation) model (or equivalently suitable assumptions and priors). We will use a formal image-formation model denoted by $I = h(\xi, \nu)$, whereby the data $I$ depend on the ``scene'' $\xi$ (the part of the data that matters) and {\em nuisance factors} $\nu$ (Section 2.3). 
\item Nuisance factors, or nuisances, include viewpoint, contrast and other {\em invertible nuisances}. \marginpar{\tiny \sc invertible vs. non-invertible nuisances}
They also include  scaling/quantization, occlusions and other {\em non-invertible nuisances}. Finally, nuisances may also include {\em intra-individual variability} and other higher-level structure/priors. One can deal with nuisances via marginalization (Bayes), extremization (Maximum Likelihood, ML), or canonization (features, detectors/descriptors, Section 2.5).
\item  What is left in the data that does not depend on the nuisances represents the (actionable) ``information'' present in the data (Section 2.6).
\item While Bayes/ML are {\em optimal} (they minimize the risk, under different priors), features are not, in general (Section 2.7). 
\item The use of features brings about the notion of {\em sufficient statistics}, the ``data processing inequality'' (or Rao and Blackwell's theorem), the notions of {\em invariance} and {\em representation}, and the interplay between the representation and the classifier. This also raises three crucial questions:
(a) When can features be used without a loss? (b) When there is a loss, can it be quantified and bounded? 
(c) Even classifiers that are optimal (Bayes, ML minimize risk) do not achieve zero risk. Are there bounds on the risk?  (Chapter 3).
\marginpar{\tiny \sc no meaningful bounds for passive visual recognition}
\item A somewhat obvious, but nevertheless key observation is that for ``passive'' visual recognition, it is not possible to arrive at ``meaningful'' bounds on the risk. Meaningful in this context means that the error bound in the task of determining the class of a certain ``object'' does not depend on the specific instance of the object and the nuisances. \marginpar{\tiny \sc active recognition brings about ``controlled recognition bounds''} (Section 10.1).
\item A second key observation is that for ``active'' visual recognition, the risk can be made arbitrarily small (asymptotically) \cite{oconnorMS10}. This brings about the role of {\em time}, and the asymmetric nature of training data (that require time) and test data (testing can be instantaneous). It also has some epistemological ramifications, and links the concept of ``information'' to ``control'' or ``action.''  (Section 10.2).
\marginpar{\tiny \sc information control}
\item For active recognition, the currency that trades off recognition performance (risk) is the control authority one can exercise over the sensing process (exploration). \marginpar{\tiny \sc control-recognition theory} (Section 10.3).
\item A notion of ``representation'' arises naturally in active recognition, and its complexity quantifies the ``complete information.''  (Sections 2.5.4 and 3.1).
\item Computing, or inferring, such a representation from the data entails the notion of {\em invariance} (for invertible nuisances) and {\em stability} (for non-invertible nuisances). \marginpar{\tiny \sc structural stability of the representation} It involves common early-vision operations, such as texture analysis and segmentation. \marginpar{\tiny \sc texture, segmentation} (Chapter 4).
\item Inferring a representation highlights the critical role of {\em time} (and motion within physical space) in the representation, and also the critical role of {\em occlusions}, without which a representation would not be justified on decision-theoretic grounds. \marginpar{\tiny \sc occlusions, motion} (Chapters 6 and 8).
\item For visual decisions regarding ``objects'' that have a temporal component (a.k.a. ``events,'' or ``actions,'' or ``activities''), time can be thought of as a nuisance factor and marginalized, max-outed, or canonized. (Chapter 7).
\end{enumerate}
}
}

\section{Related literature}

The design and computation of visual representations for recognition has a long history (see \cite{marr,romeny} and references therein). While Marr's representation using zero-crossings of differential operators was discredited because of instability in the reconstruction process (\ie obtaining images back from their representations), reconstructing {\em images} is not necessarily the purpose, if a representation is to support decision and control tasks. Many have attempted to design representations that are tailored to recognition (as opposed to image reconstruction) tasks, some using similar ideas of extrema of scale-spaces constructed from differential operators \cite{lowe99object,lindeberg98}. However, most of these designs have been performed in an ad-hoc manner, guided by intuition, common sense, and some biological inspiration. Statistical decision theory would instead call for the direct design of general ``super-classifiers'' forgoing intermediate representations altogether, unless directly tied to the task. 

{\small 
Our earlier work \cite{soatto09} aims to frame the construction of invariant/sufficient representations in the context of Active Vision, formalizing some ideas of J. J. Gibson \cite{gibson84}. Gibson's approach, however, falls short on several counts. 
First, {\em invariance is too much to ask.} In the process of being invariant to general viewpoints, shape becomes indiscriminative \cite{vedaldiS05}. And yet, we know we can discriminate objects that are deformed versions of the same material. 
Second, often priors are available, from training or otherwise, both on the nuisance and on the class, and {\em such priors should be used.} There is no point in requiring invariance to illuminations that will never be; better instead to be insensitive to common nuisances, and relax the representation where nuisances have low probability even though this opens the possibility of illusions for unlikely nuisance and scene combinations (Fig. \ref{fig-illusions}).
\begin{figure}[htb]
\begin{center}
\includegraphics[width=.6\textwidth]{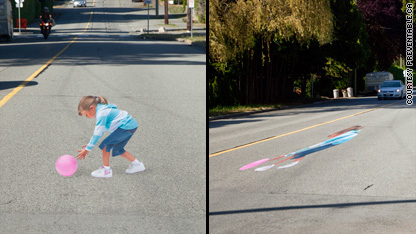}\\
\includegraphics[width=.45\textwidth,height=3cm]{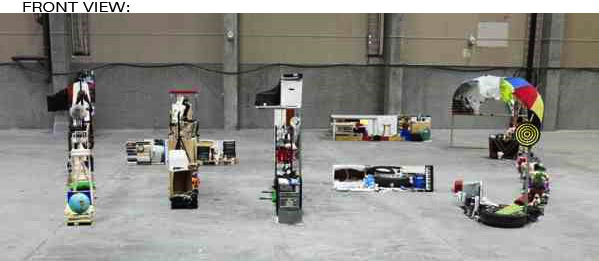}
\includegraphics[width=.45\textwidth,height=3cm]{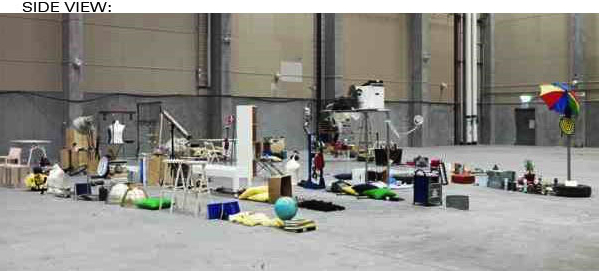}
\end{center}
\caption{\sl In the absence of sufficiently informative data ({\em e.g.}, one image), priors enable classification that, occasionally, can be incorrect if the nuisances and scene concur to form a {\em visual illusion}. The scene on the top-left looks like a girl picking up a ball (image courtesy of Preventable.com), rather than a flat painting on the road surface. Similarly, the purposeful collection of objects on the bottom looks like meaningful symbols when seen from a carefully selected (non-accidental) vantage point.}
\label{fig-illusions}
\end{figure} 
\cut{So, the invariant representations proposed are only conceptual, and not practical to compute. But, even if one could compute them, it is not clear that one would want to do so, because invariance is too strict a requirement.}
Third, even though the Active Vision paradigm is appealing, in practice {\em we often do not have control on the sensing platform.} 

{\em Therefore, there remains the need to properly treat nuisances that are {\em non-invertible}, such as occlusions, quantization and noise, in a passive sensing scenario, and to be able to exploit priors when available.}

The recent literature is studded with different approaches for low-level pre-processing of images for visual classification. These include various feature detectors and descriptors, too many to cite extensively, but the most common being \cite{lowe99object,mikolajc03survey}. These are compared empirically on end-to-end tasks such as wide-baseline matching \cite{mikolajc03survey}, categorization \cite{feifei04learning}, or category localization and segmentation \cite{martinFM04} tasks. However, an empirical evaluation tells us which scheme performs better, but gives us no indication as to the relation between different schemes, no hint on how to improve them, and no bounds on the best achievable performance that can be extrapolated to other datasets with provable guarantees. 

{\em Therefore, there remains the need to develop a framework for the analysis and design of feature detectors/descriptors, that allows rational comparison of existing descriptors, and engineering design of new ones, and understanding of the conditions under which they can be expected to perform.}

There is a sizable literature on the detection and computation of structures in images. In particular, \cite{romenyFKV91} derive detectors based on a series of axioms and postulates. However, while these explain how such low-level representations should be constructed, they give no indication as to why they would be needed in the first place. Much of the motivation in this literature stems from biology, and in particular the structure of early stages of processing in the primate visual system.\cut{ In this manuscript, we take a different approach. Rather than trying to explain biology, we ask the question of what is the best representation to support recognition tasks, regardless of the architecture or hardware limitations. It will be interesting, afterwards, to investigate whether the primate visual system behaves in a way that is compatible with the principles developed here.}

By its nature, this manuscript relates to a vast body of literature in low-level vision and also Active Vision \cite{andreopoulosT09,ballard}. Ideally it relates to every paper, by providing a framework where different approaches can be understood and compared. 
However, it is possible that some approaches may not fit into this framework. I  hope that this work provides a seed that others can grow or amend.

This manuscript also lends some analytical support for the notion of {\em embodied cognition} that has been championed by cognitive roboticists and philosophers including \cite{brooks1981symbolic,lakoff1999philosophy,varela1999embodied,pfeifer2007body,mataric2002integration,ballard}. 

Finally, the work of Naftali Tishby and co-workers, starting from \cite{tishbyPB00}, has been addressing similar questions using an information-theoretic framework; work is underway to combine and reconcile the two approaches.
}

\chapter{Formalizing Visual Decisions}

\begin{figure}[htb]
\begin{center}
\begin{tabular}{ll}
  \begin{tabular}{c} 
\begin{tabular}{c}\includegraphics[height=1in,width=.8in]{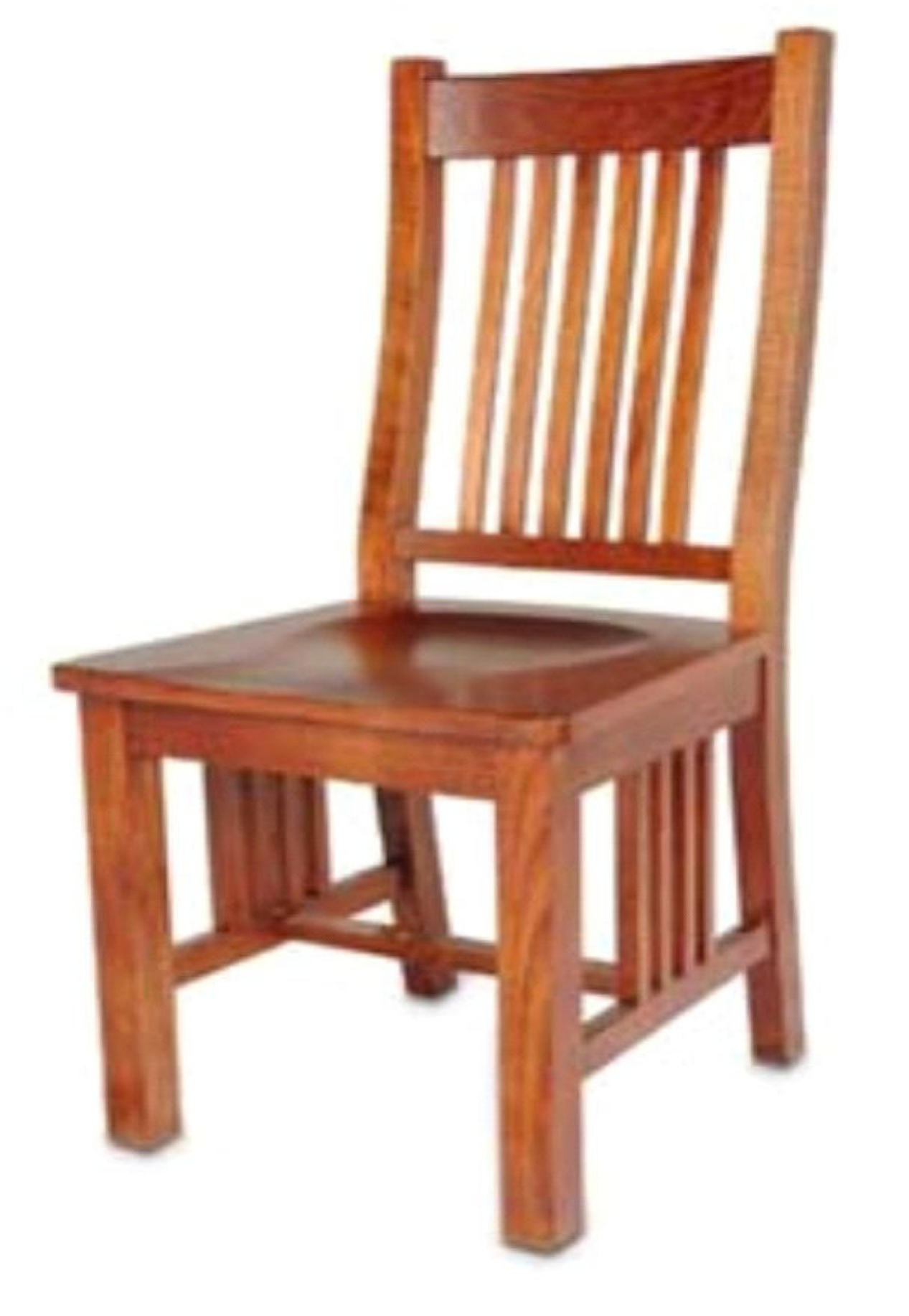}\\ $I_1$ \end{tabular} 
\begin{tabular}{c} \includegraphics[height=1in,width=1in]{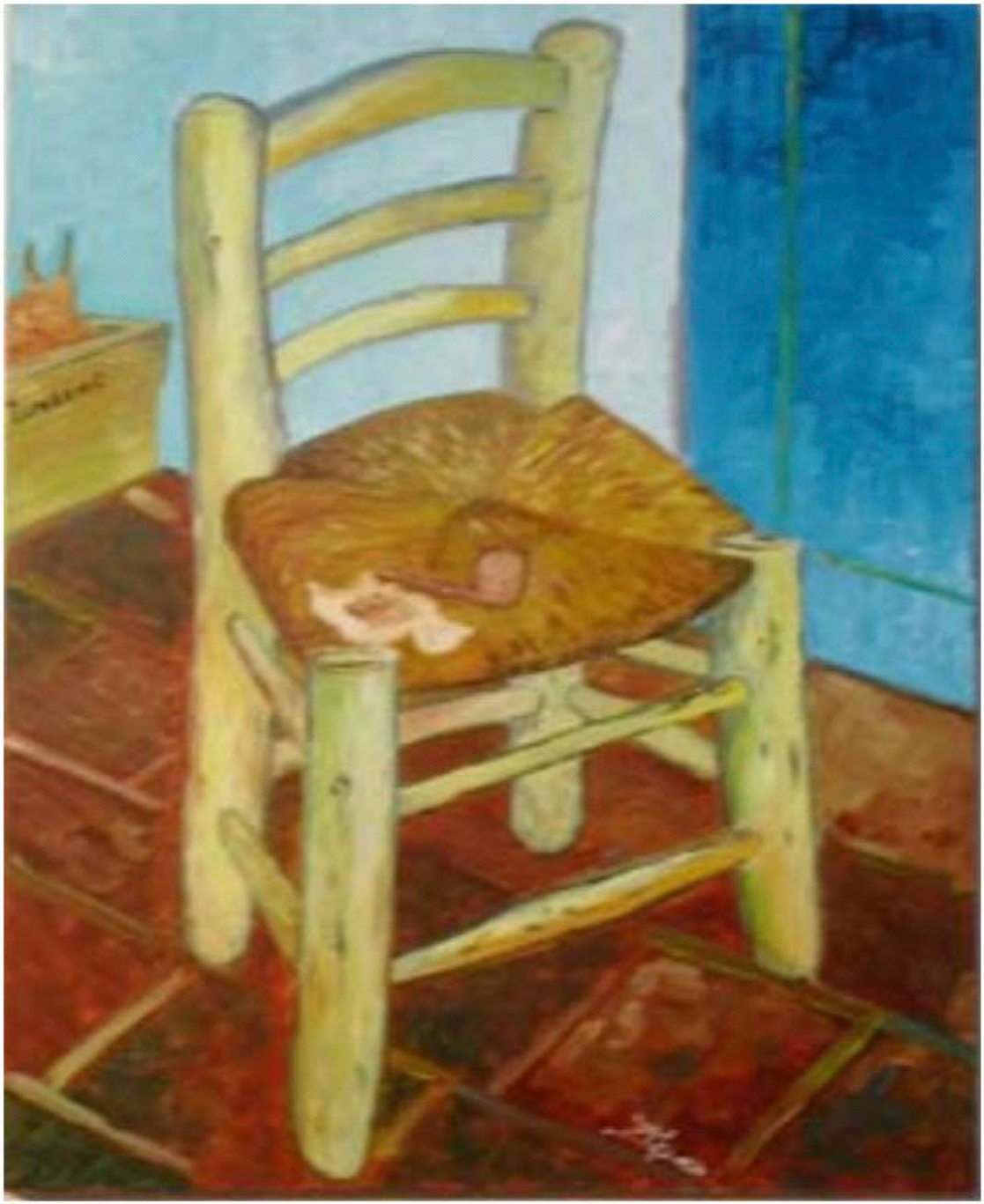}\\ $I_2$ \end{tabular}\\
\begin{tabular}{c} \includegraphics[height=1in,width=.8in]{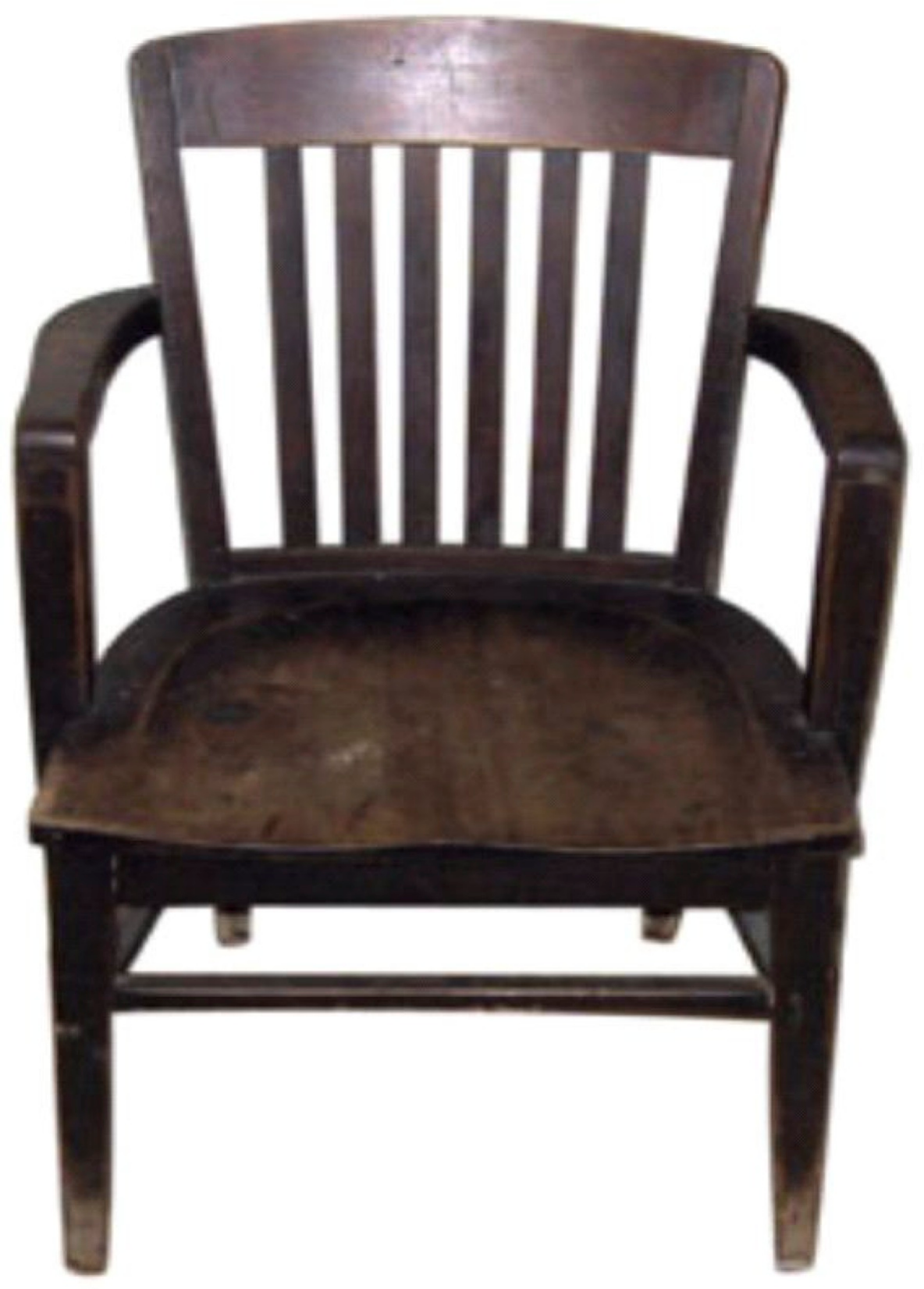}\\ $\ldots$ \end{tabular}
\begin{tabular}{c} \includegraphics[height=1in,width=1in]{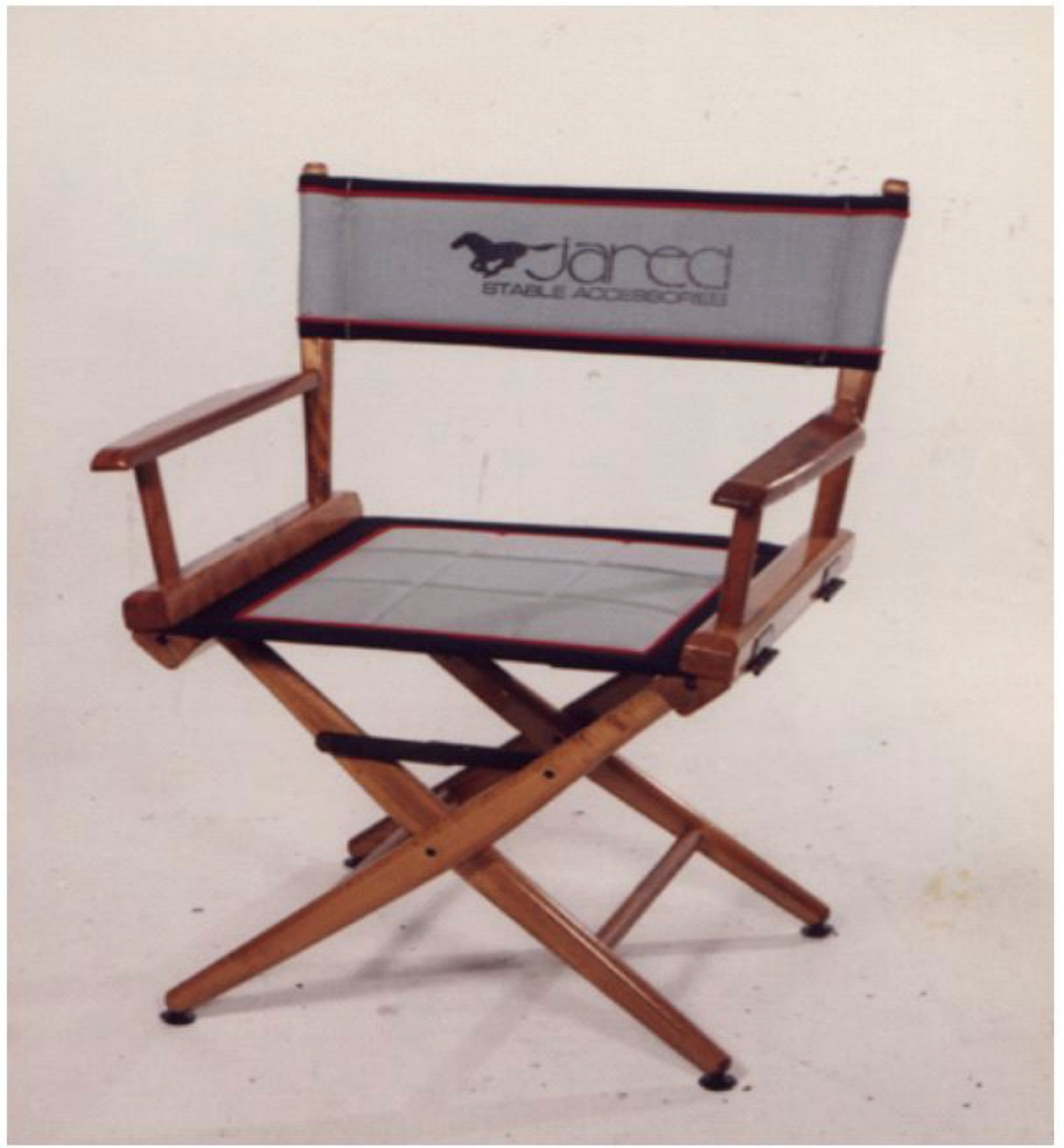}\\ $I_n$ \end{tabular}
\end{tabular} \vline
\begin{tabular}{c}
\begin{tabular}{c} \includegraphics[height=1in,width=.9in]{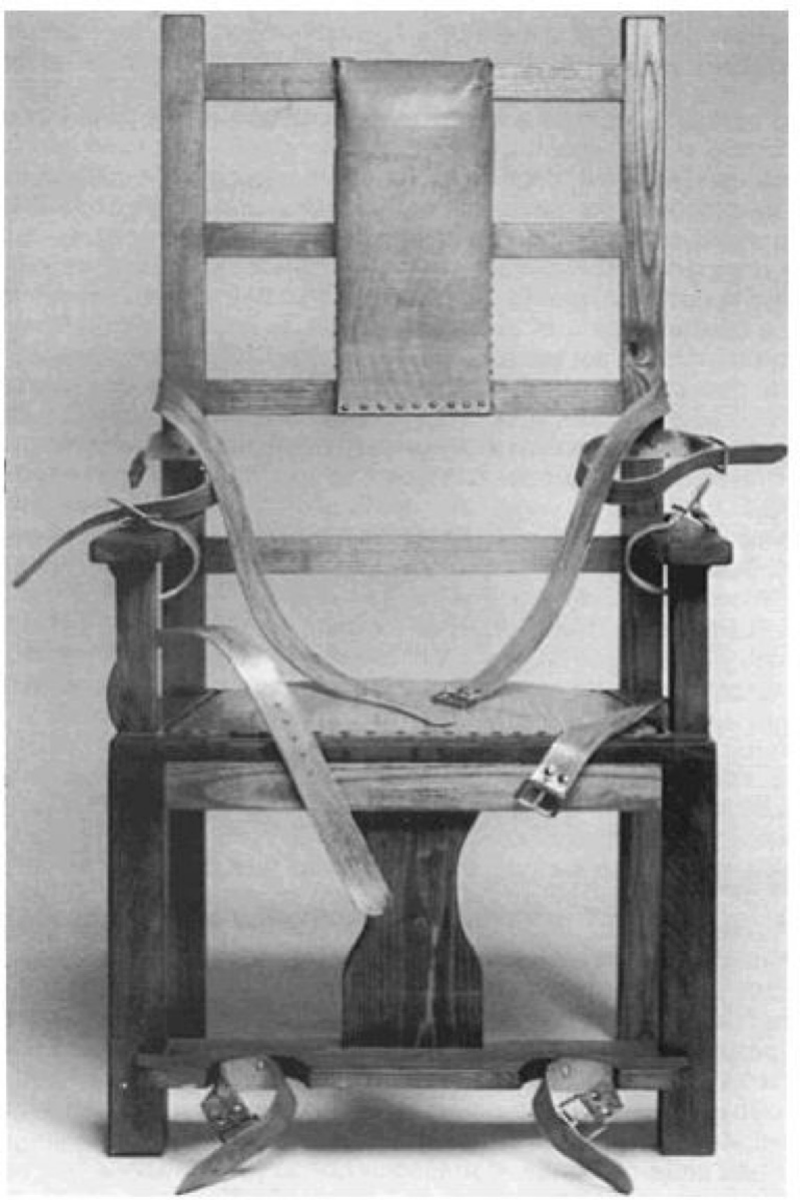}\\ similar shape?\end{tabular} 
\begin{tabular}{c} \includegraphics[height=1in,width=1in]{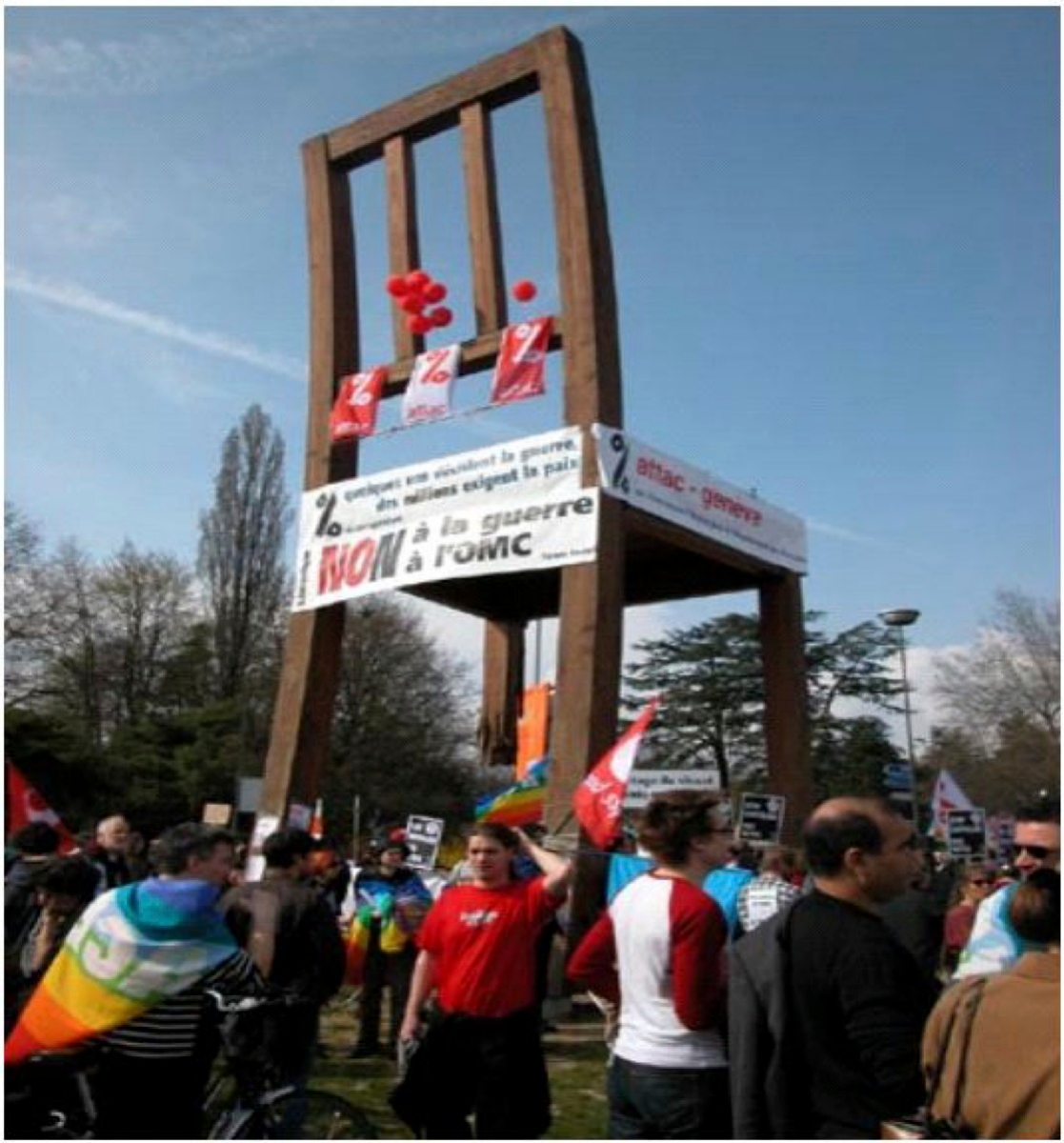}\\ similar appearance? \end{tabular}\\
\begin{tabular}{c}\includegraphics[height=1in,width=.9in]{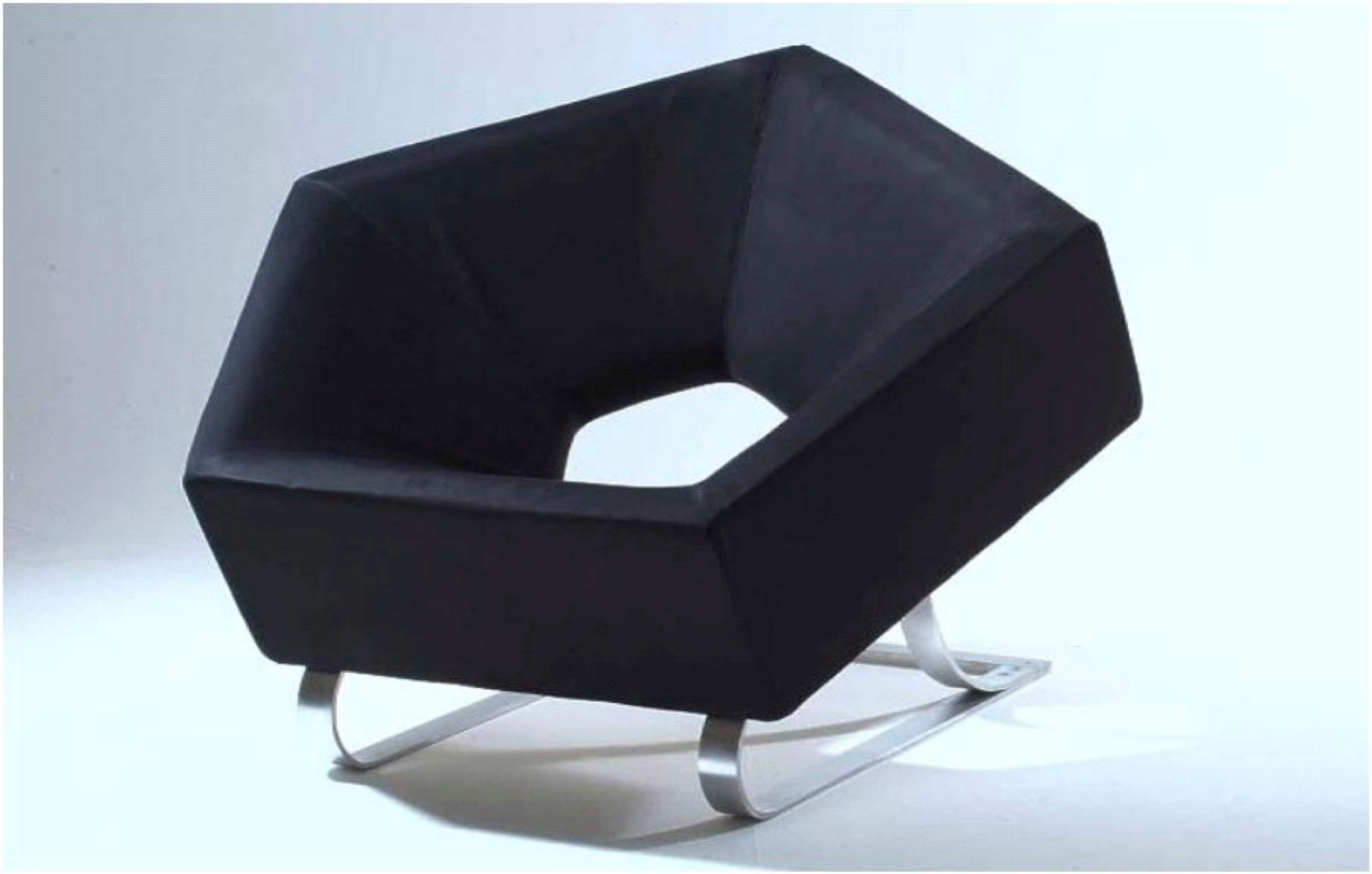} \\ similar function? \end{tabular}\begin{tabular}{c}\includegraphics[height=1in,width=1in]{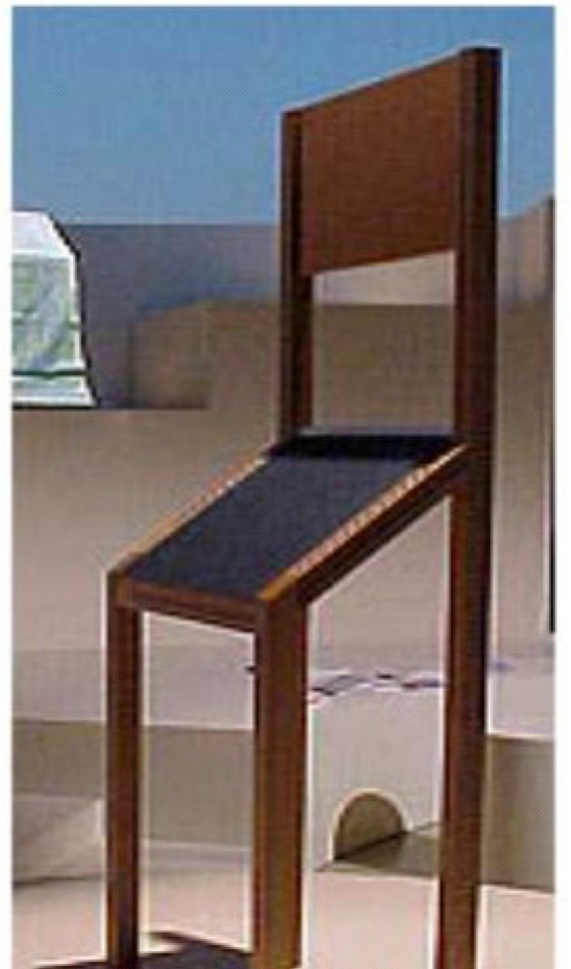} \\ ?? \end{tabular}
\end{tabular}
\end{tabular}
\end{center}
\vspace{-.51cm} \caption{\sl {\bf Category recognition:} Given images $\{I_1, \dots, I_n\}$ 
  of scenes that belong to a common class (left), does a new image
  $\tilde I$ (right) portray a scene that belongs to the same class?}
\label{fig-chairs}
\end{figure}

Visual classification tasks -- including  detection, localization, categorization, and recognition of general object classes in images and video -- are challenging because of large in-class variability. For instance, the class ``chair'', defined as ``something you can sit on'' (presumably man-made), comprises a diversity of shapes, sizes and materials that result in a wide variety of images (Figure \ref{fig-chairs}). 
Even if one sets aside within-class variability and considers the detection, localization or recognition of an individual object (\eg {\em this} chair) from a field of alternate hypotheses (\eg other chairs), the data still exhibits large variability due to {\em nuisance factors} \marginpar{\tiny \sc nuisance} such as viewpoint, illumination, occlusion, etc., that have little to do with the identity of the object (Figure \ref{fig-scene}). \cut{\color{pea} The  class variability and variability due to nuisance factors in the data often make class-conditional densities overlap.}

It is tempting to hope that a powerful classifier fed with raw data could be trained to, somehow, \marginpar{\tiny \sc learn-away} discard the variability in images due to nuisance factors\cut{ such as illumination, viewpoint, partial occlusion etc.}, and reliably recognize object, their classes and relations in pictures. The results in \cite{sundaramoorthiPS09} suggest otherwise, since {\color{orange} the volume of the quotient of the set of images modulo changes of viewpoint and contrast is infinitesimal relative to the volume of the data. This means that} a hypothetical classifier fed with raw images would spend almost all of its resources learning the nuisance variability, rather than the intrinsic variability of objects of interest. This is realistic at the phylogenic (at the evolutionary time scale), but not at the individual (ontogenic) level. This would hold {\em a-fortiori} once complex illumination phenomena, occlusions,  and quantization -- all neglected in \cite{sundaramoorthiPS09} -- were factored in.\footnote{T. Poggio recently put forward the hypothesis that this is true also in biology, in the sense that the complexity of the primate visual system is mainly to deal with nuisances, rather than to capture the intrinsic variability of the objects of interest \cite{poggio11}.} 

Moreover, if one's training consists of individual images {\em each of a different scene}, as in Fig. \ref{fig-chairs}, the process is even more problematic as a single image does not afford the ability of disentangling nuisance variability from intrinsic variability, and the complexity of the scene is infinitely more complex than the complexity of (even infinitely many) images. Therefore, a hypothetical learning machine fed with however large a dataset of individual images, each of a different scene, would never even learn that there is a {\em scene}, with shape, reflectance, illumination, etc. but instead just learn patterns of intensity in the images.

It is equally tempting to hope that one could pre-process the data to obtain some ``features,'' that do not depend on these nuisances, \marginpar{\tiny \sc features} and yet retain all the ``information'' present in the data. Indeed, \cite{sundaramoorthiPS09} suggests a construction of such features that, however, {\color{orange} requires nuisances to have the structure of a group and hence} breaks down in the presence of complex illumination effects, occlusions and quantization. One could relax such strict ``invariance'' requirement to some sort of ``insensitivity'' but, in general, pre-processing can only reduce the performance of any classifier downstream \cite{shao98}. This brings into question the role of  {\em ``vision-as-pre-processing\footnote{For pre-processing to be sound, some kind of ``separation principle'' should hold, so that different modules of a visual inference system could be designed and engineered independently, knowing that their composition or interconnection would yield sensible end-to-end performance.} for general-purpose machine learning.''} Why should one perform\footnote{As we have already pointed out in the preamble, computational efficiency alone does not justify the discretization process.} segmentation, edge detection, feature selection and other generic low-level vision (pre-processing) operations, if the classification performance {\em decreases}?

\cut{So, either early-vision operations such as edge detection, segmentation, feature selection etc. should be forgone altogether,\cutTwo{ and all of us working on visual classification should instead focus on the design of ``super-classifiers,''} or {\em the design of representations for visual classification tasks should be grounded on new principles,} or perhaps newly understood old principles. \cutTwo{Such principles should yield representations that are {\em falsifiable}.\footnote{This is not the case for most current low-level pre-processing: There is no way to falsify the assertion that a point on a digital image lies on an edge, because everything is an edge at the appropriate scale. Similarly, there is no way to ascertain that the output of a segmentation procedure is ``correct,'' since the problem is defined by the functional that the algorithm solves.}} It is our goal in this manuscript to begin this process, which is obviously too extensive to tackle conclusively in one manuscript, by proposing a framework for the analysis and design of low-level representations, where existing schemes can be understood, improved, and compared on a rational -- not just empirical -- basis.}

This goes to the heart of a notion of ``information.'' Ideally, the purpose of vision would be to ``extract information'' from images, where ``information'' intuitively relates to whatever portion of the data ``matters'' in some sense. Traditional Information Theory has been developed in the context data transmission, where one wants to reproduce as faithful as possible a copy of the data emitted by the source, after it has been corrupted by the channel. Thus the goal is {\em reproduction} (or reconstruction) of the {\em data,} with minimal distortion, and the ``representation'' simply consists of a compressed encoding of the data that exploits statistical regularity. In this context, every bit counts, and the semantic aspect of information is indeed irrelevant, as Shannon famously wrote. The theory yields a tradeoff between the minimum size of the representation as a function of the maximum amount of distortion. This tradeoff is computed explicitly for very simple cases (\eg the memoryless Gaussian channel), but nevertheless the general formulation of the problem is one of Shannon's  most significant achievements.

In our context, the data (images) are to be used for decision purposes (detection, localization, recognition, categorization). The goal is to minimize risk. In this context\cut{, following \cite{sundaramoorthiPVS09}}, there may be conditions where most of the data is useless, and the semantic aspect is fundamental, for the sufficient statistics can be discrete (symbols) even when the data lives in the continuum.

Several have advocated the development of a theory, mirroring Shannon's Rate-Distortion Theory, to describe the minimum requirements (in terms of size of the representation, computational or other ``cost'') in order to have a recognition error that is bounded above. Ideally, like in Shannon's case, this bound could be made arbitrarily small by paying a high-enough price. Despite many efforts, no theory has emerged, only special cases restricted to imaging modalities where some of the crucial aspects of image formation (scaling and occlusions) are not manifest. This may not be by chance because, in the context of visual recognition, the worst-case scenario is an arbitrarily high error rate. This is not surprising, and indeed can be considered trivial. What may be a bit more surprising is that even the {\em average-case} scenario can be arbitrarily bad, as we will argue  in Section \ref{sect-passive-bounds}. This may also shed some light on the limitations of benchmark datasets, if the performance of a given algorithm is interpreted as representative of performance on other datasets. The result of the benchmarking are meaningful to the extent in which the dataset is representative of the scenarios one wishes to capture, but no guarantee can be made on the generalization properties of these methods. Again, scale, quantization and occlusion conjure towards the failure of any ``passive'' recognition scheme to provide generalization bounds. What is surprising is that, if the data acquisition process can be controlled, then both the worst-case and average-case error can be bounded, and indeed they can be made (asymptotically) arbitrarily small (Section \ref{sect-active-bounds}). Analogously to Shannon's Rate-Distortion theory, there is a tradeoff between the ``control authority'' one can exercise over the sensing process, and the performance in a decision task (Section \ref{sect-control-recognition}). 

In the next section, we begin the formalization process necessary to answer some of the questions raised so far. For the purpose of simplicity, we will reduce visual perception to a collection of classification tasks, which is admittedly restrictive, but sufficient to commence formalization.

\section{Visual decisions as classification tasks}
\label{sect-visual-decisions}

\index{Visual decision}
By ``visual decision'' we mean tasks such as detection, localization, categorization and recognition of objects \marginpar{\tiny \sc objects} \index{Object} in images or video. These are all {\em classification} problems, where in some cases the class is a singleton (recognition), in other cases it can be quite general depending on functional or semantic properties of objects (Figure \ref{fig-chairs}). Conceptually, they all require the evaluation and learning of the  {\em likelihood} of the data (one or more images $I$) given the class label {\color{pea} $c$: $p(I|c)$. To simplify the narrative, we consider binary classifiers $c \in \{0, 1\}$ with equal prior probability $P(c) = \frac{1}{2}$. Generalizations are conceptually, although not computationally, straightforward.}

A decision rule, or a classifier, \index{Classifier} is a function $\hat c: {\cal I} \rightarrow \{0, 1\}$ mapping the set of images onto labels. It is designed to keep the average loss from incorrect decisions small. {\color{pea} A \index{Loss function} loss function $\lambda: \{0, 1\}^2 \rightarrow \real^+; (\hat c, c) \mapsto \lambda(\hat c, c)$ maps two labels to a positive real value. We will consider, for simplicity, the symmetric $0-1$ loss, \index{Loss function!symmetric 0-1} where $\lambda(\hat c, c) = 1-\delta(\hat c, c)$, where $\delta$ is Kronecker's delta, \index{Kronecker's Delta} that is $1$ if the labels are the same, $0$ if they are different. The average loss, a.k.a. {\em conditional risk}, \index{Conditional risk} is given by \marginpar{\tiny \sc risk} \index{Risk!conditional}
\begin{equation}
R(I, \hat c) = \sum \lambda(\hat c, c)P(c|I).
\label{eq-cond-risk}
\end{equation}
It can be shown \cite{dudaH73} that the decision rule that minimizes the conditional risk, that is
\begin{equation}
\hat c = \hat c(I) \doteq \arg\min_c R(I,c)
\label{eq-min}
\end{equation}
is optimal in the sense that it minimizes the expected (Bayesian) risk \index{Bayesian risk} \index{Risk!Bayesian} \index{Expected risk}\index{Risk!expected}\index{Average risk}\index{Risk!average}
\begin{equation}
R(c) \doteq \int R(I, c) dP(I).
\label{eq-bayes-risk}
\end{equation}
That is, if one could actually compute this quantity, which depends on the availability of the probability measure $dP(I)$, which is tricky to even define, let alone learn and compute \cite{mumfordG01}. However, in the context of our investigation this is irrelevant:} Whatever mathematical object $dP(I)$ is, we can easily sample from it by simply capturing images $I \sim dP(I)$, as we will see in Section \ref{sect-exploration}. We will see in Section \ref{sect-hallucination} that, even to generate ``simulated images,'' we do not need access to the ``true'' distribution $dP(I)$, but rather to a {\em representation}, which we will define in Section \ref{sect-canonization} and discuss in Section \ref{sect-hallucination}. 

Under the assumptions made, minimizing the conditional risk is equivalent to maximizing the {\em posterior} $p(c|I)$, which in turn {\color{pea} (under equiprobable priors $P(c=1) = P(c=0) = 1/2$)} is equivalent to maximizing the {\em likelihood} $p(I|c)$:  \index{Posterior distribution} \index{Likelihood}
\begin{equation}
\boxed{\hat c = \arg\max_{c \in \{0, 1\}} p(I|c).}
\label{eq-chat}
\end{equation}
So, in a sense, the problem of visual decision-making, including detection, localization, recognition, categorization, is encapsulated in (\ref{eq-chat}). That would be easy enough to solve if we could actually compute the likelihood. 

\begin{figure}[htb]
\begin{center}
\begin{tabular}{ll}
  \begin{tabular}{c} \includegraphics[height=.2\textwidth,width=.25\textwidth]{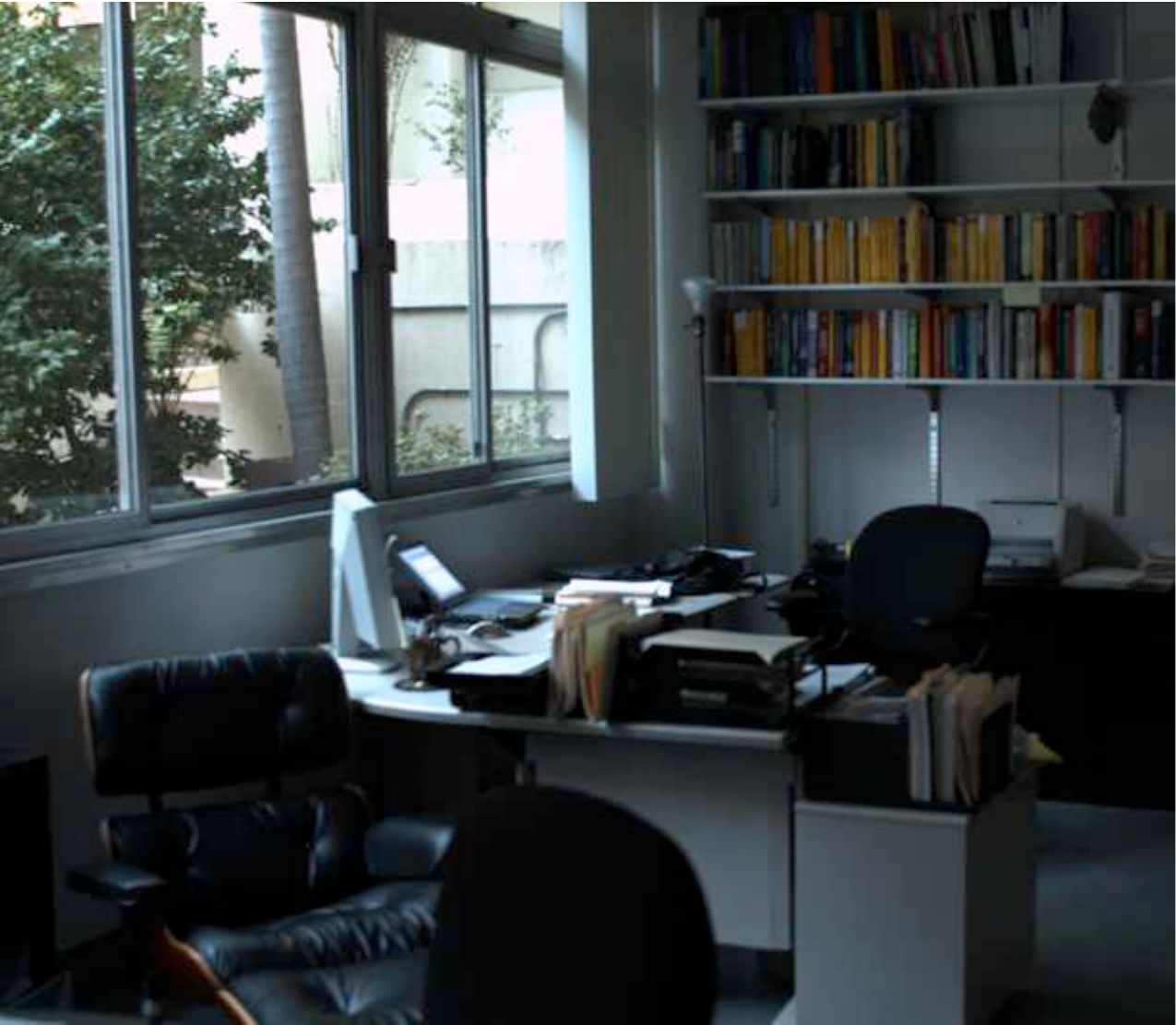} \\
$I = h(\xi, \nu)$ 
   \end{tabular} \vline &
  \begin{tabular}{c}
  \includegraphics[height=.2\textwidth,width=.25\textwidth]{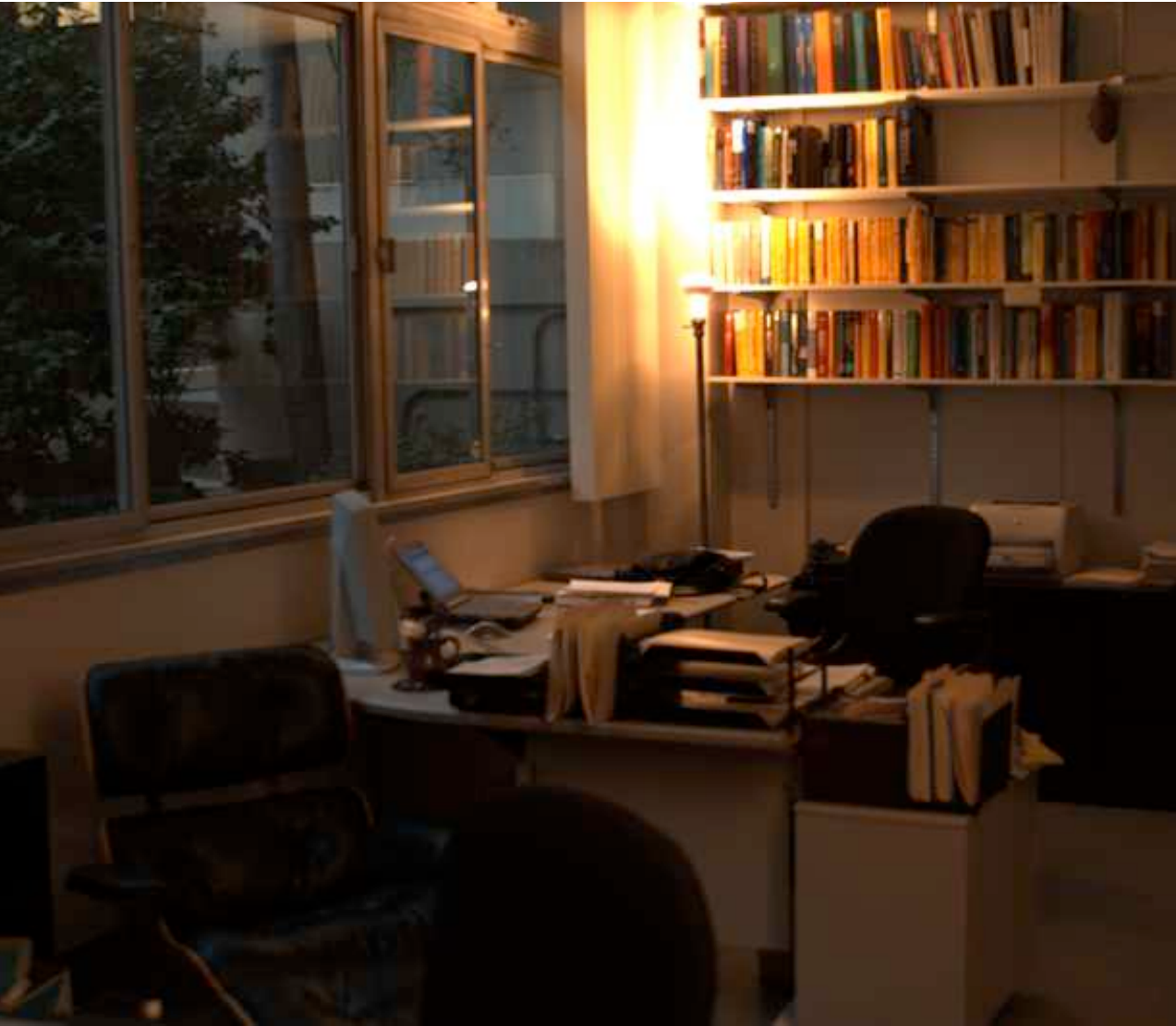} \\
  $\tilde I = h(\xi, \tilde \nu), ~~ \tilde \nu =$ illumination\\
  \includegraphics[height=.2\textwidth,width=.25\textwidth]{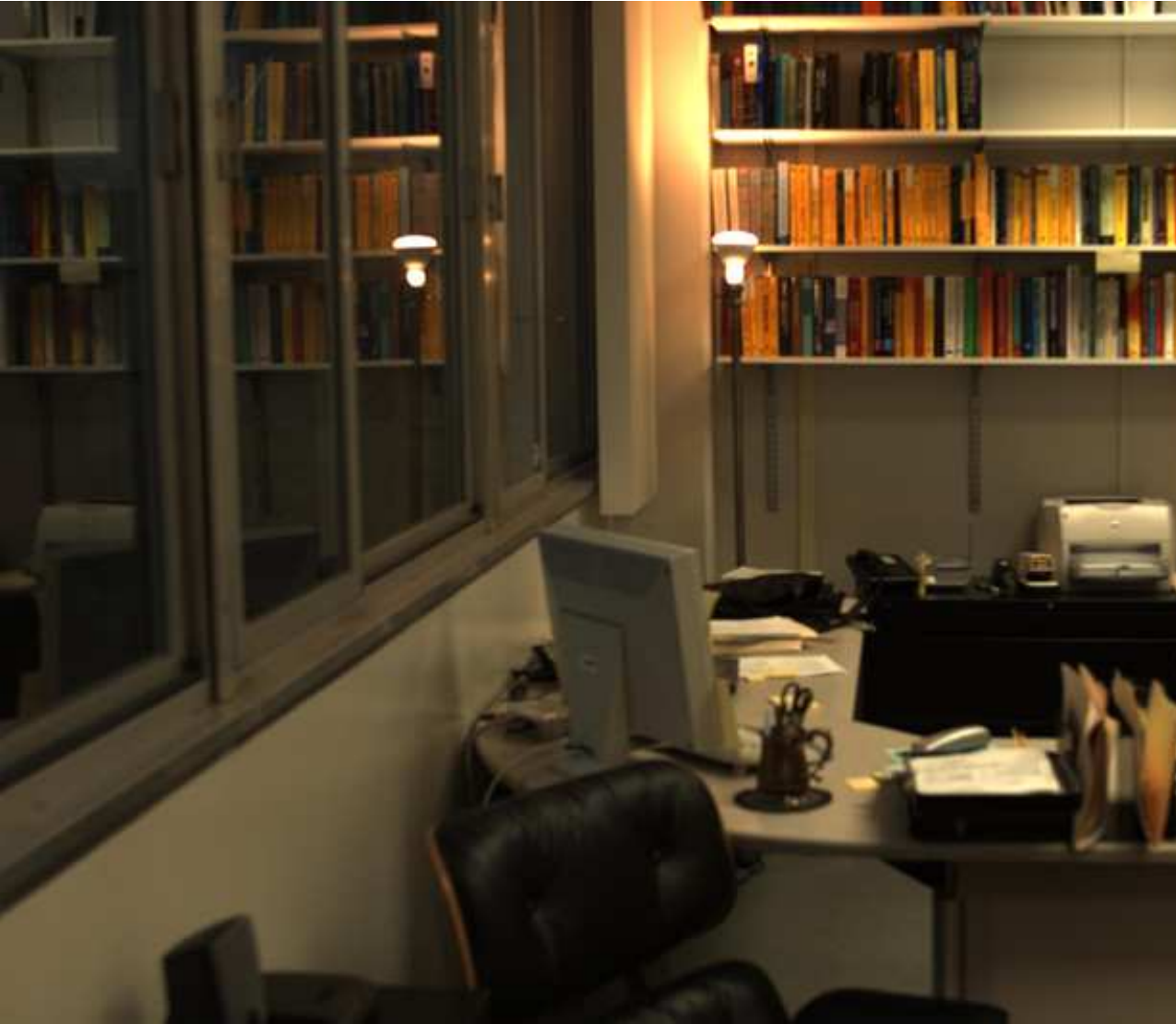} \\
  $\tilde \nu =$ viewpoint\\
  \end{tabular}
\begin{tabular}{c}
\includegraphics[height=.2\textwidth,width=.25\textwidth]{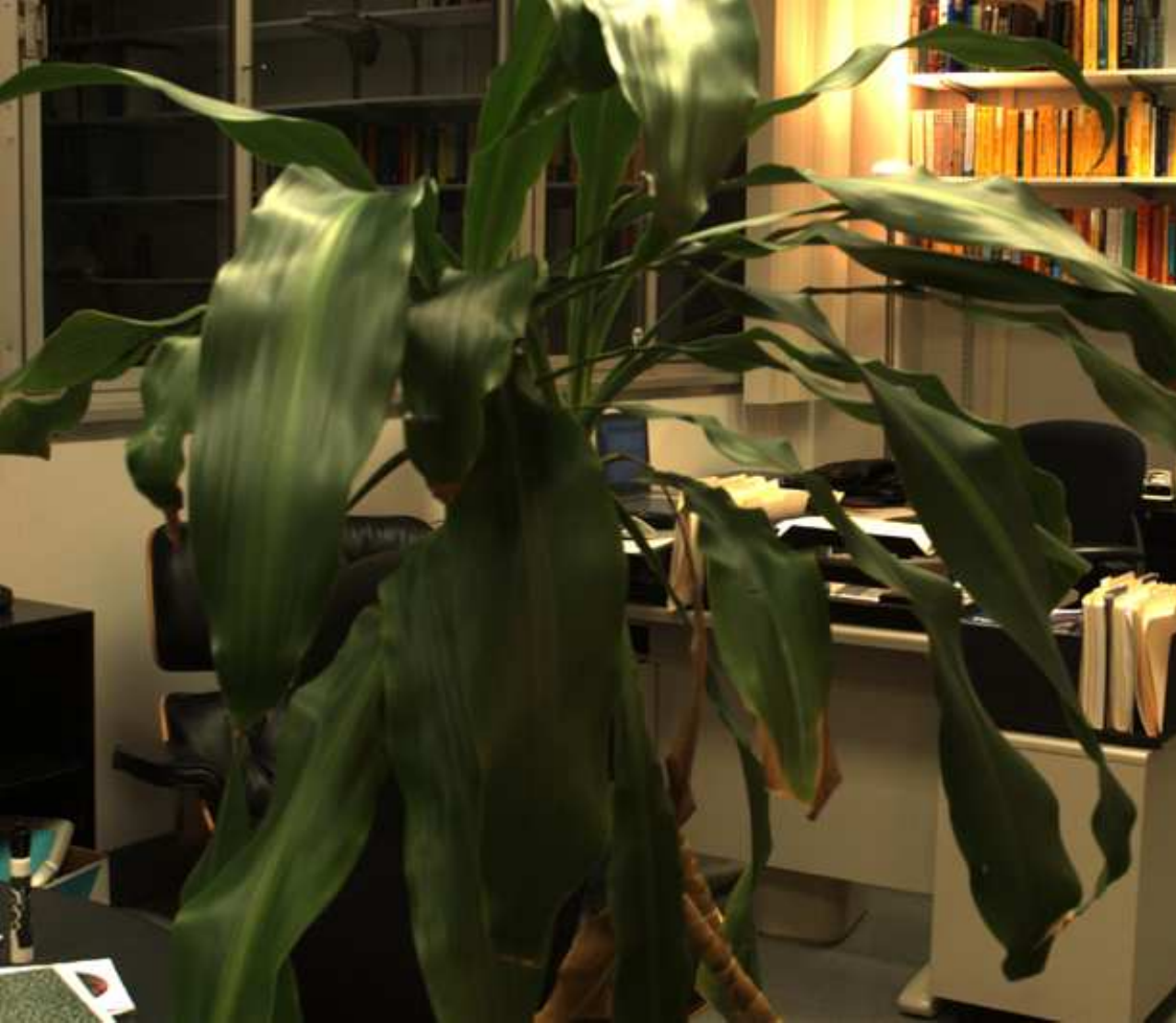} \\
  $\tilde \nu =$ visibility \\
\includegraphics[height=.2\textwidth,width=.25\textwidth]{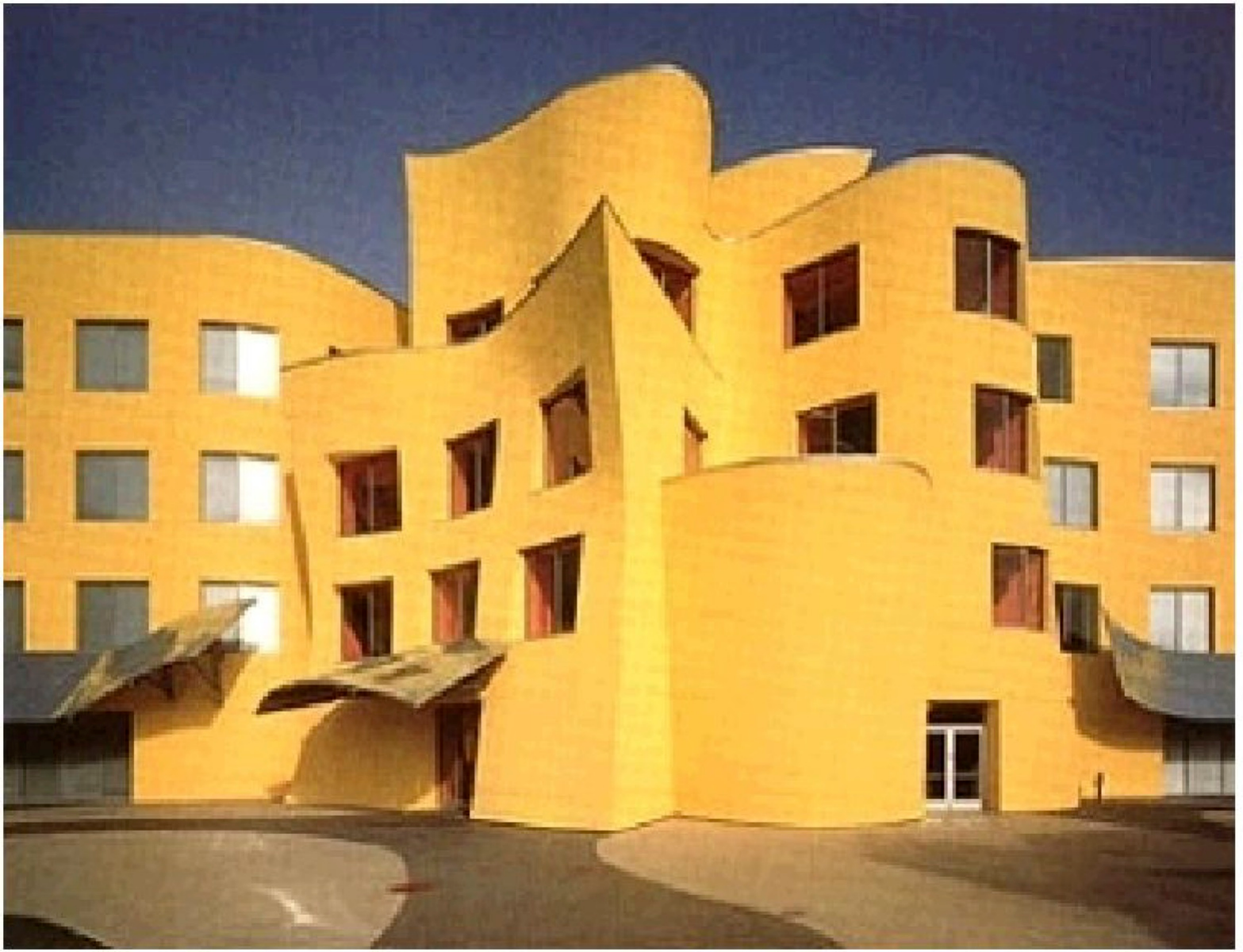}\\
$\tilde I = h(\tilde \xi, \tilde \nu), ~~ \tilde \xi \neq \xi$
  \end{tabular}
\end{tabular}
\end{center}
\caption{\sl The same scene $\xi$ can yield many different images depending on particular instantiations of the nuisance $\nu$.}
\label{fig-scene}
\end{figure}

{\em The difficulty in visual decision problems arises from} the fact that the image $I$ depends on a number of {\em nuisance factors} \marginpar{\tiny \sc nuisance} \index{Nuisance factor, or Nuisance} that do not depend on the class, and yet they affect the data. What is a nuisance depends on the task, and may include viewpoint, illumination, partial occlusions, quantization etc. (Figure \ref{fig-scene}). If we could, we would base our decision {\em not} on the data $I$, but on hidden variables $\xi$ that comprise the defining characteristics of the {\em scene} (object, category, location, event, activity etc.) that depend on the class $c$, {\color{pea} through a Markov chain $c \rightarrow \xi \rightarrow I$.  \marginpar{\tiny \sc scene} \index{Scene} \index{Markov chain} This would correspond to a data generation model whereby a sample $c$  is selected from $P(c)$, based on which a sample $\xi$ is selected from $dQ_c \doteq dP(\xi|c)$, from which a measurement $I$ is finally sampled via an image-formation functional $I = h(\xi)$.}

However, because of the nuisances, we have to instead consider a generative model of the form $I = h(\xi, \nu)$, where $h$ is a functional that depends on the imaging device and $\nu$ are all the nuisance factors. It is convenient to isolate within the nuisance $\nu$ the additive noise component $n$ arising from the compound effects of un-modeled uncertainty, although there is no added generality as $n$ can be subsumed in the definition of $\nu$. It is also useful to isolate the nuisances that act as a {\em group} on the scene, $g$, although again we could lump them into the definition of $\nu$. If we model explicitly the group and the noise, we have a model of the form \index{Image-formation model}
\begin{equation}
\boxed{I = h(g, \xi, \nu) + n}
\label{eq-gen2}
\end{equation}
This is the formal model that we will adopt throughout the manuscript (Figure \ref{fig-scene}). In the next section we make this formal notation a bit more precise with a specific instantiation, the so-called {\em Ambient-Lambert model}. More realistic instantiations are described in Appendix \ref{sect-image-formation}. The reader interested in generalizations of the simple symmetric binary decision case can consult any number of textbooks, for instance \cite{dudaH73}.

\section{Image formation: The image, the scene, the nuisance, and the Lambert-Ambient (LA) Model}
\label{sect-im-form}

\begin{figure}[htb]
\begin{center}
\includegraphics[width=.5\textwidth]{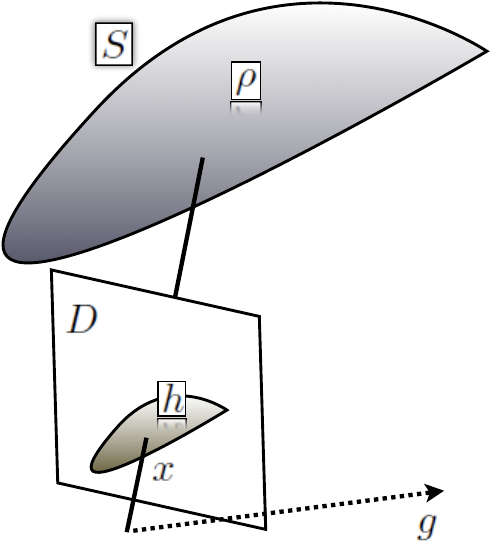}
\end{center}
\caption{\sl Image-formation model: The scene is described by objects whose geometry is represented by $S$ and photometry (reflectance) $\rho$. The image, taken from a camera moving with $g$, is obtained by a central projection onto a planar domain $D$, and a contrast transformation.}
\label{fig-image-formation}
\end{figure}
{In this section, that can be skipped at first reading, we instantiate the formal notation (\ref{eq-gen2}) for a simple model used throughout the manuscript. All the symbols used, together with their meaning, are summarized for later reference in Appendix \ref{sect-notation} in the order in which they appear. This section is necessary to make the formal notation above meaningful. However, its content will actually not be used until Sections \ref{sect-hallucination}, \ref{sect-interaction}, and will be exploited in full only starting in Section \ref{sect-detectors}. Therefore, the reader can skip this section at first reading, and come back to it, or to Appendix \ref{sect-notation}, as needed. The model we introduce in this section is the simplest instantiation of (\ref{eq-gen2}) that is meaningful in the context of image analysis. More sophisticated models, and their relation to the simplest one introduced here, are described in Appendix \ref{sect-image-formation}.}

\index{Image}
{\color{pea} An image $I:D \subset \real^2 \rightarrow \real_+^k; \ x \mapsto I(x)$ is a positive-valued map defined on a planar domain $D$; we will focus on gray-scale images, $k = 1$, but extensions to color $k=3$ or multiple spectral bands can be made. The scene, indicated formally by $\xi = \{S, \rho\}$, is described by its shape $S$ and reflectance (albedo) $\rho$. \index{Shape} The shape component $S\subset \real^3$ is described by piecewise smooth surfaces. The surface $S$ does not need to be simply connected, and can instead be made of multiple connected components, $S_i$, with $\cup_i S_i = S$, each representing an {\em object}.\marginpar{\tiny \sc object}\index{Object}\index{Reflectance}\index{Albedo} The reflectance $\rho$ is a function defined on such surfaces, with values in the same space of the range of the image, $\rho:S: \rightarrow \real^k$. In the presence of an explicit illumination model, $\rho$ denotes the diffuse albedo of the surface. In the absence of an illumination model, one can think of the surfaces as ``self-luminous'' and $\rho(p)$ denotes \index{Self-luminous} the energy\cut{ \marginpar{\tt check energy, irradiance, density}} 
emitted by an infinitesimal area element at the point $p$ isotropically in all directions. 
We call the space of scenes $\Xi$. Deviations from diffuse reflectance  (inter-reflection, sub-surface scattering, specular reflection, cast shadows) will not be modeled explicitly and lumped as additive errors $n$. \index{Diffuse reflectance}}

Nuisance factors in the image-formation process are divided into two components, $\{ g, \nu\}$, one that has the structure of a group $g\in G$, and a component that is not a group, \eg quantization, occlusions, cast shadows, sensor noise etc. We denote the image-formation model formally with a functional $h$, so that
$ I = h(g,\xi, \nu) + n$ as in (\ref{eq-gen2}). This highlights the role of group nuisances $g$, the scene $\xi$, non-invertible nuisances, and the additive residual that lumps together all unmodeled phenomena including noise and quantization (non-additive noise phenomena can be subsumed in the non-invertible nuisance $\nu$). \index{Nuisance!group} \index{Nuisance!invertible} \index{Nuisance!non-invertible} We often refer to the group nuisances as {\em invertible} \marginpar{\tiny \sc invertible nuisance} and the non-group nuisances as {\em non-invertible}.

The simplest instantiation of this model is the so-called Lambert-Ambient-Static (LAS) model, \index{Lambert-ambient-static model} that approximates a static Lambertian scene seen under constant diffuse illumination, describing reflectance via a diffuse albedo function and thus neglecting complex reflectance phenomena (specularity, sub-surface scattering), and describing changes of illumination as a global contrast transformation and thus neglecting complex effects such as vignetting, cast shadows, inter-reflections etc. 
{\color{pink}
Under these assumptions, the radiance $\rho$ \index{Radiance} emitted by an area element around a visible point $p\in S$ is modulated by a \index{Contrast transformation} contrast transformation $k$ (a monotonic continuous transformation of the range of the image)  to give the \index{Irradiance} irradiance $I$ measured at a pixel element $x$, except for a discrepancy $n:D \rightarrow \real_+^k.$ The correspondence between the point $p\in S$ and the pixel $x\in D$ is due to the motion of the viewer $g\in SE(3)$, the special Euclidean group of rotations and translations in three-dimensional (3-D) space:
\begin{equation}
\begin{cases}
I(x) = k \circ \rho(p) + n(x); \\
x = \pi(g p); ~~~ p \in S
\end{cases}
\label{eq-lambert-ambient}
\end{equation}
\marginpar{\tiny \sc lambert-ambient model}
\index{Lambert-ambient-static model} where if $p$ is represented by a vector $X\in \real^3$, then $\pi:\real^3 \rightarrow \real^2; X \mapsto x = [X_1/X_3, \ X_2/X_3]^T$ is a central perspective projection (Figure \ref{fig-perspective}). \index{Perspective projection} 
\begin{figure}[htb]
\begin{center}
\includegraphics[width=.8\textwidth]{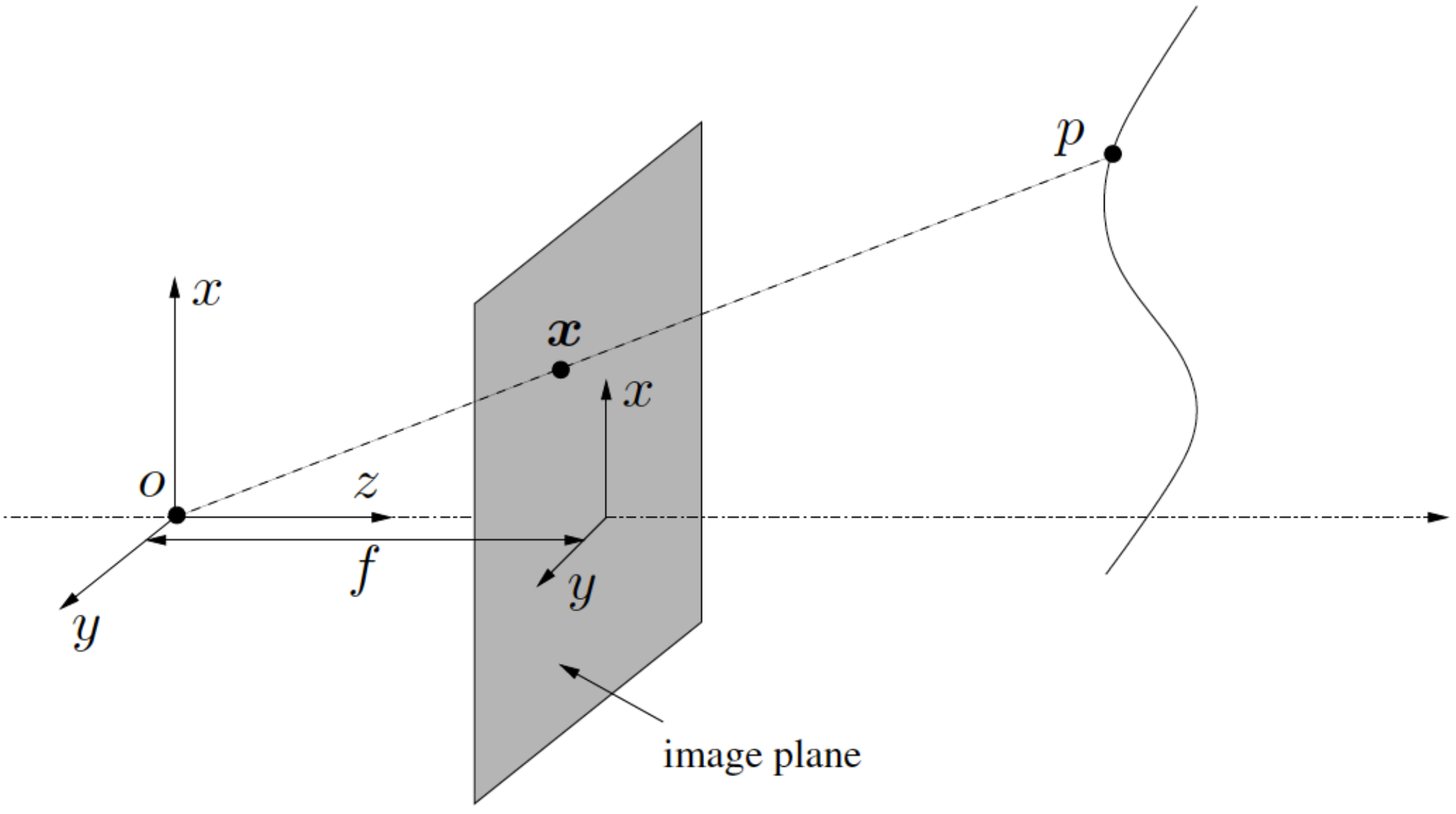}
\end{center}
\caption{\sl Perspective projection: $x$ is a point on the image plane, $p$ is its pre-image, a point in space. Vice-versa, if $p$ is a point in space, $x$ are the coordinates of its image, under perspective projection.}
\label{fig-perspective}
\end{figure}
This equation is not satisfied for all $x \in D$, but only for those that are projection of points on the object, $x\in D \cap \pi(gS)$. Also, not all points $p\in S$ are visible, but only those that intersect the projection ray \index{Projection ray} \marginpar{\tiny \sc projection ray} closest to the optical center\cut{ (Figure \ref{fig-perspective})}. If we call $p_1$ the intersection of $S$ with the projection ray through the origin of the camera reference frame and the pixel with coordinates $x\in \real^2$, represented by the vector $\bar x$, we can write $p_1$ as the graph of a scalar function $Z: \real^2 \rightarrow \real^+; x \mapsto Z(x)$, the {\em depth map}\index{Depth map}\marginpar{\tiny \sc depth map}. Here a bar $\bar x \in {\mathbb P}^2 \simeq \real^2$ denotes the \index{Homogeneous coordinates} \index{Projective coordinates} homogeneous (projective) coordinates of the point with Euclidean coordinates $x\in \real^2$: $\bar x = \ba{c} x \\ 1 \ea$. 
}
\index{Euclidean coordinates}  More in general, the {\em pre-image} \index{Pre-image}\marginpar{\tiny \sc pre-image} of the point $x$ (on the image plane) is a collection of points (in space) given by 
\begin{equation}
\pi^{-1}_S(x) = \{p_1, \dots, p_N \in S \} = \{ g^{-1} \bar x Z_i(x)\}_{i=1}^{N(x)}.
\label{eq-preimage}
\end{equation}
Note that the number of points in the pre-image depends on $x$, and is indicated by $N(x)$. The pixel locations where two pre-images coincide are the {\em occluding boundaries}, \marginpar{\tiny \sc occluding boundaries} \index{Occluding boundaries} also known as {\em silhouettes}. \marginpar{\tiny \sc silhouette} \index{Silhouette! see occluding boundary} {\color{pea} For instance,  $\{x \ | \ Z_1(x) = Z_j(x)\}$ for some $j = 2, \dots, N$ is an occluding boundary.}  If we sort the points in order of increasing depth, so that $Z_1(x) \le Z_2(x) \le \dots \le Z_N(x)$, then the pre-image, {\em restricted to the point of first intersection}, is indicated by
\begin{equation}
\pi^{-1}(x) = p_1 = g^{-1} \bar x Z(x)
\label{eq-pi-1}
\end{equation}
where we indicate $Z_1(x)$ simply with $Z(x)$. We omit the subscript $S$ in the pre-image for simplicity of notation, although we emphasize that it depends on the geometry of the surface $S$. We also omit the temporal index $t$, although we emphasize that, even when the scene $S$ is static (does not depend on time), changes of viewpoint $g_t$ will induce changes of the range map $Z_t(x)$. We also omit the contrast transformation $k$, that can also change over time, although we will re-consider all these choices later.

{\color{pink} 
Note that the image only exists for $x\in D \cap \pi(gS)$, and the pre-image only exists for $p \in S \cap \pi^{-1}(D)$.\cut{ If we put the two together, we have that the image-formation model we have derived is valid only for $x\in D \cap \pi(g \{S \cap \pi^{-1}(D)\})$.} For the region of the image where the scene is visible, say at time $t = 0$, we can represent $S$ as the graph of a function defined on the domain of the image $I_0$, so $p = \pi^{-1}(x_0) = \bar x_0 Z(x_0)$. Here $k$ can be taken to be the identity function, $k_0 = Id$, so that $I_0(x_0) = \rho(p) ~~~ p = \pi^{-1}(x_0).$  Therefore, combining this equation with (\ref{eq-lambert-ambient}), we have $I(x) = k\circ \rho\circ \pi^{-1}(x_0)$, with the relation between $x$ and $x_0$ being 
\begin{equation}
x = \pi g p = \pi g \pi^{-1}(x_0) \doteq w(x_0)
\label{eq-motion-field}
\end{equation}
where $w:\real^2 \rightarrow \real^2$ denotes the domain deformation induced by a change of viewpoint $g$, and the photometric relation being
\begin{equation}
I\circ w(x_0) = k \circ I_0(x_0).
\label{eq-opt-fl}
\end{equation}
This is know as the ``brightness constancy constraint equation.'' \marginpar{\tiny \sc brightness constancy constraint} \index{Brightness constancy constraint} Note that the two previous equations are simultaneously valid only in 
\begin{equation}
x_0 \in D \cap w(D) = D \cap \pi g \pi^{-1}(D)
\end{equation}
which is called the {\em co-visible region}. \marginpar{\tiny \sc co-visible region}\index{Co-visible region}
It can be shown that the composition of maps $w: \real^2 \rightarrow \real^2; \ x_0 \mapsto x =  w(x_0) \doteq \pi(g \bar x_0 Z(x_0))$ spans the entire group of diffeomorphisms (Theorem 2 of \cite{sundaramoorthiPVS09}) as the function $Z(\cdot)$ is allowed to vary (which is necessary, unless it is known). 
}
In this model, visibility \index{Visibility} is captured by the map $\pi$ and its inverse. In particular, $\pi: \real^3 \rightarrow \real^2$ maps any points in space $p$ into one location $x = \pi(p)$ on the image plane, assumed to be infinite. So the image of a point $p$ is unique. \index{Pre-image} However, the {\em pre-image} of the image location $x$ is not unique, as there are infinitely many points that project onto it. If we assume that the world is populated by opaque objects, then the pre-image $\pi^{-1}_S(x)$ consists of all the points on the surface(s) $S$ that intersect the projection ray from the origin of the camera reference frame through the given image plane location $x$. This model does not include quantization, noise and other phenomena that are discussed in more detail in the appendix.

\section{Marginalization, extremization (max-out), blurring}
\label{sect-ml}
\index{Marginalization}
\index{Extremization}
\index{Max-out}

If we have prior knowledge on all the hidden variables, $\xi, g, \nu, n$, we can compute the likelihood in (\ref{eq-chat}) by marginalization. This is conceptually trivial, but computationally prohibitive. Prior knowledge on the nuisance is encoded in a distribution $dP(\nu)$, {\color{pea} which may have a density $p(\nu)$ with respect to a base measure $d\mu(\nu)$.} Prior knowledge on the scene is encoded in the {\em class-conditional} \index{Class-conditional distribution} distribution $dQ_c(\xi)$, {\color{pea} which may again have a density $q(\xi | c)$ with respect to a base measure $d\mu(\xi)$.} The same goes with the group $g$.\footnote{One may encode complete ignorance of some of the group parameters by allowing a subgroup of $G$ to have a uniform prior, a.k.a. {\em uninformative}, and possibly {\em improper,} \ie not integrating to one, if the subgroup is not compact.}  \index{Prior!uninformative} \index{Prior!improper} We can reasonably\footnote{If this is not the case, that is if the residual exhibits significant spatial or temporal structure, such structure can be explicitly modeled, thus leaving the residual white.} assume that the residual, after all relevant aspects of the problem have been explicitly modeled, is a white zero-mean Gaussian noise $n \sim {\cal N}(0; \Sigma)$ with covariance $\Sigma$. Marginalization then consists of the computation of the following integral
\begin{equation}
{p(I|c) = \int {\cal N}(I-h(g, \xi, \nu);\Sigma)dP(g)dP(\nu)dQ_c(\xi).}
\label{eq-marg}
\end{equation}
\index{Gaussian density}
\index{Normal density}
The problem with this approach is not just that this integral is difficult to compute. Indeed, even for the simplest instantiation of the image formation model (\ref{eq-lambert-ambient}), it is not clear how to even define a base measure \index{Base measure} on the sets of scenes $\xi$ and nuisances $\nu$, let alone putting a probability on them, and learning a prior model. It would be tempting to discretize the model (\ref{eq-lambert-ambient}) to make everything finite-dimensional; unfortunately, because of scaling and quantization phenomena in image formation, any reasonable discretization of the scene would yield a very large scale inference problem. This integral is costly to compute even for the simplest characterizations of the scene and the nuisances. Indeed, the space of scenes (shape, radiance distribution functions) does not admit a ``natural'' or ``sufficient'' discretization.  However, if one was able to do so, then a threshold on the likelihood ratio, depending on the priors of each class, yields the optimal (Bayes) classifier \cite{robert}. \index{Likelihood ratio} Recently, a {\em local} approximation of the above integral has been proposed in \cite{dongS15}, and named R-HOG (reconstructive HOG).

An alternative to marginalization, where all possible values of the nuisance are considered with a weight proportional to their prior density, is to find the class together with the value of all the hidden variables that maximize the likelihood:
\begin{equation}
{\tilde p(I|c) = \sup_{g, \nu, \xi} {\cal N}(I-h(g, \xi, \nu))p(g)p(\nu)q(\xi|c).}
\label{eq-ML}
\end{equation}
\index{Maximum likelihood}
This procedure of eliminating nuisances by solving an optimization problem is called ``extremization'' or sometimes ``max-out'' of the hidden variables. A threshold on the result yields the {\em maximum likelihood} (ML) classifier. Needless to say, this is also a complex procedure. This procedure also relates to {\em registration} or {\em alignment} between (test) data and a ``template,'' a common practice in pattern recognition.  Recently, a {\em local} approximation of the above integral has been proposed in \cite{dongS15}, and named MV-HOG (multi-view HOG).

A third alternative to obtain statistics that reduce the variability of the data with respect to nuisances is to {\em average} the data with respect to the group action. For instance, a function of the image that is invariant to planar rotations can be easily computed by averaging rotated versions of the image. This is clearly lossy, as there are many different scenes that produce the same invariant, so one may want to restrict the averaging to {\em small} transformations. If the averaging is done properly \cite{mallatB11}, the process can be made lossless and therefore provide true invariance, at least to a small group of transformations, but in general produces statistics that are {\em insensitive} (as opposed to ``invariant'' or ``stable'').
The next section describes an alternative, further elaborated in Section \ref{sect-canonization}.
\index{Decision time} Note that the max-out procedure can be understood as a special case of marginalization when the prior is uniform (possibly improper). Therefore, we will use the term {\em marginalization} to refer to either procedure, depending on whether or not a prior is available.

\section{Features}
\index{Feature}

\index{Statistic} A feature is any deterministic function of the data, $\phi: {\cal I} \rightarrow \real^K; I \mapsto \phi(I)$, or sometimes $\phi \circ I$. In general, it maps onto a finite-dimensional vector space $\real^K$, although in some cases the feature could take values in an infinite-dimensional (function) space. Obviously, there are many kinds of features, so we are interested in those that are ``useful'' in some sense. The decision rule itself, $\hat c(I)$ is a feature. However, it does not just depend on the datum $I$, but also on the entire training set. Therefore, we reserve the nomenclature ``feature'' only for deterministic functions of the (sample) test data, but not on the training set. Deterministic functions of (ensemble) data are called ``statistics.'' We call any statistic of the training set (but not of the test datum) a {\em template}. \index{Template} For instance, the class-conditional mean is a template.\cut{ One could build a classifier as a combination of a feature and a template, but in general the classifier can operate freely on training and test data, and does not necessarily compute statistics on each set independently.} 

One can think of a feature as any kind of ``pre-processing'' of the data. \index{Pre-processing} The question as to whether such pre-processing is useful is addressed by the data processing inequality. 

\subsection{Data processing inequality}
\label{sect-dpi}
\index{Data processing inequality}
\index{Rao-Blackwell Theorem}

{\color{pea} Let $R(I, c)$ be the conditional risk (\ref{eq-cond-risk}) associated with a decision $c$, and $\hat c: {\cal I} \rightarrow \{0, 1\}; I \mapsto \hat c(I)$ be the optimal classifier, defined as $\hat c = \arg\min_c R(I, c)$. Let $\phi: {\cal I} \rightarrow \real^K$ be any feature. Then, if $R(c)$ is the Bayes risk (\ref{eq-bayes-risk}) associated with the classifier $c$, we have that  
\begin{equation}
\min_cR(c) \le \min_{\tilde c} R(\tilde c \circ \phi).
\label{eq-dpi}
\end{equation}
}
\cut{Note that the two functionals, $c(I)$ and $\tilde c\circ \phi(I) = \tilde c(\phi(I))$ belong to two different function spaces.}
In other words, there is no benefit in pre-processing the data. This results follows simply from the Markov chain dependency $c \rightarrow I \rightarrow \phi$, 
known as ``data processing inequality'' (\cite{coverT} Theorem 2.8.1, page 32, and the following corollary on page 33). Thus, it seems that the best one can do is to forgo pre-processing and just use the raw data. Even if the purpose of a feature $\phi(I)$ is to reduce the complexity of the problem, this is in general done at a loss, because one could include a complexity measure in the risk functional $R$, and still be bound by (\ref{eq-dpi}). However, there are some statistics that are ``useful'' in a sense that we now discuss.
\index{Complexity}

\subsection{Sufficient statistics}
\index{Sufficient statistic}

Those statistics that maintain the ``='' sign in the data processing inequality (\ref{eq-dpi}) are called {\em sufficient statistics} \marginpar{\tiny \sc sufficient statistics} for the purpose of this manuscript.\footnote{This is a less restrictive condition than the independence in Neyman's factorization.} Thus, $\phi: {\cal I} \rightarrow \real^K$ is a sufficient statistic if $\min_c R(c) = \min_{\tilde c}R(\tilde c \circ \phi)$. Of course, what is a sufficient statistic depends on the task, encoded in the risk functional $R$. A trivial sufficient statistic is the identity functional $\phi(I) = I$. Of all sufficient statistics, we are interested in the ``smallest,'' \ie, the one that is a function of all other sufficient statistics. This is called the {\em minimal sufficient statistic}, which we indicate with $\phi^\vee(I)$. 
\index{Minimal sufficient statistic}
\index{Sufficient statistic!minimal}

Clearly, sufficiency and minimality represent useful properties of a statistic. A minimal sufficient statistic contains everything in the data that matters for the decision; no more (minimality) and no less (sufficiency). Unfortunately, (finite-dimensional) sufficient statistics do not always exist, so relaxed versions of the notion of sufficient statistics have been developed \cite{tishbyPB00}, and will be discussed later in this manuscript. Another useful property of a feature is that it contains nothing that depends on the nuisances.

\subsection{Invariance}
\index{Invariant}
\index{Invariance}

In the image-formation model (\ref{eq-gen2}), we have isolated nuisances that have the structure of a group $g$, those that are not, $\nu$, and then the additive ``noise'' $n$. The latter is a very complex process that has a very simple statistical description (\eg independent and identically distributed samples from a Gaussian process in space and time). Instead, we focus on the group nuisances, $g\in G$ and on the {\em non-invertible}\footnote{The nomenclature ``non-invertible'' for nuisances that are not group stems from the fact that the crucial property of group nuisances that we will exploit is their invertibility.} nuisances $\nu$. \marginpar{\tiny \sc non-invertible nuisances}
\index{Maximal invariant}
\index{Invariant!maximal}
\index{Feature!invariant}
A statistic is $G$-invariant if it does not depend on the nuisance: $\phi(I) = \phi\circ h(g,\xi, \nu) = \phi\circ h(\tilde g, \xi, \nu)$ for any two nuisances $g, \tilde g$ and for any $\nu$ and $\xi\in \Xi$ 
. 
One could similarly define an invariant feature for the non-invertible nuisances.

Any constant function is an invariant feature, $\phi(I) = {\rm const}, \ \forall  \ I$; obviously it is not very useful. Of all invariant features, we are interested in the ``largest,'' in the sense that all other invariants are functions of it. We call this the {\em maximal invariant}, and indicate it with the symbol $\phi_G^\wedge(I)$ or $\phi^\wedge(I)$ when the group $G$ is clear from the context. \marginpar{\tiny \sc maximal invariant}

In general, there is no guarantee that an invariant feature, even the maximal one, be a sufficient statistic. In the process of removing the effects of the nuisances from the data, one may lose discriminative power, quantified by an increase in the expected risk. Vice-versa, there is no guarantee that a sufficient statistic, even the minimal, be invariant. In the best of all worlds, one could have a minimal sufficient statistic that is also invariant, \cut{\tt (in which case it would be the maximal invariant, otherwise if there was another invariant that contained it, it would not be minimal)} or vice-versa that a maximal invariant that is sufficient. \cut{\tt (in which case it would be the minimal sufficient statistic)} In this case, the feature $\phi^\vee(I) = \phi^\wedge(I)$ would be the best form of pre-processing one could hope for: It contains all and only the ``information'' the data contains about $\xi$, and have no dependency on the nuisance. We call a minimal sufficient invariant statistic \cut{\tt (or a maximal invariant sufficient statistic)} a {\em complete feature}. \marginpar{\tiny \sc complete feature}
\index{Complete feature}
\index{Feature!complete}
Note that, in general, a complete feature is still not equivalent to $\xi$ itself, for the map $\xi \rightarrow \phi$ may not be injective (one-to-one). 
However, it can be shown that when the nuisance has the structure of a group, one can define an invariant statistical model (Definition 7.1, page 268 of \cite{robert}) and design a classifier, called {\em equi-variant}, \marginpar{\tiny \sc equi-variant classifier} that achieves the minimum (Bayesian) risk (see Theorem 7.4, page 269 of \cite{robert}). {\color{pea} This can be done even in the absence of a prior, assuming a uniform (``un-informative'' and possibly improper) prior.} \cut{\tt Indeed, there is a question as to whether having a prior that is not uniform matters, or if it increases the risk.} For this reason, we will be focusing on the design of invariants to the group component of the nuisance, an refer to $G$-invariants as simply {\em invariants.}

Therefore, if all the nuisances had the structure of a group $G$, \ie when $\nu = 0$, the maximal invariant would also be a sufficient statistic, and this would be a very fortunate circumstance (Example \ref{example-invertible}). Unfortunately, in vision this does not usually happen, since the nuisance groups of interest act on the {\em scene}, rather than the image. Nevertheless, it may be possible to compute invariants if we are willing to {\em act} (Section \ref{sect-exploration}). 

\subsection{Representation}
\label{sect-representation2}
\index{Representation}

Given a scene $\xi$, we call ${\cal L}(\xi)$ the set of all possible images that can be generated by that scene up to an uninformative\footnote{For instance, a spatially and temporally independent homoscedastic noise process.} residual:
\begin{equation}
{\cal L}(\xi) \doteq \{ I \simeq h(g,\xi, \nu), ~~~ g \in G, \nu \in {\cal V} \}.
\end{equation}
We have omitted the additive noise term since, in general, it describes the compound effect of multiple factors that we do not model explicitly, and therefore it is only described in terms of its ensemble properties from which it can be easily sampled. The $\simeq$ sign above indicates that the scene image is determined up to the additive residual $n$, which is assumed to be spatially and temporally white, identically distributed with an isotropic density (homoscedastic). \index{Homoscedastic} It is, by definition, uninformative.\footnote{If it is not, as previously pointed out, the phenomena that cause violations of these assumptions can be modeled explicitly.}

Given an image $I$, in general there are infinitely many scenes $\hat \xi$ that could have generated it under {\em some} unknown nuisances, so that 
\begin{equation}
I \in {\cal L}(\hat \xi).
\label{eq-compatibility0}
\end{equation}
We call any such scene a {\em representation} compatible with that image. \marginpar{\tiny \sc representation} \index{Representation} A representation is a feature, \ie, a function of the data: It is the pre-image (under $\cal L$) of the measured image $I$: $\hat \xi \in {\cal L}^{-1}(I)$, and takes values in the space of all possible scenes $\Xi$.

A trivial example\footnote{Another example can be constructed from any partition of the image domain \label{note-partition} (a partition of the domain $D$ is a collection of sets $\{\Omega_i\}_{i=1}^K$ that are disjoint $\Omega_i \cap \Omega_j = \delta_{ij}$ and whose union equals $D = \cup_i \Omega_i$) \index{Partition} by assigning to each region $\Omega_i$ an arbitrary depth $Z_i$, and constructing a piece-wise planar surface $S$, on which to back-project the image via $\rho(p) = I(x)$ for all $p = \bar x Z_i(x), \ x\in \Omega_i$.} of a representation is the image itself glued onto a planar surface, that is $\hat \xi = (S, \rho)$ with $S = D\subset \real^2$ and $\rho = I$. Depending on how we define the nuisance $\nu$ in relation to the additive noise $n$, the ``true'' scene may {\em not} actually be a viable representation, which has subtle philosophical implications. 

Note that {\em there is no requirement that the representation $\hat \xi$ be unique, or have anything to do with the ``true'' scene} $\xi$; the relation between the two is explored in Chapter \ref{sect-exploration}.\footnote{Indeed, some philosophers would argue that for a theory to be viable it must not rely on the existence of the ``true scene.''}

While this concept of representation is pointless for a single image, it is important when considering multiple images of the same scene. In this case, the requirement is that the {\em single} representation $\hat \xi$ simultaneously ``explains'' an entire set of images $\{ I \}$ of the same scene: 
\begin{equation}
\hat \xi \in {\cal L}^{-1}(\{ I\}) \in \Xi.
\end{equation}
Of all representations, we will be interested in either the ``most probable,'' if we are lucky enough to have a prior on the set of scenes $dQ(\xi)$, which is rare, or in the {\em simplest} one, which we call the {\em minimal representation} (a minimal sufficient statistic), and indicate with ${\hat \xi}^\vee$. 

Given a scene $\xi$, we are particularly interested in the minimal representation that can generate all possible images that the original scene $\xi$ can generate, up to uninformative residuals. We call this a {\em complete representation}:\marginpar{\tiny \sc complete representation}\index{Representation!complete}\index{Complete representation}
\begin{equation}
{\hat \xi} {\rm \ is \ complete \ iff \ } {\cal L}(\hat \xi) = {\cal L}(\xi).
\label{eq-compatibility}
\end{equation}
When clear from the context, we will omit the superscript ${}^\vee$, and refer to a minimal complete representation as simply the {\em representation}. The symbol ${\cal L}$ for the set of images that are generated by a representation is chosen because, as we will see, a complete representation is related to the {\em light field} of the underlying scene. \index{Light field} \marginpar{\tiny \sc light field}
Indeed, with any representation $\hat \xi$ \marginpar{\tiny \sc hallucination} one could synthesize, or {\em hallucinate}, infinitely many images via the image-formation model (\ref{eq-gen2})
\begin{equation}
 \hat I \simeq  h(g,\hat \xi, \nu), ~~~ \forall \ g, \nu.
\end{equation}
In other words, a representation is a scene from which the given data can be hallucinated. We will elaborate the issue of hallucination in Section \ref{sect-hallucination}, where we will describe the relation to the light field. \marginpar{\tiny \sc light field} 
For the purpose of a visual decision task, a complete representation is as close as we can get to reality (the scene $\xi$) starting from the data.

\section{Actionable Information and Complete Information}
\label{sect-actinf}
\index{Actionable information}
\index{Complete information}
\index{Information!actionable}
\index{Information!complete}
The complexity of the data, measured in various ways, \eg via coding length or algorithmic complexity \cite{liV97}, has traditionally been called ``information'' in the context of data compression and transmission.\cut{ If we think of an image as a distribution of pixels, then the entropy of this distribution is sometimes used as a measure of its complexity, or ``information'' content \cite{coverT}.} Although the complexity of an image may be relevant to transmission and storage tasks (the most costly signal to transmit and store is white noise), it is in general not relevant to decision or control tasks. Extremely complex data, where all the complexity arises from nuisance factors, is useless for visual decisions, so one could say that such data is ``uninformative.'' Vice-versa, there could be very simple data that are directly relevant to the decision. So, the complexity of the data itself is not a viable measure of the information content in the data {\em for the purpose of visual decisions}. A natural image is not just a random collection of pixels; rather, an image is a {\em sample} from a distribution of (natural) images, $p(I)$, not a distribution in itself. So, if we want to measure the ``informative content'' of an image, we have to do so {\em relative to the scene}. \index{Information}

Setting aside technicalities that arise when the distributions are over continuous infinite-dimensional spaces, we define entropy formally as $H(I) \doteq E[\log p(I)]$ where the expectation is with respect to $p$ itself; that is, $H(I) = \int \log p(I)dP(I)$ (see \cite{coverT} for details). Entropy measures the ``uncertainty'' or ``information'' about the random variable $I$. Other measures of complexity, such as coding length \cite{coverT} or algorithmic complexity  \cite{kolmogorov68,liV97}, can be also related to entropy.  \index{Entropy} \index{Coding length} \index{Complexity} The mutual information ${\mathbb I}(I;\xi)$ \cite{coverT} between the image and the scene  is given by $H(\xi) - H(\xi|I)$. It is the residual uncertainty on the scene $\xi$ when the image $I$ is given. This  would be a viable measure of the information content of the image, if one were able to calculate it. As we have seen in the previous section, the ``true'' scene $\xi$ is an elusive concept. It certainly cannot be easily discretized or endowed with a base measure. In a sense, this manuscript explores ways to compute such a mutual information. \index{Mutual information} \index{Information!mutual}


In particular, using the properties of mutual information, we have that ${\mathbb I}(\xi; I) = H(I) - H(I|\xi)$, and the latter denotes the uncertainty of the image given a description of the scene. This only depends on the sensor noise and other unmodeled phenomena, for all other nuisances are encoded in $\xi$, so it is not indicative of the informative content of the image. On the other hand, if most of the uncertainty in the image $I$ is due to nuisance factors, the quantity $H(I)$ is also not indicative of the informative content of the image.  So, instead of $H(I)$, what we want to measure is  $H(\phi^\wedge(I))$, that discounts the effects of the nuisances. This is called {\em actionable information} \cite{soatto09} and formalizes the notion of information proposed by Gibson \cite{gibson84}: \marginpar{\tiny \sc actionable information}
\begin{equation}
\boxed{{\cal H}(I) \doteq H(\phi^\wedge(I)).}
\end{equation}
In Section \ref{sect-control-recognition} we will explore the relation between Actionable Information and the conditional entropy of the image given an estimated description of the scene.
Now, it is possible that the maximal invariant of the data $\phi^\wedge(I)$ contains no information at all about the object of inference. What would be most useful to perform the task would be a {\em complete representation}, $\hat \xi$. Of all statistics of the complete representation we are interested in the {\em smallest}, so we could measure the {\em complete information} as the entropy of a minimal sufficient statistic of the complete representation. 
 \marginpar{\tiny \sc complete information}
\begin{equation}
\boxed{{\cal H}_\xi = H({\hat \xi})}
\end{equation}
We will defer the issue of computing these quantities to Section \ref{sect-exploration}, although one could already conjecture that 
\begin{equation}
{\cal H}(I) \le {\cal H}_\xi.
\end{equation}
It is important to notice that, whereas the scene $\xi$ consists of complex objects (shapes, reflectance functions) that live in infinite-dimensional spaces that do not admit simple base measures, let alone distributions of which we can easily compute the entropy, the representation $\hat \xi$ is a function of the measured data, which lives in a benign finite-dimensional space.\cut{ For instance, in \cite{sundaramoorthiPVS09} it is shown that even when the images are thought of as surfaces (with infinite resolution) the representation is a tree with a finite number of nodes (for a bounded region of space), called Attributed Reeb Tree (ART).}
\begin{example}
\label{example-invertible}
It is interesting to notice that there are cases when ${\cal H}(I) = {\cal H}_\xi$. This happens, for instance, when the only existing nuisances have the structure of a group. For instance, if contrast is the only nuisance, then the geometry of the level lines (or equivalently the gradient direction) is a complete contrast invariant. Likewise, for viewpoint and contrast nuisances, the ART \cite{sundaramoorthiPVS09} is a complete feature.\footnote{In general, however, this is not the case, and the ART should be considered just a conceptual construction to illustrate the critical role of nuisance factors in the variability of imaging data.}
\end{example}
\section{Optimality and the relation between features and the classifier}
\label{sect-optimality}
\index{Optimality}
\index{Classifier!optimal}
Eliminating nuisances via marginalization yields classification that is ``optimal'' by definition. Eliminating nuisances via the design of invariant features yields optimal classification only if the nuisances have the structure of a group, \ie $\nu = 0$. That is, there are no other nuisances other than the group $G$ and the latter admits a representation on the image. In this case, one can ``pre-process'' both the data and the training set to eliminate the effects of $G$, and design an equivalent statistical model (called an ``invariant'' model), and an equi-variant classifier. \index{Classifier!equivariant} In Section \ref{sect-detectors} we will show constructive ways of designing invariant features, via the use of {\em co-variant detectors} and their corresponding {\em invariant descriptors}.
\index{Invariant descriptor}
\index{Co-variant detector}

Optimality, as we have defined it in (\ref{eq-min}), does not impose restrictions on the set of classifiers, and the data processing inequality (\ref{eq-dpi}) stipulates that any pre-processing can at best keep the Bayesian risk constant, but not decrease it. In the presence of {\em only invertible} nuisances, $\nu = 0$, it is sensible to compute the maximal invariant to eliminate group nuisances, but non-invertible nuisances can only be eliminated {\em without a loss} at decision time, via marginalization or extremization. This puts all the burden on the classifier, that at decision time has to compute a complex integral (\ref{eq-marg}), or solve a complex optimization problem (\ref{eq-ML}).

An alternative to this strategy is to constrain the choice of classifiers by limiting the processing to be performed at decision time. For instance, one could constrain the classifiers to \marginpar{\tiny \sc nearest-neighbor} nearest-neighbor rules, \index{Nearest-neighbor} \index{Classifier!nearest-neighbor} with respect to the distance between statistics computed on the test data (features) and statistics computed on the training data (templates). Two questions then arise naturally: What is the ``best'' template, if there is one? Of course, even choosing the best template, a feature-template nearest neighbor is not necessarily optimal. Therefore, the second question is: When is a template-based approach optimal? We address these questions next.

\subsection{Templates and ``blurring''}
\label{sect-best-template}
\index{Template}

This section refers to a particular instantiation of visual decision problems, where the set of allowable classifiers is constrained to be on the form 
\begin{equation}
{
\hat c = \arg\min_{c\in \{0, 1\}} d_\phi(I, \hat I_c) = \|\phi(I) - \phi(\hat I_c) \|}
\label{eq-templ1}
\end{equation}
for some statistic $\phi$ and some choice of distance or norm in the space of images. Here $\hat I_c$ is a function of the likelihood $p(I|c)$ that can be pre-computed, and is called a {\em template}. \marginpar{\tiny \sc template} If the likelihood is given in terms of samples (training set) \marginpar{\tiny \sc training set} \index{Training set} $\{I_k\}_{k=1}^K \sim p(I|c)$, then the template can be any statistic of the (training) data. In particular, a distance can be defined by designing features $\phi$ that are {\em invariant} to the group component of the nuisance $g \in G$. {\color{orange} In this case, the space $\phi({\cal I}) = {\cal I} /  G$ is the quotient \index{Quotient} of the set of images modulo the group component of the nuisance, \index{Orbit} which is in general {\rm not} a linear space even when both $\cal I$ and $G$ are linear. The distance above is a {\em cordal} distance, \index{Cordal distance}\index{Distance!cordal} that does not respect the geometric structure of the quotient space ${\cal I}/G$. A better choice would be to define a {\em geodesic distance}, \index{Geodesic distance} \index{Distance!geodesic} or a distance between equivalence classes, $d_G([\phi(I)], [\phi(\hat I_c)])$. The structure of orbit spaces of scenes under the action of finite-dimensional groups is conceptually clear and will not be further discussed here (see Appendix \ref{sect-differential-geometry}). Instead,} we focus on the two critical questions: First, {\em what is the ``best'' template $\hat I_c$, and how can it be computed from the training set?} Second, {\em are there situations or conditions under which this approach can yield the same performance of the Bayes or ML classifiers?}  (Section \ref{sect-canonization}). In Section \ref{sect-descriptors} we will show that one can build the equivalent of a template also for the test set, provided that a {\em ``sufficiently exciting''} test sample is available. \marginpar{\tiny \sc sufficiently exciting sample} \index{Sufficiently exciting sample}

\index{Best template}
\index{Template!best}
The first thing to acknowledge is that the ``best'' template depends on the class of discriminants (or distance functions) one chooses. We will therefore answer the question for the simplest case of the squared Euclidean distance in the embedding (linear) space of images $\real^{N\times M}$. More in general, the distance and its corresponding optimal template have to be {\em co-designed} \cite{leeS10,leeS11}. In all cases, one can choose as optimal template the one that induces the smallest expected distance for each class. For the case of the Euclidean distance we have
\begin{equation}
\hat I_c = \arg\min_{I_c} E_{p(I|c)}[ \| I - I_c \|^2] = \int_{\cal I} \| I - I_c \|^2 dP(I|c)
\end{equation}
that is solved by the conditional mean and approximated by the sample mean obtained from the training set
\begin{equation}
\hat I_c = \int_{\cal I} I dP(I|c) = \sum_{I_k \sim p(I|c)} I_k = \sum_{\begin{array}{c} g_k \sim dP(g) \\ \nu_k \sim dP(\nu)\\ \xi_k \sim dQ_c(\xi)\end{array}} h(g_k, \xi_k, \nu_k).
\label{eq-template}
\end{equation}
Note that the distribution of the training samples, {\color{pea} that in the integral above acts as an importance distribution,} has to be ``sufficiently exciting'' \index{Sufficient excitation} \cite{bartlett56}, in the sense that the training set must be a fair sample from $p(I|c)$. If this is not the case, for instance in the trivial instance when all the training samples are identical, then the optimal template cannot be constructed.  Different instantiations of this notation (corresponding to different choices of groups $G$, scene representation $\Xi$, and nuisances $\nu$, often not explicit but latent in the algorithms) yield {Geometric Blur \cite{berg01geometric}, where the priors $dP(g)$ are not learned but sampled in a neighborhood of the identity, 
and DAISY \cite{brown02invariant,tola2008fast}, where instead of the intensity the template is comprised of quantized gradient orientation histograms.} More in general, many models of early vision architectures include filtering steps, that mimic the quotienting operation to generate the invariant $\phi^\wedge(I)$, followed by a pooling or averaging operation, akin to computing the template above \cite{mallatB11}. Note also that the choice of norm, or more in general of classification rule, affects the form of the template. For instance, if instead of the $\ell^2$ norm we consider the $\ell^1$ norm, the resulting template is the {\em median}, rather than the mean, of the sample distribution. Other statistics are possible, including the (possibly multiple) modes, or the entire sample distribution \cite{leeS11}.

Note that the relationship between a template-based nearest-neighbor classifier and a classifier based on the proper likelihood is not straightforward, even if the class consists of a singleton -- which means that it could be captured by a single ``template'' if there were no nuisances. The marginalized likelihood is
 \begin{equation}
 \int \exp\left(-\| I-h(g, \xi, \nu)\|_\Sigma^2\right)dP(g)dP(\nu)
 \end{equation}
 assuming a normal density for the additive residual $n$, with covariance $\Sigma$, where $\| \cdot \|_\Sigma$ is a Mahalanobis norm, $\| v \|^2_\Sigma \doteq v^T\Sigma^{-1} v$; the nearest-neighbor template-based classifier would instead try to maximize 
 \begin{equation}
 \exp\left(-\| I-\underbrace{\int h(g, \xi, \nu)dP(g)dP(\nu)} \|^2\right).
 \end{equation}
The quantity bracketed is called the {\em blurred template} $\hat I_c$
\marginpar{\tiny \sc blurred template}
\index{Blurred template}
\index{Template!blurred}
\begin{equation}
{
\hat I_c \doteq \int h(g,\xi, \nu)dP(\nu)dP(g)}
\end{equation}
which does not depend on the nuisance {\em not} because it has been marginalized or max-outed, but because it has been ``blurred''  or ``smeared'' all over the template. This strategy does not rest on sound decision-theoretic principles, and yet it is one of the most commonly used in a variety of classification domains, from nearest neighbor \cite{berg01geometric} to support vector machines (SVM) \cite{chapelleWBV01} to neural networks \cite{simard01transformation}, to boosting \cite{vedaldiFS08}. 

It should be clear that {\em blurring is a lossy process}, as aggregating the training set into a single statistic decreases the discriminative power of the approach. One could attempt to minimize the loss, as done by Mallat and co-workers \cite{mallatB11}, who devise an iterative coding/blurring process that is lossless in the limit, for sufficiently concentrated nuisance distributions; the process is then truncated after two stages and not carried to the limit, so the loss remains.

Note that if instead of computing distance in the embedding space $\| I - \hat I \|_{\cal I}$ we compute the distance in the quotient, ${\cal I / G}$, we do not need to blur out the group in the template. In fact, the expectation of the quantity
\cutTwo{\begin{equation}
\exp\left( -d_G(\phi(I), \phi(\hat I)) \right)
\end{equation}
is minimized by} 
\begin{equation}
\phi(\hat I) \doteq \int \phi\circ h(\xi, \nu)dP(\nu).
\label{eq-blurred-template}
\end{equation}
Note that we have implicitly assumed that $\phi$ acts linearly on the space $\cal I$, lest we would have to consider ${\widehat{\phi(I)}}$, rather than $\phi(\hat I)$. We discuss linear features in more detail in Section \ref{sect-detector-linear}.\cut{ In either case, the analysis above shows that, whenever possible, group nuisances should be factored out in a template-based approach.}

 \begin{figure}[htb]
 \begin{center}
 \includegraphics[height=4cm]{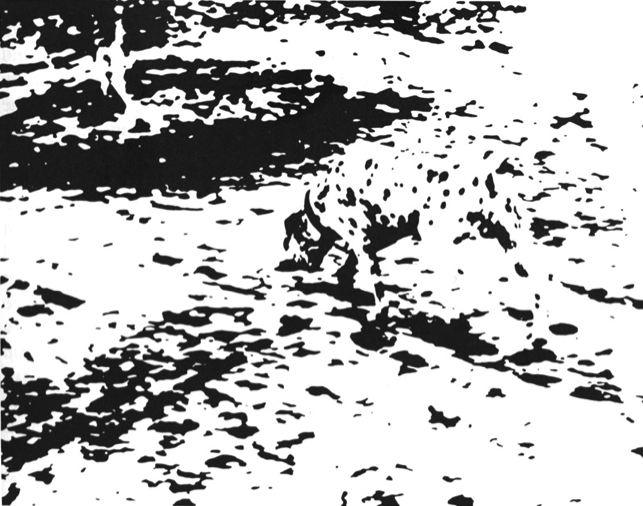}
\includegraphics[height=4cm]{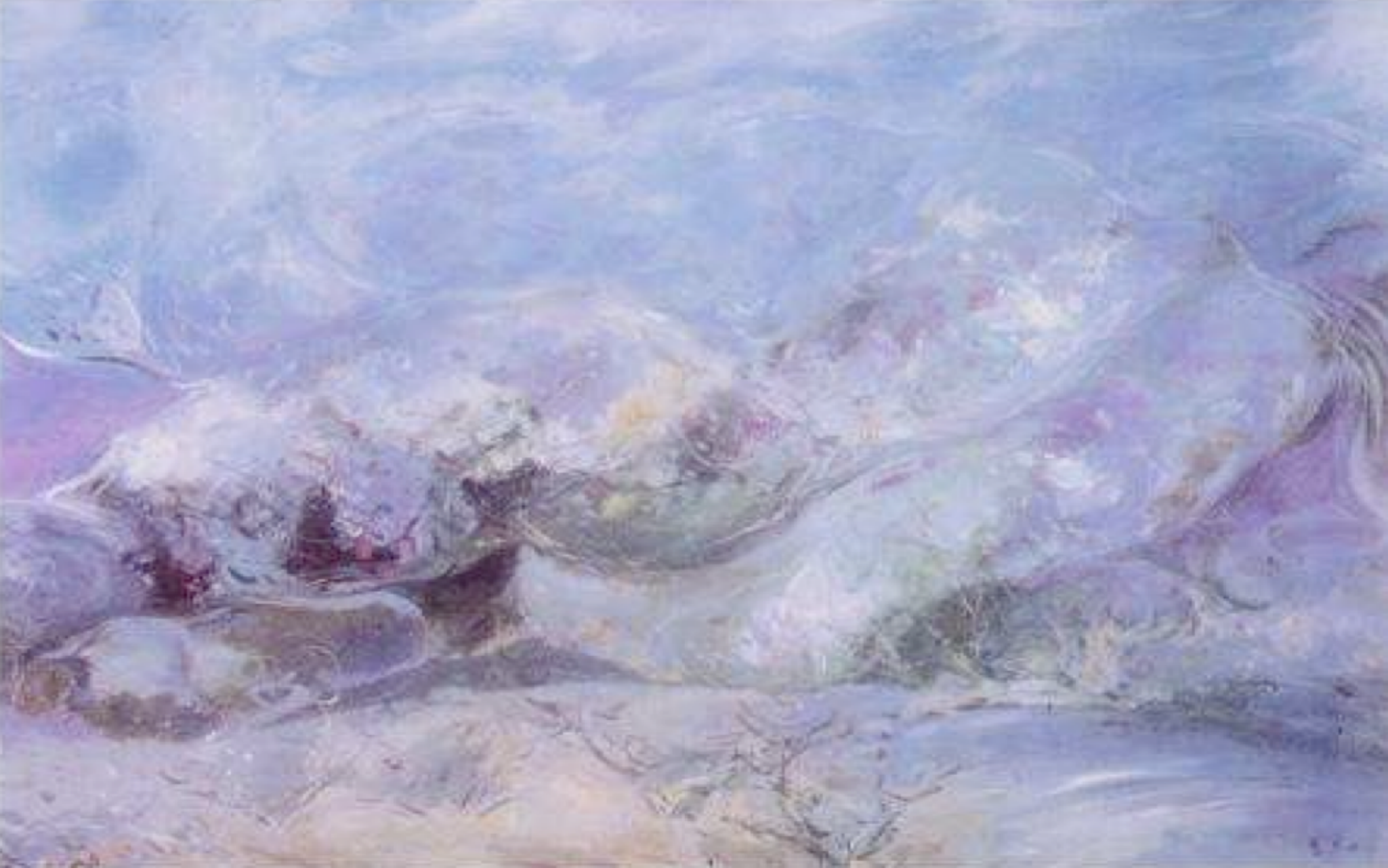}
 \end{center}
 \caption{\sl Complex nuisances (quantization, contrast, partial occlusions) can be so severe as to render a (single) template, $\hat I_c$ above, largely indiscriminative. Marginalization may enable recognition, especially if guided by priors. In some cases, even that does not work. However, extremization of the nuisances {\em given} the class can still be performed from (\ref{eq-ML}). For instance, given the class $c=$``dalmatian'' (left) and $c=$``back of a nude'' (right), one can easily determine viewpoint (pose) $\hat g$, and occlusions (location) $\hat \nu$ from (\ref{eq-ML}).}
\label{fig-dalmatian}
\end{figure}

As for how the priors can be learned from the data, we defer the answer to Section \ref{sect-learning} since it is not specific to the use of templates. For now, we note that -- should a prior be available -- it can be used according to (\ref{eq-template}) for the case of classifiers based on features/templates, (\ref{eq-marg}) for the case of Bayesian and (\ref{eq-ML}) for the case of ML classifiers.

The second question, which is under what conditions this template-based approach is optimal, is somewhat more delicate and will be addressed in Section \ref{sect-interaction}. The short answer is {\em never}, in the sense that if it were optimal, there would be no need for averaging. However, a template approach can be advocated when there are constraint on decision-time, and at least some of the nuisances can be eliminated via canonization, as described in Section \ref{sect-canonization}, and the residual uncertainty is described by a uni-modal class-conditional density that is well captured by its mode or mean (\ref{eq-blurred-template}). 

When the template is computed on the test data, it can be used as a {\em ``descriptor''},  \marginpar{\tiny \sc descriptor} that is, an invariant feature. \index{Descriptor} More general descriptors will be discussed in the next chapter, and how to construct them from data will be described in Section \ref{sect-descriptors}.

\begin{rem}[On the use of the word ``template'']
The term {\em template} refers to a large variety of approaches, only some of which are captured by the definition we have given in this chapter. In particular, {\em Deformable Templates}, studied in depth by Grenander and coworkers \cite{grenander93}, are based on premises that are not valid in our setting. In fact, in Deformable Templates there is an underlying hidden variable, the ``template'' (which could be given or learned), that is acted upon by a ``deformation'' (typically an infinite-dimensional group of diffeomorphisms). However, in Grenander's approach the group acts transitively on the template, meaning that from the template one can ``reach'' any object of interest. In other words, there is a single orbit that covers the entire space of the measurements. In this case, all the ``information'' is contained in the deformation (the group), that is therefore {\em not} a nuisance.
\end{rem}

To introduce the next chapter, we consider the simple example of classification of hand-written digits. This is simple because the data formation process lacks the two fundamental phenomena we discussed in the previous chapter: Scaling and occlusion. The only nuisance variability is due to small deformations of the domain (geometric deformations, including misalignment), and range (stroke thickness). Fig. \ref{fig-nist} shows a collection of images from the MNIST Digit dataset and the corresponding blurred templates. Then further templates are shown when the dataset is blurred with respect to additional translation, Euclidean transformations, and affine transformations. It is clear that discriminative power has been lost, since the digits can hardly be recognized from the blurred template. However, classification is increasingly insensitive to the transformations being considered, as the distance to a template changes little as the test sample is, respectively, translated, translated and rotated, or transformed with an affine map.

The likelihood of a test sample under the blurred template changes depending on whether the template has been blurred, or whether the nuisance transformation has been marginalied, or eliminated with max-out. The latter two operations entail an integral or a search at test time. When the nuisance is canonized, through a process that we describe in the next chapter, test-time complexity is the same as when the template is blurred, but the likelihood is similar to the case when the nuisance is marginazlied or max-outed.
\begin{figure}[htb]
\caption{\sl Samples from the MNIST dataset of hand-written digits that have been aligned with respect to small affine transformations \cite{vedaldiS06NIPS}. The resulting templates still exhibit some discriminative power, as the digits are discernible from the templates. If, however, nuisance transformations are included in the computation of the blurred template, the discriminative power reduces significantly, as the digits are barely discernible from the template. On the flip-side, the comparison is insensitive to misalignment of the test digit. Marginalizing or max-outing the nuisance transformations retains the discriminative power, but at an expense in run-time complexity. Canonizing the nuisance transformations achieves the best compromise of retaining discriminative power, achieving invariance to nuisance transformations, and minimizing run-time complexity.}
\label{fig-nist}
\end{figure}

\chapter{Canonized Features}
\label{sect-canonization}
\index{Canonization}
\index{Feature!canonized}

Invariant features can be designed in a number of ways. In this section we describe a constructive approach called {\em canonization},\footnote{The name ``canonization'' comes from the fact that a co-variant detector determines a canonical frame, or a canonical element of the group. In an ecclesiastic context, canonization is the elevation of an individual to one of the steps to the ladder of sanctitude. Similarly, a co-variant detector has the authority to elevate a group element to be ``special'' in the sense of determining the reference frame around which the data is described. However, this must be followed by additional steps,  that are discussed in future chapters. Canonical elements are {\em not} unique, but they are {\em isolated} and sparse.} that leverages on the notion of {\em co-variant detector} and its associated {\em invariant descriptor}.  This is also related to the notions of {\em alignment} and {\em registration}. \marginpar{\tiny \sc alignment} \marginpar{\tiny\sc registration} \index{Alignment}\index{Registration} The basic idea is that a group $G$ acting on a space $\Xi$ organizes it into orbits, \index{Orbit} $[\xi] \doteq \{ g \xi \ \forall \ g \in G\}$ each orbit being an equivalence class \marginpar{\tiny \sc equivalence class} \index{Equivalence class} (reflexive, symmetric, transitive) representable with any one element along the orbit. Of all possible choices of representatives, we are looking for one that is {\em canonical}, in the sense that it is isolated and can be determined consistently for each orbit. {\color{orange} This corresponds to cutting a section (or {\em base}) of the orbit space. All considerations (defining a base measure, distributions, discriminant functions) can be restricted to the base, which is now independent of the group $G$ and effectively represents the quotient space ${\cal I} / G$.} Alternatively, one can use the entire orbit $[\xi]$ as an invariant representation, and then define distances and discriminant functions among orbits, for instance via max-out, $d([\xi_1], [\xi_2]) = \min_{g_1, g_2\in G} d(g_1 \xi_1, g_2 \xi_2)$, as we discussed in the previous chapter.

The name of the game in canonization is to design a functional -- called {\em feature detector} -- that chooses a canonical representative for a certain nuisance $g$ that is insensitive to (ideally independent of) other nuisances. We will discuss the issue of interaction of nuisances in canonization in Section \ref{sect-interaction}. Before doing so, however, we recall some nomenclature.
\begin{defn}[Invariant Feature]
A {\em feature} $\phi: {\cal I} \rightarrow \real^K$ is any deterministic function of the data taking values in some vector space, $I \mapsto \phi(I)$. Considering the formal generative model (\ref{eq-gen2}), a feature is {\em $G$-invariant} if 
\begin{equation}
\phi \circ h(g,\xi, \nu) = \phi \circ h(e, \xi, \nu), ~~~ \forall \ g\in G
\end{equation}
and for all $\xi, \nu$ in the appropriate spaces, where $e\in G$ is the identity transformation.
\end{defn}
 In other words, an invariant feature is a function of the data that does not depend on the nuisance. Note that we are focusing on the group component of the nuisance, for reasons explained in the previous chapter and further elaborated in Section \ref{sect-interaction}. We recall the definition of representation from Section \ref{sect-representation2}:
\begin{defn}[Representation]
Given a collection of images $\{ I\}$, a feature $\hat \xi \in \Xi$ is a {\em representation} if $\{ I \} \in {\cal L}(\hat \xi)$, where ${\cal L}(\hat \xi) = \{ h(g,\hat \xi, \nu), ~~ g \in G, \nu \in {\cal V} \}$. Equivalently, $\hat \xi \in {\cal L}^{-1}(\{ I\})$. Given a scene $\xi$, a representation $\hat \xi$ is {\em complete} if it satisfies the compatibility condition ${\cal L}(\hat \xi) = {\cal L}(\xi)$; a minimal complete representation is a minimal sufficient statistic of ${\cal L}(\xi)$. 
\end{defn}
A representation is three things at once: It is a {\em statistic}, that is a function of the images. However, it is embedded in the space of {\em scenes}, so it can be thought of as a scene itself. For instance, given a single image, under the Lambert-Ambient-Static model (\ref{eq-lambert-ambient}) of Section \ref{sect-im-form}, one can construct a representation that is a plane (shape $S$) with the image $I$ glued onto it, so $\rho = I$. Finally, the representation is a finite-complexity {\em data structure} that can be stored in the memory of a digital computer.\cut{ Even if the data had infinite complexity (infinite-resolution images), the analysis of \cite{sundaramoorthiPVS09} suggests that the representation occupies an infinitesimal volume in the space of scenes.}  We call it a ``feature,'' even though it lives in the embedding space of the scene, because, as we will show in Section \ref{sect-exploration}, it can be computed from data.

A minimal complete representation, which we refer to as a ``representation'' without additional qualifications when clear from the context,  would be the ideal feature, in the sense that it captures everything about the scene that can be gathered from the data except for the effect of the nuisances. When non-invertible nuisances are absent, $\nu = 0$, a representation can be used as a representative of the orbits (equivalence classes) $[\xi]$:
\begin{equation}
\hat \xi = \phi^\wedge (h(g, \xi, 0)) \sim  h(e, \xi,0).
\end{equation}
Clearly the absence of non-invertible nuisances is a rare phenomenon, especially in vision where scaling and occlusion phenomena are dominant. Nevertheless, there are some cases of practical relevance where non-invertible nuisances are by and large absent (other than the additive ``noise'' component), such as the classification of hand-written digits of Fig. \ref{fig-nist}. The case where other nuisances are present, $\nu \neq 0$, requires some attention and will not be fully addressed until after Section \ref{sect-interaction}. In the next section, however, we pause to elaborate on the notion of representation and what it entails. Then, we study the groups $G$ for which complete features exist (Section \ref{sect-interaction}).

\section{Hallucination and representation}
\label{sect-hallucination}
\index{Hallucination}

 \begin{figure}[h!]
 \begin{center}
 \includegraphics[width=.6\textwidth]{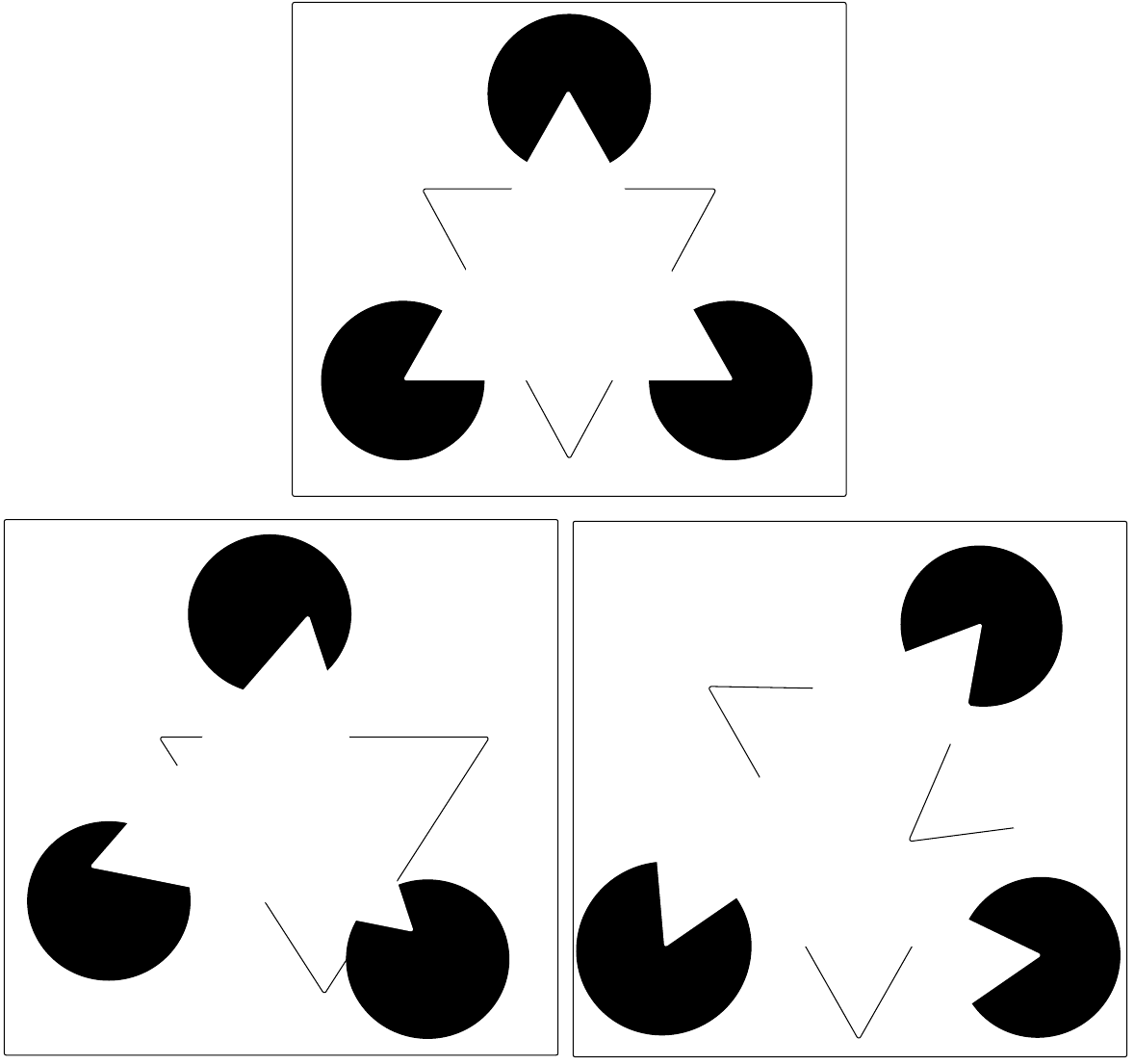}
 \end{center}
 \caption{\sl \small A representation enables the synthesis of an arbitrary number of images, and an image is compatible with infinitely many representations. {\bf Kanisza's triangle} (top) is an image that is compatible with multiple representations (\eg the two scenes on the bottom). There is no way, from an image alone or without further exploration, to ascertain which representation is ``correct.'' A unique interpretation can be {\em forced} by imposing priors or other model selection criteria (\eg minimum description), but the process is problematic from an epistemological stance. On the other hand, if one is allowed to gather more data, for instance as part of an {\em exploration} process as described in Section \ref{sect-exploration}, the set of representations that is compatible with the data shrinks. For instance, of all the representations compatible with the top image, only a subset will also be compatible with the bottom-right, and another subset with be compatible with the bottom-left. While it is not possible to say, even asymptotically with an infinite amount of data, whether the representation approaches in some sense the ``true'' scene, it is possible to validate whether the representation is compatible with the true scene in the sense of ${\cal L}(\hat \xi) = {\cal L}(\xi)$. In the case above, given the top image and one of the bottom two, it is possible to determine whether the representation is compatible with two triangles and three discs or three pac-man figures and three wedges. }
\label{fig-kanisza}
\end{figure}

If there were only invertible nuisances, there would be no need for a notion of representation, since the equivalence class $[h(g,\xi,0)]$ is a complete feature and it can be inferred from a single datum. One would only have to learn the intra-class variability, that some have argued is considerably simpler \cite{poggio11}. To begin understanding the notion of representation in the presence of non-invertible nuisances, we need to go back to the image formation model (\ref{eq-lambert-ambient}), and in particular to the pre-image of a point $x\in D$ on the domain of an image (\ref{eq-preimage}):
\index{Pre-image}
\index{Representation}
\begin{equation}
\pi^{-1}_S(x) = \{p \in S \ | \ \pi(p) = x \}.
\end{equation}
We note that this set may contain multiple elements, and in particular all points that lie on the same projection ray $\bar x$: \index{Projection ray}
\begin{equation}
\pi^{-1}_S(x) = \{\bar x Z_i(x) \in S\}_{i=1}^{N(x)}.
\end{equation}
We recall from (\ref{eq-preimage}) that we sort the depths $Z_i$ in increasing order, $Z_1(x) \le Z_2(x) \le \dots \le Z_N(x)$, and when we want to restrict the pre-image to the point of first intersection, $Z(x) = Z_1(x)$, we indicate the (unique) pre-image via $\pi^{-1}(x) = \bar x Z(x)$ where $Z(x) = Z_1(x)$ (\ie, we forgo the subscript $S$, even though the pre-image of course does depend on the shape of the scene $S$). Note that both $\pi^{-1}(x)$ and $\pi^{-1}_S(x)$ depend on the viewpoint $g$ and on the geometry and {\em topology} of the scene. They depend on how many simply connected components\footnote{A detached object is a simply connected component $S_i$ of the scene. The scene is made of multiple components, $S = \cup_i S_i$, as defined in Section \ref{sect-im-form}.}  $S_i$ there are (``detached objects''), \index{Object} \marginpar{\tiny \sc detached objects} how many holes, openings, folds, occlusions etc.
\index{Topology}
Note that we have sorted the depths while allowing the possibility of multiple points at the same depth. This happens in the limit when $x$ approaches\footnote{Note that occluding boundaries depend on the vantage point, which is usually treated as a nuisance $g$. }  an {\em occluding boundary}.

\begin{rem}[The ``ideal image'']
\label{rem-equiv-geom-top}
In Section \ref{sect-representation2} we have hinted at the fact that the scene itself may not be a viable representation of the images that it has generated, which may appear strange at first. Indeed, the ``true'' scene may never be known, because it exists at a level of granularity that is finer than any instrument we have available to measure it. So, we only have access to the ``true'' scene via the data formation process (visual or otherwise). This is unlike a representation $\hat \xi$, that can be arbitrarily manipulated in order to generate any image $\hat I \in {\cal L}(\hat \xi)$. 

In the presence of occlusions, evaluating the (hypothetical) ``ideal image'' $h(e, \xi, 0)$, that is the image that would be obtained if there were no nuisances,  \index{Ideal image} requires the ``inversion'' of the occlusion process, including a description of the scene both in the visible and in the non-visible portions. In formulas, from (\ref{eq-lambert-ambient}) and (\ref{eq-preimage}), we have \index{Ideal image}
\marginpar{\tiny\sc ideal image}
\begin{equation}
\boxed{h(e, \xi, 0) = \rho\left(\pi^{-1}_S(D)\right).}
\label{eq-repr2}
\end{equation}
Note that what this object returns is {\em not} the shape $S$ of the scene, but the {\em radiance} of the scene in both the visible and the occluded portions of the scene. So, in a sense, it is an image of a ``semi-transparent world'' made of many simply connected surfaces. Note, however, that the 3-D {\em geometry} may be recovered if we can control the data acquisition process via $h(g,\xi, 0)$, with $g = g(u)$ as described in Chapter \ref{sect-exploration}, because by doing so we can generate the collection of all possible occluding boundaries, which is equivalent to an approximation of the 3-D geometry of the scene \cite{yezziS03IJCV}. In any case, the hypothetical image $h(e, \xi, 0)$ captures the {\em topology} of the world, reflected in the number of layers $N$ present in the pre-image. While the notion of ideal image is awkward, and we will soon move beyond it, we need to consider it for a little while longer to ascertain what would be possible (or may be possible, depending on the sensing modality) if there were only invertible nuisances.
\end{rem}
\cut{The issue of representation has been investigated by many, and in particular by Koenderink and van Doorn \cite{koenderinkVD08}; we will return to their work later. Now, if we were given given $h(\xi, 0)$, we could generate infinitely many images from it. We cannot just introduce nuisances (for instance, change the viewpoint $g$ or the occlusion structure $\nu$) to get an image $I = h(g,\xi, \nu)$, since we do not have direct access to $\xi$. However, we have seen in Section \ref{sect-representation2} that given any one image $I$, generated by a certain scene $\xi$ under nuisances $g, \nu, n$, we can construct at least one scene (in fact, infinitely many) hypothetical scenes $\hat \xi$, not necessarily equal to the ``true'' scene $\xi$, but such that 
\begin{equation}
I = h(g,\xi, \nu) + n  = h(\hat \xi, 0).
\label{eq-compatibility}
\end{equation}
For instance, if we think of the representation as consisting of multiple layered radiance functions according to (\ref{eq-repr2}), and think of each as an image, $I_j(x) = \rho(\bar x Z_j(x))$, we can collapse all layers onto one and generate an image $I(x) = \sum_j I_j(x) = h(\hat \xi, 0)$ as if the scene was semi-transparent, by defining $\hat \xi = \{\hat S, \hat \rho\}$ with $\hat S = {\mathbb S}^2$, the unit sphere, and $\hat \rho(p) = I(x)$ with $p = \frac{\bar x}{\| \bar x\|}$. This scene would not be very realistic. A more realistic ``hallucination'' can be constructed by 
partitioning\footnote{A partition of the domain $D$ is defined in Footnote \ref{note-partition}.  It is a collection of disjoint sets $\{\Omega_j\}$ whose union is equal to $D$, that is $\Omega_i \cap \Omega_j = \delta_{ij}$ and $\cup_j \Omega_j = D$.} the domain of the image into regions $\Omega_j$ and then assigning to the pre-image of each region, $\pi^{-1}(\Omega_j)$, the radiance of a layer $I_j(x) = \rho(\bar x Z_j(x))$, supported on a piecewise planar surface at an arbitrary depth $Z_j$. 
\index{Virtual images}
\index{Image!virtual}}

As we have anticipated in Section \ref{sect-representation2}, building a representation from a single image would not be very useful, because in general the space of all scenes $\Xi$ is much larger than the space of all images $\cal I$, and lifting the image onto the scene yields no practical benefit. However, one may be able to construct a representation $\hat \xi$ that is compatible with a collection of images, $\{I_k\}_{k=1}^K$, in the sense that
\begin{equation}
I_k = h(g_k, \xi, \nu_k) + n_k \simeq  h(\hat g_k, \hat \xi, \hat \nu_k)
\end{equation}
for all $k = 1, \dots, K$ and for some $\hat g_k, \hat \nu_k$, but the {\em same} $\hat \xi$. 
In fact, once we have an hallucinated scene $\hat \xi$, we can produce infinitely many ``virtual'' images, that we can then compare with any real image that is actually measured. In fact, ${\cal L}(\hat \xi)$ produces the set of all possible hallucinated images of the scene $\hat \xi$, and is therefore equivalent to its {\em light field}\footnote{A representation enables generating images with any nuisance combinations. In this sense it is ``equivalent'' to the light field, because it can generate any image that a sampling of the light field would produce. How to {\em store} this representation, and how it is compatible with the computational architecture of the primate's brain, is well outside the scope of this manuscript. However, others have tackled this question, most notably Koenderink and van Doorn, who have studied in depth the structure of the representation (even though it is not called a representation) and its relation to the light field, and also coined the phrase ``images as controlled hallucinations.'' We refer the reader to their work (\cite{koenderinkVD08} and references therein) for further investigations on this topic. In our case, we will give a constructive example on how to build and store a representation through the exploration process in Section \ref{sect-exploration}.}
 \cite{koenderinkVD08}. While the hypothesis $\hat \xi = \xi$ cannot be tested, and a meaningful distance $d(\xi, \hat \xi)$ cannot be computed (we do not even have a metric in the space $\Xi$), the hypothesis $h(g,\xi, \nu) \simeq h(\hat g, \hat \xi, \hat \nu)$ for some $\hat g, \hat \nu$ can be very simply tested by comparing two images, which is a task that poses no philosophical or mathematical difficulties. In compact notation, what we can test is $\hat \xi \in {\cal L}^{-1}\left( {\cal L}(\xi) \right)$. We will return to this issue in Sections \ref{sect-correspondence} and \ref{sect-exploration}.

\index{Hallucination}
\index{Ambiguity}
So far we have described the {\em hallucination} process, by which a representation can be used to generate images. The inverse process of hallucination is {\em exploration}, \marginpar{\tiny \sc exploration} \index{Exploration} by which one can use images to construct a representation. In Section \ref{sect-exploration} we will show how to perform this limit operation. Exactly how the representation $\hat \xi$, built through exploration, is related to the actual ``real'' scene $\xi$, for instance whether they are ``close'' in some sense, hinges on the characterization of visual ambiguities, which we discuss in Appendix \ref{sect-ambiguity}. 
\index{Exploration}

We will come back to the notion of representation after establishing what could be done if only invertible nuisances were present, and establishing exactly what nuisances {\em are} indeed invertible.

\section{Optimal classification with canonization }
\index{Canonization}
\index{Classifier!optimal}

In this section and the next we elaborate on what would be possible if all nuisances were invertible.\footnote{Note that this situation is hypothetical, for even a group $g$ (\eg vantage point) acting on the scene ($\xi$) can generate non-invertible nuisances (self-occlusions $\nu$). According to the Ambient-Lambert model, a group transformation $g_t \in SE(3)$ due to a change of vantage point induces on the image domain an epipolar transformation that, depending on the shape of the scene, can be arbitrarily complex. In particular, such a transformation need not be a {\em function} in the sense that self-occlusion will cause the pre-image of a point $p$ via the domain deformation $w^{-1}(x)$ to have multiple values. This issue will be resolved in later sections; for now, we assume that the scene is such that a group nuisance does not induce non-invertible transformations of the image domain. In particular, it has been shown in \cite{sundaramoorthiPVS09} that -- in the absence of occlusion phenomena -- the closure of epipolar domain deformations has as closure the entire group of planar diffeomorphisms.} So, we assume that $\nu = 0$, and focus on the role of $g$. We will address $\nu \neq 0$ in Section \ref{sect-interaction}. For now, we are interested in how to design invariant and sufficient statistics for $g$ alone, as if all other nuisances were absent. In this case, we can assume that the group $g$ acts on the space of images, or that the space of scenes and the space of images coincide.

One of the many possible ways of designing an invariant feature is to use the data $I$ to ``fix'' a particular group element $\hat g(I)$, and then ``undo'' it from the data. \cutTwo{If the data does not allow fixing a group element $\hat g$, it means it is already invariant to $G$.} So, we define a {\em (co-variant) feature detector} to be  a functional designed to choose a particular group action $\hat g$, from which we can easily design an invariant feature, often referred to as an {\em invariant (feature) descriptor}. Note that both the detector and the descriptor are deterministic functions of the data, hence both are features.
\index{Co-variant detector}
\index{Detector!co-variant}
One can think of this process as some kind of registration, or alignment, but rather than registering to a target, it is a form of ``self-alignment.'' 

{\color{orange} We consider the set of digital images ${\cal I}$ to be functions $I:\real^2 \rightarrow \real^+; \  x \in {\cal B}_\epsilon(x_{ij}) \mapsto I_{ij}$. Because of spatial quantization, they are piecewise constant and thus can be identified with the set of matrices $\real^{N\times M}$. A differentiable functional $\psi:{\cal I} \times G \rightarrow \real; (I, g) \mapsto \psi(I, g)$ is said to be {\em local}, with  {\em effective support} $\sigma$ if its value at $g$ only depends on a neighborhood of the image of size $\sigma > 0$, up to a residual that is smaller than the mean quantization error. For instance, for a translational frame\footnote{Since a group element can be identified with a reference frame, we often refer to a group element as a ``frame.''} $g$, if we call $I_{|_{{\cal B}_\sigma(g)}}$ an image that is identical to $I$ in a neighborhood of size $\sigma$ centered at position $g \equiv T$, and zero otherwise, then $\psi(I_{|_{{\cal B}_\sigma(g)}}, g) = \psi(I, g) + \tilde n$, with $|\tilde n| \le \frac{1}{NM} \sum_{i,j} | n_{ij}|$ the spatial quantization error. For instance, a functional that evaluates the image at a pixel $g \equiv T = x \in {\cal B}_\epsilon(x_{ij})$, is local with effective support $\epsilon$. For groups other than translation, we consider the image in the reference frame determined by $g$, or equivalently consider the ``transformed image'' $I\circ g^{-1}$, in a neighborhood of the origin, so $\psi(I, g) = \psi(I\circ g^{-1}, e)$.

If we call $\nabla \psi \doteq \frac{\partial \psi }{\partial g}$ the gradient of the functional $\psi$ with respect to (any) parametrization of the group,\footnote{The following discussion is restricted to finite-dimensional groups, but it could be extended with some effort to infinite-dimensional ones.} then under certain (so-called ``transversality'') conditions on $\psi$, the equation $\nabla \psi = 0$ locally determines $g$ as a function of $I$, $g = \hat g(I)$, via the Implicit Function Theorem. Such conditions are independent of the parametrization and consist of the Hessian matrix $H(\psi) \doteq \nabla \nabla \psi$ (\ie the Jacobian of $\nabla \psi$) being non-singular,  $\det(H(\psi)) \neq 0$. The function $\hat g$ is unique in a neighborhood where the transversality condition is satisfied, and is called a (local) {\em canonical representative} of the group. If the canonical representative co-varies with the group, in the sense that $\hat g(I \circ g) = (\hat g \circ g)(I)$, then the functional $\psi$ is called a {\em co-variant detector}. Each co-variant detector determines a local reference {\em frame}  so that, if the image is transformed by the action of the group, a hypothetical observer attached to the co-variant frame (i.e., a ``Lagrangian'' observer) would see no changes. We summarize this in the following definition:
 \begin{defn}[Co-variant detector]
 \label{defn-covariant}
 A differentiable functional $\psi: {\cal I} \times G \rightarrow \real; \ (I,g) \mapsto \psi(I,g)$  
is a co-variant detector if 
 \begin{enumerate}
 \item The equation $\det\left(H(\psi(I, g))\right) = 0$ locally determines a unique isolated extremum in the frame $g\in G$, and 
 \item if $\nabla \psi(I,\hat g) = 0$, then $\nabla \psi(I\circ g, \hat g \circ g) = 0 \ \forall \ g\in G$, \ie, $\psi$ co-varies with $G$.
 \end{enumerate}
 \end{defn}
}
\noindent The notation $I\circ g$ indicates the map $(I, g) = (h(e,\xi, 0), g) \mapsto h(g, \xi, 0) \doteq I \circ g$. This may seem a little confusing at this point, since the group is acting on the scene, but is being canonized by acting on the image. This will be clarified in Sect. \ref{sect-clarify}. 
{\color{pea} The first ``transversality'' condition \cite{guilleminP74} corresponds to the Jacobian of $\nabla \psi$ with respect to $g$ being non-singular:
\index{Transversality}
\begin{equation}
| J_{\nabla \psi} | \neq 0.
\label{eq-jacobian} 
\end{equation}}
\index{Jacobian}
\index{Canonizability}
In words, a co-variant detector is a function that determines an isolated group element in such a way that, if we transform the image, the group elements is transformed in the same manner. Examples of co-variant detectors will follow shortly. 
\begin{defn}[Canonizability]
We say that the image $I$ is $G$-canonizable (is canonizable with respect to the group $G$), and $\hat g \in G$ is the canonical element, if there exists a covariant detector $\psi$ such that $\nabla \psi(I,\hat g) = 0$. 

\index{Locality}
\marginpar{\tiny \sc locality}

Note that, depending on the functional $\psi$, the canonical element may be defined only {\em locally}, \ie the functional may only depend on $I(x)_{|_{x\in {\cal B}\subset D}}$, that is a restriction of the image to a subset $\cal B$ of its domain $D$. In the latter case we say that $I$ is locally canonizable, or, with an abuse of nomenclature, we say that the {\em region} $\cal B$ is canonizable.
\index{Local feature}
\index{Feature!local}
\index{Canonizable region}
\marginpar{\tiny \sc canonizable region}
\end{defn}
{\color{orange} The transversality condition (\ref{eq-jacobian}) guarantees that $\hat g$, the canonical element, is an isolated (Morse) \index{Morse critical point} critical point \cite{Milnor} \index{Critical point} of the derivative of the function $\psi$ via the \index{Implicit function theorem} Implicit Function Theorem \cite{guilleminP74}.} So a co-variant detector is a statistic (a feature) that ``extracts'' a group element $\hat g$.
With a co-variant detector we can easily construct an invariant descriptor, or {\em local invariant feature}, by considering the data itself in the reference frame determined by the detector: \marginpar{\tiny \sc local invariant feature} \index{Local invariant feature}\index{Feature!local}
\index{Canonized descriptor}
\index{Descriptor!canonized}
\begin{defn}[Canonized descriptor]
For a given co-variant detector $\psi$ that fixes a canonical element $\hat g$ via $\nabla \psi(I, \hat g(I)) = 0$ we call the statistic
\begin{equation}
\boxed{\phi(I) \doteq I \circ {\hat g}^{-1}(I) ~~ | ~~  \nabla \psi(I,\hat g(I))= 0.}
\label{eq-can-descr}
\end{equation}
an invariant descriptor. 
\end{defn}
Note that the descriptor is invariant with respect to the same group $G$ that is canonized by the detector. Obviously it is not invariant to other nuisances, including groups other than the one chosen to design the detector.

A trivial example of canonical detector and its corresponding descriptor can be designed for the translation group. Consider for instance the detector $\psi$ that finds the brightest pixel in the image. Relative to this point, that is, if we assign that point to the coordinate $(0, 0)$, the image is translation-invariant. This is because, as the image translates, so does its brightest pixel, and in the moving frame the image does not change as we translate. Similarly, we can assign the value of the brightest pixel to $1$, the value of the darkest pixel to $0$, linearly interpolate pixel values in between, and we have an affine contrast-covariant detector and the associated contrast-invariant descriptor.

As far as eliminating the effects of a group, all covariant detectors are equivalent.\footnote{Since canonization is possible only for groups acting on the domain of the {\em data} (\ie, images), in the presence of more complex nuisances one has to exercise caution in that the group $g$ acting on the scene may {\em induce} a different transformation on the domain of the image, which may not even be a group. For instance, a spatial translation along a direction parallel to the image plane does not induce a translation of the image plane, unless the scene is planar and fronto-parallel. We will come back to this issue later.}
Where they differ is in how they behave relative to all other nuisances. Later we will give more examples of detectors that are designed to ``behave well'' with respect to other nuisances. In the meantime, however, we state more precisely the fact that, as far as dealing with a group nuisance, all co-variant detectors do the job.
\begin{theorem}[Canonized descriptors are complete features]
\label{thm-canonized}
Let $\psi$ be a co-variant detector. Then the corresponding
canonized descriptor (\ref{eq-can-descr}) is an invariant sufficient statistic. 
\end{theorem}
\noindent {\bf Proof:}
{\em 
To show that the descriptor is invariant we must show that $\phi(I\circ g) = \phi(I)$. But $\phi(I\circ g) = (I \circ g) \circ {\hat g}^{-1}(I \circ g) = I \circ g \circ (\hat g g)^{-1} = I \circ g \circ g^{-1} {\hat g}^{-1} (I)= I\circ {\hat g}^{-1}(I).$ To show that it is complete it suffices to show that it spans the orbit space ${\cal I}/G$, which is evident from the definition $\phi(I) = I \circ g^{-1}$.} \\
{\color{orange} The notation $I\circ g$ is slightly ambiguous at this point, but will be clarified in Sect. \ref{sect-clarify}.
}

~\\ Note that invertible nuisances $g$ may act on the domain of the image (\eg affine transformations due to viewpoint changes), as well as on its range (\eg contrast transformations due to illumination changes).
\begin{example}[SIFT detector and its variants]
To construct a simple translation-covariant detector, consider an isotropic bi-variate Gaussian function \\ ${\cal N}(x; \mu, \sigma^2) = \frac{1}{{2\pi}\sigma}\exp(-\frac{\|x - \mu \|^2}{2\sigma^2} )$; then for any given scale $\sigma$, the  Laplacian-of-Gaussian (LoG)  $\psi(I,g) \doteq  \nabla^2 {\cal N}(x; g, \sigma^2) * I(x)$ is a linear translation-covariant detector. If the group includes both location and scale, so $\bar g = (g, \sigma^2)$, then the same functional can be used as a translation-scale detector. Other examples are the difference-of-Gaussians (DoG)  $\psi(I, g) \doteq \frac{{\cal N}(x; g, \sigma^2) - {\cal N}(x; g, k^2 \sigma^2)}{k-1} * I(x)$, with typically $k = 1.6$, and the Hessian-of-Gaussian (HoG) is $\psi(I, g) = \det H({\cal N}(x; g, \sigma^2))$. Among the most popular detectors, SIFT uses the DoG, as an approximation of the Laplacian. 
\end{example}
\begin{example}[Harris' corner and its variants]
Harris' corner and its variants (Stephens, Lucas-Kanade, etc.) replace the Hessian with the second-moment matrix: 
\begin{equation}
\psi(I, g) \doteq \det\left(\int_{{\cal B}_\sigma(g)} \nabla^TI \nabla I(x)dx\right).
\label{eq-harris}
\end{equation}
 One can obtain generalizations to groups other than translation in a straightforward manner by replacing ${\cal N}(x; g, \sigma^2)$ with $\frac{1}{{2\pi}\sigma \det J}\exp(-\frac{\|g^{-1}(x) \|^2}{2\sigma^2\det(J_g)^2} )$ where $J_g$ is the Jacobian of the group. For instance, for the affine group $g(x) = Ax + b$, we have that $\psi(I, g) = \nabla^2\left( \frac{1}{2\pi\sigma \det A}\exp(-\frac{\|A^{-1}(x-b) \|^2}{2\sigma^2\det(A)^2} )\right)$ is an affine-covariant (Laplacian) detector. One can similarly obtain a Hessian detector or a DoG detector. The Euclidean group has $A \in SO(2)$, so that $\det{A} = 1$, and the similarity group has $\tilde \sigma A$, with determinant $\tilde \sigma$. 

As we will see in Example \ref{ex-harris2}, this functional has some limitations in that it is not a linear functional, and therefore it does not commute with additive nuisances such as quantization or noise.
\end{example}

\begin{example}[Harris-Affine]
The only difference from the standard Harris' corner is that the region where the second-moment matrix is aggregated is not a spherical neighborhood of location $g$ with radius $\sigma$, but instead an ellipsoid represented by a location $T \in D \subset \real^2$ and an invertible $2\times 2$ matrix $A \in {GL}(2)$. In this case, $g = (T, A)$ is the affine group, and the second-moment matrix is computed by considering the gradient with respect to all $6$ parameters $T, A$, so the second-moment matrix is $6\times 6$. However, the general form of the functional is identical to (\ref{eq-harris}), and shares the same limitations, as we will see, in that affine-covariant canonization necessarily entails a loss.
\end{example}

Although these detectors are not local, their effective support can be considered to be a spherical neighborhood of radius a multiple of the standard deviation $\sigma>0$, so they are commonly treated as local. Varying the scale parameter $\sigma$ produces a {\em scale-space}, whereby the locus of extrema of $\psi$ describes a {\em graph} in $\real^3$, via $(x,\sigma) \mapsto \hat x = \hat g(I; \sigma)$. Starting from the finest scale (smallest $\sigma$), one will have a large number of extrema; as $\sigma$ increases, extrema will merge or disappear. Although in two-dimensional scale space extrema can also appear as well as split, such {\em genetic} effects (births and bifurcations) have been shown to be increasingly rare as scale increases, so the locus of extrema as a function of scale is well approximated by a tree, which we call the {\em co-variant detection tree} \cite{leeS10}.

\index{Sketch}
\index{Sketchability}
\begin{rem}[Canonizability, saliency and the ``sketch'']
The notion of canonizability is related to the notion of ``sketchability'' defined in \cite{guoZW03}, although the latter is introduced without ties to a specific task, \cut{which raises the issue of falsifiability,} and  motivated in \cite{marr}. In the case of this manuscript, the notion of canonizability arises naturally from visual classification tasks, in the sense of providing a vehicle to design an invariant descriptor. It also relates to the notion of ``saliency'' defined in \cite{tsotsos95,ittiK01}, although again the latter stems from resource limitations rather than from a direct tie to a visual classification task.
\end{rem}

\begin{rem}[Invariant descriptors without co-variant detectors]
\label{rem-invariance-no-covariance}
~\\
The assumption of differentiability in a co-variant detector can be easily lifted; in Section \ref{sect-superpixels}  we will show how to construct co-variant detectors that are not differentiable. Indeed, canonization itself is not necessary to design invariant descriptors. We have already mentioned ``blurring'' as a way to reduce (if not eliminate) the dependency of a statistic on a group, although that does not yield a sufficient statistic. However, even for designing complete (\ie invariant and sufficient) features, canonization is not necessary. \index{Feature!complete} For instance, the geometry of the level curves -- or its dual, the gradient direction -- is a complete contrast-invariant which does not require a contrast-detector.   Indeed, even the first condition in the definition of a co-variant detector is not necessary in order to define an invariant descriptor: {\color{orange}  Assume that the image $I$ is such that for any functional $\psi$, the equation $\nabla \psi(I,g) = 0$ does {\em not} uniquely determine $\hat g = \hat g(I)$. That means that $| J_{\nabla \psi} | = 0$ for all $\psi$, and therefore all statistics are already (locally) invariant to $G$. More in general, where the structure of the image allows a ``stable'' and ``repeatable'' detection\footnote{``Stability'' will be captured by the notion of Structural Stability, and ``repeatability'' by the notion of Proper Sampling.} of a frame $\hat g$, this can be inverted and canonized $\phi(I) = I \circ {\hat g}^{-1}$. Where the image does not enable the detection of a frame $\hat g$, it means that the image itself is already invariant to $G$. These statements will be elaborated in Section \ref{sect-texture} where we introduce the notion of texture.}

Intuitively, if a covariant detector is {\em ``unstable''}, \ie,  $| J_{\nabla \psi} | \simeq 0$, then {\em any} function $\phi(I)$ is ``insensitive'' to $g$, in the sense that, assuming $\psi\circ I$ to be smooth, we have $\phi(I) \simeq \phi(I\circ g)$. This means that {\em what we cannot canonize does not matter} towards the goal of designing invariant (insensitive) statistics; it is already invariant (insensitive). Of course, these statistics will not be {\em localized}. In particular, the definition of canonizability, and its requirement that $\hat g$ be an {\em isolated} critical point, would appear to exclude edges and ridges, {\color{orange} and in general co-dimension one critical loci that are not Morse critical points.} However, this is not the case, because the definition of critical point depends on the group $G$, which can include discrete groups (thus capturing the notion of ``periodic structures,'' or ``regular texture'') and sub-groups of the ordinary translation group, for instance planar translation along a given direction, capturing the notion of ``edge'' or ``ridge'' in the orthogonal direction. We will come back to the notion of stability in Section \ref{sect-stability}.
\end{rem}
\begin{rem}[Dense descriptors]
We emphasize that detectors' only purpose is to avoid marginalizing the invertible component of the group $G$. However, at best such detectors can yield no improvement over marginalizing the action of $G$, that is to {\em use no detector at all} (Section \ref{sect-dpi}). Therefore, one should always marginalize or max-out the nuisances if this process is viable given resource constraints. This is a design choice that has been explored empirically. In visual category recognition, some researchers prefer to use features selected around ``keypoints,'' whereas others prefer to compute ``dense descriptors'' at each pixel, or at a regular sub-sampling of the pixel grid, and let the classifier sort out which are informative, at decision time.
\end{rem}
\index{Canonizability}
\index{Proper Sampling}
\index{Aliasing}
\marginpar{\tiny\sc aliasing}
\begin{rem}[Preview: Aliasing and Proper Sampling]
\label{rem-proper-sampling}
Canonizability, as we have defined it, entails the computation of the Jacobian, which is a differential operation. However, images are {\em discrete}, merely a sampled version of the underlying signal (assuming {\em that} is piecewise differentiable), that is the radiance of the scene. In any case, the differentiable approximation, or the computation of the Jacobian, entails a choice of {\em scale}, depending on which any given ``structure'' (isolated extremum) may or may not exist: A differential operator such as the Jacobian could be invertible at a certain scale, and not invertible at a different scale {\em at the same location}. Because the ``true'' scale is unknown (and it could be argued that it does not exist), canonizability alone is not sufficient to determine whether a region can be {\em meaningfully} canonized. ``Meaningful'' in this context indicates that a structure detected in an image corresponds to some structure in the {\em scene}, and is not instead a {\em sampling} artifact ({\em ``aliasing''}) due to the image formation process, for instance quantization and noise. Therefore, an additional condition must be satisfied for a region to be ``meaningfully'' canonized. This is the condition of {\em Proper Sampling} that we will introduce in Section \ref{sect-correspondence}.
\end{rem}

\section{Optimality of feature-based classification}
\index{Insensitivity}
\index{Equivariant classifier}
\index{Classifier!equivariant}

The use of canonization to design invariant descriptors requires the image to support ``reliable'' (in the sense of Definition \ref{defn-covariant}) co-variant detection. As we have discussed in Remark \ref{rem-invariance-no-covariance}, the challenge in canonization is {\em not} when the co-variant detector is unreliable, for that implies the image is already ``insensitive'' to the action of $G$. Instead, the challenge is when the covariant detector reliably detects the {\em wrong} canonical element $\hat g$, for instance where there are multiple repeated structures that are locally indistinguishable, as is often the case in cluttered scenes. We will come back to this issue in Section \ref{sect-matching}. 

The good news is that, when canonization works, it simplifies visual classification by eliminating the group nuisance without any loss of performance. This was already illustrated empirically in Fig. \ref{fig-nist}, and is formalized by the following theorem.
\begin{theorem}[Invariant classification]
\label{thm-equivariant}
If a complete $G$-invariant descriptor $\hat \xi = \phi(I) $ can be constructed from the data $I$, it is possible to construct a classifier based on the class-conditional distribution $dP(\hat \xi | c)$ that attains the same minimum (Bayesian) risk as the original likelihood $p(I|c)$.
\end{theorem}
The proof follows from the definitions and Theorem 7.4 on page 269 of \cite{robert}. The classifier based on the complete invariant descriptor is called {\em equi-variant}.

\cutTwo{Based on this result, one would surmise that it is always desirable to canonize group nuisances, whether or not they come with a prior $dP(g)$.\cut{ In other words, {even if a prior is available, it does not help improving classification performance,} since the equi-variant estimator already attains the optimal risk.}\footnote{This may seem confusing if one consider classification of hand-written digit modulo planar rotation. However, in the case of digits such as ``6'' and ``9'', the class identity depends on the group, which is therefore {\em not} a nuisance and should instead be part of the description $\xi$.}} 
\index{Commutative nuisance}
\index{Nuisance!commutative}
An important caveat is that, so far in this section, we have assumed that the non-invertible nuisance is absent, \ie $\nu =0$, or that, more generally, the canonization procedure for $g$ is independent of $\nu$, or ``commutes'' with $\nu$, in a sense that we will make precise in Definition. \ref{def-commute}. This is true for some nuisances, but not for others, even if they have the structure of a group, as we will see in the next section. 
There we will show that many group nuisances indeed can{\em not} be canonized. These include the affine and projective group (scale, skew, deformation) as well as more complex nuisances that have to be dealt with either by marginalization (\ref{eq-marg}), or by extremization (\ref{eq-ML}).

\section{Interaction of  nuisances in canonization: what is {\em really} canonizable?}
\label{sect-interaction}

The previous section described canonization of the group nuisance $g\in G$ in the absence of other nuisances $\nu = 0$. Unfortunately, some nuisances are clearly not invertible (occlusions, quantization, additive noise), and therefore they cannot be canonized. What is worse, even group nuisances may lose their invertibility once composed with non-invertible nuisances. 

In this section, we deal with the interaction between invertible and non-invertible nuisances, so we relax the condition $\nu = 0$ and describe feature detectors that ``commute'' with $\nu$. We show that the only subgroup of $G$ that has this property is the isometric group of the plane. \marginpar{\tiny \sc isometric group} \index{Isometric group} That is, planar rotations, translations and reflections. Other nuisances, groups or not, have to be dealt with by marginalization or extremization if one wishes to retain optimal performance. This includes the similarity group of rotations translations and scale, that is instead canonized in \cite{lowe99object}, \marginpar{\tiny \sc similarity group} \index{Similarity group} and the affine group, \marginpar{\tiny \sc affine group} \index{Affine group} that is instead canonized in \cite{mikolajczykSb}.
\index{Commutative nuisance}
\index{Nuisance!commutative}

In order to simplify the derivation, we introduce the following notation, in part already adopted earlier in the proof of Theorem \ref{thm-canonized}. The notation will be fully clarified in Sect. \ref{sect-clarify}. If $\hat I$ is the ``ideal image'' \marginpar{\sc\tiny ideal image}\index{Ideal image} (without nuisances, see Remark \ref{rem-equiv-geom-top}), $\hat I = h(e, \xi, 0)$, then
\begin{eqnarray}
\hat I \circ g &\doteq& h(g, \xi, 0) \\
\hat I \circ \nu &\doteq& h(e, \xi, \nu).
\end{eqnarray}
The operators $(\cdot \circ g)$ and $(\cdot \circ \nu)$ can also be composed, $\hat I \circ g \circ \nu = h(g,\xi, \nu)$ and applied to an arbitrary image; for instance, if $I = h(g, \xi, \nu)$, then for any other $\tilde g, \tilde \nu$ we have
\begin{equation}
I \circ \tilde g \circ \tilde \nu = h(\tilde g g,  \xi, \tilde \nu \oplus \nu)
\end{equation}
where $\oplus$ is a suitable composition operator that depends on the space where the nuisance $\nu$ is defined. Note that, in general, the action of the group and the other nuisances do not commute: $I \circ g \circ \nu \neq I \circ \nu \circ g$. When this happens we say that the group commutes with the (non-group) nuisance:
\marginpar{\tiny \sc commutative nuisance}
\begin{defn}[Commutative nuisance]
\label{def-commute}
A group nuisance $g \in G$ {\em commutes} with a (non-group) nuisance $\nu$ if
\begin{equation}
I\circ g \circ \nu = I \circ \nu \circ g.
\end{equation}
\end{defn}
Note that commutativity does not coincide with invertibility: A nuisance can be invertible, and yet not commutative (\eg the scaling group does not commute with quantization).

For a nuisance to be canonizable {\em without a loss}, (\ie, eliminated via pre-processing, or via a complete invariant feature) it not only has to be a group, but it also has to commute with the other nuisances. In the following we show that the only nuisances that commute with quantization are the isometric group of the plane. While it is common, following the literature on scale selection, to canonize it, scale is {\em not} canonizable without a loss, so the selection of a single representative scale is not advisable.\footnote{Even rotations are technically not invertible, because of the rectangular shape of the pixels; however, this is a second-order effect that can be neglected for practical purposes.} Instead, a description of a region of an image at all scales should be considered, since scale, in a quantized domain, is a {\em semi-group}, rather than a group.

\subsubsection{Interaction of group nuisances with quantization}
\index{Quantization}

Note that, per Theorem \ref{thm-equivariant}, only for canonizable nuisances can we design an equi-variant classifier via a co-variant detector and invariant descriptor. All other nuisances should be handled via marginalization or extremization in order to retain optimality (minimum risk), or via a template if one is willing to sacrifice optimality in favor of speed at decision time. 
\begin{theorem}[What to canonize]
\label{thm-what-to-canonize}
The only nuisance that commutes with quantization is the isometric group of the plane, that is the group of rotations, translations and reflections. 
\end{theorem}
{\bf Proof:}~{\em We want to characterize the group $g$ such that $I \circ g \circ \nu = I \circ \nu \circ g$ where $\nu$ is quantization. For a quantization scale $\sigma$, we have the measured intensity (irradiance) at a pixel $x_i$
\begin{multline}
I\circ \nu (x_i) \doteq \int_{{\cal B}_\sigma(x_i)} I(x) dx = \int \chi_{{\cal B}_\sigma(x_i)}(x) I(x) dx = \int \chi_{{\cal B}_\sigma(0)}(x-x_i) I(x) dx \\ \doteq \int {\cal G}(x -x_i; \sigma) I(x) dx
\end{multline}
where ${\cal B}_\sigma(x)$ is a ball of radius $\sigma$ centered at $x$, $\chi$ is a characteristic function that is written more generally as a kernel ${\cal G}(x; \sigma)$, for instance the Gaussian kernel ${\cal G}(x; \sigma) \doteq {\cal N}(x; 0, \sigma^2)$, allowing the possibility of more general quantization or sampling schemes, including soft binning based on a partition of unity of $D$ rather than simple functions $\chi$. Now, we have
\begin{equation}
(I \circ \nu)\circ g(x_i) = \left(\int{\cal G}(x-x_i; \sigma)I(x)dx \right)\circ g = \int{\cal G}(x-gx_i; \sigma)I(x)dx
\end{equation}
whereas, with a change of variable $x' \doteq gx$, we have
\begin{equation}
(I \circ g)\circ \nu (x_i) = \int{\cal G}(x-x_i; \sigma)I(gx)dx = \int{\cal G}(g^{-1}(x'-gx_i); \sigma)I(x')| J_g | dx'
\end{equation}
where $| J_g |$ is the determinant of the Jacobian of the group $G$ computed at $g$, so that the change of measure is $dx' = | J_g | dx$. From this it can be seen that the group nuisance commutes with quantization if and only if
\begin{equation}
\begin{cases}
{\cal G} = {\cal G}\circ g \\
|J_g| = 1.
\end{cases}
\end{equation}
That is, the quantization kernel has to be $G$-invariant, ${\cal G} (x; \sigma) = {\cal G}(gx; \sigma)$, and the group $G$ has to be an {\em isometry}. The only isometry of the plane is the set of planar rotations and translations (the Special Euclidean group $SE(2)$) and reflections. The set of isometries of the plane is often indicated by $E(2)$.
}

\begin{corollary}[Do not canonize scale (nor the affine group)]
The affine group does not commute with quantization, and in particular the scaling and skew sub-groups. As immediate consequence, neither do the more general projective group and the group of general diffeomorphisms of the plane. Therefore, scale should not be (globally) canonized and the scaling sub-group should instead be {\em sampled}. We will revisit this issue in Section \ref{sect-matching} where we introduce the selection tree.
\end{corollary}
So, although \cite{soatto09} suggests that invariant sufficient statistics can be devised for general viewpoint changes, this is only theoretically valid in the limit when there is no quantization and the data is available at infinite resolution. In the presence of quantization, canonization of anything more than the isometric group is not advisable. Because the composition of scale and quantization is a {\em semi-group}, the entire semi-orbit -- or a discretization of it -- should be retained. This corresponds to a {\em sampling} of scales, rather than a scale {\em selection} procedure.

\cut{\begin{example}[The SIFT detector revisited]
\end{example}}

It should be noted that the sampling of the scale group could be performed {\em uniformly}, by selecting a number of equally spaced scales, or {\em adaptively}, depending on the signal. For instance, SIFT's scale ``selection'' procedure \cite{lowe04distinctive}, based on Lindeberg's work \cite{lindeberg93scale}, should be interpreted as an adaptive scale {\em sampling}, rather than a {\em selection} (canonization), since scale is not canonizable. Correctly, SIFT allows the selection of multiple scales at the same location, which are signal-dependent samples of scale at that location.

The additive residual $n(x)$ does not pose a significant problem in the context of quantization since it is assumed to be spatially stationary and white/zero-mean, so quantization actually reduces the noise level:
\begin{equation}
n(x_i) = \int_{{\cal B}_\sigma(x_i)} n(x)dx \stackrel{\sigma}{\longrightarrow} 0.
\end{equation}
Instead, the other important nuisance is occlusion. 

\subsubsection{Interaction of group nuisances with occlusion: Local invariant features}
\label{sect-local}
\index{Occlusion}

Planar rotations do not generate occlusions, so canonization of rotation is, at least to first approximation, unaffected by occlusion. Translation, however, is. \cutTwo{In particular, if planar translations are due to parallax, \marginpar{\tiny \sc parallax} \index{Parallax} consisting of a translation of the optical center in front of a scene that is not flat and fronto-parallel, \marginpar{\tiny \sc fronto-parallel} \index{Fronto-parallel} there will be occlusions unless the scene is concave.} Therefore, translation can{\em not} be globally canonized either. This means that one should consider the set of all possible translations and defer the treatment of the nuisance to training or testing. This is indeed an approach that has recently taken hold in the recognition literature \cite{dalalT05,triggs04detecting}.

However, owing to the statistics of natural images \cite{mumfordG01,olshausen98sparse}, the response of any translation-co-variant detector will be neither a unique global extremum nor the entire lattice $\Lambda = D \cap {\mathbb Z}^2$,  but instead a discrete set of (multiple, isolated)  critical points $T_i \in \Lambda$. Therefore, one could hypothesize that each of them is a viable canonical representative, and then test the hypothesis at decision-time. This means that one should construct {\em multiple} descriptors for each canonical translation, at multiple scales, and then marginalize or eliminate occlusions by a max-out procedure at decision time, that now corresponds to a (combinatorial) search among all possible canonical locations in putatively corresponding images. This is just like the case of (multiple) scale selection in SIFT, that we interpreted as adaptive sampling of the scale group. One can think of the multiple selections of location and scales as adaptive sampling of the group, or multiple local canonization procedures. 

In either case, the procedure provides multiple similarity frames, one per each canonical translation $T_i$, and at that translation for each sample scale $\sigma_{j}$, and a canonized rotation $R_{ij}$:
\begin{equation}
\hat g_{ij} \doteq \{T_i, \sigma_{j}, R_{ij}, m_{ij}\}
\end{equation}
where we have indicated with $m_{ij}$ a canonized contrast transformation, although contrast can equivalently be eliminated by considering the level lines or the gradient direction as described in Section \ref{sect-nuisance-taxonomy}. 

The image, or a contrast-invariant, {\em relative to the reference frame identified by} $\hat g_{ij}$ is then, by construction, invariant {\em modulo a selection process}, in the sense that the corresponding frame may or may not be present in another datum (\eg a test image) depending on whether it is {\em co-visible}. Thus, we have a collection of {\em local} invariant features\footnote{If contrast invariance is desired, the image $I$ can be replaced by the gradient direction $\frac{\nabla I}{\|\nabla I\|}$.} of the form \marginpar{\tiny \sc local invariant features} \index{Co-visible region} \index{Invariant feature} \index{Local canonization}
 \begin{eqnarray}
\phi_{ij} (I) &=& I \circ {\hat g}^{-1}_{ij} = I(R_{ij}^T(S_j^{-1} x - T_i)), ~~~   \label{eq-local-feature}  \\ 
~~~~~~~~ && i,j =1, \dots N_T, N_S | {\cal B}_{\sigma_{j}}(x + T_i ) \cap \Omega  = \emptyset \nonumber
\nonumber
 \end{eqnarray}
where $S: \real^2 \rightarrow \real^2; \ x \mapsto S(x) = \sigma x$, with $\sigma >0$ is a scale transformation, with $S^{-1}(x) = x/\sigma$, $T\in \real^2$ is a planar translation, and $R \in SO(2)$ is a planar rotation \cite{maSKS}. Here $N_T$ and $N_S$ are the number of canonical locations and sample scales respectively. Note that the selection of occluded regions, which is excluded from the descriptor, is not known a-priori and will have to be determined at decision time as part of the matching process. In particular, given one image (\eg the training image) $I_1$ and its features $\phi_{ij}(I_1)$, and another image (\eg the test image) $I_2$, and its features $\phi_{lm}(I_2)$, marginalizing the nuisance amounts to testing whether $\phi_{ij}(I_1) \simeq \phi_{lm}(I_2)$, for all possible combinations of $i,j,l,m$, corresponding to the hypothesis that the canonical region around the canonical translation $T_i$ at scale $\sigma_j$ in image $I_1$ is {\em co-visible} with the region around the canonical translation $T_l$ at scale $\sigma_m$ in image $I_2$. \marginpar{\tiny \sc marginalizing occlusion via combinatorial selection}

In this sense, we say that {\em translation is locally canonizable}: The description of an image around each $t_i$ at each scale $\sigma_j$ can be made invariant to translation, unless the region of size $\sigma_j$ around $t_i$ intersects the occlusion domain $\Omega \subset D$, which is a binary choice that can only be made at decision time, not by pre-processing test images independently.

This is particularly natural when translation is canonized using a partition of the image into $\epsilon$-constant statistics (``segments''), such as normalized intensity, color spectral ratios, or normalized gradient direction histograms. In fact, each region (a node in the adjacency graph) or each junction (a face in the adjacency graph) can be used to canonize translation, and the image can be described by the statistics of adjacent neighbors, as we illustrate in Section \ref{sect-superpixels}.

{So, although translation is {\em not} globally canonizable, we will refer to its treatment as {\em local canonization} modulo a selection process to detect whether the region around the canonical representative is subject to occlusion.} What we have in the end, for each image $I$, is a set of multiple descriptors (or templates), one per each canonical translation, and for each translation multiple scales, canonized with respect to rotation and contrast, but still dependent on deformations, complex illumination and occlusions. One can also think of local canonization as a form of {\em adaptive sampling}, where regions that are not covariant are excluded. The same reasoning can also be applied to scale, where one can think of (multiple) scale selection as adaptive sampling of scale.

In Section \ref{sect-detectors} we will show how to construct local co-variant detectors, and in Section \ref{sect-descriptors} how to exploit this process to construct local invariant descriptors.

\begin{rem}[When to avoid canonization]
Canonization is just a way to factor out simple nuisances. Because of occlusions, this canonization process reduces to feature selection (in each individual image) and combinatorial matching. This process represents an approximation of the ``correct'' marginalization or max-out procedure described in Section \ref{sect-ml}. Therefore, if computational resources allow, the best course of action is to max-out or marginalize the nuisance, \ie to {\em avoid the canonization process altogether.} This choice is usually dictated by the application constraints, and should be made with knowledge of the tradeoffs involved: Canonization reduces the complexity of the classifier, but at a cost in discriminative power. Marginalization is the best option if one has a prior available. Otherwise, max-out is the best option if the classifier can be computed in time useful for the task at hand.
\end{rem}

\subsection{Clarifying the formal notation $I\circ g$}
\label{sect-clarify}

So far we have used the notation $I\circ g$ to indicate that, if $I = h(e, \xi, \nu)$, then $I\circ g = h(g,\xi, \nu)$. This may seem inconsistent, as the scene and the image live in different spaces, and therefore $g$ acting on $\xi$ is not the same as $g$ acting on $I$. We are now ready to explain this inconsistency.

{\color{orange}
For a nuisance $g$ to be eliminated in pre-processing without a loss, it has to be a group, as well as commute with non-group nuisances. After the discussion in the previous sections, this can be expressed as $h(g,\xi, \nu) = g h(e, \xi, \nu)$ for all $\nu$. Therefore, if we divide the group nuisance $g$ into an ``invertible'' component $g^i$ and a ``non-invertible'' component $g^n$, so that $I = h(g^i g^n, \xi, \nu)$, we then have that
\begin{equation}
I = h(g^i g^n , \xi, \nu) = g^i h(g^n ,\xi, \nu).
\end{equation}
But because $g^n$ is not invertible, and therefore it cannot be eliminated in pre-processing without a loss, its group structure is of no use, and we might as well lump $g^n$ with the non-invertible nuisance $\nu$. Therefore, from this point on, we can assume as a formal model of image formation the following: $I_t = g_t h(e, \xi, \nu_t) + n_t$, and for simplicity we can omit the identity $e$. Therefore, we can represent this equivalently a $I = g h(\xi, \nu)$ or $I = h(g, \xi, \nu)$. Now it should be clear that the (invertible) group $g$ can act on the image $I$, so the notation $I\circ g \doteq g I$ is justified.} 

For instance, in the Lambert-Ambient model, $I(x) = \rho(p)$, contrast normalization is commutative: A contrast transformation applied to the albedo $k(\rho(p))$ induces a contrast transformation on the image, and therefore it can be neutralized (``inverted'') by performing co-variant detection on the {\em image}. So, if $I(x) =  k(\rho(p))$, then ${\hat k}^{-1}(I(x)) = {\hat k}^{-1} k(\rho(p))$, where $\hat k = \hat k(I)$, is invariant to contrast. Similarly, a planar rotation applied to the scene, $\rho(Rp)$, where 
\[ R = \ba{ccc} \cos \theta & - \sin \theta & 0 \\ \sin \theta & \cos\theta & 0 \\ 0 & 0 & 1 \ea
\]
induces a rotation on the image plane, $I(r x)$ where
\[r = \ba{cc} \cos \theta & - \sin \theta \\ \sin \theta & \cos \theta \ea
\]
so $I(r x) = \rho (R p)$. A choice of orientation on the image plane, corresponding to a canonical rotation $\hat r = \hat r (I)$ can then be used to canonize $R$. Calling
\[
\hat R = \ba{cc} \hat r & \begin{array}{c} 0 \\ 0 \end{array} \\ 0 ~~~ 0 & 1 \ea
\]
we have that $I(\hat r^{-1} x) = I(\hat r^T x) = \rho(\hat R^T R p)$ is invariant to a planar rotation $R$. The story is considerably different for translation, and for general rigid motions, including out-of-plane rotation. 

A translation along the optical axis, $T = [0, \ 0, \ s]^T$ induces a transformation on the image plane that depends on the shape of the scene $S$. If the scene is planar and fronto-parallel, translation along the optical axis induces a re-scaling of the image plane, also a group: $I(s x) = \rho(p + T)$. However, if the scene is not planar and fronto-parallel, forward translation can generate very complex deformations of the image domain, including  (self-) occlusions. Similarly, for translation along a direction parallel to the optical axis, $T = [u, \ v, \ 0]^T$,  only if the scene is planar and fronto-parallel does this result in a planar translation, so that $I(x + [u, \ v]^T) = \rho(p + T)$. 

It should be mentioned that different data modalities yield different commutativity properties, so it is possible that for certain data types the entire group nuisances may be commutative. For instance, cast shadows are non-invertible nuisances in grayscale images. However, if one has multiple spectral bands available, then several contrast-invariant features, such as spectral ratios, can be employed to eliminate the cast shadows.


\subsection{Local canonization of general rigid motion}

As we have discussed, parallax motion, including translation and out-of-plane rotation, does not commute with rotation, therefore one can at best locally canonize it. If one wishes to eliminate the effects of an arbitrary rigid motion, the entire group of diffeomorphisms has to be canonized; as shown in \cite{sundaramoorthiPVS09}, this comes at a loss of discriminative power, since scenes of different shape are lumped onto the same equivalence class. But since occlusions force locality, one can also  approximate general diffeomorphisms locally, and therefore only canonize sub-groups of planar diffeomorphisms. There is then a tradeoff between the sub-group being canonized and the size of the domain where the canonization is valid. Given an arbitrary margin, there will be a domain where the deformation is approximated by a translation, a larger domain where it is approximated by an isometry, a larger domain yet where it is a similarity, an affine transformation, and a projective transformation. 

{\color{pea}
To see that, consider a general rigid motion in space $(R, T)$. That induces a deformation of the image domain that can be described, in the {\em co-visible} regions, by equation (\ref{eq-motion-field}). In more explicit form, if we represent shape $S$ as the graph of a function in one of the images, say at time $t=0$, so that $p = p(x_0) = \bar x_0 Z(x_0)$, with $x_0 \in D$, then under the assumptions of the Lambert-Ambient model, $I_t(x_t) = \rho(p) = I_0(x_0)$ where 
\begin{equation}
x_t = w(x_0) = \pi(R \bar x_0 Z(x_0) + T) = \frac{R_{(1:2,:)} \bar x_0 Z(x_0) + T_{(1:2)}}{R_{(3,:)} \bar x_0 Z(x_0) + T_3}
\end{equation}
in the co-visible region $D \cap w(D)$. Here we have used Matlab notation, where $R_{(1:2,:)}$ means the first two rows of $R$.}

Of course, the size of the region where the approximation is valid cannot be determined a-priori, and should instead be estimated as part of the correspondence hypothesis testing process \cite{vedaldiS06cvpr}. However, it is common practice to select the size of the regions adaptively as a function of the {\em scale} of the translation-covariant detector. This will be further elaborated in Chapter \ref{ch-correspondence}. 

\cut{For now, we wish to substantiate the approximation statement made before. Let $\epsilon > 0$ be a tolerable approximation error. For a given depth map $Z(\cdot)$, and rigid motion $(R, T)$, around a position $x\in D$, assuming that $w(x) = x$ (that is, we have canonized translation), the domain deformation induced in a neighborhood of $x$ of the form $x+v$ with $v\in \Omega$ a neighborhood of $x$, we have, to first-order approximation
\begin{equation}
w(x+v) =   x + v + \left(\frac{\partial w}{\partial x}_{|_{x}} -Id \right) v
\end{equation}
From this one can see that, for the translational approximation to be valid, $v$ has to be small enough that $\| w(x-v) - \left(\frac{\partial w}{\partial x}_{|_{x}} -Id \right) v \| \le \epsilon$.}

In the next section we establish the relation between local canonization and texture.

\section{Textures}
\label{sect-texture}
\index{Texture}

This section establishes the link between the notion of canonization, described in the previous section, and the design of invariant features, tackled in the next section.

\subsection{Defining ``textures''}
\label{sect-def-texture}

\begin{figure}[htb]
\begin{center}
\includegraphics[width=.5\textwidth]{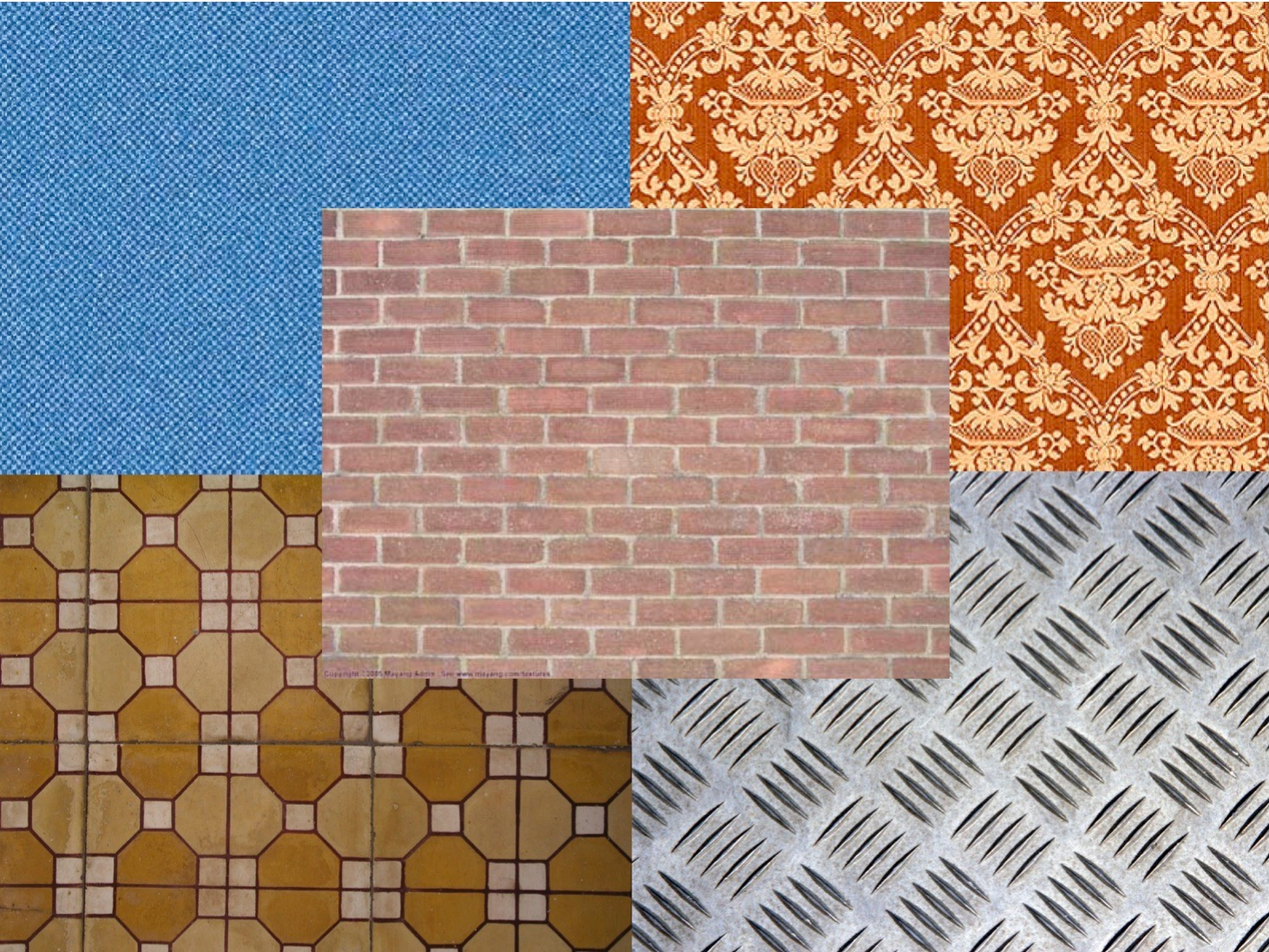}
\end{center}
\caption{\sl Regular textures}
\label{fig-texture1}
\end{figure}

\begin{figure}[htb]
\begin{center}
\includegraphics[width=.7\textwidth]{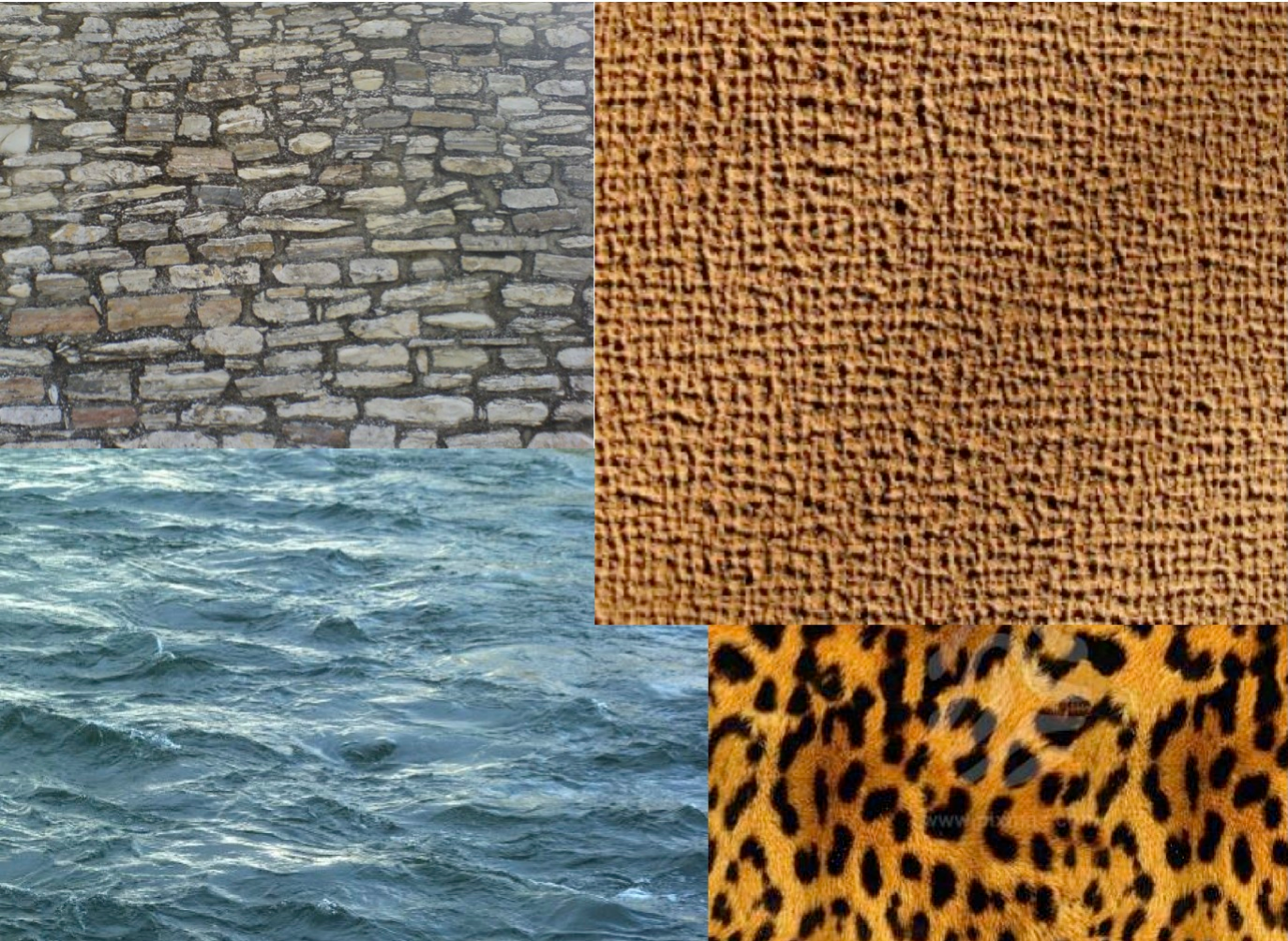}
\end{center}
\caption{\sl Stochastic textures}
\label{fig-texture2}
\end{figure}

A ``texture'' is a region of an image that exhibits some kind of spatial regularity. Figure \ref{fig-texture1} shows some examples of what are often called {\em regular textures}. They share the spatial regularity of some {\em elementary structure}, or ``texture element'', or ``texton'' \cite{julesz75}. The images in Figure \ref{fig-texture2}, on the other hand, do not exhibit regular repetition of any such structure. Instead, they are characterized by the fact that some ensemble property of the image is spatially homogeneous, \ie, some statistic is translation-invariant. Such ensemble properties are pooled in a region whose minimal size plays the role of the elementary texture element. Translation invariance of statistics is captured by the  notion of {\em stationarity} \cite{shao98}. For instance, a wide-sense stationary process has translation-invariant mean and correlation function. A strict-sense stationary process has translation-invariant statistics of all orders. 
The concept of stationarity can be generalized from simple translation invariance to invariance relative to some group:  In Figure \ref{fig-texture5} (top) the images do not have homogeneous statistics; however, in most cases one can apply an invertible transformation to the images that yields spatially homogeneous statistics.

The following characterization of texture is adapted from \cite{georgiadisCS13DCC}, where the basic definitions of stationarity, ergodicity, and Markov sufficient statistics are described in some detail.  The important issue for us is that, if a process is stationary and ergodic, a statistic $\phi$ can be predicted from a realization of the image in some region $\bar \omega$. Once established that a process is stationary, hence spatially predictable, we can inquire on the existence of a statistic that is {\em sufficient} to perform the prediction. This is captured by the concept of Markovianity: 
\begin{equation}
I(x) \perp I_{|_{{\omega}^c}} ~ | ~ \phi_{\omega\backslash x}.
\label{eq-suff}
\end{equation}
Equation (\ref{eq-suff}) establishes $\phi_{{\omega}}$ as a {\em Markov sufficient statistic}. In general, there will be many regions $\omega$ that satisfy this condition; the one with the smallest area $|\omega | = r$, determines a {\em minimal sufficient statistic} and the statistic defined on it is the {\em elementary texture element}. From now on, we will refer to $\phi_\omega$ as the minimal {\em Markov sufficient statistic.} 

We can then define {\em a texture as a region of an image that can be rectified into a sample of a stochastic process of a planar lattice that is locally stationary, ergodic and Markovian.} More precisely, assuming for simplicity the trivial (translation) group, a region $\Omega \subset D \subset \real^2$ of an image is a texture at scale $\sigma > 0$ if there exist regions $\omega \subset \bar \omega \subset \Omega$ such that $I$ is a realization of a stationary, ergodic, Markovian process locally within $\Omega$, with $I_{|_{\omega}}$ a Markov sufficient statistic and $\sigma = | \bar \omega |$ the stationarity scale. We then have that 
\begin{equation}
H(I(x) | \omega - \{x\}) = H(I(x) | \Omega - \{x\}).
\end{equation}
We can therefore seek for $\omega \subset \Omega$ that satisfies the above condition. Without a complexity constraint, there are many regions that satisfy the above condition. To find $\omega$, we therefore seek for the {\em smallest} one, by solving
\begin{equation}
\hat \omega = \arg\min_{\omega} H(I(x) | \omega - \{x\}) + \frac{1}{\beta} | \omega |.
\label{eq-minent}
\end{equation}
Note that this is a consequence of the Markovian assumption and the Markov sufficient statistic satisfies the Information Bottleneck principle \cite{tishbyPB00} with $\beta \rightarrow \infty$.
As a special case, we can choose $\bar \omega$ to belong to a parametric class of functions, for instance square neighborhoods of $x$, excluding $x$ itself, of a certain size $\sigma$, ${\cal B}_\sigma(x)$, so the optimization above is only with respect to the positive scalar $\sigma$.
The tradeoff will naturally settle for $1 < r < \sigma$. Therefore, we can simultaneously infer both $\sigma$ and $r $ by minimizing the sample version of (\ref{eq-minent}) with a complexity cost on $\sigma = |\bar \omega|$:
\begin{equation}
\hat \omega, \sigma = \arg\min_{\omega, \sigma = |\bar \omega|} \hat H(I(x) | \omega - \{x\}) + \frac{1}{\beta} | \bar \omega |.
\label{eq-minent-sample}
\end{equation}
Note that both $\omega$ and $\bar \omega$ are necessary for extrapolation: $\omega$ defines the Markov neighborhood used for comparing samples, and $\bar \omega$ defines the region where such samples are sought to approximate the probability distribution $p(I(x) | \omega - \{x\})$.

Eq. (\ref{eq-minent-sample}) provides means to infer both the Markov sufficient statistic as well as the scale $\sigma = | \bar \omega|$ of a texture, assuming that the stationarity and ergodicity assumptions are satisfied. Testing for stationarity (ergodicity must be assumed and cannot be validated) amounts to inferring $\Omega$, a {\em texture segmentation} problem\cut{, described in Sect. \ref{sect-segm}}. Estimating the group element $g \in G$ amounts to a {\em canonization}, or {\em alignment} process \cite{tilt}. 

Given $\{I(x), x\in \Omega\}$, compression is achieved by inferring the (approximate) minimal sufficient statistic $\omega$ and the stationarity scale $\sigma$ by solving (\ref{eq-minent-sample}). Then $I(\bar \omega)$, for any $\bar \omega \subset \Omega$ with $| \bar \omega| = \sigma$ is stored. 

Given a compressed representation $I(\bar \omega)$, we can in principle synthesize novel instances of the texture by sampling from $dP(I_{|_{\omega}})$ within $\bar \omega$. In a non-parametric setting this is done by sampling directly neighborhoods $I(\omega)$ within $\bar \omega$. To extrapolate the texture from a given sample $I(\bar \omega)$ \cut{, or inpaint a missing portion, }compatibility conditions have to be ensured at the boundaries of $\bar \omega$.\cut{, which causes the sampling complexity to increase. Efros and Leung \cite{efrosL99} proposed a {\em causal} sampling scheme that satisfies the boundary conditions. However, it fails to respect the Markov structure: To synthesize $I(x)$ one needs $I(\omega - \{x\})$, with $\omega \ni x$, whereas \cite{efrosL99} only use {\em half} of the region $\omega$. If we split $\omega$ into a subregion $\tilde \omega$ and its complement $\tilde \omega^c = \omega - \tilde \omega$, they find regions around the boundary of $\bar \omega$ that resemble $I(\tilde \omega)$, and therefore copy $I(\tilde \omega^c)$ outside the boundary. Unfortunately, $I(\omega)$ is not a sufficient statistic for $I(\tilde \omega^c)$. This causes the ``blocky'' artifacts in Fig. \ref{fig-synth}. } 

To find $\Omega$ we can solve
\begin{equation}
\Omega = \arg\min_{\bar \omega}  H(I(x \in \partial \bar \omega) | \bar \omega ) + \frac{\beta}{|\bar \omega|}
\end{equation}
where the entropy is normalized by the length of the boundary $|\partial \bar \omega|$ (entropy rate).
Since we do not know the distribution outside $\bar \omega$, the above is approximated by
\begin{equation}
\Omega = \arg\min_{\bar \omega} \hat H(I(x \in \partial \bar \omega) | \bar \omega, \hat I(x \in \delta \bar \omega - \{x\})) + \frac{\beta}{|\bar \omega|}.
\end{equation}
In the continuum, the problem above can be solved using variational optimization as in \cite{sundaramoorthiSY10}. In the discrete, the reader can refer to \cite{georgiadisCS13DCC} for a description of compression, synthesis, segmentation and rectification algorithms. 

\subsection{Textures and Structures}
\label{sect-tex-str}
In this section we establish the relation between textures, defined in the previous section, and structures, defined  by co-variant detectors. Consider a point $x\in D$ and its neighborhood. If it is canonizable at a scale $\epsilon$, there is a co-variant detector with support $\epsilon$ (a statistic) that has an isolated extremum. This implies that the underlying process is not stationary at the scale $| \bar \omega | = \epsilon$. Therefore, it is not a texture. It also implies that any region $\omega$ of size $\epsilon = | \omega |$ is not sufficient to predict the image outside that region. This of course does not prevent a region that is canonizable at $\epsilon$ to be a texture at a scale $\sigma  \neq \epsilon$. Within a region $\sigma$ there may be multiple frames of size $\epsilon$, spatially distributed in a way that is stationary/Markovian, so the region may be a texture at a scale $\sigma > \epsilon$. Vice-versa, if a region of an image is a texture with $\sigma = \bar \omega$, it cannot have a unique (isolated) extremum within $\bar \omega$, lest it would not be a sample of a stationary process. Of course, it could have multiple extrema, each isolated within a region of size $\epsilon < \sigma$. The above argument is a sketch of a proof of the following:
\begin{theorem}[Structure-Texture]
\label{thm-1}
For any given scale of observation $\sigma$, a region $\bar \omega$ with $|\bar \omega| = \sigma$ is either a structure or a texture.
\end{theorem}
Hence one can detect textures for each scale, as the residual of the canonization process described in the earlier part of this chapter. One may have to impose boundary conditions so that the texture regions fill around structure regions seamlessly. This can be accomplished by an explicit generative model that enforces boundary conditions and matches marginal statistics in the texture regions, following \cite{jaynes57}, as illustrated by \cite{zhuWM96}. However, this has to be done across all scales, unlike \cite{zhuWM96}, since whether a region is classified as texture or structure depends critically on the scale $\sigma$.

\begin{figure}[htb]
\begin{center}
\includegraphics[width=.7\textwidth]{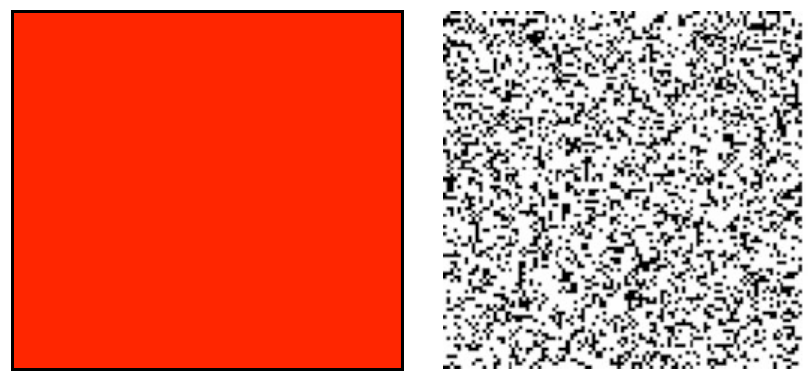}
\end{center}
\caption{\sl {\bf Testing the definition of texture}. Both images above satisfy the definition of texture. Images like the one on the left, however, are often called ``textureless.'' Therefore, textureless is a limiting case of a trivial texture where any statistic $\phi$, computed in any region $\omega$, is constant with respect to any group $G$. Images like the one on the right are called {\em random-dot displays}, and Julesz showed that one can establish binocular correspondence between such displays (see Figure \ref{fig-rds}). 
}
\label{fig-texture4}
\end{figure}

\begin{figure}[htb]
\begin{center}
\includegraphics[width=.7\textwidth]{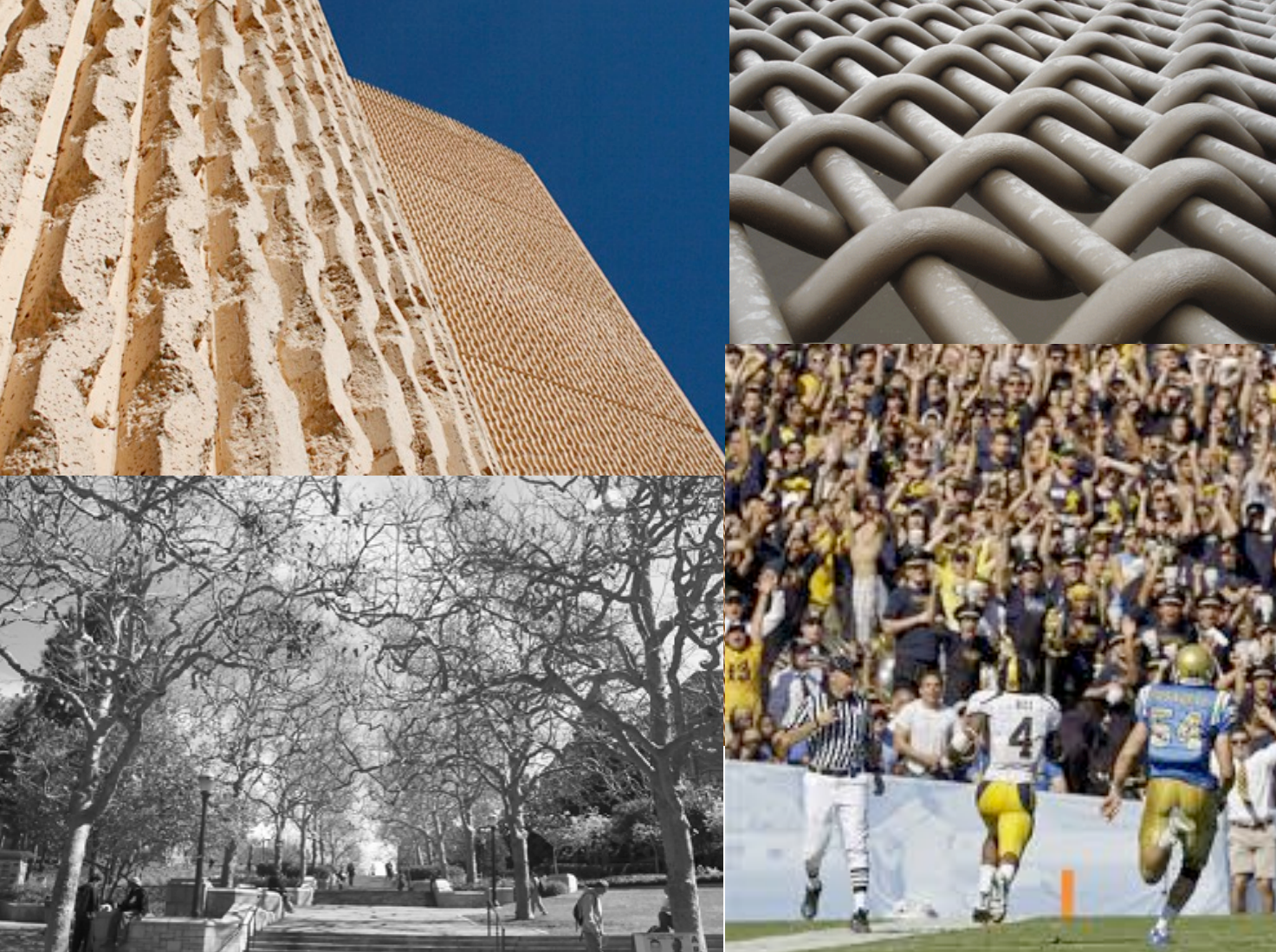}
\end{center}
\caption{\sl {\bf Interplay of texture and scale}. Whether something is a texture depends on scale, and ``structures'' can appear and disappear across scales. Such ``transitions'' can be established in an ad-hoc manner using complexity measures as in Figure \ref{fig-texture6}, or they can be established unequivocally by analyzing multiple views of the same scene through the notion of proper sampling, introduced in Section \ref{sect-correspondence}.}
\label{fig-texture5}
\end{figure}

\begin{figure}[htb]
\begin{center}
\includegraphics[width=.7\textwidth]{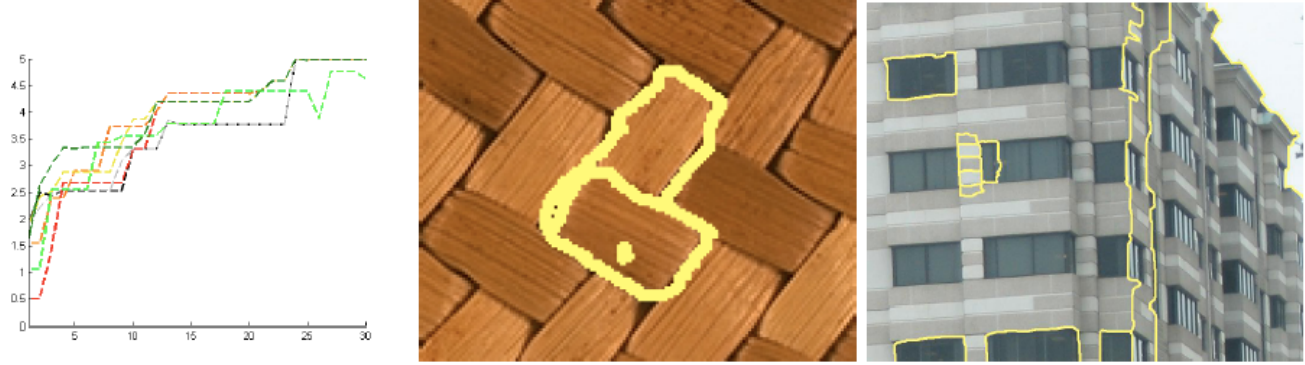}
\end{center}
\caption{\sl {\bf Texture-structure transitions} (from \cite{boltzNS10}). The left plot shows the entropy profile for each of the regions marked with yellow boundaries in the middle and right figure. The abscissa is the scale of the region where such entropy is computed, starting from a point and ending with the entire image. As it can be seen, the entropy profile exhibits a staircase-like behavior, with each plateau bounded below by the scale corresponding to the small $\omega$, and above by the boundary of $\cal B$. This also shows that the neighborhood of a given point can be interpreted as texture at some scales (all the scales corresponding to flat plateaus), structure at some other scale (the non-flat transitions), then texture again, then structure again etc. }
\label{fig-texture6}
\end{figure}

\index{Intrinsic scale}
\index{Scale!intrinsic}

\begin{rem}[``Stripe'' textures]
The reader versed in low-level vision will notice that edges are {\em not} canonizable with respect to the groups we have discussed so far, including the translation group. This should come at no surprise, as we have anticipated in Remark \ref{rem-invariance-no-covariance}, because of the {\em aperture problem:} \marginpar{\tiny \sc aperture problem} \index{Aperture problem} While one can fix translation along the direction normal to the edge (at a given scale), one cannot fix translation along the edge (Figure \ref{fig-aperture}). However, if one considers the sub-group of translations normal to the edge, the corresponding region is indeed canonizable. 
\begin{figure}[htb]
\begin{center}
\includegraphics[width=.5\textwidth]{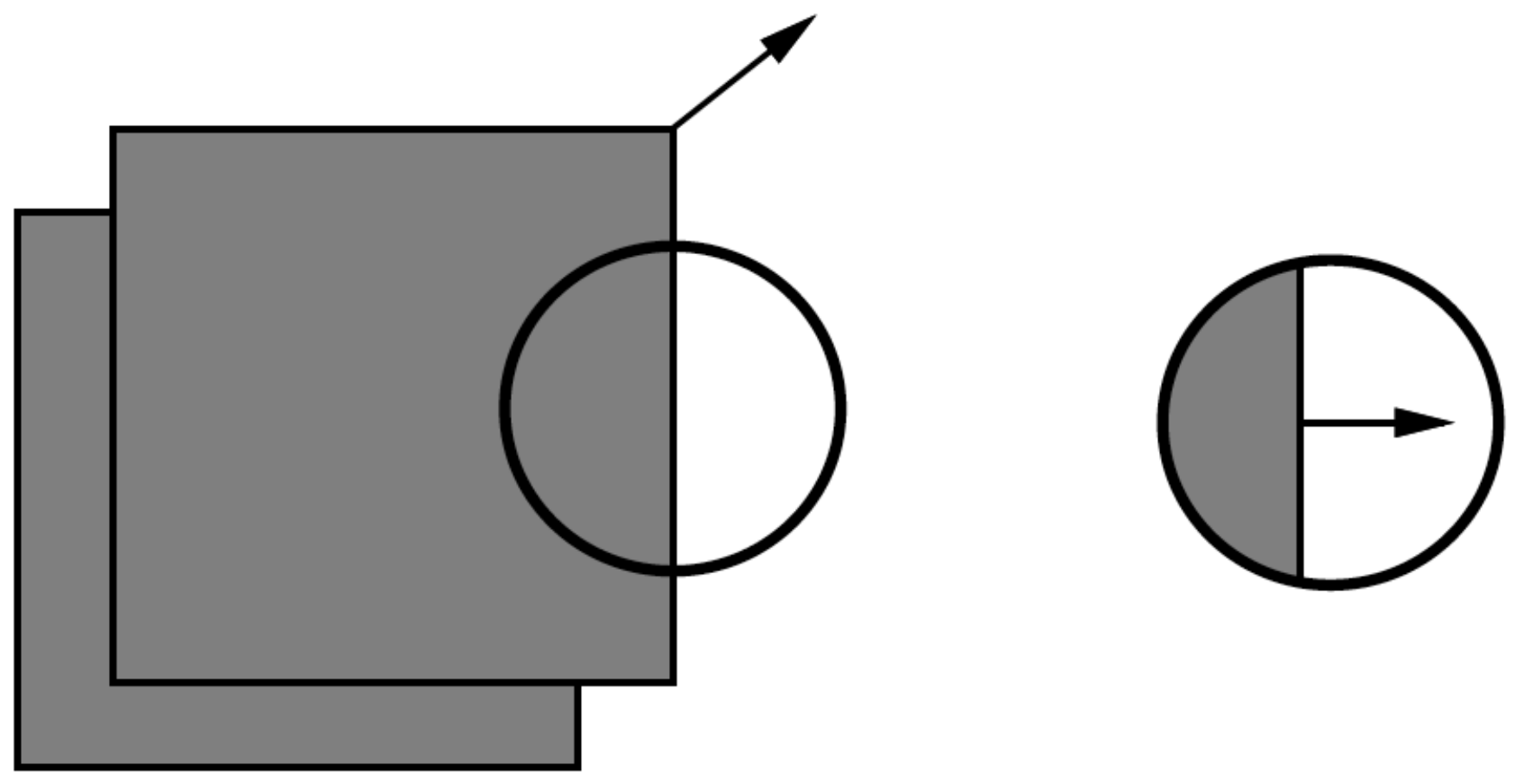}
\includegraphics[width=.1\textwidth]{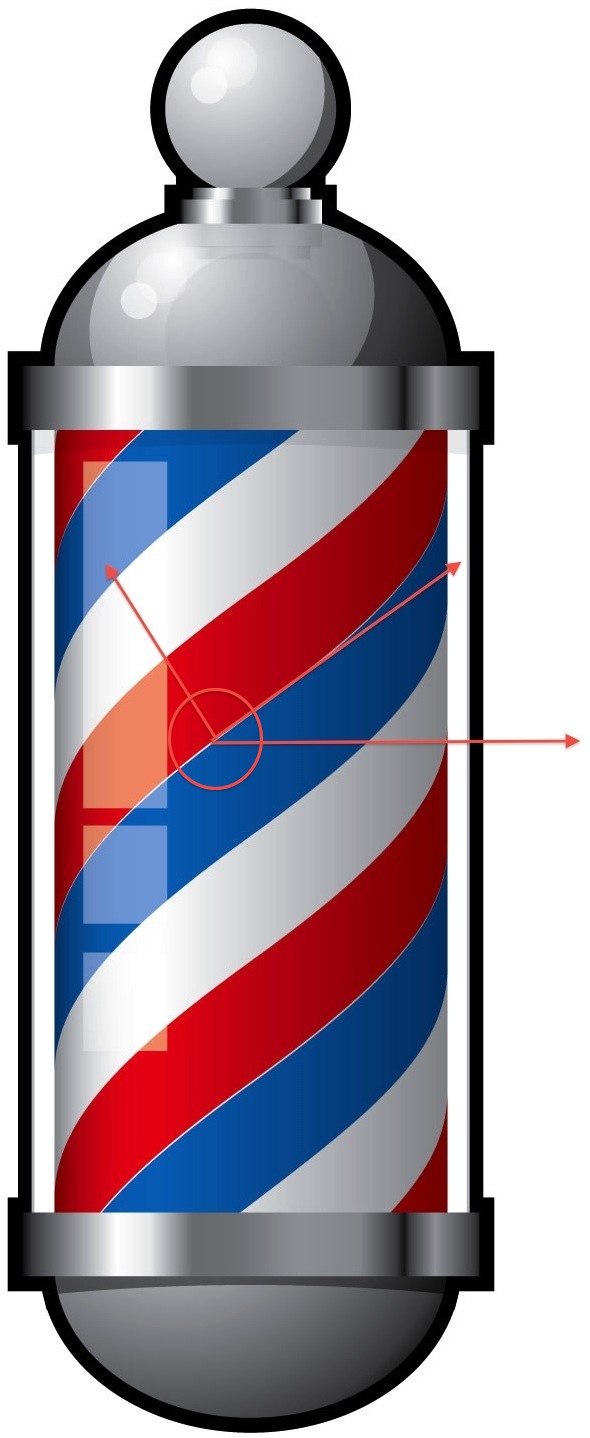}
\end{center}
\caption{\sl The aperture problem: Locally, in a ball of radius $\sigma$, ${\cal B}_\sigma$, motion can only be determined in the direction of the non-zero gradient of the image. The component of motion parallel to an edge cannot be determined. This causes several visual illusions, including the so-called ``barber pole'' effect where a rotating cylinder appears to translate upward.}
\label{fig-aperture}
\end{figure}
\end{rem}

Since the decision of what is a structure depends on time (or, more in general, on the presence of multiple images), consequently, what is a texture is also a decision that requires multiple images. For instance, a random-dot stereogram \cite{julesz71} is everywhere canonizable and properly sampled at some scale $\sigma$, \cut{witness the fact that}since one can establish correspondence. It is, however, a texture at all coarser scales, where any statistic is translation invariant.

The ``random-dot display'' of Figure \ref{fig-texture4} (right) seems to provide a counter-example to Theorem \ref{thm-1}. In fact, Julesz \cite{julesz71} showed that random-dot displays can be successfully fused binocularly, which means that pixel-to-pixel correspondence can be established, which means the the image is canonizable at every pixel which, by Theorem \ref{thm-1}, means that it is not a texture. But the random-dot display satisfies the definition of texture given in the previous section. How can this be explained?

First, there is no contradiction, since random-dot displays {\em are} indeed canonizable {\em at the scale of the pixels} $\sigma = 1$. However, they are stationary at any coarser scale, hence they are textures {\em at those scales}. 

More importantly, however, whether two regions of an image can be matched {\em at a given scale} depends not only on whether they are canonizable, but also whether they are {\em properly sampled at that scale}. We have anticipated the concept of proper sampling in Remark \ref{rem-proper-sampling}, and we will introduce and discuss the notion of proper sampling in Section \ref{sect-matching}, where we will revisit the random-dot display. There, we will see that the critical scales within which regions can be successfully matched (hence {\em not} called ``textures'' per Theorem \ref{thm-1}) can{\em not} be decided by analyzing one image alone. Instead, the ``thresholds'' for the transition from ``matchable'' to ``non-matchable'' require having multiple images of the same scene available.

\chapter{Designing feature detectors}
\label{sect-detectors}

Section \ref{sect-interaction} showed that rotation and contrast can be canonized, and that translation and scale can only be locally canonized. The definition of a canonized feature requires the choice of a functional $\psi$ (a co-variant detector), and an invariant statistic $\phi$ (a feature descriptor). Here we will focus on the location-scale group $g = \{x, \sigma\} \subset \real^2\times \real^+$, assuming that rotation and contrast have been canonized separately (later we advocate using gravity as a global canonization reference for orientation). The goal, then, is to design functionals $\psi$ that satisfy the requirements in Definition \ref{defn-covariant} that they yield isolated critical points in $G$. At the same time, however, we want them to be ``unaffected'' as much as possible by other nuisances.\footnote{So, we want $\nu$-stable $G$-covariant detectors.} Such ``insensitivity'' is captured by the notions of commutativity, stability, which we introduce next, and proper sampling, that we introduce in the next chapter.

\section{Sensitivity of feature detectors}
\label{sect-stability}
\index{Stability}
\index{Feature!stable}
\index{Stability!BIBO}
\index{Stability!structural}

We consider two qualitatively different measures of sensitivity. Note that we use the (improper) term ``stability'' because it is most commonly used in the literature, even though there is no {\em equilibrium} involved, and therefore the proper system-theoretic nomenclature would really be {\em sensitivity}.\footnote{An even worse name used for sensitivity is ``invariance,'' that suggest that a statistic could be more or less invariant. Invariance, as it has been defined here, is a binary property of a statistic: It either depends on the nuisance, or it does not. Sensitivity, on the other hand, comes in shades, as a statistic can be more or less sensitive to a nuisance, regardless of whether the latter is a group.}
\cut{BIBO stability and structural stability. The former imposes that small perturbations of the non-invertible nuisances yield small changes in the canonical element. Many existing detectors are designed to be structurally stable (although many have a large BIBO gain, for instance, SIFT, as it can be seen in Figure X of \cite{lindeberg98}, where small changes in scale produce significant changes in translation.) Structural stability imposes that small perturbations of non-invertible nuisances does cause catastrophic failure of the canonization mechanism (the detector).} We will start from the most common notion of bounded-input bounded-output stability.
\index{BIBO stability}
\begin{defn}[BIBO stability]
\label{def-bibo}
A $G$-covariant detector $\psi$ (Definition \ref{defn-covariant}) is {\em bounded-input bounded-ouput} (BIBO) stable if small perturbations in the nuisance cause small perturbations in the canonical element. {\color{pea} More precisely, $\forall \ \epsilon > 0 \ \exists \ \delta = \delta(\epsilon)$ such that for any perturbation $\delta \nu$ with $\| \delta \nu \| < \delta$ we have $ \| \delta \hat g \| < \epsilon$.}
\end{defn}
{\color{pea}
Note that $\hat g$ is defined implicitly by the equation $\nabla \psi(I, \hat g(I)) = 0$, and a nuisance perturbation $\delta \nu$ causes an image perturbation $\delta I = \frac{\partial h}{\partial \nu}\delta \nu$. Therefore, we have from the Inverse Function theorem\footnote{\color{pea} One has to exercise some care in defining the proper (Frech\`et) derivatives depending on the function space where $\psi$ is defined. The implicit function theorem can be applied to infinite-dimensional spaces so long as they have the structure of a Banach space (Theorem A.58, page 246 of \cite{kirsch96}). Images can be approximated arbitrarily well in $L^1(\real^2)$, that is Banach.} \cite{guilleminP74} 
\begin{equation}
\delta \hat g = - 
|J_{\hat g}|^{-1} \frac{\partial h}{\partial \nu} \delta \nu \doteq K \delta \nu
\label{eq-BIBO}
\end{equation}
where $J_g$ is the Jacobian (\ref{eq-jacobian}) and $K$ is called the {\em BIBO gain.} As a consequence of the definition, $K < \infty$ is finite.} The BIBO gain can be interpreted as the sensitivity of a detector with respect to a nuisance. Most existing feature detector approaches are BIBO stable with respect to simple nuisances. Indeed, we have the following
\begin{theorem}[Covariant detectors are BIBO stable]
\label{thm-covariant}
Any covariant detector is BIBO-stable with respect to noise and quantization.
\end{theorem}
\cutTwo{{\bf Proof\cut{ of thm \ref{thm-covariant}}:} {\em Noise and quantization are additive, so we have $\frac{\partial h}{\partial \nu}\delta \nu = \delta \nu$, and the gain is just the inverse of the Jacobian determinant, $K = | J_{\hat g} |^{-1}$. Per the definition of co-variant detector, the Jacobian determinant is non-zero, so the gain is finite.
}}

\begin{figure}[htb]
\begin{center}
\includegraphics[width=.4\textwidth]{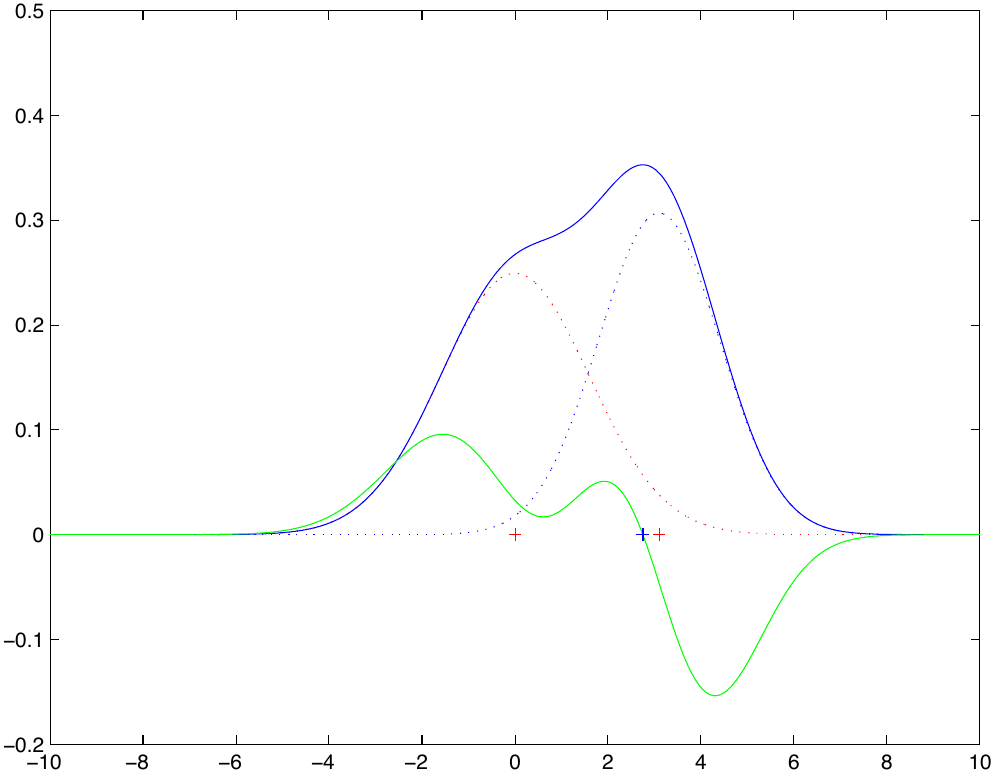}
\includegraphics[width=.4\textwidth]{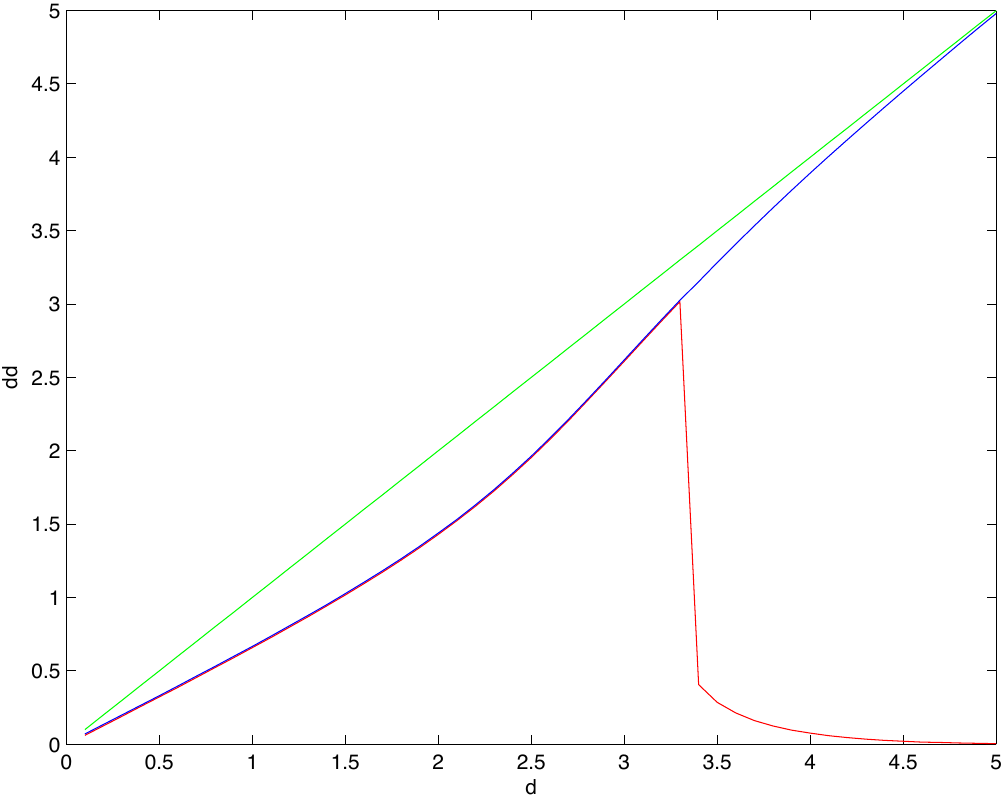}
\end{center}
\caption{\sl Catastrophic approximation: The original data is the sum of two Gaussians, one centered at $\mu_1 = 0$ (left, red dotted line), the other at a distance $\mu_2 = d$ (left, blue dotted line) that increases from $0$ to $5$ (green line). The translation-detector $\hat x$ initially detects one extremum whose position (blue) deviates from $\mu_2$ as $d$ increases. However, at a certain point the detector $\hat x$ splits into two, $\hat x_2$ that follows $\mu_2$, and $\hat x_1$ that converges towards $\mu_1 = 0$. Note that at no point does the translation detector $\hat x$ coincide with the actual modes $\mu$ (\ie, this detector is not BIBO insensitive). Furthermore, only for sufficiently separated, or sufficiently close, extrema does the location of the detector approximate the location of the actual extrema.}
\label{fig-topology-diagram}
\end{figure}

\noindent BIBO stability is reassuring, and it would seem that a near-zero gain is desirable, because it is ``maximally (BIBO)-stable.'' {\color{pea} However, simple inspection of (\ref{eq-BIBO}) shows that $K=0$ is not possible without knowledge of the ``true signal.'' In particular, this is the case for quantization, when the operator $\psi$ must include spatial averaging with respect to a shift-invariance kernel (low-pass, or anti-aliasing, filter).} However, {\em a non-zero BIBO gain is irrelevant for recognition,} because it corresponds to an additive perturbation of the domain deformation (domain diffeomorphisms are a vector space), which is a nuisance to begin with (corresponding to changes of viewpoint \cite{sundaramoorthiPVS09}).  On the other hand, structural instabilities are the plague of feature detectors. \cutTwo{When the Jacobian is singular, $| J_{\nabla \psi}| \rightarrow 0$, we have a {\em degenerate critical point}, a catastrophic scenario \cite{postonS78} whereby a feature detector returns the {\em wrong} canonical frame.}
\index{Structural stability}
\index{Stability!structural}
\marginpar{\tiny \sc structural stability margin}
\begin{defn}[Structural Stability]
\label{def-structural-stability}
A $G$-covariant detector $\psi \ | \ \nabla \psi(I, \hat g(I)) = 0$ is Structurally Stable if small perturbations $\delta \nu$ preserve the rank of the Jacobian matrix:{\color{pea}
\begin{equation}
\exists \ \delta > 0 \ | \ | J_{\hat g} 
| \neq 0 
\Rightarrow 
| J_{\hat g + \delta \hat g} 
| \neq 0 ~~ \ \forall \ \delta \nu \ | \ \| \delta \nu \| \le \delta
\label{eq-margin}
\end{equation}
with $\delta \hat g$ given from (\ref{eq-BIBO}).}
\end{defn}
 In other words, a detector is structurally stable if small perturbations do not cause singularities in canonization. We define the maximum norm of the nuisance that does not cause a catastrophic change \cite{postonS78} in the detection mechanism the {\em structural stability margin}. This can serve as a score to rank features. 
\index{Stability margin}
\begin{defn}[Structural Stability Margin]
We call the largest $\delta$ that satisfies equation (\ref{eq-margin}) the {\em structural stability margin}:
\begin{equation}
\delta^* = \sup \| \delta \nu \| ~~~~ | ~~~~ |J_{\hat g + K \delta \nu} | \neq 0
\end{equation}
\end{defn}
\cutThree{\begin{example}[Structural stability for the translation-scale group]
\label{example-gaussian}
~\\ Consider the set of images, approximated by a sum of Gaussians as described in Section \ref{sect-light-gauss}. The image is then represented by the centers of the Gaussians, $\mu_i$, their variance $\sigma_i^2$ and the amplitudes $\alpha_i$, so that $I(x) = \sum_i \alpha_i {\cal G}(x-\mu_i; \sigma_i^2)$. Consider a detection mechanism that finds the extrema $\hat g = \{\hat x, \sigma\}$ of the image convolved with a Gaussian centered at $\hat x$, with standard deviation $\sigma$: $\psi(I, \hat g) = I * \nabla {\cal G}(x-\hat x; \sigma^2) = 0$. Among all extrema, consider the two $\hat x_1, \hat x_2$ that are closest. Without loss of generality, modulo a re-ordering of the indices, let $\mu_1$ and $\mu_2$ be the ``true'' extrema of the original image. In general $\hat x_1 \neq \mu_1$ and $\hat x_2 \neq \mu_2$. Let the distance between $\mu_1$ and $\mu_2$ be $d = | \mu_2 - \mu_1|$, and the distance between the detected extrema be $\hat d = | \hat x_2 - \hat x_1 |$. Translation nuisances along the image plane do not alter the structural properties of the detector ($\hat d$ does not change). However, translation nuisance orthogonal to the image plane do. These can be represented by the scaling group $\sigma$, and in general $\hat d = \hat d(\sigma)$ is a function of $\sigma$ that starts at $\hat d = d$ when $\sigma = 0$ and becomes $\hat d = 0$ when $\sigma = \sigma^*$, \ie when the two extrema merge in the scale-space. In this case, $\delta^* = \sigma^*$ is the structural stability margin. It can be computed analytically for simple cases of Gaussian sums, or it can be visualized as customary in the scale-space literature. It is the maximum perturbation that can be applied to a nuisance that does not produce bifurcations in the detection mechanism (Figure \ref{fig-topology-diagram}). Note that one could also compute the structural stability margin using Morse's Lemma, or the statistics of the detector (\eg the second-moment matrix). Finally, the literature on Persistent Topology \cite{edelsbrunnerLZ02,cohenEH07,chazalGOS09} also provides methods to quantify the life-span of structures, which can be used as a proxy of the structural stability margin. Indeed, the notion of structural stability proposed above is a special case of persistent topology.
\end{example}}
\index{Aliasing}
\index{Proper sampling}
\index{Sampling!proper}
A sound feature detector is one that identifies Morse critical points in $G$ that are as far as possible from singularities. Structural instabilities correspond to {\em aliasing errors}, or {\em improper sampling} \marginpar{\tiny \sc proper sampling} (Section \ref{sect-correspondence}), where spurious extrema in the detector $\psi$ arise that do not correspond to extrema in the underlying signal. 
Proper sampling depends on the detector functional $\psi$, that in the presence of quantization depends on the scale $\sigma$ (the area of the support of the quantization kernel). Thus the ideal detector is one that chooses $\hat g$ that is as far as possible from singularities in the locus $\{\hat g  \ | \nabla \psi(I, \hat g) = 0\}$. The selection of the best canonical frames according to this principle is described in Section \ref{sect-matching}. 

\cutThree{Note that a canonical frame $\hat g$ is often called a {\em ``feature point''} or {\em ``keypoint''} (or {\em ``corner''}), an inappropriate nomenclature unless $G$ is restricted to the translation group. {Note also that one should not confuse a (canonical reference) {\em frame} $\hat g$ from a {\em (video) frame}, which is an image $I_t$ that is part of a sequence $\{I_t\}_{t=1}^T$ obtained sequentially in time. Which ``frame'' we are referring to should be clear from the context.}}

Structural instabilities would be fatal if one insisted on {\em uniqueness} of the canonical element. However, as we have pointed out, occlusions prevent uniqueness in the first place; instead, we will have to select multiple canonical elements, each of which will have to be validated. Thus, in a sense, canonization is just a way of sampling the similarity group in a manner that is adapted to the signal, as opposed to being generic.

\section{Maximum stability and linear detectors}
\label{sect-detector-linear}
\index{Detector!linear}

In this section, as a way of example, we introduce detectors designed to be maximally BIBO stable. In other words, we look for classes of functionals $\psi$ that yield {\em maximally isolated critical points.} This seems to be an intuitive appealing criterion to follow. However, we will see in Section \ref{sect-correspondence} that this does not guarantee structural stability. Nevertheless, we describe this approach because it is one of the most common in the literature, and it is better than just testing for the transversality condition (\ref{eq-jacobian}), which is fragile in the sense that, in the presence of noise, it will almost always be satisfied. In Section \ref{sect-correspondence} we will introduce an approach to feature detection that is better suited to the notion of structural stability. Therefore, the reader can skip this section unless he or she is interested in understanding the relation between the approach proposed there and the ones used in the current literature.

\begin{footnotesize}
{\color{orange} To design a maximally BIBO-stable detector, we look for functionals that yield critical points where the Jacobian \index{Jacobian determinant} determinant is {\em not just non-zero}, but it is {\em largest}. Following the most common approaches in the literature, we try to capture this notion using differential operators.\footnote{An alternative approach not requiring differentiability is presented in Section \ref{sect-superpixels}.} In this case, we look for points that are at the same time zeros of $\nabla \psi\doteq \Psi$, and also critical points of $| \nabla \Psi |$. Now, in general, for a given class $\Psi$, there is no guarantee that the first set (zeros of $\Psi$) intersects the second (zeros of $\nabla | \nabla \Psi |$). However, we can look for classes of functions where this is the case, for all possible images $I$. These will be functionals that satisfy the following partial differential equation (PDE):
\begin{equation}
\Psi = \nabla | \nabla \Psi|.
\label{eq-PDE}
\end{equation}
Note that designing feature detectors by finding extrema of functionals requires continuity, which is something digital images do not possess. However, instead of finding extrema of the (discontinuous) image one can find extrema of operators, exploiting a duality as customary \cite{donohoC95,lindeberg98,romeny} especially for translation-invariance. {Note that the ``shift-plus-average'' in \cite{donohoC95} corresponds to template blurring, a lossy process that is not equivalent to proper marginalization or extremization.} 

We can start our quest for such functionals among {\em linear}\footnote{The advantage of linear functionals is that they commute with all additive nuisance operations, such as quantization and noise.} ones, \marginpar{\tiny \sc linear detectors} \index{Linear detector}\index{Detector!linear} \ie of the form $\Psi(I, \{x, \sigma\}) = {\cal G}*I(x, \sigma) = [0, 0, 0]^T$, where $*$ is a convolution product. \index{Convolution} In particular, ${\cal G}$ can be thought of as the gradient of a scalar kernel \index{Kernel} $k:\real^2\times \real^3 \rightarrow \real^+; (x, \{y,\sigma\}) \mapsto k(x-y; \sigma)$, acting \index{Convolution!kernel} on the image via a \index{Convolution!product} convolution $k * I(x,\sigma) \doteq \int k(x-y; \sigma) I(y)dy$, so the zeros of ${\cal G} * I$ are the critical points of $k * I$, so ${\cal G} \doteq \nabla k^T$:
\begin{equation}
\hat g = \{x, \sigma \} \ | \ \int(\nabla k^T(x-y; \sigma)I(y)dy \doteq {\cal G}(x,\sigma) * I = 0.
\end{equation}
Written in terms of $\cal G$, the PDE above becomes an integro-partial differential equation (I-PDE)
\begin{equation}
{\cal G}*I = \nabla | \nabla {\cal G} * I | 
\label{eq-IPDE}
\end{equation}
a condition that should be satisfied for all $I\in {\cal I}$. Note that the condition is only required at the zero-level set, so technically speaking a solution of the entire (\ref{eq-IPDE}) is not necessary. 
We can approximate a generic function $I$ with a linear combination of (basis) vectors $\{b_i(x)\}$, $I(x) = B(x) \alpha$, where $B(x) = [b_1(x), \dots, b_N(x)]$ are, for instance, a complete orthonormal basis, then the condition above has to be satisfied for all functions $b_i(x)$. Of the many choices of basis, we favor one that follows a result of Wiener \cite{wiener33}, that states that any positive $L^1$ distribution can be approximated arbitrarily well with a positive combination of Gaussian kernels. So, $b_i(x) = {\cal N}(x-x_i; \sigma_i)$ are Gaussian kernels centered at $x_i$ with standard deviation $\sigma_i$. Therefore, the condition above becomes
\begin{equation}
{\cal G}*{\cal N}(x,\sigma) = \nabla | \nabla {\cal G} * {\cal N}(x, \sigma) | \ ~~~ \forall \ (x, \sigma). 
\end{equation}
All the gradient operators can thus be transferred to the Gaussian kernel by linearity, and derivatives of Gaussian are products of Gaussians with Hermite polynomials, that form a complete orthonormal family in $L^2$. Using Jacobi's formula \index{Jacobi's formula} for the gradient of the determinant of a function with respect to its elements, we get
\begin{equation}
{\cal G}*{\cal N}(x,\sigma) = {\rm trace} \left( {\rm adj}({\cal G}*{\nabla \cal N}) {\cal G} * D^2 {\cal N} \right) 
\end{equation}
where $D^2$ denotes the Hessian matrix and $\rm adj$ is the adjugate matrix \index{Adjugate matrix} of co-factors (each element $i,j$ is the determinant of the minor obtained by removing row $i$ and column $j$). This is another way of looking at (\ref{eq-PDE}) and (\ref{eq-IPDE}), but now as an ordinary (non-linear) functional equation in the unknown $\cal G$.
}
\cut{{\tt For the separable case, this reduces to 
\begin{equation}
\int \exp\left(\frac{(y-x)^2}{\sigma^2}\right)\left(1 + \frac{2}{\sigma^2} - \frac{4 (x-y)^2}{\sigma^4}\right){\cal G}(y)dy = 0
\end{equation}
which implicitly defines ${\cal G} = {\cal G}(y; x, \sigma)$. It can be verified that any odd-order derivative of the Gaussian kernel satisfies the equation (this follows from the properties of Hermite polynomials):
\begin{equation}
{\cal G}(y; x, \sigma) = \nabla {\cal N}(x-y; \sigma).
\end{equation}
So, the set of canonizing features is the set of all odd-derivatives computed at all scales, at all locations. 
{\tt There is a question as to what kind of detectors arise if one chooses different bases, for instance sinusoids or Gabors.  Also, are there other linear canonization mechanisms?}
}}

The equation (\ref{eq-PDE}) is related to \index{Monge-Ampere equation} Monge-Ampere's equation in optimal transport; \index{Optimal transport} an approximation of the determinant of the Hessian has been used for a Gaussian Kernel in approximate form (using Haar wavelet bases and the Integral Image) in the SURF detector \cite{surf}. \marginpar{\tiny \sc monge-ampere equation}\index{Monge-Ampere equation}
\end{footnotesize}



\marginpar{\tiny \sc harris' corner}
\index{Harris' corner}
\begin{example}[Harris' corner revisited]
\label{ex-harris2}
The discussion above legitimizes the use of Hessian-of-Gaussian operators for feature detection, for instance \cite{surf}. The Laplacian-of-Gaussian operator, used in the popular SIFT \cite{lowe99object}, can also be partially justified on the grounds of first-order approximation of the Hessian, as customary in Newton methods. The other popularly used detector, Harris' corner detector, however, is not explained by the derivation above. In fact, note that Harris' operator
\begin{equation}
H(I,0) \doteq | \int \nabla^T I \nabla(I) dx| - \lambda {\rm Trace}\left( \int \nabla^T I \nabla(I) dx \right) 
\end{equation}
is not a linear functional of the image. It is still possible, however, to define the canonizing operator
\begin{equation}
\psi(I,g) =  H(I,g)
\end{equation}
and verify whether it yields isolated critical points. This is laborious and not relevant in the context of our discussion. We only mention that, because $H$ is {\em non-linear,} it does not commute with quantization, and therefore {\em it does not meet the conditions of a proper co-variant detector} (it is not commutative). Instead, below we suggest a different procedure to detect corners (or, more in general, junctions) that also provides a canonization procedure.
\end{example}

\cut{Derivatives of Gaussian are Gaussian modulated by Hermite polynomials. Note functions that violate condition $|\nabla G * I| \neq 0$ are ruled surfaces; these have degenerate singularities if they are smooth. However, they can have isolated extrema at non-differentiable points (\eg tip of a cone). However, the regularized derivative prevents this from happening, so the tip of a cone would come out as a proper isolated extremum (show example) that would not be picked up by Harris?? but yes by LoG?.}

\section{Non-linear detectors and the segmentation tree}
\label{sect-superpixels}
\index{Superpixel}
\index{Superpixel tree}
\index{Segmentation}
\index{Segmentation tree}

\begin{figure}[htb]
\begin{center}
\includegraphics[width=.3\textwidth]{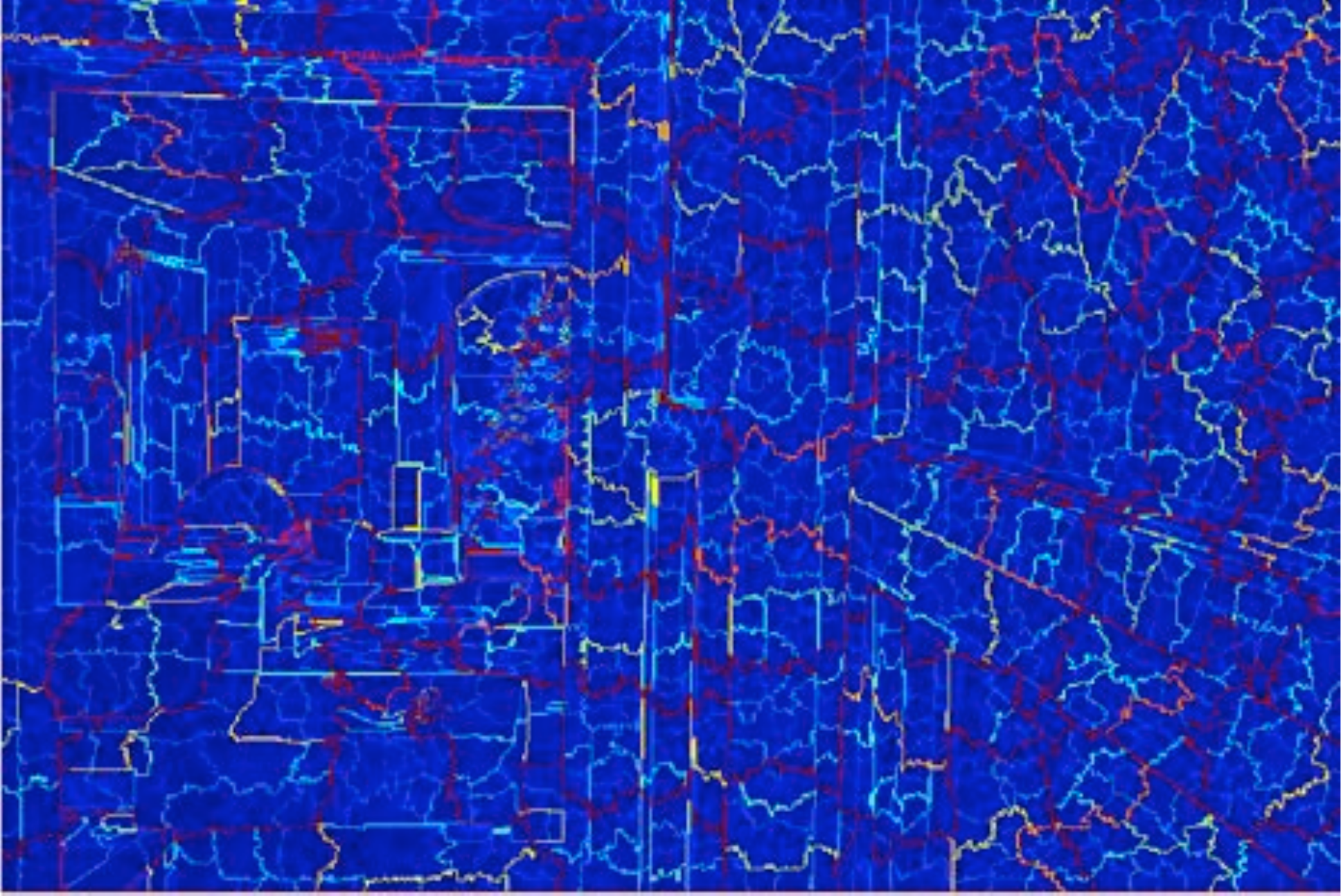}\\\includegraphics[height=3cm,width=.3\textwidth]{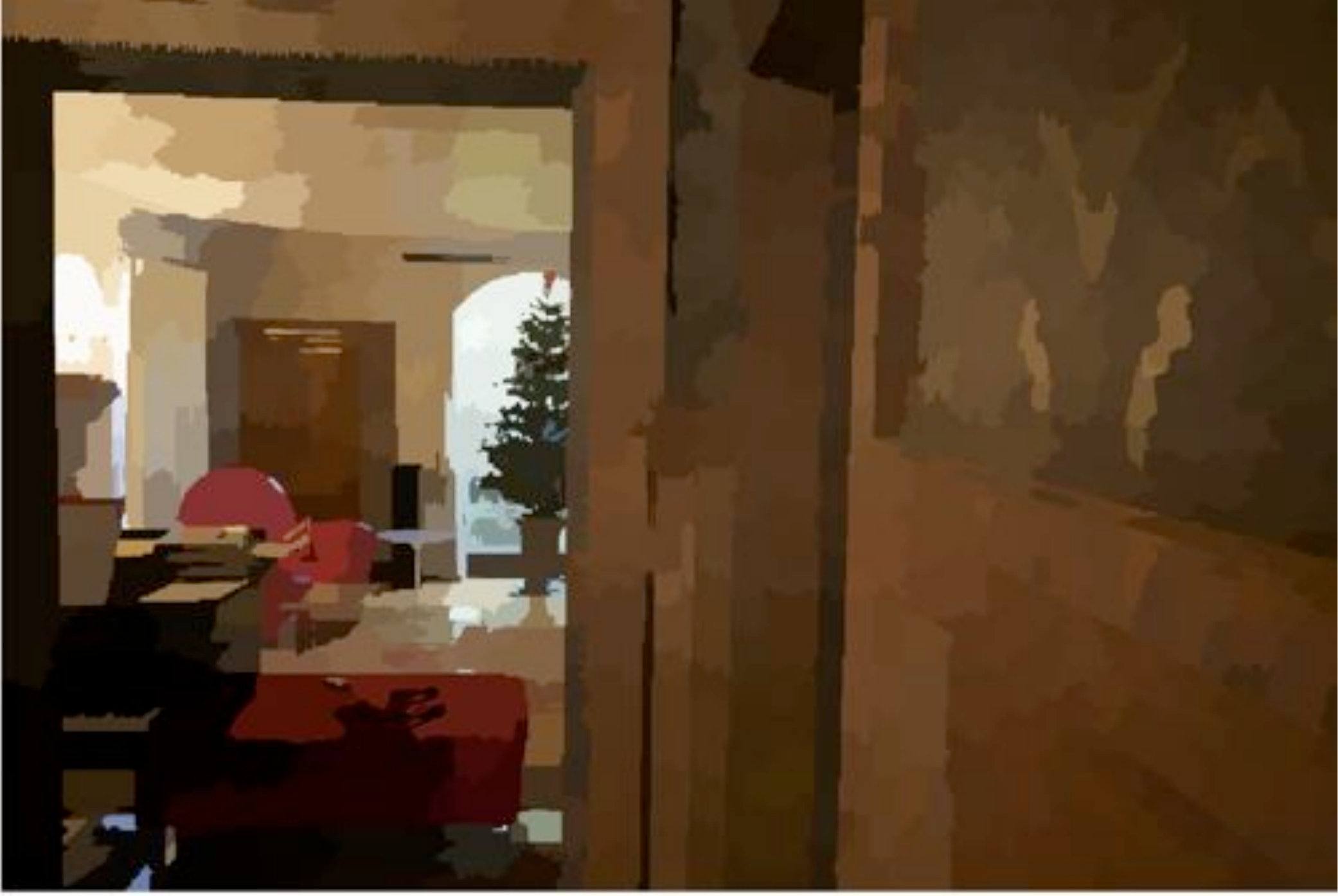}~\includegraphics[height=3cm,width=.3\textwidth]{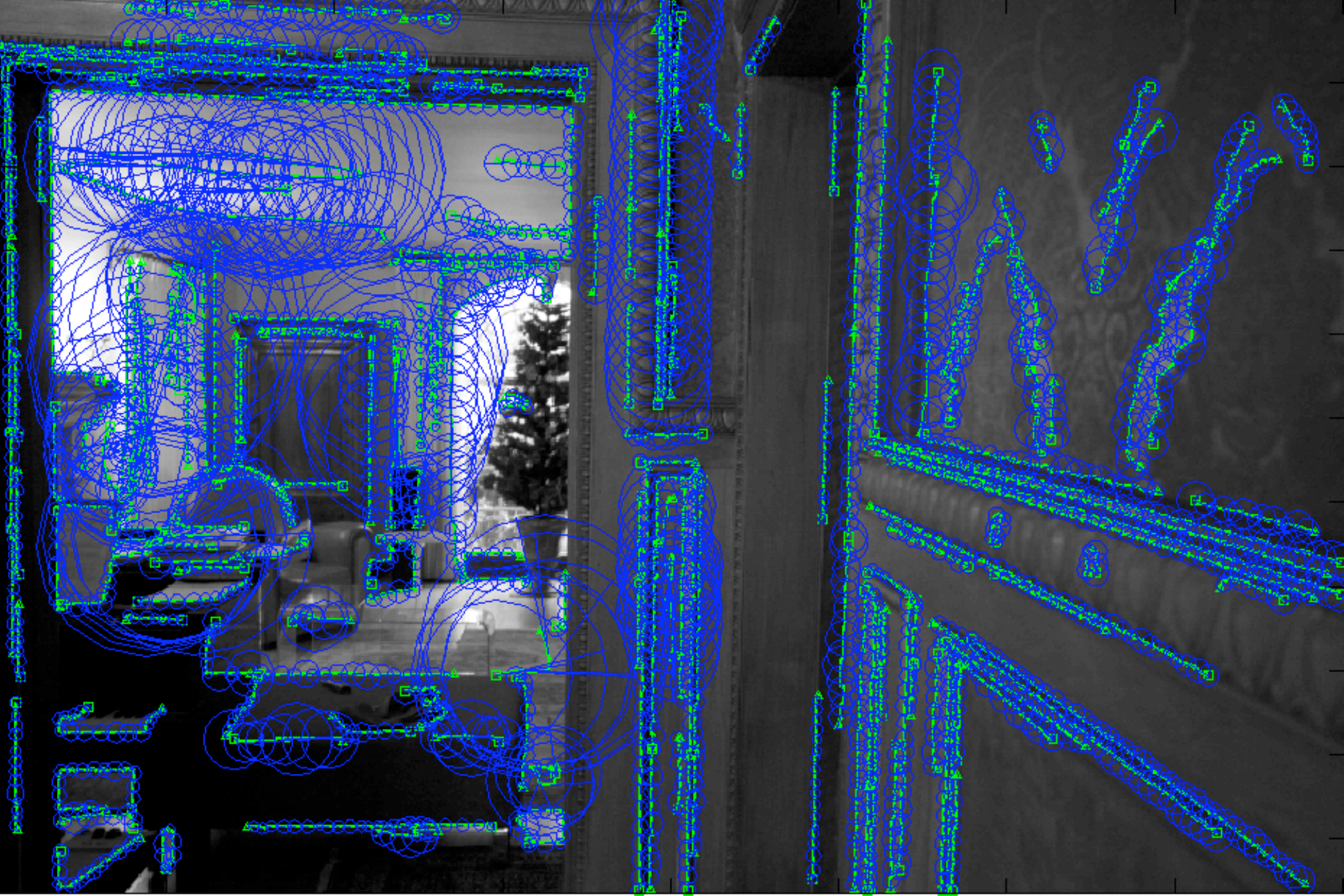}\\\includegraphics[height=3cm,width=.3\textwidth]{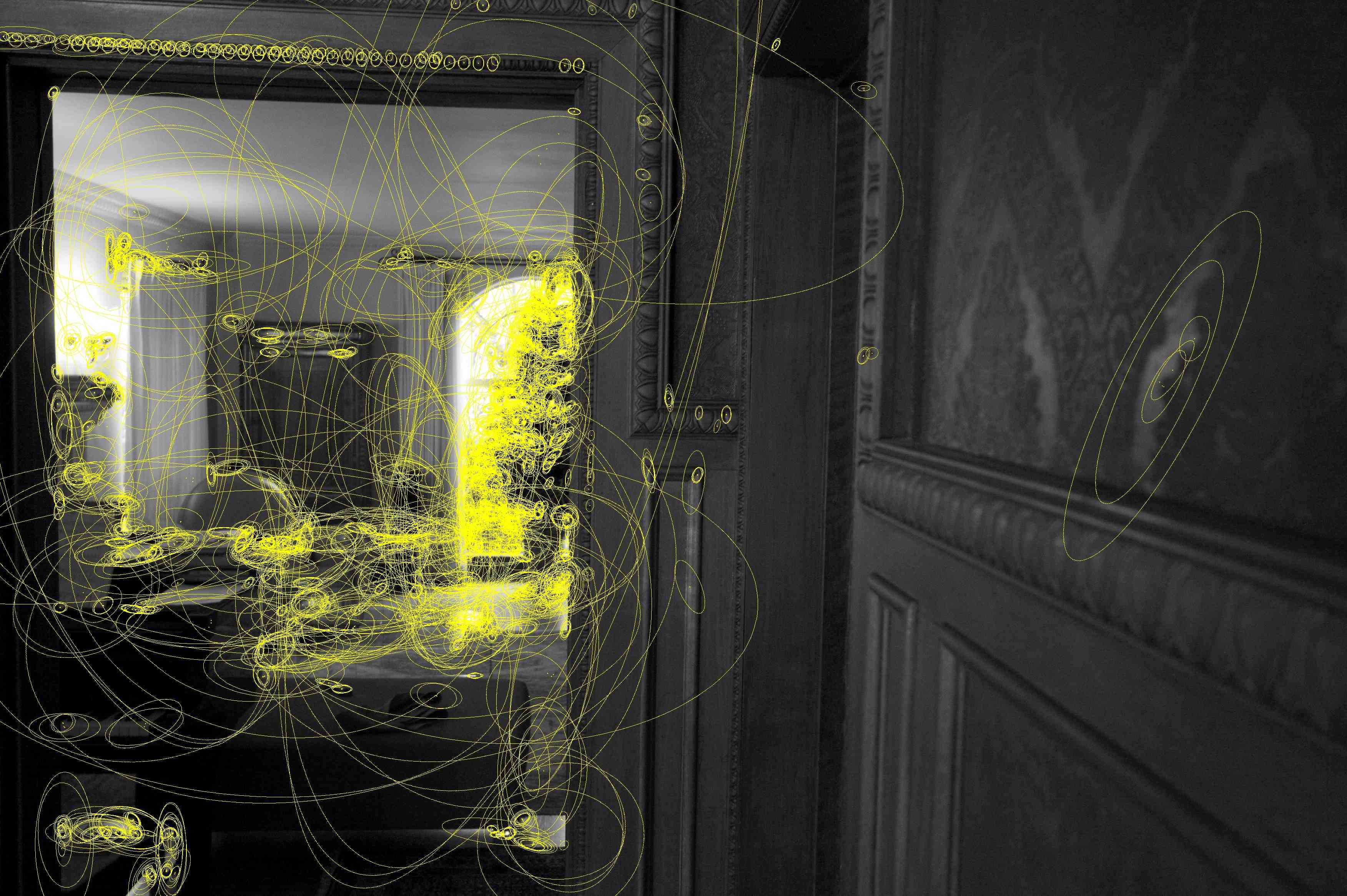}~\includegraphics[height=3cm,width=.3\textwidth]{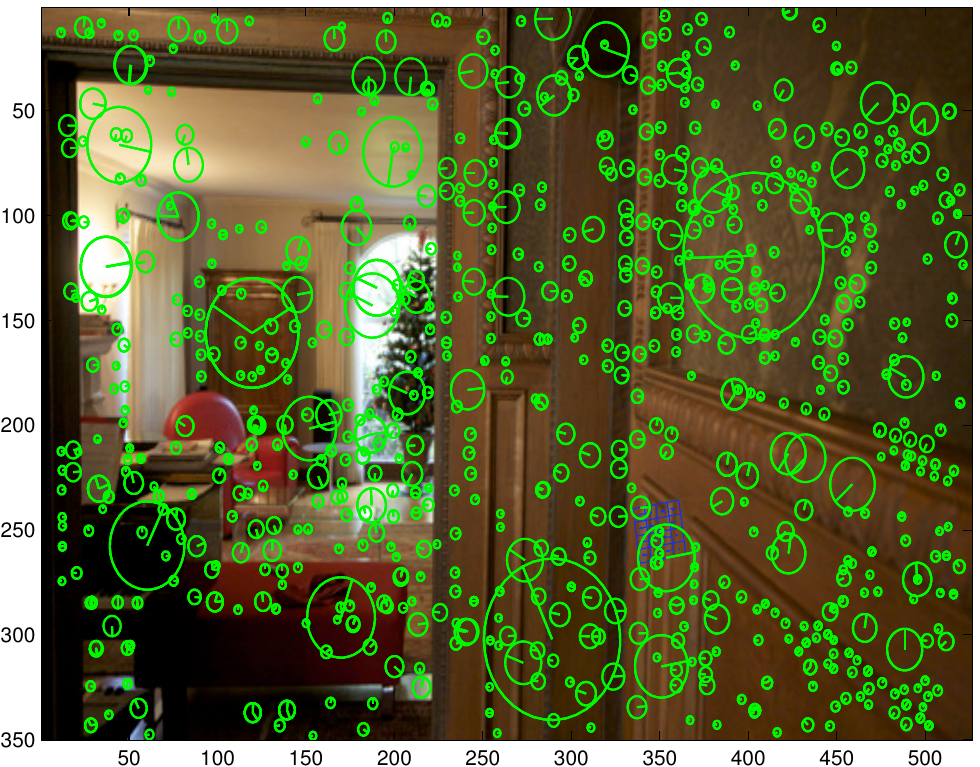}
\end{center}
\caption{\sl {\bf Representational structures:} Superpixel tree (top), dimension-two structures (color/texture regions), dimension-one structures (edges, ridges), dimension-zero structures (Harris junctions, Difference-of-Gaussian blobs). Structures are computed at all scales, and a representative subset of (multiple) scales are selected based on the local extrema of their respective detector operators (scale is color-coded in the top figure, red=coarse, blue=fine).\cutThree{ Only a fraction of the structures detected are visualized, for clarity purposes. All structures are supported on the Representational Graph, described in the next figure.}}
\label{fig-repr}
\end{figure}

Linear functionals are {\em not} the only feature detectors. Indeed, Theorem \ref{thm-1} establishes the link between feature detection and (generalized) texture segmentation. Therefore, rather than testing for canonizability (as done customarily in feature detection) one can test for stationarity (as done customarily in segmentation) and then construct features from the segmentation tree. The caveat is that, because of the interplay of the scale group with quantization, and of the translation group with occlusion, {\em no single segmentation} can be used as a viable canonization procedure, and instead the entire {\em segmentation tree} must be considered.


\begin{figure}[h!]
\begin{center}
\includegraphics[height=3cm,width=.3\textwidth]{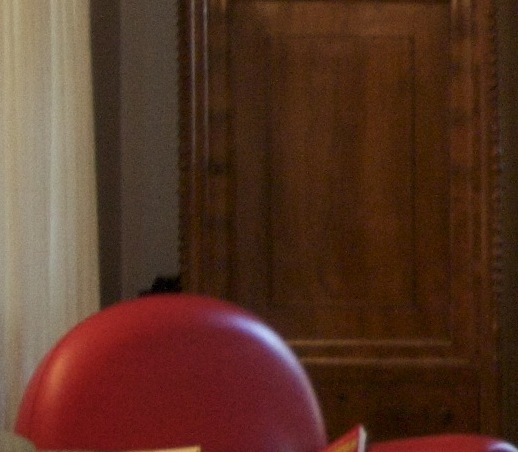}~\includegraphics[height=3cm,width=.3\textwidth]{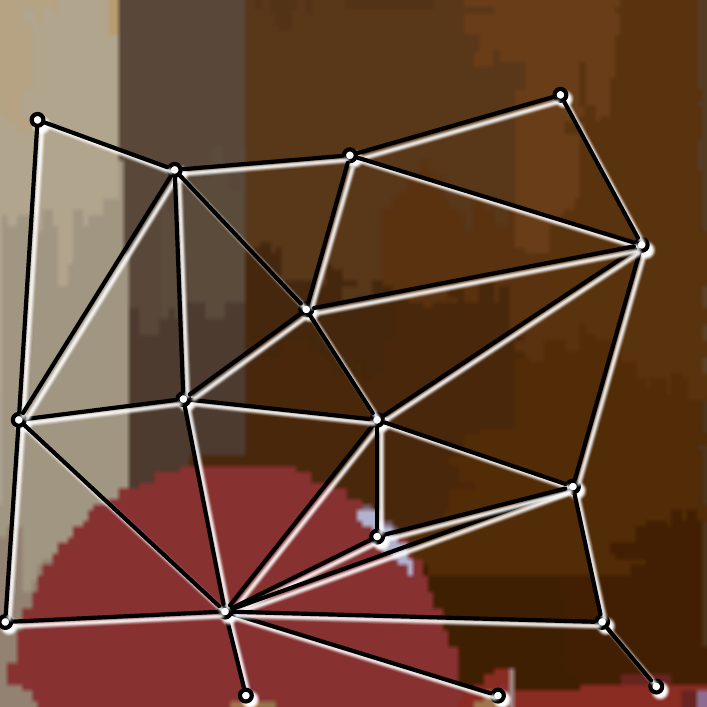}\\\includegraphics[height=3cm,width=.3\textwidth]{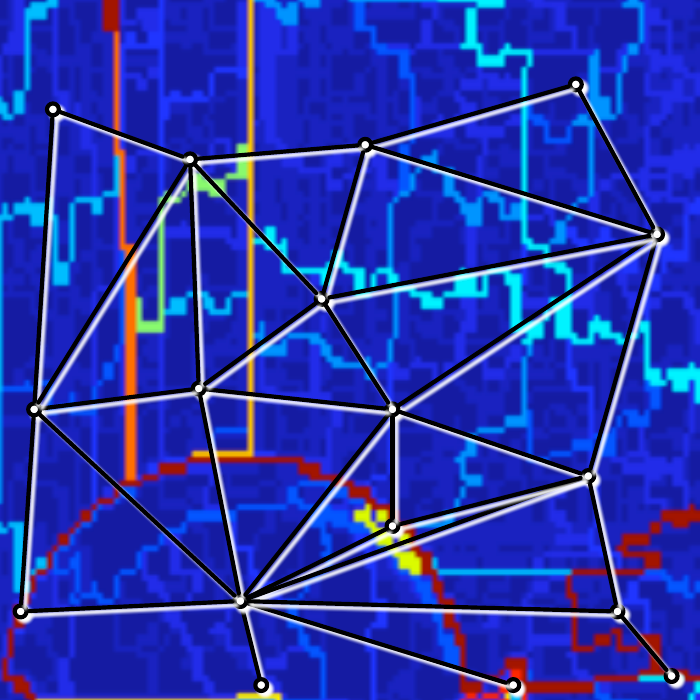}~\includegraphics[height=3cm,width=.3\textwidth]{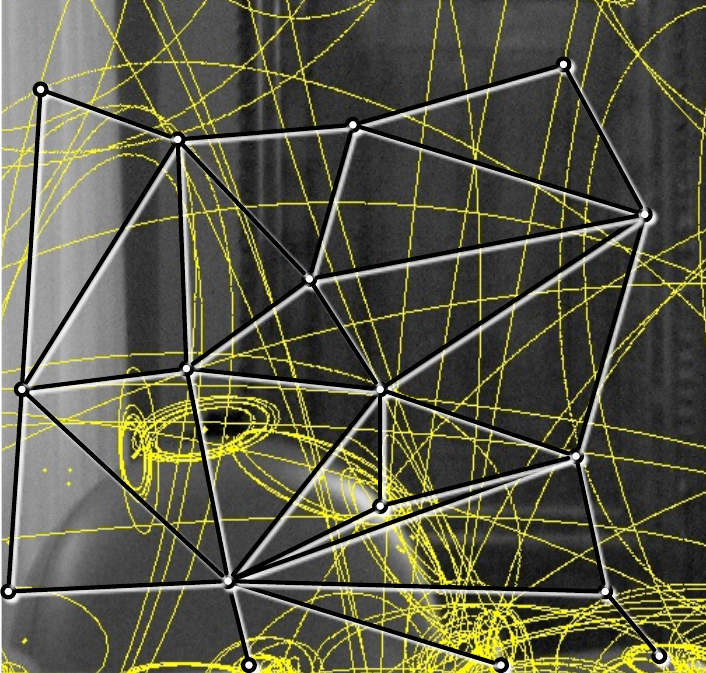}
\end{center}
\caption{\sl {\bf Representational Graph} (detail, top-left) Texture Adjacency Graph (TAG, top-right); {\bf nodes} encode (two-dimensional) region statistics\cutTwo{ (vector-quantized filter-response histograms),\cutOne{ or the ART of chromaticity within the region);}} pairs of nodes, represented by {\bf graph edges}, encode the likelihood computed by a multi-scale (one-dimensional) edge/ridge detector between two regions; pairs of edges and their closure ({\bf graph faces}) represent (zero-dimensional) attributed points (junctions, blobs).\cutThree{ For visualization purposes, the nodes are located at the centroid of the regions, and as a result the attributed point corresponding to a face may actually lie outside the face as visualized in the figure. This bears no consequence, as geometric information such as the location of point features is discounted in a viewpoint-invariant statistic.}}
\label{fig-RG}
\end{figure}
\begin{footnotesize}
{\color{orange} The starting point for this approach to canonization is a different approximation model of the image. Rather than a linear combination of globally-defined basis vectors such as Gaussians or sinusoids, we use simple functions. These are constant functions on a compact domain, whose universal approximation properties in several measures are guaranteed by Weierstrass' theorem \cite{rudin73}. In particular, let $\sigma > 0$ be a given ``scale.'' Then, given an image, we can find a partition of the domain $D$ and constant values such that a combination of simple functions approximates the original image (or any statistic computed on it) to within $\sigma$ in each region (all filter channels can then be combined into a vector description of a region). {\color{pea} Specifically, for a given $I \in {\cal I}$ and $\sigma$, we assume there is $N$ and constants $\{\alpha_1, \dots, \alpha_N\}$ and a partition of the domain $\{S_1, \dots, S_N\}$ such that
\begin{equation}
|I(x) - \alpha_j| \le \sigma \ \forall \ x\in S_j; ~~~ S_i \cap S_j = \delta_{ij}; \ \cup_{j=1}^N S_j = D,
\end{equation}
where $\delta_{ij}$ is Kronecker's Delta. Then, if we define the simple functions as characteristic functions of $S_j$
\begin{equation}
\chi_{S_j}(x) =  \begin{cases}
1, \ \forall \ x\in S_j \\
0 \ {\rm otherwise}
\end{cases}
\end{equation}
we can approximate the image with
\begin{equation}
\hat I(x) = \sum_{j=1}^N \chi_{S_j}(x) \alpha_j.
\end{equation}
This is true for scalar-valued images, but a similar construction can
be followed to partition the domain into regions (often called
{\em superpixels}), \marginpar{\tiny \sc superpixels} based on $\sigma$-constancy of any other \index{Superpixel} 
statistic, such as color or any other higher-dimensional feature $\phi$. In any case, a superpixelization algorithm can be thought of as a quantizer, \marginpar{\tiny \sc quantizer} \index{Quantizer} that is an operator that takes an image $I$ and a parameter $\sigma$ and returns a family of domain partitions $N = N(\sigma), \{\hat S_j\}_{j=1}^N$ with
\begin{equation}
\{\hat S_j\}_{j=1}^N = \phi_\sigma(I); ~~~ \hat \alpha_j = \frac{1}{|\hat S_j|} \int I(x)dx
\end{equation}
where $|S_j|$ is the area of $S_j$. We can now use this functional to determine a co-variant detector $\psi(I,g)$. For the case of translation, $g = T$, we define (multiple) canonical elements $T_j$ to be the centroids of the regions $S_j$, $\hat T_j = \frac{1}{|S_j|}\int_{S_j} xdx$. Making the dependency on the image explicit, we have 
\begin{equation}
\hat T(I, \sigma) = \frac{\int_{\phi_\sigma(I)} x dx}{\int_{\phi_\sigma(I)} dx}
\end{equation}
to which there corresponds the canonizing functional
\begin{equation}
\psi(I,\{T,\sigma\}) = \int_{\phi_\sigma(I)} x dx - T \int_{\phi_\sigma(I)}dx.
\label{eq-cov-superpix}
\end{equation}
It can be easily verified that this functional is co-variant. For the case of translation (fixing $\sigma$), $g = T$, we have that, for $\hat g$ that solves $\psi(I, \hat g) = 0$, and for any $g$
\begin{eqnarray}
\psi(I\circ g,\hat g \circ g) &=& \int_{\phi(I\circ g)} xdx - g \hat g \int_{\phi(I\circ g)} dx =  \int_{g \phi(I)} xdx - g \hat g \int_{\phi(I)} dx =  \nonumber \\
&=& \int_{\phi(I)} g x'dx' - g \hat g \int_{\phi(I)} dx = g\left( \int_{\phi(I)}  x'dx' -  \hat g \int_{\phi(I)} dx \right) = \nonumber \\
&=& g \psi(I, \hat g) = 0
\label{eq-covariant-nonlinear}
\end{eqnarray}
where we have used the fact that the group $g$ is isometric, so $dx = dx'$.} 

One may also believe that this functional yields isolated extrema, based on the fact that
\begin{equation}
| \nabla \psi| = |\frac{\partial \psi}{\partial T}| = \int_{\psi_\sigma(I)}dx > 0.
\label{eq-covariant-nonsingular}
\end{equation}
However, this result, as well as (\ref{eq-covariant-nonlinear}), is misleading because it assumes that the superpixelization $\phi_\sigma(I)$ is independent of (small variations in) $T$. More precisely, in order for translation to be canonizable, the canonization process {\em has to commute with quantization.} If we assume that the underlying ``ideal image'' \index{Ideal image} $I(x), \ x\in D$ is continuous, then the ``discrete'' (quantized) image $\hat I(x_i) = \int_{{\cal B}_\epsilon(x_i)} I(x)dx/\epsilon, x_i \in \Lambda$ defined on the lattice $\Lambda$, is related to it via the mean-value theorem, that guarantees the existence, for each $x_i$, of a translation $\delta_i$ such that 
\begin{equation}
\hat I(x_i) \doteq \frac{1}{|{\cal B}_\epsilon|} \int_{{\cal B}_\epsilon(x_i)} I(x)dx = I(x_i + \delta_i) = I(x_i) + n_i \doteq I \circ \nu 
\label{eq-superpix-stable}
\end{equation}
where $\nu$ denotes the quantization nuisance. For the canonization process to be viable we must have
\begin{equation}
\phi_\sigma(I) = \phi_\sigma(I\circ \nu)
\end{equation}
which is clearly not the case in general for a superpixelization algorithm. In fact, if we apply small perturbations to the levels $n_i$, from (\ref{eq-superpix-stable}) we get small perturbations in the location of the boundaries $\partial S_j$, and in the location of their centroid, $T_j \rightarrow T_j + \delta_i$. Since $n_i = \nabla I(x_i)\delta_i$, we see that for a superpixelization procedure $\phi_\sigma$ to provide a viable canonization mechanism, it has to place the boundaries in such a way that $\frac{\partial I}{\partial x}$ is negligible within each region, and as large as possible at region boundaries. We will see this spelled out more in detail shortly. The sensitivity of the boundary location as a function of a perturbation can be phrased in terms of BIBO stability, introduced in Definition \ref{def-bibo}.  Specifically, if $\phi_\sigma$ is a quantization/superpixelization operator acting on an $\epsilon$-quantized image (\ref{eq-superpix-stable}), and $\{S_j\}_{j=1}^N\doteq \phi_\sigma(I)$ and $\{\tilde S_j\}_{j=1}^N \doteq \phi_\sigma(I+n)$ the corresponding partitions, then $\phi_\sigma$ is BIBO stable if, according to (\ref{eq-BIBO}), 
\begin{equation}
\| n \| \le \sigma \Rightarrow | \tilde S_j - S_j | \le \epsilon
\end{equation}
where $| S_1 - S_2|$ denotes the area of the set-symmetric difference of the two sets $S_1$ and $S_2$.

Note that, in general, the operator $\phi_\sigma$ is not continuous, since it depends on $N$. When $N$ is fixed, it is possible to design a procedure that implements an operator $\phi_\sigma$ that is guaranteed to be stable even if the image is not continuous (but piece-wise differentiable). For instance, in a variational multi-phase region-based segmentation approach, one has an energy functional $E(I)$ that is continuous and minimized with respect to the (infinite-dimensional) partition $\{S_j\}$, so in this case we have that $\phi_\sigma$ is defined implicitly by the first-order optimality conditions (Euler-Lagrange) $\delta E(I) =0$, yielding a partial differential equation \cite{sundaramoorthiSY10}.

More in general, consider a perturbation of the image $\tilde I(x) = I(x) + n(x)$, with $n(x)$ assumed to be small in some norm. Then after quantization of the domain, using the Mean Value Theorem, we have
\begin{equation}
\tilde I(x_i) = \int_{{\cal B}_\epsilon(x_i)} \tilde I(x)dx = \int_{{\cal B}_\epsilon(x_i)} I(x)dx + \int_{{\cal B}_\epsilon(x_i)} \nabla I(x) dx \delta_i
\end{equation}
where we have assumed that $\delta(x)$ is constant within each quantization region ${\cal B}_\epsilon(x_i)$. Now, if we approximate the piecewise constant function in each ${\cal B}_\epsilon(x_i)$ into a piecewise constant function in the partition $\{S_j\}_{j=1}^N$, for a given $N$, we have
\begin{equation}
\sum_{j=1}^N \chi_{\tilde S_j}(x_i)\tilde \alpha_j = 
\sum_{j=1}^N \chi_{S_j}(x_i)\alpha_j + \sum_{j=1}^N \chi_{S_j}(x_i) \int_{{\cal B}_\epsilon(x_i)} \nabla I(x)dx \delta(x_i)
\end{equation}
from which we can obtain, defining $\delta S_j = \tilde S_j - S_j$ the set-symmetric difference between corresponding regions, and measuring the area in each region, 
\begin{equation}
\sum_{j=1}^N \sum_{x_i\in S_j} \chi_{\delta S_j}(x_i) \alpha_j = 
\sum_{j=1}^N \int_{S_j} \nabla I(x)dx \delta_j
\end{equation}
where we have now assumed that $\delta(x_i)$ is constant within each $x_i \in S_j$, and we have called that constant $\delta_j$. So, we have that for a partitioning to be BIBO Stable, we must have
\begin{equation}
\sum_{j=1}^N | \delta S_j | \le \sum_{j=1}^N \int_{S_j} \| \nabla I(x)\| dx
\end{equation}
which is guaranteed so long as the image is smooth within each region $S_j$ (but it can be discontinuous across the boundary $\partial S_j$). Indeed, any reasonable segmentation procedure would attempt, {\em for any given (fixed) $N$}, to place the discontinuities of the (true underlying) image $I$ at the boundaries $\partial S_j$, therefore guaranteeing stability of the boundaries with respect to small perturbations of the image per the argument above. In particular, for $N=2$, there are algorithms that guarantee a globally optimal solution \cite{chan} that is, by construction, the most stable with respect to small perturbations of the image.
}
\end{footnotesize}
These results are summarized into the following statement.
\begin{theorem}[BIBO Stability of the segmentation tree]
A quantization/superpixelization operator, acting on a piecewise smooth underlying field and subject to additive noise, is BIBO stable at a fixed scale for a fixed tolerance $\delta$ 
(or complexity level $N$) if and only if it places the boundary of the quantized regions/superpixels at the discontinuities of the underlying field.
\end{theorem}
However, our concern is not just stability with respect to the partition $\{S_j\}_{j=1}^N$ {\em for a fixed $N$}, but also stability with respect to {\em singular perturbations} \cite{dai89} that change \marginpar{\tiny \sc singular perturbations} the cardinality of the partition (a phenomenon linked to scale since $N = N(\sigma)$). So, the superpixelization procedure should be designed to be stable with respect to $N$, which is not a test we can write in terms of differential operations on the image. However, one can construct a greedy method that is designed to be stable with respect to both regular (bounded) and singular perturbations.
\begin{small}
\begin{enumerate}
\item Start at level $k=0$ with each pixel representing as a region, $S_i(0) = {\cal B}_\epsilon(x_i)$ with $i=1, \dots, N_0 = \# \Lambda$, the number of pixels in the image.
\item Construct a tree by creating a node representing the merging of the two regions that have the smallest average gradient along their shared boundary: for each $k$, obtain a new merged region $S_i(k+1) = S_{i_1}(k) \cap S_{i_1}(k)$ where 
\begin{multline}i_1, i_2 = \arg\min_{l,m \ | \ S_i(j) \subset S_l(k) \cup S_m(k) } E_{l,m}(k) \doteq \\ \doteq \int_{\partial S_l(k) \cap \partial S_m(k)} \| \nabla I (x)\| dx/\int_{\partial S_l(k) \cap \partial S_m(k)} ds
\end{multline}
\item Continue until the gradient between regions being merged fails the test (\ref{eq-BIBO}), that is 
\begin{equation}
\hat k, \hat l  = \arg\min_{l,m} E_{l,m}(k)  \ge \sigma. 
\label{eq-min-cost}
\end{equation}
\item At the end of the procedure, we have $N = N(\sigma) = N_0 -k$ regions.

These regions are, by construction, BIBO stable for any given value of $N$. Now to obtain a partition that is also stable with respect to singular perturbations, we have to choose regions that are unaffected by changes in $N$. To this end:
\item Consider each pixel $S_i(0)$, and follow its path in the tree, $\{S_i(k)\}$ as it is merged with other regions, together with the cost of each merging. Note that the cost is zero except when a merging occurs; call the mergings $\{ k_{i_1}, k_{i_2}, \dots \}$.  If at a certain $k_i$ we have that $S_i(k_i) = S_{l}(k_i) \cap S_m(k_i)$, then the cost is $E_i(k_i) \doteq E_{l,m}(k_i)$ given by (\ref{eq-min-cost}). 
\item Let $\{k_1(i), k_2(i), \dots \}$ be the instances when the region $S_i$ is merged with one of its neighbors, and $E_i(k_j(i))$ the corresponding costs. Furthermore, let $\delta k_j(i) \doteq k_{j+1}(i)-k_j(i)$ be the ``iteration gaps'' between two merges involving the region $S_i$.
\item For each pixel $i$ and each set of indices $\{k_{i_j}\}_{j=1}^L$ sort the gaps $\delta k_j(i) \doteq k_{i_j}-k_{i_{j-1}}$ in decreasing order of their minimum gap 
\begin{equation}
\min(\delta k_j(i), \delta k_{j+1}(i)).
\end{equation}
The level $\hat k$ with the largest minimum gap corresponds to the region $S_i(\hat k)$ containing the pixel $x_i$ that is least affected by singular perturbations. Indeed, for all $k_j(i) \le k \le k_{j+1}(i)$, any singular perturbation is a change in the value of $N = N_0-k$ that does not change the region $S_i(k)$.
\end{enumerate}
\end{small}
If we follow this procedure for every pixel $i$, sorting their paths in decreasing order of gaps, until a minimum gap $\sigma$ is reached, then we have that each initial region $S_i(0)$ is now included in a number of (overlapping) regions $S_i(k_j(i))$ that are not only BIBO stable, but also stable with respect to singular perturbations. This follows by construction and is summarized in the following statement.
\begin{theorem}[Stable segments]
Let $S_i(0)$, with $i=1, \dots, N_0$ be each pixel in an image. Then let $\{k_j(i)\}_{i,j}$ be ($j$-multiple) indices where $S_i$ is merged with neighboring regions, 
\begin{equation}
k_j(i)  = \arg\min_{l,m \ | S_i(k-1) \subset S_l(k-1)\cap S_m(k-1)} E_{l,m}(k) \le \sigma
\end{equation}
then the regions $\{S_i(k_j(i))\}_{i=1}^{N_0}$ are BIBO stable and stable with respect to singular perturbations.
\end{theorem}
Clearly many of these regions will be overlapping, and many indeed identical, so some heuristics can be devised to reduce the number of regions that, as it stands, can be more than the number of pixels in the image. A principle to guide such agglomeration is the so-called Agglomerative Information Bottleneck \cite{fulkersonVS08,tishbyPB00}.

This procedure relates to MSER \cite{matasCUP}, where however regions are created from the watershed, and are selected based on the variation of their boundary relative to their area as the watershed progresses. It also relates to other attempts to define ``stable segmentations'', for instance \cite{galleguillosBRB07}. However, in these cases stability is characterized empirically, and the algorithm is not guaranteed to be stable against a formal definition. Also, we are not necessarily interested in a partition of the domain, so long as a sufficient number of regions are present to enable local canonization.

So we have shown that canonizing translation via the centroid of superpixels is automatically viable, per (\ref{eq-covariant-nonlinear}) and (\ref{eq-covariant-nonsingular}), but only so long as the superpixelization is stable with respect to additive noise, for instance generated by quantization mechanisms.
\begin{theorem}[Canonization via superpixels]
Let $\phi_\sigma(I) = \{S_j\}_{j=1}^N$ be a partition of the domain into superpixels. Then the functional (\ref{eq-cov-superpix}) is a (local, translation) co-variant detector so long as it is BIBO stable.
\end{theorem}
That (\ref{eq-cov-superpix}) is covariant follows from (\ref{eq-covariant-nonlinear}), provided that (\ref{eq-BIBO}) is satisfied.

By the same token, one could canonize rotation by considering the principal axis of the sample covariance approximation of the regions $S_j$. Although \cite{soatto09} advocates canonizing general viewpoint changes by considering only the adjacency graph of the regions $S_j$, as we have discussed in Section \ref{sect-canonization}, one should canonize no further.

An alternative (and dual) procedure for canonization is to use {\em not} the centroid of the superpixels, but their {\em junctions}, which are the points of intersection of the boundaries of adjacent superpixels. This provides a {\em ``corner detection mechanism''} that is consistent with the theory, and can be used as an alternative to Harris' corner detector discussed in the previous section. It should be noted that critical point filters \cite{shinagawa08} have been proposed for the detection of junctions at multiple scales.

Detection and canonization are not important {\em per se}, and can be forgone if one is willing to marginalize nuisances at decision time. Simple (conservative) detection can be thought of as a way {\em not} to select canonizable regions, but to quickly reject obviously non-canonizable ones. In this, the approach can be related to Boosting \cite{friedman00aditive}. Where detection becomes important is when no marginalization can be performed, for instance in two adjacent frames in video, where one {\em knows} that the underlying scene is the same. As it turn out, this impoverished recognition problem, often called {\em tracking}, plays an important role in recognition, to the point where we devote a chapter to it, the next.

Before moving on, however, we would like to re-iterate the fact that the best mechanisms to eliminate nuisances is via marginalization or max-out. When decision-time constraints dictate that this is not feasible, canonization provides a useful way to reduce complexity, ideally keeping the expected risk in the overall decision problem unchanged (Section \ref{sect-dpi}).

\chapter{Correspondence}
\label{ch-correspondence}

In this chapter we explore the critical role that multiple images play in visual decisions. ``Correspondence'' refers to a particular case of visual decision problem, whereby two adjacent images are assumed to portray the same scene (owing to the physical constraints of causality and inertia, the scene is not likely to change abruptly from one image to the next), and this ``bit'' is used to prime the construction of models of the scene for recognition. Whether correspondence can be established depends on a variety of conditions that we describe next.

\begin{figure}[htb]
\begin{center}
\includegraphics[height=.5\textwidth,width=.3\textwidth]{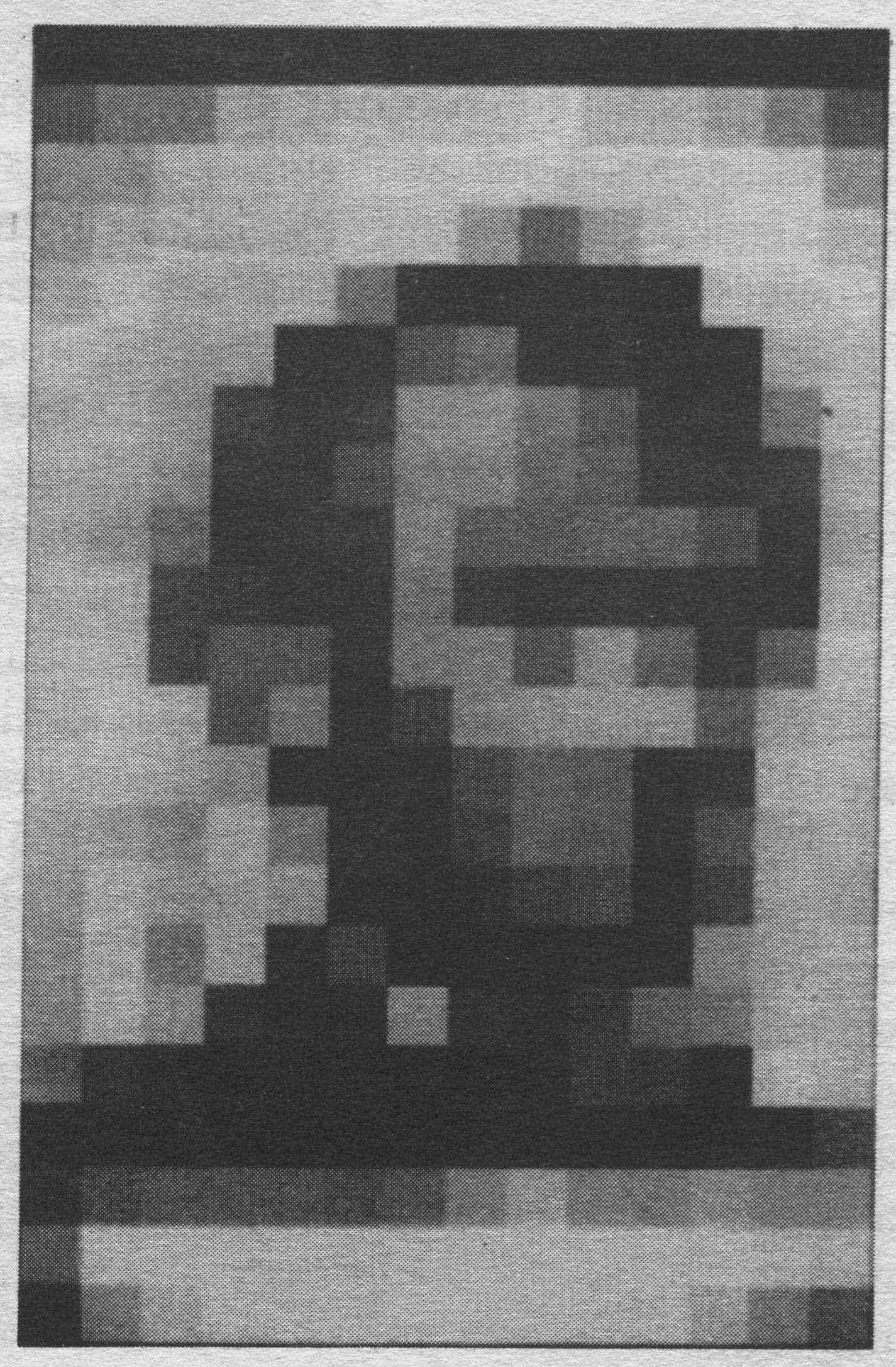}
\includegraphics[height=.5\textwidth,width=.3\textwidth]{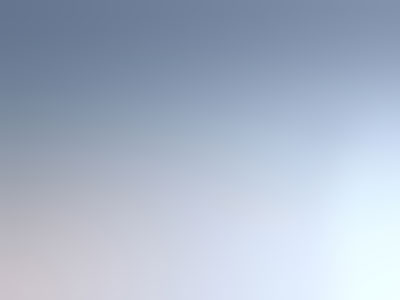}
\end{center}
\caption{\em A region of an image can be canonizable, but not properly sampled (left), or properly sampled, but not canonizable (right). Only when a region of the image is both properly sampled and canonizable can we establish {\em meaningful} correspondence between structures in the image and structures in the scene.}
\label{fig-X}
\end{figure}

In the next section we will introduce the notion of Proper Sampling, that has been anticipated in Chapter \ref{sect-canonization}, and is illustrated in Figure \ref{fig-X}: A signal can be canonizable, but not properly sampled (\ie, canonizability arises from {\em aliasing} effects, such as the block-structure due to quantization). Or, it can be properly sampled, but not canonizable (\eg a constant region of the image), it can be not canonizable, and not properly sampled (\eg a spatially and temporally independent realization of Gaussian ``noise''), and finally it can be properly sampled {\em and} canonizable, for instance a ``blob'' or the fine-scale structure in a random-dot stereogram (Figure \ref{fig-rds}), not to be confused with ``noise.''

In Chapter \ref{sect-canonization} we have seen that an invariant descriptor can be constructed via canonization, but canonizability is a necessary, not sufficient, condition for meaningful correspondence. In order to be meaningful, a structure on the image needs to {\em correspond} to a structure in the {\em scene}, as we discussed in Remark \ref{rem-proper-sampling}. Unfortunately, this condition cannot be tested on one image alone. It can either be determined at decision time, by marginalization or extremization, or it can be determined -- under the Lambertian assumption -- by looking at {\em different images of the same scene}, as we describe next.

\section{Proper Sampling}
\label{sect-correspondence}
\index{Correspondence}
\index{Proper sampling}
\index{Sampling!proper}

 \begin{figure}[htb]
 \begin{center}
 \includegraphics[width=.5\textwidth]{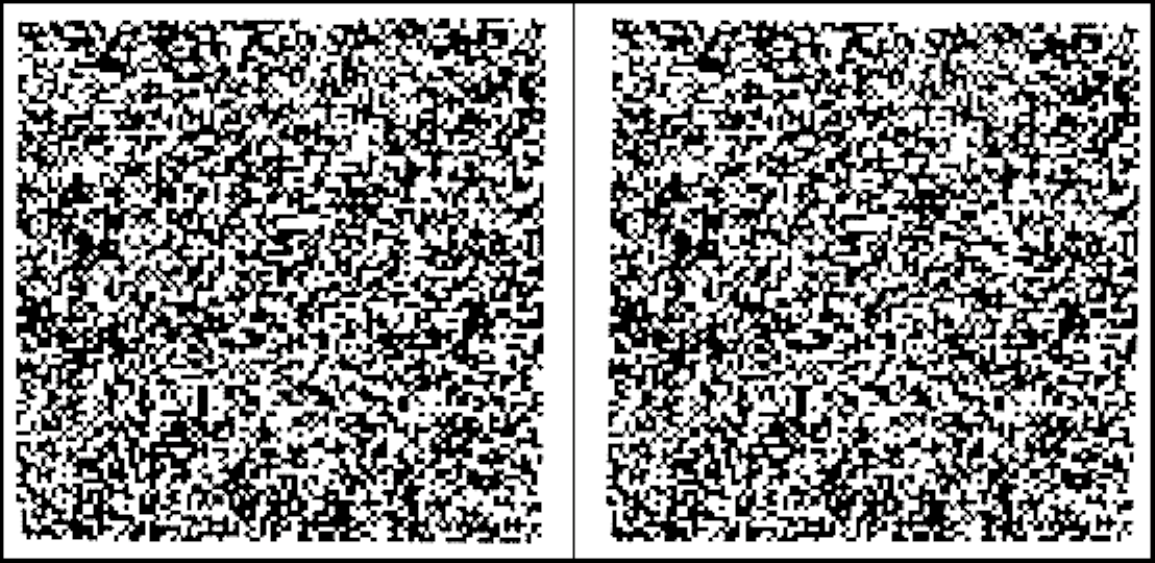}
 \end{center}
 \caption{\sl {\bf Texture or structure?} This image helps understanding the importance of the notion of ``proper sampling.'' Since it is possible to fuse these two images in the {\em random-dot stereogram} binocularly and get a dense depth map \cite{julesz71}, this means that point-to-point correspondence is possible, which would {\em not} be possible if this were a texture {\em at the pixel scale}, according to the definition in Section \ref{sect-texture}. However, the important question for correspondence is not whether something is a texture (stationary) or structure (canonizable), but whether it is {\em properly sampled}. This cannot be decided in a {\em single image}, but by testing the {\em proper sampling} hypothesis of Definition \ref{defn-proper-sampling}. In the specific case of the random-dot stereogram, this is a {\em properly sampled} pair of images, that just happen to have a very complex radiance. \index{Random-dot stereogram} Recall that proper sampling depends not just on the scene, but also on the imaging condition, including the distance from the scene and the point-spread function of the eye, both of which affect the scale of the representation. 
}
\label{fig-rds}
\end{figure}
\cut{\tt Consider the difference between parallax and occlusions: red-green (stereo) glasses generate the appearance of parallax, but as you move relative to the glasses, you can see things moving but no occlusions. Study phenomenon also in relation to 3D displays} 

Feature detection is a form of sampling, but one that is rather different from the classical sampling theory of Nyquist and Shannon.  In traditional signal processing, proper sampling refers to regular sampling at twice the Nyquist rate, since a band-limited signal can be reconstructed exactly from the samples under these conditions. This assumes that the signal is {\em band-limited,} \marginpar{\tiny \sc band-limited} \index{Band-limited} which is as much an idealization as the Lambert-Ambient static model (strictly band-limited signals do not exist, as they would have to have infinite spatial support).

But we are not interested in using the data to reconstruct a replica of the ``original signal.''  Instead, we are interested in using the data to solve a {\em decision or control} task as if we had the ``original signal'' at hand. But what is the ``original signal'' in our context? It is the image before any quantization phenomena occur. For a single image, since we cannot detect occlusions (Section \ref{sect-occlusion}), quantization represents the main non-invertible nuisance (neglecting complex illumination effects). Similarly, as we have seen in  Theorem \ref{thm-what-to-canonize}, the translation-scale group $g = \{x, \sigma\}$ is the main invertible nuisance (since more complex groups are not invertible once composed with quantization). Therefore, following  (\ref{eq-pi-1}) in Section \ref{sect-im-form}, we have that if the image is $I = h(g,\xi, \nu) + n$, then the ``ideal image'' is \index{Ideal image} 
\begin{equation}
h(e,\xi, 0) \doteq \rho(\pi^{-1}(D)).
\end{equation}
Therefore, in the context of visual decisions, we can define {\em proper sampling} as a discretization that yields a response of co-variant detector functionals having the same number, type and connectivity of critical points that the ``ideal image'' would produce.
\begin{defn}[Proper Sampling]
\label{defn-proper-sampling}
A signal $I$ is properly sampled at a scale $\sigma$ if a co-variant detector operating on the image yields the same critical points as if it operated on the radiance of the scene. {\color{orange} In other words, if $\psi(I,g)$ is a co-variant detector for the location-scale group $g = \{x, \sigma\}$, and $h(e,\xi, 0) \doteq \rho \circ \pi^{-1}(D)$, then $\nabla \psi(I,\hat g) = 0 \Leftrightarrow \nabla \psi(h(\hat g, \xi, 0)) = 0$, and corresponding extrema are of the same type.}
\end{defn}
\index{ART}
\index{Attributed Reeb Tree (ART)}
\marginpar{\tiny \sc attributed reeb tree (art)}
In other words, an image is properly sampled if it is possible to reconstruct, from the samples, {\em not} the ``ideal image'' \index{Ideal image} (the radiance of the scene), but one that is topologically equivalent to to it, in the sense of yielding the same extrema via a co-variant detector operator. The attributed Reeb tree (ART) is a topological construction that can be assembled from the image at any given scale. {\color{orange} It is a tree with nodes at every isolated extremum (maxima, minima, saddles), with edges connecting extrema that are ``adjacent'' in the sense of the Morse-Smale complex. The value of the image at the extrema is not relevant, but the ordering is, and the position of these extrema is not relevant, but the connectivity is.} The ART was introduced in \cite{sundaramoorthiPVS09} as a maximal contrast-viewpoint invariant away from occlusions, and is described in more detail in Section \ref{sect-topology}. It can be shown that the outcome $\hat g = \{\hat x_i, \sigma\}$ of any feature detector $\nabla \psi(I,\hat g) = 0$ operating at a scale $\sigma$ can be written in terms of the $ART$: $\{\hat x_i\}_{i=1}^N = ART(I* {\cal G}(x; \sigma^2))$, which leads to the following claim.
\begin{theorem}
A signal $I$ is properly sampled at $\sigma$ if and only if $ART(h(e, \xi, 0) * {\cal G}(x; \sigma^2)) = ART(I* {\cal G}(x; \sigma^2))$.
\end{theorem}
Thus, any of a number of efficient techniques for critical point detection \cite{shinagawaKK91} can be used to compute the $ART$ and test for proper sampling.
\cutThree{Note that, in general, any feature detection mechanism alters the position of the extrema relative to the original signal, as illustrated in Example \ref{example-gaussian}. This is not an issue in the context of visual decisions, because extrema move as a result of viewpoint changes, and their motion would therefore be discarded as a nuisance anyway.} 

{\em The problem with the definition of proper sampling is that it requires knowledge of the ``ideal image'' $h(e, \xi, 0)$,} \index{Ideal image} which is in general not available (indeed, it may be argued that it does not exist). \cutThree{Unlike classical sampling theory,\footnote{In reality, even in classical sampling theory there is no critical sampling, since no real-signal is strictly band-limited, so in strict terms Nyquist's frequency does not exist.} there is no ``critical frequency'' beyond which one is guaranteed success in canonization, because of the scaling/quantization phenomenon. }Therefore,\cut{ consistent with the theory of Actionable Information, } to test for proper sampling on the image $I_t(x)$, rather than comparing it to the ``ideal image''(which is a function of the unknown radiance distribution of the scene), we compare it to {\em the next image} $I_{t+1}(x)$, which is equivalent to it under the Lambertian assumption in the co-visible region \cite{soattoY02cvpr}, as done in Section \ref{sect-im-form}. \index{Co-visible region} Therefore, the notion of proper sampling (now in space {\em and} time) relates to the notion of correspondence, co-detection, or ``trackability'' as we explain below. 
\index{Proper sampling}
\index{Sampling!proper}
\begin{defn}[Proper Sampling]
We say that a signal $\{I_t\}_{t = 1}^T$ can be {\em properly sampled} at scale $\sigma_t$ at time $t$ if there exists a scale $\sigma_{t+1}$ such that $ART(I_t*{\cal G}(x; \sigma_t^2)) = ART(I_{t+1}*{\cal G}(x; \sigma_{t+1}^2))$. 
\end{defn}
In other words, a signal is properly sampled in space and time if the feature detection mechanism is topologically consistent in adjacent times. Alternatively, we can impose that $\sigma_t = \sigma_{t+1}$ and therefore require topological consistency {\em at the same scale}. Sometimes we refer to proper sampling of corresponding regions in two images as being {\em co-canonizable}.

\cutThree{Note that in the complete absence of motion, proper sampling cannot be ascertained. However, complete absence of motion is only real when one has {\em one} image, as a continuous capture device will always have some changes, for instance due to noise,  making two adjacent images different, and therefore the notion of topological consistency over time meaningful, since extrema due to noise will not be consistent. Note that the position of extrema will in general change due to {\em both} the feature detection mechanism, and also the inter-frame motion. Again, what matters in the context of visual decisions is the structural integrity (stability) of the detection process, \ie its topology, rather than the actual position (geometry). }If a catastrophic event happens between time $t$ and $t+1$, for instance the fact that an extremum at scale $\sigma$ splits or merges with other extrema, then tracking cannot be performed, and instead the entire $ART$s have to be compared across all scales in a complete graph matching problem. 

\begin{figure}[htb]
\begin{center}
\includegraphics[width=.6\textwidth]{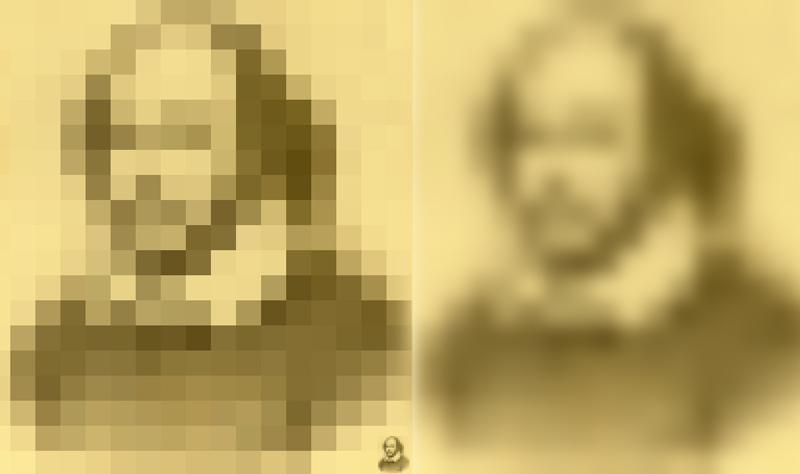}
\end{center}
\caption{\sl Whether an image is properly sampled depends on scale. The image on the left is not properly sampled at the finest scale afforded by the reproduction medium of this manuscript. However, it is properly sampled at a considerably coarser scale (right).}
\label{fig-shakeok}
\end{figure}

While establishing correspondence under these circumstances can certainly be done (witness the fact that we can recognize the top-left image in Figure \ref{fig-X} as a quantized photograph of Abraham Lincoln), this has to be done by marginalization, extremization or canonization, treating correspondence as a full-fledged recognition problem (a.k.a. {\em wide-baseline matching}) \marginpar{\tiny \sc wide-baseline matching} \index{Wide-baseline matching} rather than exploiting {\em knowledge that the underlying scene is the same.} This is the crucial ``bit'' provided by temporal continuity. Motivated by this reasoning, we introduce the notion of ``trackability.'' 
\index{Trackability}\marginpar{\tiny \sc trackability}
\index{Tracking}
\begin{defn}[Trackability]
A region of the image $I_{|_{\cal B}}$ is {\em trackable} at a given scale if it is canonizable and properly sampled at that scale.
\end{defn}
It may seem at first that any region of an image can be made trackable by considering it at a sufficiently coarse scale. Unfortunately, this is not the case. 
\index{Occlusion}
\begin{rem}[Occlusions]
\label{rem-occlusions}
We first note that occlusions do, in general, alter the topology of the feature detection mechanism, hence the $ART$. Therefore, they cannot be properly sampled at any scale. This is not surprising: We know that we cannot track through an occlusion, as correspondence cannot be established for regions that are visible in one image (either $I_t$ or $I_{t+1}$) but not the other. In the context of feature detectors/descriptors, occlusion detection reduces to a combinatorial matching process at decision time. Pixel-level occlusion detection will be discussed in Section \ref{sect-occlusion}.
\end{rem}
\begin{rem}
Note that a signal is not  properly sampled {\em per-se}. In general, only trivial signals are globally properly sampled. However, there is a scale at which a signal may be properly sampled, as we will see shortly.
As an illustration, Figure \ref{fig-shakeok} shows the same image that is {\em not} properly sampled at the pixel level (left) but it is properly sampled when seen at a significantly coarser scale (small in-set image on the left, or blurred version on the right).
\end{rem}
Note that the notion of proper sampling relates to texture, in the sense that even if two images independently exhibit stationary statistics (and therefore they are independently classified as textures), if they are {\em co-canonizable} they are properly sampled, and therefore point-correspondence can be established (Figure \ref{fig-rds}). Also, we recall that we consider constant-color regions to be (trivial) textures, even though they are often referred to as ``textureless.'' Note that a constant-intensity region is, in general, properly sampled, but not canonizable. A stochastic texture is not canonizable (by definition) at the native scale (sensor resolution), where it is usually not properly sampled, because canonizability arises from sampling/aliasing phenomena. Note, again, that all these considerations depend on {\em scale}, as we have discussed in Section \ref{sect-texture}. Also note that we are not {\em assuming} that the signals we measure are properly sampled. Instead, we use the definition to {\em test} the hypothesis of proper sampling at a given scale, and therefore determine co-visibility and ascertain trackability.

{\color{pea} Although, as we have pointed out before, two-dimensional scale-spaces do not enjoy a causality property, typically the number of extrema diminishes at coarser scales, and the extrema that persist are, usually, corresponding to some structure in the image.  So, although one cannot prove that for {\em any} image there exists a scale at which it can be properly sampled, in co-visible regions, one typically finds this to be the case for most natural images. This is because, at large enough scales, extrema will coalesce, and eventually quantization phenomena become negligible.\footnote{However, at coarser scales, the landscape of the response of a co-variant detector becomes flatter, and therefore the BIBO gain smaller, and feature detection less reliable.} 

Therefore, one can conjecture that, for most natural images, in the absence of occlusions, assuming continuity and a sufficiently slow motion relative to the temporal sampling frequency, {\em any image region is trackable.} 
This means that there exists a large-enough scale $\sigma_{max}$ such that $ART(I_t * {\cal G}(x; \sigma_{max}^2) = ART(I_{t+1} * {\cal G}(x; \sigma_{max}^2)$.}

Thus {\em anti-aliasing in space can lead to proper sampling in time}. This is important because, typically, temporal sampling is performed at a fixed rate, and we do not want to perform temporal anti-aliasing by artificially motion-blurring the images, as this would destroy spatial structures. {\color{pea} Note, however, that once a large enough scale is found, so correspondence is established at the scale $\sigma_{max}$, the motion ${\hat g}_t$ computed at that scale can be compensated for, and therefore the (back-warped) images $I_t \circ {\hat g}^{-1}_t$ can now be properly sampled at a scale $\sigma \le \sigma_{max}$. This procedure can be iterated, until a minimum $\sigma_{min}$ can be found beyond which there is no topological consistency.  Note that $\sigma_{min}$ may be smaller than the native resolution of the sensor, leading to a {\em super-resolution} phenomenon. \marginpar{\tiny \sc super-resolution} \index{Super-resolution} \cut{Note also that this makes sense in the explorative framework of Actionable Information \cite{soatto09}, where one can make $\sigma_{min}$ smaller and smaller by getting closer and closer to objects of interest in the scene.}}
\index{Scale selection}

This suggests a procedure for tracking, whereby one first selects structurally stable features via proper sampling. The structural stability margin  determines the neighborhood in the next image where detection is to be performed. If the procedure yields precisely one detection in this neighborhood, topology is preserved, and proper spatio-temporal sampling is achieved, hence trackability. Otherwise, a topological change has occurred, and the track is broken.\cutThree{ This procedure is performed first at the coarsest level, and then propagated at finer scales by compensating for the estimated motion, and then {\em re-selecting} at the finer scales \cite{leeS10}. Note that this procedure, described in more detail in the next section, is different from traditional multi-scale feature tracking, where each feature detected at the finest scale is tracked at all coarser scales. In this framework, feature detection is initiated at each scale, in the region back-warped from coarser scales.}

\cut{\begin{rem}[The Actionable Information Paradox]
\label{ex-paradox}
Consider a large dataset of images, of the kind common in modern benchmarks. Each image is usually represented by a collection of feature descriptors, each associated with the response of a co-variant detector. If the images are sampled independently from the class-conditional density (for instance, for the category ``house'' we have $N$ images of $N$ different houses), then we have no way of determining whether the co-variant frames are properly sampled. Therefore {\em we have to retain all descriptors} (or a compressed/quantized version thereof). However, if {\em multiple images of the same scene} are available (for instance, $K$ images for each of the $N$ house, for an overall dataset of size $KN$), then we can test the hypothesis that each frame be {\em properly sampled}, and only retain the descriptors that corresponds to properly sampled frames. These are a subset of the set of descriptors, that is smaller than the number of descriptors in each individual image. Therefore, having a larger training data actually results in a {\em smaller storage requirement}, and inevitably a higher performance of the resulting classifiers, since the aliased frames, that are not a function of a scene but of the nuisances in image formation, are discarded.
\end{rem}}

\section{The role of tracking in recognition}
\label{sect-matching}

The goal of tracking is to provide correspondence of (similarity or isometric) reference frames $\hat g_{ij} = \{x_i, \sigma_{ij}, R_{ij}\}$, centered at $x_i$, with size $\sigma_{ij}$ and orientation $R_{ij}$. This is the outcome of feature detection based on structural stability and proper sampling. Because of temporal continuity,\footnote{This temporal continuity of the class label does not prevent the data from being discontinuous as a function of time, owing for instance to occlusion phenomena. However, in general one can infer a description of the scene $\xi$, and of the nuisances $g, \nu$ from these continuous data, including occlusions \cite{duciYMS06,jinSY05IJCV}, as we show in Section \ref{sect-occlusion}. It this were not the case, that is if the scene and the nuisances cannot be inferred from the training data, then the dependency on nuisances cannot be learned.} the class label $c$ is constant under the assumption of co-visibility and false otherwise. In this sense, time acts as a ``supervisor'' or a ``labeling device'' that provides ground-truth training data. The local frames $g_k$ are now effectively {\em co-detected} in adjacent images. Therefore, the notion of structural stability and ``sufficient separation'' of extrema depends not just on the {\em spatial scale}, but also on the {\em temporal scale}. For instance, if two $5$-pixel blobs are separated by $10$ pixels in one image, they are not sufficiently separated for tracking under temporal sampling that yields a $20$-pixel inter-frame displacement. 
\index{Temporal scale}
\index{Spatial scale}
\index{Scale!temporal}
\index{Scale!spatial}

\cut{\begin{figure}[htb]
\begin{center}
\includegraphics[width=.2\textwidth,height=.3\textwidth]{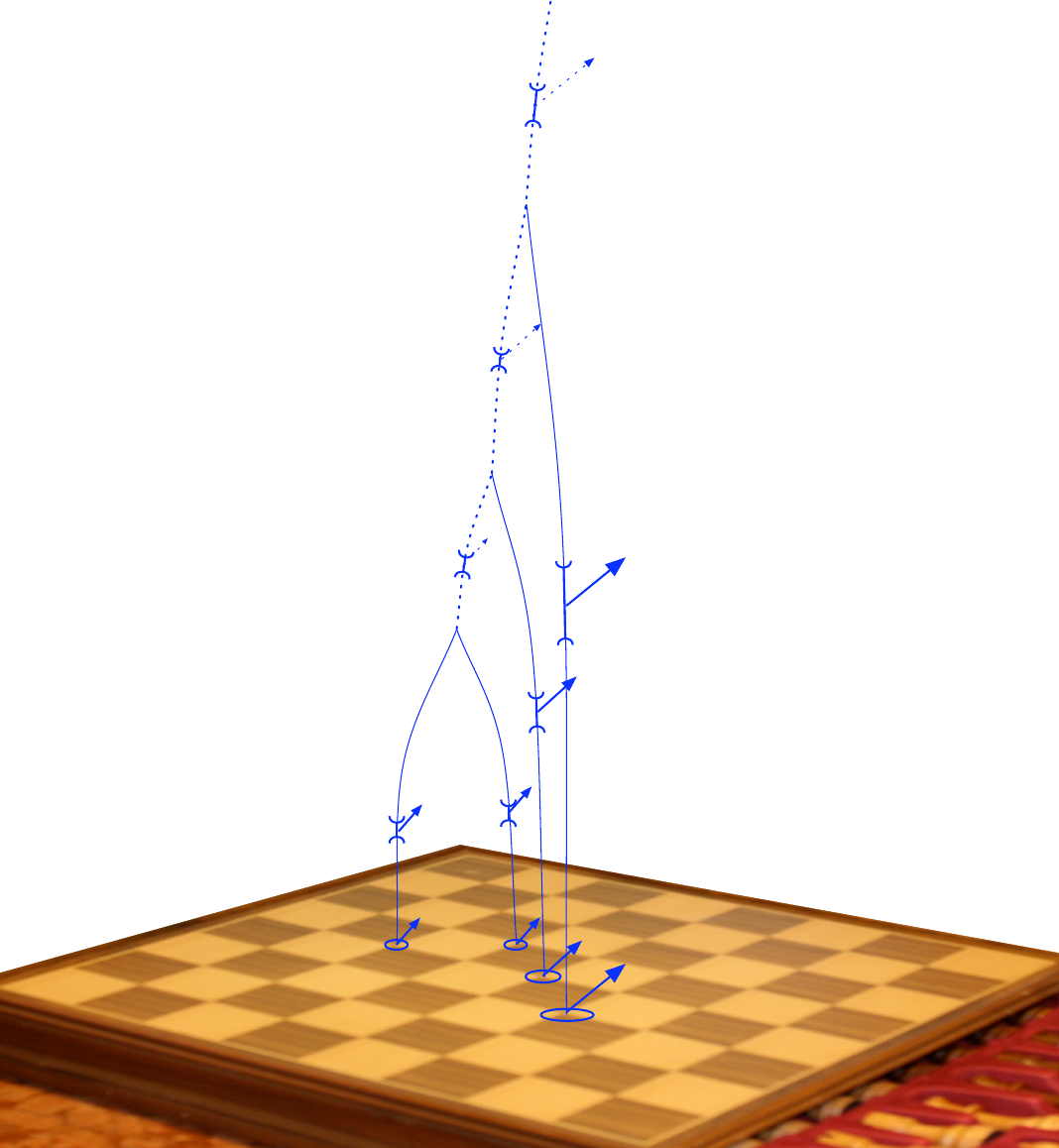}\hspace{-.2cm}\includegraphics[width=.2\textwidth,height=.3\textwidth]{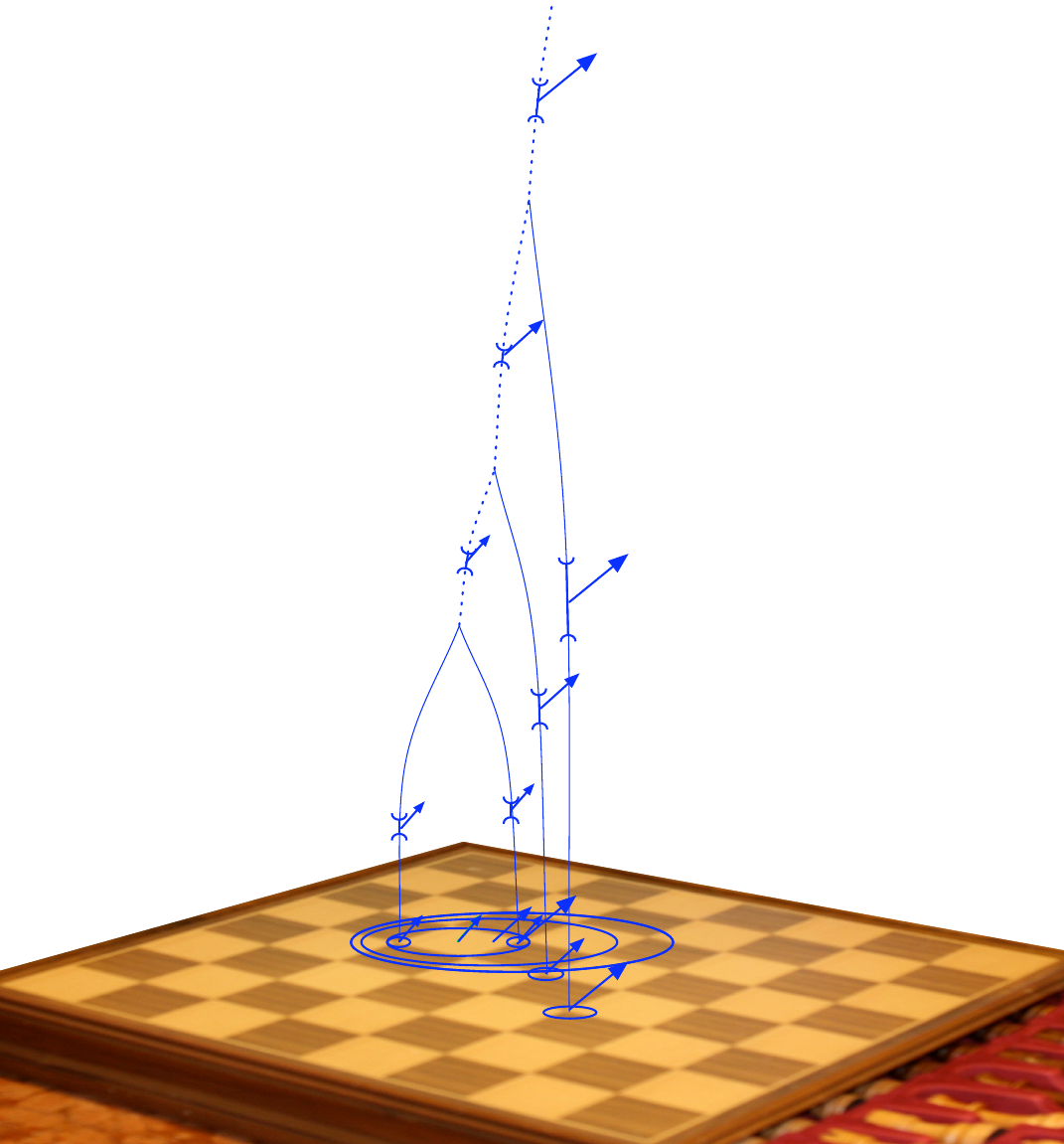}\hspace{-.2cm}\includegraphics[width=.2\textwidth,height=.3\textwidth]{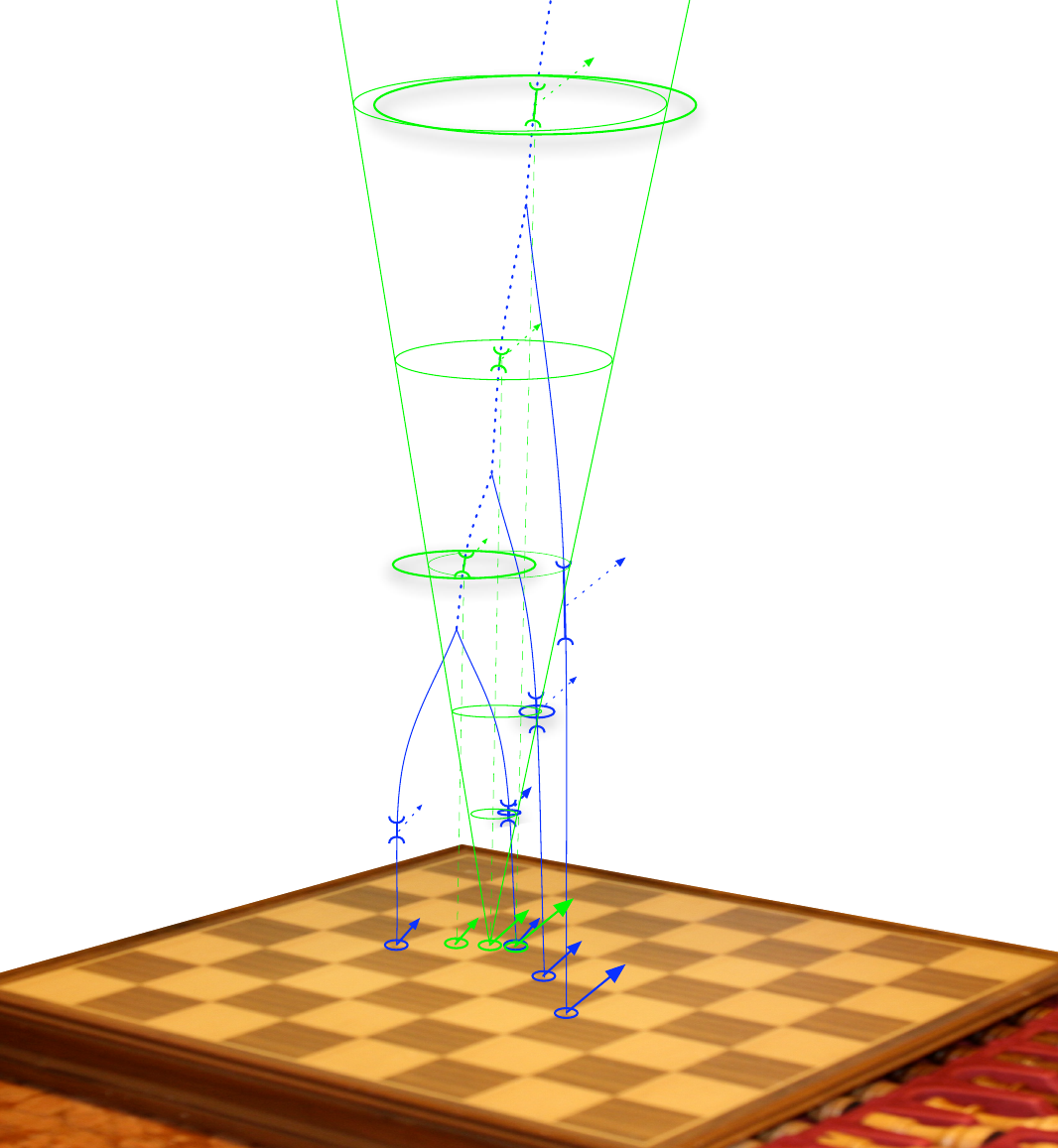}\hspace{-.2cm}
\vline
\includegraphics[width=.2\textwidth,height=.08\textwidth]{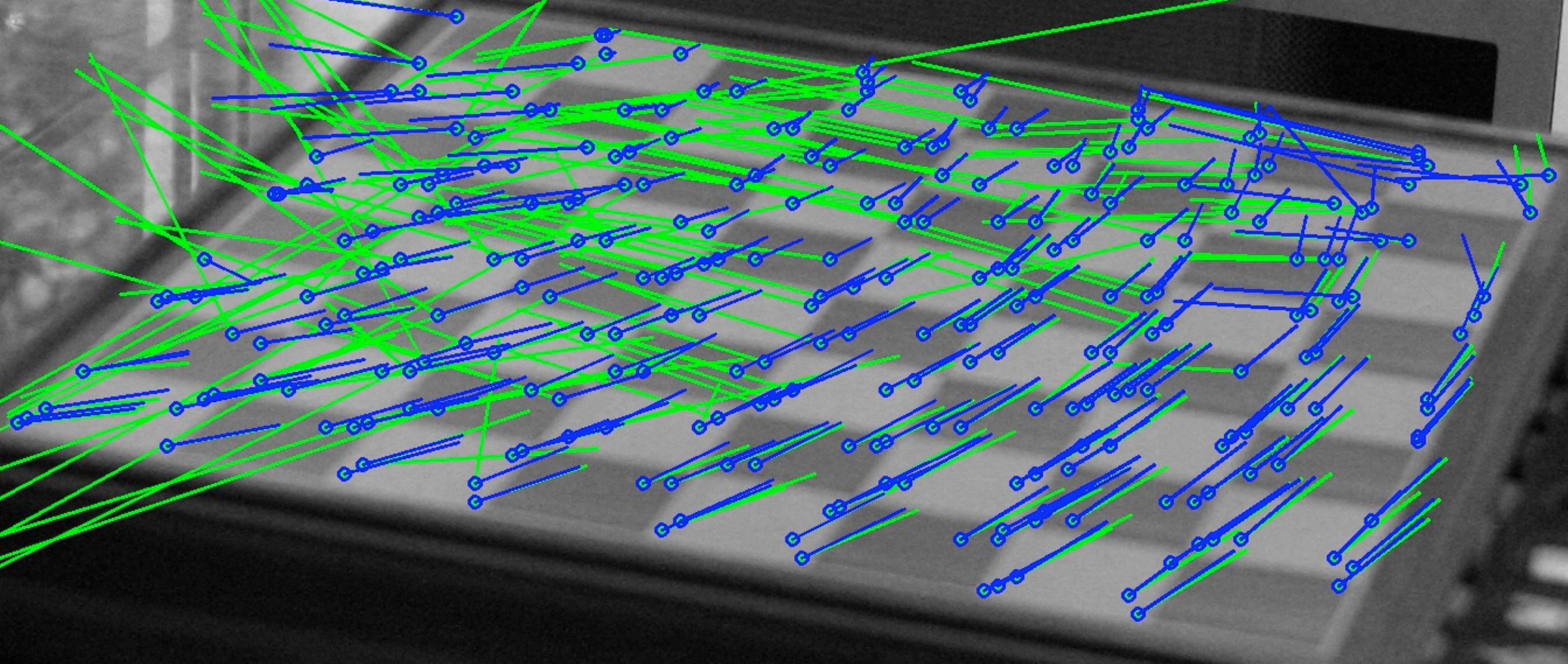}
\end{center}
\caption{\sl {\bf Tracking on the selection tree} Proper sampling only provides motion estimates at the terminal branches (finest scale); the motion estimated at inner branches is used to back-warp the images so large motion would yield properly-sampled signals at finer scales (left). As an alternative, the motion estimated at inner branches can also be returned, together with their corresponding scale (middle). Traditional multi-scale detection and tracking, on the other hand, first ``flattens'' all selections down to the finest level (dashed vertical downwards lines), then for all these points consider the entire multi-scale cone above (shown only for one point for clarity). As a result, multiple extrema at {\em inconsistent} locations in scale-space are involved in providing coarse-scale initialization (right). Motion estimates at a scale {\em finer} than the native selection scale (thinner green ellipse), rather than improving the estimates, degrade them because of the contributions from spurious extrema (blue ellipses).}
\label{fig-selection-tree}
\end{figure}}

\cut{\begin{figure}[htb]
\begin{center}
\includegraphics[height=.25\textheight]{figures/street_tracking1.png}
\includegraphics[height=.25\textheight]{figures/street_tracking2.png}
\includegraphics[height=.25\textheight]{figures/feature_depth_limit2.png}
\end{center}
\caption{\sl {\bf Feature tracking in multi-scale in an urban scene} Green trails show the tracks for forward motion (left), and rotational motion (middle). Detail of selection and outlier rejection (empty circles).}
\label{fig-tracking-corners-streets}
\end{figure}}

\cut{\begin{figure}[htb]
\begin{center}
\includegraphics[width=.3\textwidth]{figures/tracking2.png}
\includegraphics[width=.3\textwidth]{figures/tracking3.png}
\end{center}
\caption{\sl {\bf Tracking corner features on a mobile phone} Green trails show multi-scale FAST corners being tracked a mobile phone as moving the phone camera downward, and to the right. }
\label{fig-tracking-corners-phone}
\end{figure}}

Thus the ability to track depends on {\em proper sampling} in both space and time. 
This suggests the approach to multi-scale tracking 
used in \cite{leeS10}:
\begin{enumerate}
\item Construct a spatial scale-space of feature detectors, until the signal is properly sampled in time. \cut{\tt This can always be guaranteed\cutThree{ as diffusion will eventually make the signal arbitrarily low-pass}.} 
\item Estimate motion at the coarser scale, with whatever feature tracking/motion estimation/optical flow algorithm one wishes to use \cite{leeS10}. This is now possible because the proper sampling condition is satisfied both in space and time.\cutThree{\footnote{In practice, there is a trade-off, as in the limit too smooth a signal will fail the transversality condition (\ref{eq-jacobian}) and will not enable establishing a proper frame $\hat g$.}}
\item Propagate the estimated motion in the region determined by the detector to the next scale. At the next scale, there may be only one selected region in the corresponding frame, or there may be more (or none), as there can be singular perturbations (bifurcations, births and deaths). 
\item For each region selected at the next scale, repeat the process from 2. 
\end{enumerate}
This is illustrated in Figure \ref{fig-selection-tree}. 
\begin{figure}[htb]
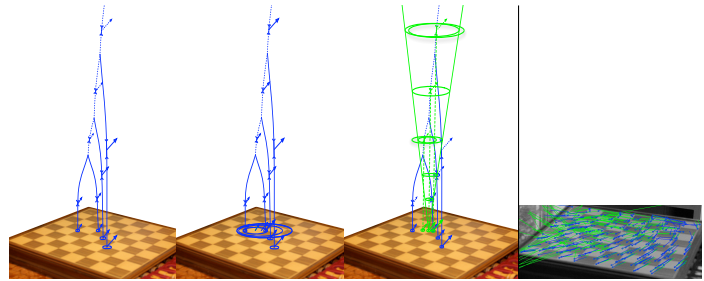

\begin{center}
\includegraphics[width=.2\textwidth,height=.3\textwidth]{figures/selection_tree.pdf}\hspace{-.2cm}\includegraphics[width=.2\textwidth,height=.3\textwidth]{figures/selection_tree_v2.pdf}\hspace{-.2cm}\includegraphics[width=.2\textwidth,height=.3\textwidth]{figures/selection_tree_v4.pdf}\hspace{-.2cm}
\vline
\includegraphics[width=.2\textwidth,height=.08\textwidth]{figures/track-1-3-detail-fine.pdf}
\end{center}
\caption{\sl {\bf Tracking on the selection tree.} The approach we advocate only provides motion estimates at the terminal branches (finest scale); the motion estimated at inner branches is used to back-warp the images so large motion would yield properly-sampled signals at finer scales (left). As an alternative, the motion estimated at inner branches can also be returned, together with their corresponding scale (middle). Traditional multi-scale detection and tracking, on the other hand, first ``flattens'' all selections down to the finest level (dashed vertical downwards lines), then for all these points considers the entire multi-scale cone above (shown only for one point for clarity). As a result, multiple extrema at {\em inconsistent} locations in scale-space are involved in providing coarse-scale initialization (right). Motion estimates at a scale {\em finer} than the native selection scale (thinner green ellipse), rather than improving the estimates, degrade them because of the contributions from spurious extrema (blue ellipses). Motion estimates are shown on the right (blue = \cite{leeS10}, green = multi-scale Lucas-Kanade \cite{lucasK81,bakerGM04}).}
\label{fig-selection-tree}
\end{figure}

Note that only the terminal branches of the selection scale-space provide an estimate of the frame $\hat g$, whereas the hidden branches are used only to initialize the lower branches. Nevertheless, one can report each motion estimate at the native selection scale (Figure \ref{fig-selection-tree} middle), even if they are subsumed by the terminal branches of the selection tree.

\cut{This is different than multi-scale tracking as traditionally done\cutThree{, whereby features are selected at multiple scales, then the scale-space is {\em flattened} (dashed line in Figure \ref{fig-selection-tree} right), and then each of the resulting feature points (now the nomenclature of ``point'' is appropriate, as scale has been removed from the detected frames) are tracked in scale-space. This is what is done in the multi-scale version of} \cite{lucasK81} described in \cite{bakerM01} and implemented in \cite{openCV}, illustrated in Figure \ref{fig-selection-tree} (right).}

{This approach is called {\em tracking on the selection tree} (TST) \cite{leeS10}, because it is based on proper sampling conditions and tracking is performed at each native selection scale.\cut{ Not only is the proper algorithm more accurate, but it is also faster because it avoids computations up and down the multi-scale pyramid where it is not necessary and where, because of improper sampling, it is detrimental due to aliasing errors.}\cutThree{ Indeed, note that the ``second-moment-matrix'' test commonly used for tracking \cite{lucasK81,shiT94}, even if performed at each scale, cannot be relied upon to reject features that cannot be tracked. This is because it is possible, and indeed typical,  that due to singular perturbations, the region passes the second-moment test (\eg harris' \cite{harris88}) at each scale, but {\em not} because the feature of interest is trackable, but because additional structures have appeared in the new scale (Figure \ref{fig-selection-tree} right).} This is a form of {\em structural aliasing} phenomenon. \marginpar{\tiny \sc structural aliasing} \index{Structural aliasing}

\cut{Note that once active exploration has been performed at least once, and is encoded in the training set, classification can be performed on a snapshot datum, for instance a single picture. \cut{We would not be able to recognize faces in one photograph if we had not seen (possibly different) faces from different vantage points, under different illumination etc. Similarly, if we were never allowed to move during learning (or, more in general, to adapt the controlled nuisance to the scene), we would not be able to develop templates that are discriminative and at the same time be insensitive to the nuisances.}}

Tracking provides a time-indexed frame $\hat g_t$ that can be canonized in multiple images to provide samples of the canonized statistic $\phi^\wedge h(\xi, \nu_t)$, where $\nu_t$ lumps all other nuisances that have not been canonized (including the group nuisances that do not commute with non-invertible nuisances). These can then be used to build a descriptor, for instance a template (\ref{eq-blurred-template}), or more general ones that we discuss next.

Note that the process of selection by maximizing the structural stability margin can be understood as a scale canonization process, even though in Section \ref{sect-interaction} we argued against canonization of scale. Note, however, that here we are not making an association of features across scales other than for the purpose of initializing tracks. In particular, consider the example of the corner in Figure \ref{fig-selection-tree}: The corner only exists at the finest scale. Features detected at coarser scales serve to initialize tracking at the finest scale, but features selected at coarse scales are not associated to the corner point, they are simply different local features. Aggregating multiple TSTs can also be done, when some ``side information'' is available that enables to group them together. For instance, in \cite{wnukS11}, detachability is used as side information \cite{ayvaciS11EMMCVPR}, exploiting, again, knowledge of occlusions \cite{ayvaciRS11IJCV}.

\cut{\subsection{Selecting the best canonical frames}
\label{sect-selecting}
\index{Canonical frame}
\index{Frame!canonical}
From the previous section we have that {\bf translation} is locally canonizable within an unoccluded domain. In Section \ref{sect-implementation} we use \cite{rosten_2006_machine} to find multiple canonical representatives for translation, sorted in order of separation from other extrema to guarantee (local) structural stability; {\bf scale} is not canonizable in the presence of quantization. Instead, the scale-quantization semi-group can be sampled at multiple scales starting from the native resolution of the sensor. Three options are possible at this point: (a) one can jointly locally canonize translation and scale, as customary in the scale-selection literature, or (b) one can first canonize translation (by feature selection at the finest scale) and then sample scale (by computing a local description across scales of the {\em same} point in space), or (c) first canonize or sample scale, and then for each sample scale canonize translation. It should be noted that (a) is a good selection strategy only for blobs, because in the presence of corners a feature detector based on joint canonization of the translation-scale group results in a {\em non-Morse} function that has a continuum of extrema at different positions and different scales. This is reflected in the ``traveling feature'' phenomenon observed in Figure 1.14 on page 26 of \cite{lindeberg98}. Therefore, this popular procedure \cite{lowe99object} is not consistent with the principles of optimal canonization. This might explain why others have advocated forgoing feature selection altogether \cite{dalalT05}, or have simply moved to sampling translations, along with scale \cite{fulkersonVS09}.
In Section \ref{sect-implementation} we use a fixed sampling of scale dictated by the computational limitations; {\bf rotation} is canonizable in the local frame determined by translation and scale, with a variety of canonization mechanisms. However, the projection of the {\em gravity vector} onto the image plane provides a natural canonical direction, which we use in Section \ref{sect-implementation}, since an estimate of gravity is available from inertial sensors. Note that, unlike advocated in the scale-selection literature \cite{lindeberg98,lowe04distinctive}, this procedure yields a number of detected translations (locations) $\{x_i\}_{i=1}^N$, and for each of them a collection of scales {\em at that location}, $\{\sigma_{ij}\}_{i,j = 1}^{N,M}$. Once rotation is canonized, we have a collection of {\em similarity (reference) frames} $\hat g_{ij} = \{x_i, \sigma_{ij}, R_{ij}\}$ each identifying a region ${\cal B}_{\sigma_{ij}}(x-R_{ij}x_i)$, where a complete {\bf contrast} invariant can be computed:
\begin{equation}
\phi(I) =  {\large \{ } \frac{\nabla h(\hat g_{ij} \xi, \nu)}{\|  \nabla h(\hat g_{ij} \xi, \nu) \| } = I\circ \hat g_{ij} (x)  \doteq \phi_{ij}(I) ~~~ \ \forall \ x \in {\cal B}_{\sigma_{i,j}}(x-R_{i,j}x_i). {\large \}}_{i,j = 1}^{N,M}
\label{descriptor}
\end{equation}
The feature $\phi(I)$ is now a multi-component descriptor for the entire image $I$. Non-invertible nuisances such as {\bf complex illumination effects, noise, quantization} etc. are un-modeled phenomena that do not enter the representation $\phi(I)$. All other nuisances are not canonizable and must be {\em marginalized} (or eliminated via maximum-likelihood). 
\index{Occlusion}
In particular,  {\bf occlusions} can be marginalized via a combinatorial matching test between collections of (reference) features $\{\phi_{ij}(I)\}$ in different images. Arbitrary {\bf changes of viewpoint} correspond to diffeomorphic domain deformations that do not constrain the frames $\{\hat g_{ij}\}$ \cite{sundaramoorthiPVS09}, making the collection $\{\phi_{ij}(I)\}$ a {\em bag of features}. This in part explains the surprising success of this simplistic model, which we adopt in Section \ref{sect-implementation}. One could restrict the allowable scenes geometry, and correspondingly the domain deformations to be {\bf affine} or {\bf projective}, in which case marginalization can be performed as a {\em geometrically-validated matching} test, by comparing configurations of local reference frames $\{ \hat g_{ij}(t) \}$ in different images. In particular, one makes the hypothesis that there exists a homography $H$ such that $x_i(t+1) = H x_i(t)$ for a subset of $i = 1, \dots, N$, with $\hat g_i(t) = \{x_i(t), \sigma_{ij}(t), R_{ij}(t)\}$. Putative correspondences $(i,j)_{(t)} \leftrightarrow (l, m)_{(t+1)}$ that violate this hypothesis are rejected as outliers, and the hypothesis testing can be performed with a variety of schemes, \eg variants of RANSAC \cite{ransac}. This stage is too costly to implement on a hand-held device and is therefore replaced with a simpler test in our implementation in Section \ref{sect-implementation}.
}

\chapter{Designing feature descriptors}
\label{sect-descriptors}
\index{Feature descriptor}
\index{Descriptor}

A feature detector provides a $G$-covariant reference frame, relative to which the image is, by construction, $G$-invariant. So, the simplest invariant descriptor is the image itself, expressed in the reference frame determined by the detector (\ref{eq-can-descr}). For the case of Euclidean and contrast transformations, $G = SE(2) \times {\cal M}$, where $\cal M$ denotes the set of monotonic continuous transformations of the range of the image. 
In the case in which a sequence of images is available, correspondence of reference frames is provided by tracking $\{g_t\}_{t = 1}^T$, as described in Section \ref{sect-correspondence}, and again the entire time series in the normalized frame, $I\circ g_t^{-1}$ is by construction $G$-invariant. Of course, all the non-invertible nuisances, as well as the group nuisances that do not commute with them, are {\em not} eliminated by this procedure, and therefore they have to be dealt with at decision time via either  marginalization (\ref{eq-marg}), or extremization (\ref{eq-ML}). 
Using the formal image-formation model $h$, and the maximal $G$-invariant $\phi^\wedge$, we have that
\begin{equation}
\phi^\wedge(\{I_t \}_{t=1}^T) = \{ h({\hat g}^{-1}_t(I_t),  \xi, \nu_t)\}_{t=1}^T.
\label{eq-phi-t}
\end{equation}
However, marginalizing or max-outing the residual effects of the nuisance $\{\nu_t\}_{t=1}^T$ -- that include occlusions, quantization, noise, and all un-modeled photometric phenomena such as specularities, translucency, inter-reflections etc. -- can be costly at decision time.\cut{\footnote{In particular, marginalization of occlusions reduce, in the presence of locally-detected features, to a combinatorial matching corresponding to testing the hypothesis that the same frame is {\em co-visible} in two or more images (\eg test-image and template-image).}} Therefore, in Section \ref{sect-best-template}, we have justified the introduction of an alternate approach that attempts to simplify the decision run-time by aggregating the training set into a {\em template}. Of course, one could do the same on the test set, which may consist of a single image $T=1$, or of video. The result is what is called a {\em feature descriptor}.  In this chapter we will explore alternate descriptors, more general than the best template described in Section \ref{sect-best-template}, which we show how to compute in Section \ref{sect-btd}. We also discuss relationships to {\em sparse coding} in Section \ref{sect-sparse}, which links to the discussion of linear detectors in Section \ref{sect-detector-linear}. Recall that the ``best template'' in Section \ref{sect-best-template} is only the best among (static) templates. Therefore in Section \ref{sect-time-hog} we show that one can, in general, achieve better results by {\em not} eliminating the time variable $t$ as part of the feature description process, but retaining the time variable, and instead marginalizing it or max-outing it as part of the decision process. This requires the introduction of techniques to marginalize time, which we describe in Section \ref{sect-twdc}.

Before doing all that, however, we summarize the role of various nuisances and how they are handled before the descriptor design process commences.

\section{Nuisance taxonomy (summary)}
\label{sect-nuisance-taxonomy}

This section summarizes material from previous chapters, and can therefore be skipped. 
\begin{footnotesize}
In Section \ref{sect-im-form} we have divided the nuisance into those that have the structure of a group, $g$, and those that are {\em non-invertible}, $\nu$, including the additive component $n$, which we refer to as ``noise.'' Then, in Section \ref{sect-interaction}, we have shown that of the group nuisance, only a sub-group commutes with $\nu$ and $n$. These are the group-commutative nuisances, consisting of the isometric group of the plane. \index{Isometric group} Indeed, these are only {\em locally} canonizable \index{Local canonization} and therefore can be eliminated at the outset via co-variant detection and invariant description (\ref{eq-can-descr}), modulo a selection process corresponding to combinatorial matching at decision time, to marginalize the occlusion nuisance, as described in (\ref{eq-local-feature}) in Section \ref{sect-local}. In this section we elaborate on how to treat the various nuisances, thus expanding (\ref{eq-local-feature}).

\begin{description}
\item[Translation:] Canonization corresponds to the selection of a particular point (feature detection) that is chosen as (translational) reference. The image in this new reference is, by construction, invariant to planar translation, assuming that the feature detection process is structurally stable, commutes with non-invertible nuisances, and there is no occlusion. There are many mechanisms to canonize translation. Examples include the extrema of the determinant of the Hessian of the image, or of the convolution with a Laplacian of Gaussian, extrema of the determinant of the second-moment matrix\cut{, or the nodes (region centroids) or faces (junctions) of a superpixel adjacency graph, as we have illustrated in Section \ref{sect-detectors}}. In any case, one must choose multiple translations $T_i$ at each sample scale $\sigma_j$. As we have discussed in Section \ref{sect-detectors}, scale should be sampled, {\em not canonized} (it does not commute with quantization), and it can be either sampled regularly, or adaptively in a way that yields maximum structural stability margins, as described in Section \ref{sect-stability}. 
If the image is encoded by a ``segmentation tree'' \index{Superpixel tree} (a partition of the domain into regions that are constant to within a certain tolerance $\sigma$), then the centroid of each region (a node in the region adjacency graph) or the junctions of boundaries of three or more regions (faces in the adjacency graph) are also viable canonization procedures that can be designed to be structurally stable.
\item[Scale:] Scale is not canonizable in the presence of quantization. Whatever descriptor one chooses should be represented at multiple scale. 
This is described in Section \ref{sect-matching}.

Note that a variety of sampling options is possible, depending on the interplay  with translational feature detection. One could first sample all available scales, and then independently canonize translation in each one. Or, one could canonize translation at the native scale of the image, and then sample the image at multiple scales at that location. Or, one can simultaneously sample scale and detect translational frames by performing maximally-structurally stable translation selection, where the maximization of the stability margin is performed relative to scale. This is accomplished in a manner similar to scale selection as prescribed by \cite{lindeberg98}, described in \cite{leeS10}. It is conceptually important to understand that {\em multiple scales} can be detected at the same location, therefore the scale selection process is more like an adaptive scale sampling as opposed to a (unique) scale canonization procedure.

\item[Rotation:] Planar rotation is canonizable, and therefore it should be canonized. In the presence of measurements of gravity, a natural canonical orientation is provided by the projection of the gravity vector onto the image plane (assuming it is not aligned with the optical axis). \cut{\tt Knowledge of the distribution, however, does not make a difference in the representation, after Theorem \ref{thm-equivariant}.} 

If we have $N_TN_S$ regions available from sampling translations and scale, we can canonize each one with a co-variant detector and its corresponding invariant descriptor $\phi_R$. Co-variant detection can be performed in a number of ways, using extrema of the gradient direction \cite{lowe99object} (at each location, at each scale), or the principal direction of the second moment matrix \cite{mikolajczykSb}  at each location and scale, or nodes and faces of the adjacency graph  of $\epsilon$-constant regions \cite{vedaldiS05}, at each location and scale. The result is an unchanged number $N_T N_S$ of descriptors that are now invariant to rotation as well as (locally) translation. Alternatively, rotation can be canonized by using gravity and longitude (or latitude) as canonical references.
\item[Contrast:] Contrast can be canonized independently of other nuisances by replacing the intensity at every pixel with the gradient direction:
\begin{equation}
\phi(I) = \frac{\nabla I}{\| \nabla I \| }.
\end{equation}
This is conceptually straightforward, except that it assumes that the image is differentiable, which in general is not the case (one can define discrete differentiability which, however, entails a choice of scale, thus making this choice still dependent on scale). It also raises issues when $\| \nabla I \| =0$, \ie, when the image is constant in a region. As an alternative mechanism, one can canonize contrast, in a local region determined by a translational frame $T_i$ at a scale $\sigma_j$, by normalizing the intensity by subtracting the mean and dividing by the standard deviation of the image, or of the logarithm of the image. If we call $B_{ij} \doteq {\cal B}_{\sigma_j}(x-T_i)$ a neighborhood of size $\sigma_j$ around the canonical reference $T_i$, then 
\begin{equation}
\phi_{ij}(I) = \frac{I - \mu(I)}{{\rm std}(I)}
\end{equation} 
where $\mu(I) \doteq \frac{\int_{B_{ij}} I(x)dx}{\int_{B_{ij}} dx}$ and ${\rm std}(I) \doteq \sqrt{\frac{\int_{B_{ij}} |I(x) - \mu(I)|^2dx}{\int_{B_{ij}} dx}}.$
In this latter case, the canonization procedure interacts with scale, and therefore contrast normalization should be done independently at all scales. Additional options if multiple spectral bands are available is to consider spectral ratios, \marginpar{\tiny \sc spectral ratio} \index{Spectral ratio} for instance in an $(R,G,B)$ color space one can consider the ratio $R/B$ and $G/B$, or the normalized color space, or in an $(H,S,V)$ color space one can consider the $H$ and $S$ channels, etc.
\end{description}

\noindent Among the nuisances that cannot be canonized, in addition to scale and occlusion that should be sampled and marginalized as in Section \ref{sect-local}, we have:
\begin{description}
\item[Illumination:] 
Complex illumination effects other than contrast cannot be canonized, and therefore their treatment should be deferred to decision time.
\item[Quantization and noise:] 
Quantization is intertwined with scale and cannot be canonized. At each scale, quantization error can be lumped as additive noise. Detectors for canonizable nuisances should be designed to commute with quantization and noise. Linear detectors do so by construction. 
\item[Skew:] One could treat the (non-canonizable) group of skew transformation in the same way as scale, but since there is no meaningful sampling of this space, we lump the skew with other deformations and defer their treatment to marginalization or blurring. 
\item[Deformations and quantization:] 
Domain deformations other than rotations and translations, including the affine and projective group or more general diffeomorphisms, cannot be canonized in the presence of quantization, and therefore their handling should be deferred to either training (blurring) or testing (marginalization, extremization). 
\item[Occlusion:] 
Occlusions, finally, are not invertible and cannot be canonized, so they will have to be explicitly detected during the matching process (via max-out or marginalization), which in this case corresponds to a selection of a subset of the $N_TN_S$ local descriptors as described in Section \ref{sect-local}. 
\end{description}
\end{footnotesize}
What we have in the end, for each image $I$, is a set of multiple descriptors (or templates), one per each canonical translation and, for each translation, multiple scales, canonized with respect to rotation and contrast, but still dependent on deformations, complex illumination and occlusions:
\begin{eqnarray}
\phi(I) &=& \{ k_{ij}\circ\rho(S_{j}R_{ij} x + T_i + v_{ij}(x)) + n_{ij}(x),  \label{descriptor} \\ \nonumber && ~~~~~ i,j =1, \dots N_T, N_S | {\cal B}_{\sigma_{j}}(x + T_i) \cap D  = \emptyset \}
\end{eqnarray}
where $v_{ij}$ is the residual of the diffeomorphism $w(x)$ after the similarity transformation $SRx+T$ has been applied, \ie, $v_{ij}(x) = w(x) - S_j R_{ij}x - T_i$. Here $S = \sigma I$ is a scalar multiple of the identity matrix that represents an overall scaling of the coordinates, and $(R, T) \in SE(2)$ represent a rigid motion with rotation matrix $R\in SO(2)$ and translation vector $T\in \real^2$.  If we call the frame determined by the detector $\hat g_{ij} = \{S_j, T_i, R_{ij}, k_{ij}\}$, we have that
\begin{equation}
\phi(I) = \{ I \circ {\hat g}^{-1}_{ij}\}_{i,j=1}^{N_T, N_S}.
\end{equation}
Note that the selection of occluded regions, which is excluded from the descriptor, is not known a-priori and will have to be determined as part of the matching process. 
As we have discussed in Section \ref{sect-stability}, the selection process should be designed to be invariant with respect to group-commutative nuisances, and ``robust'' (in the sense of structural stability) with respect to all other nuisances. The selection and tracking process described in Section \ref{sect-correspondence} provides one such design.

In the case of video data, $\{I_t\}_{t=1}^T$, one obtains a {\em time series} of descriptors, \marginpar{\tiny \sc time series} \index{Time series}
\begin{equation}
\phi(\{I_t\}_{t=1}^T) = \{ I_t \circ {\hat g}^{-1}_{ij}(t) \}_{i,j,t=1}^{N_T, N_S,T}
\end{equation}
where the frames $\hat g_{ij}(t)$ are provided by the feature detection mechanism that, in the case of video, includes tracking via proper sampling (Section \ref{sect-correspondence}). Once that is done, one should store the time series of descriptors $\phi(\{I_t\}_{t=1}^T)$ for later marginalization, if sufficient storage and computational power is available, as we describe in Section \ref{sect-time-hog}. Otherwise, a static descriptor can be devised, as we discuss in Section \ref{sect-btd}. Before doing so, however, we discuss the interplay between the scene and the nuisance in the next section. 

\section{Disentangling nuisances from the scene}
\label{sect-disentangle}

In the image-formation model described in Section \ref{sect-im-form}, and in the more general model in Section \ref{sect-image-formation}, some of the nuisances interact with the scene to form an image. If the nuisances are marginalized or max-outed, as in (\ref{eq-marg}) or (\ref{eq-ML}), this is not a problem. However, if we want to canonize a nuisance, in the process of making the feature invariant to the nuisance, we may end up making it also invariant to some components of the scene. In other words, by abusing canonization we may end up throwing away the baby (scene) with the bath water (nuisances).

The simplest example is the interaction of {\em viewpoint and shape}. \marginpar{\tiny \sc viewpoint-shape interaction} {\color{orange} In the model (\ref{eq-lambert-ambient}), we see immediately that the viewpoint $g$ and shape $S$ interact in the motion field (\ref{eq-motion-field}) via $w(x) = \pi g \pi^{-1}(x)$, where $p = \pi^{-1}(x) \in S$ depends on the shape of the scene. It is shown in \cite{sundaramoorthiPVS09} that the group closure of domain warpings $w$ spans the entire group of diffeomorphisms, which can therefore be canonized -- if we exclude the effects of occlusion and quantization. However, necessarily the canonization process {\em eliminates the effects of the shape $S$ in the resulting descriptor}, which is the ART described in Section \ref{sect-correspondence}. This had already been pointed out in \cite{vedaldiS05}.} This means that if we want to perform recognition using a strict viewpoint-invariant, then we will lump all objects that have the same radiance, modulo a diffeomorphism, into the same class. That means that, for instance, all white objects are indistinguishable using a viewpoint invariant statistic, no matter how such an invariant is constructed. Of course, as pointed out in \cite{vedaldiS05}, this does not mean that we cannot recognize different objects that have the same radiance. It just means that we cannot do it {\em with a viewpoint invariant,} and instead we have to resort to marginalization or extremization.

The same phenomenon occurs with reflectance (a property of the scene) and illumination (a nuisance). \marginpar{\tiny \sc reflectance-illumination interaction} This is best seen from a more general model than (\ref{eq-lambert-ambient}), for instance one of the models discussed in Section \ref{sect-image-formation}. In the case of moving and possibly deforming objects, there is also an ambiguity between the deformation (a characteristic of the scene) and the viewpoint (a nuisance).

Deciding how to manage the scene-nuisance interaction is ultimately a modeling choice, that should be guided by two factors. The first is the priority in terms of speed of execution (biasing towards canonizing nuisances) {\em vis-a-vis} discriminative power (biasing towards marginalization to avoid having multiple scenes collapse into the same invariant descriptor). The second is a thorough understanding of the interaction of the various factors and the ambiguities in the image formation model. This means that one should understand, given a set of images, what is the set of all possible scenes that, under different sets of nuisances, can have generated those images. This is the set of {\em indistinguishable scenes}, that therefore cannot be discriminated from their images. This issue is very complex and delicate, and a few small steps towards a complete analysis are described in Section \ref{sect-ambiguity}.

In this chapter, we will set this issue aside, and agree to canonize (locally) translation, rotation and contrast, and sample scale. This means that scenes that are equivalent up to a similarity transformation are indistinguishable, as shown in \cite{maSKS}, which is not a major problem, but leaving the reflectance-illumination ambiguity unresolved, other than by removing local contrast transformations that represent a coarse model of illumination changes. In the next section, we move on to describe some of the descriptors that can be constructed under these premises.

\section{Template descriptors for static/rigid scenes}
\label{sect-btd}
\index{Template}
\index{Best Template Descriptor}
\index{Descriptor!invariant}

If we are given a sequence of images $\{I_t\}$ of a static scene, or a rigid object, then the only temporal variability is due to viewpoint $g_t$, which is a nuisance for the purpose of recognition, and therefore should be either marginalized/max-outed or canonized. In other words, there is no ``information'' in the time series $\{ {\hat g}_t \}$, and once we have the tracking sequence available, the temporal ordering is irrelevant. This is not the case when we have a deforming object, say a human, where the time series contains information about the particular action or activity, and therefore temporal ordering is relevant. We will address the latter case in Section \ref{sect-twdc}. For now, we focus on rigid scenes, where $S$ does not change over time, or rigid objects, which are just a simply connected component of the scene $S_i$ (detached objects). The only role of time is to enable correspondence, as we have seen in Section \ref{sect-correspondence}.

The simplest descriptor that aggregates the temporal data is the best template descriptor introduced in Section \ref{sect-best-template}. However, after we canonize the invertible-commutative nuisances, via the detected frames $\hat g_t$, we do not need to blur them, and instead we can construct the template (\ref{eq-blurred-template}) where averaging is only performed with respect to the nuisances $\nu$, rather than (\ref{eq-template}), where all nuisances are averaged-out. The prior $dP(\nu)$ is generally not known, and neither is the class-conditional density $dQ_c(\xi)$. However, if a sequence of frames\footnote{We use $k$ as the index, instead of $t$, to emphasize the fact that the temporal order is not important in this context.} $\{\hat g_k\}_{k=1}^T$ has been established in multiple images $\{I_k\}_{k=1}^T$, with $I_k = h(g_k, \xi_k, \nu_k)$, then it is easy to compute the best (local) template via\footnote{As pointed out in Section \ref{sect-best-template}, this notation assumes that the descriptor functional acts linearly on the set of images $\cal I$; although it is possible to compute it when it is non-linear, we make this choice to simplify the notation. }
\begin{equation}
\phi(\hat I_c) = \int_{\cal I} \phi(I) dP(I|c) = \sum_{\begin{array}{c} \nu_k \sim dP(\nu)\\ \xi_k \sim dQ_c(\xi)\end{array}} \phi\circ h(\hat g_k \xi_k, \nu_k)  = \sum_k I\circ \hat g_k = \sum_{k,i,j} \phi_{ij}(I_k)
\label{eq-tmplt}
\end{equation}
where $\phi_{ij}(I_k)$ are the component of the descriptor defined in eq. (\ref{descriptor}) for the $k$-th image $I_k$. 
A sequence of canonical frames $\{\hat g_k\}_{i=1}^T$ is the outcome of a {\em tracking} procedure (Section \ref{sect-correspondence}). Note that we are tracking reference frames $\hat g_k$, not just their translational component (points) $x_i$, and therefore tracking has to be performed on the selection tree (Figure \ref{fig-selection-tree}). 
{\em The template above $\hat I_c$, therefore, is an averaging of the gradient direction, in a region determined by $\hat g_k$, according to the nuisance distribution $dP(\nu)$ and the class-conditional distribution $dQ_c(\xi)$, as represented in the training data.}
This {\em ``best-template descriptor''} (BTD) is implemented in \cite{leeS10}. It is related to \cite{dalalT05,berg01geometric,tola2008fast} in that it uses gradient orientations, but instead of performing spatial averaging by coarse binning, it uses the actual (data-driven) measures and average gradient directions weighted by their standard deviation over time. The major difference is that composing the template {\em requires local correspondence}, or tracking, of regions $\hat g_k$, in the training set. Of course, it is assumed that a {\em sufficiently exciting sample} is provided, lest the sample average on the right-hand side of (\ref{eq-tmplt}) does not approximate the expectation on the left-hand side. Sufficient excitation is the goal of active exploration described in Chapter \ref{sect-exploration}.
Note that, \cut{once the template descriptor is learned, with the entire scale semi-group spanned in $dP(\nu)$\footnote{Either because of a sufficiently rich training set, or by extending the data to a Gaussian pyramid in post-processing.} recognition can be performed by computing the descriptors $\phi_{ij}$ {\em at a single scale} (that of the native resolution of the pixel). This significantly improves the computational speed of the method, which in turn enables real-time implementation even on a hand-held device \cite{leeS10}. It should also be noted that,} once a template is learned from multiple images, recognition can be performed on a {\em single} test image.

It should be re-emphasized that the best-template descriptor is only the best among templates, and only relative to a chosen family of classifiers (\eg nearest neighbors with respect to the Euclidean norm). For non-planar scenes, the descriptor can be made viewpoint-invariant by averaging, but that comes at the cost of losing shape discrimination. If we want to recognize by shape, we can marginalize it, but that comes at a (computational) cost.

It should also be emphasized that the template above is a first-order statistic (mean) from the sample distribution of canonized frames. Different statistics, for instance the median, can also be employed \cite{leeS10}, as well as multi-modal descriptions of the distribution \cite{wnukS11} or other dimensionality reduction schemes to reduce the complexity of the samples.

\section{Time HOG and Time SIFT}
\label{sect-time-hog}
\index{Time HOG}
\index{Time SIFT}

The best-template descriptor in the previous section is a {\em first-order} statistic of the conditional distribution $p(I|c)$. As an alternative, instead of {\em averaging} the sample, one can compute a {\em histogram}, thus retaining all statistics. For this to be done properly, one has to consider each image to be a sample from the class-conditional distribution. A wildly simplifying assumption is to assume that every pixel is independent (obviously not a realistic assumption) so that the conditional distributions can be aggregated at each pixel. 
This is akin to computing, for every location $x$ in a canonized frame $\hat g_k$, the {\em temporal histogram} of gradient orientations. This statistic eliminates time and discards temporal ordering.
\begin{example}[SIFT and HOG revisited]
If instead of a sequence $\{I_k\}$ one had only one image available, one could generate a pseudo-training set by duplicating and translating the original image in small integer intervals. The procedure of building a temporal histogram described above then would be equivalent to computing a {\em spatial histogram} of gradient orientations. SIFT \cite{lowe99object} and HOG \cite{dalalT05} describe one particular way of binning such histograms. So, one can think of SIFT and HOG as a special case of template descriptor where the nuisance distribution $dP(\nu)$ is {\em not} the real one, but a simulated one.
\end{example}
We call the distributional aggregation, rather than the averaging, of $\{ \phi_{ij}(I_k) \}$ in (\ref{eq-tmplt}) the Time HOG or Time SIFT, depending on how the samples are aggregated and binned into a histogram\marginpar{\tiny \sc time hog}\marginpar{\tiny \sc time sift}\index{Time HOG}\index{Time SIFT} \cite{raptisS10}. 

As an alternative to temporal aggregation via a histogram, one could perform aggregation by dimensionality reduction, for instance by using principal component analysis, or a kernel version of it as done in \cite{meltzerS06}, or using sparse coding as we describe in the next section.

Although a step up from template descriptors, Time SIFT and Time HOG still discard the temporal ordering in favor of a static descriptor. In cases where the temporal ordering is important, as in the recognition of temporal events, one should instead retain the time series $\{ \phi(I_t)\}$ and compare them as we describe in Section \ref{sect-twdc}, which corresponds to marginalizing, or max-outing, time. This process is considerably more onerous, computationally, at decision time. Before doing so, we illustrate an alternate approach to build local descriptors based on ideas from sparse coding \cite{olshausen98sparse}.

\section{Sparse coding and linear nuisances}
\label{sect-sparse}
\index{Sparse coding}
\index{Nuisance!linear}

Many nuisances, whether groups or not, act {\em linearly} on the (template) representation. Take for instance the instantiation of the Lambert-Ambient model (\ref{eq-lambert-ambient}), and assume that we have canonized contrast $m$ (or, equivalently, after canonization, assume $m = Id$). The diffeomorphic warping $w$ includes canonizable nuisances, group nuisances that are not canonizable, and nuisances that are not invertible, such as occlusions, quantization, and additive noise. Complex as it may be, $w$ is a {\em linear functional}\footnote{A linear functional is a map $W$ from a function space to a vector space that satisfies the linearity assumption, that is $W(\alpha f_1 + \beta f_2) = \alpha W(f_1) + \beta W(f_2)$ for any functions $f_1, f_2$ an scalars $\alpha, \beta$. In general, linear maps can be written as the integral of the argument against a {\em kernel}, \cite{rudin73}.} acting on the radiance function $\rho$: 
\index{Linear functional}
\begin{equation}
W: {\cal I} \rightarrow {\cal I}; \ \rho \mapsto W\rho \doteq \rho \circ w^{-1}.
\end{equation} 
{\color{pea}
The kernel that \index{Kernel} represents it is degenerate, in the sense that it is not an ordinary function but a distribution (Dirac's Delta) \index{Distribution} \index{Dirac's Delta}
\begin{equation}
W\rho = \rho(w^{-1}x) = \int_{\real^2} \delta(w^{-1} x - y) \rho(y)dy ~~~ x \in D \backslash \Omega 
\end{equation}
where $\Omega$ denotes the occluded region. The kernel an ordinary function once we introduce sampling into the model:
\begin{eqnarray}
I(x_i) &=& \int_{{\cal B}_\sigma (x_i)} W\rho(x)dx + n(x_i) = \\
&=& \int_{{\cal B}_\sigma (x_i)}  \int_{\real^2} \delta(w^{-1} x - y) \rho(y)dx dy + n(x_i) \\
&=& \int_{\real^2} {\cal G}(w^{-1} x_i - y; \sigma) \rho(y)dy + n(x_i)
\label{eq-model-before-sparse}
\end{eqnarray}
where the kernel $\cal G(\cdot; \sigma)$ is the convolution of $\delta$ with the characteristic function of the sampling domain ${\cal B}_\sigma$. This shows that the compound effect of quantization and domain deformations is the convolution with a deformed sampling kernel. The kernel may also include an anti-aliasing filter.\footnote{One may notice that the integral is performed over the entire plane $\real^2$, which may raise the concern that the procedure has obvious memory and computational limitations. The fact that the domain is not bounded can be addressed by noticing that  the value of the radiance $\rho$ outside a ball of radius $3 \sigma$ around the point $w^{-1}x_i$ is essentially irrelevant since $\cal G$ is typically (exponentially) low-pass. One could also discretize $\rho$, but that is also fraught with difficulties: How many samples should we store? Would $\rho$ obey the conditions of Nyquist-Shannon's sampling theorem? In general we cannot expect the radiance $\rho$ to be effectively band-limited: Real-world reflectance has sharp discontinuities due to material transitions, cast shadows and occlusions. One could coerce the problem into the narrow confines of the sampling theorem, but then low-pass filtering would be needed to fit the existing memory and computational constraints, leading us back to the place we started.}

An alternative is to approximate $\rho$ with piecewise linear or even piecewise constant functions, which can be done arbitrarily well under mild assumptions as described in Section \ref{sect-superpixels}, and encode the domain partition where the function is $\epsilon$-constant:
\begin{equation}
\{ S_i \}_{i = 1}^{N_S} \ | \ \cup_{i=1}^{N_S} S_i = D \backslash \Omega , ~~~ S_i \cap S_j = \delta_{ij}.
\end{equation}
The function $\rho$ can then be written as
\begin{equation}
\rho(x) = \sum_{i=1}^{N_S} \rho_i \chi_{S_i}(x).
\end{equation}
While the group $w(x)$ is, in general, not piecewise constant, its variation within each $S_i$ is not observable, and therefore we can without loss of generality assume that $w(x) = w_i \ \forall \ x \in S_i$. Naturally, there remains to be determined which indices $i$ correspond to regions that are visible in both the template and the target image, and which ones are partially or completely occluded.

A third alternative, motivated by the statistics of natural images \cite{olshausen98sparse}, is to assume that $\rho(y)$, while not smooth, is {\em sparse} with respect to some basis. This means that there are functions $b_1(y), \dots, b_N(y)$ and coefficients $\alpha_1, \dots, \alpha_N$ such that, for all $y \in \real^2$, we have 
\begin{equation}
\rho(y) = \sum_{i=1}^N b_i(y) \alpha_i \doteq b(y) \alpha
\end{equation}
for some finite $N$, where we have used the vector notation $b = [b_1, \dots, b_N]$ and $\alpha = [\alpha_1, \dots, \alpha_N]^T$. Plugging the equation above into (\ref{eq-model-before-sparse}), we have
\begin{equation}
I(x_i) = \int_{\real^2} {\cal G}(w^{-1} x_i - y; \sigma) b(y)dy \alpha + n(x_i).
\end{equation}
If we consider the joint encoding of every pixel $x_i$ in a region centered at the canonized locations $T_j$ of size corresponding to the sample scales $S_{jk}$, we have
\begin{equation}
I_{jk} \doteq \left[ \begin{array}{c}
I(x_1) \\ \vdots \\ I(x_N)
\end{array}\right] = {\Large{\int}}_{\real^2}
\left[ \begin{array}{c}
{\cal G}(T_j^{-1} x_1 - y; \sigma_{jk}) \\ 
\vdots \\ {\cal G}(T_j^{-1} x_N - y; \sigma_{jk})
\end{array}\right]  b(y)dy \alpha \doteq W b \alpha_{jk}
\end{equation}
from which one can see that the coefficients representing the ``ideal image'' \index{Ideal image} $\rho$ with respect to an over-complete basis $b$ are the same as the coefficients representing the ``actual'' image $I$, relative to a transformed basis $B = W b$, whose elements are $B_{lm} = \int {\cal G}(T_j^{-1}x_l - y; \sigma_{jk}) b_m(y)dy$. Note that the dependency on $j$ (the location) and $k$ (scale) is reflected in the coefficients $\alpha_{jk}$. To obtain a description, one can estimate, for each $j, k$, the coefficients 
\begin{equation}
\hat \alpha_{jk} = \arg\min_{w, \alpha} \| I_{jk} - Wb \alpha \|
\end{equation}
from which the representation is given by \index{Representation}
\begin{equation}
\hat \xi_{jk} = b \hat \alpha_{jk}.
\end{equation}
This requires knowing the basis $b$, which in turn requires solving the above problem for all corresponding local regions of all images in the training set. This can be done by joint alignment as in \cite{vedaldiGS08,vedaldiS06NIPS}.}

Obviously, the linear model does not hold in the presence of occlusions. Therefore, the representation $\hat \xi_{jk}$ is only {\em local} where co-visibility conditions are satisfied. Occlusions, as usual, have to be marginalized or max-outed at decision time as described in Section \ref{sect-local}.

\subsection{Dictionary Hypercolumns}
\index{Dictionary}
\index{Dictionary hypercolumn}

As we have argued, it is possible to estimate an invariant representation from the image by assuming that the radiance function is sparse. This led to the observation that the representation of the ``true'' (but unknown) radiance $\rho$ with respect to an over-complete basis $b$ is the same of the representation of the known image $I$ with respect to a transformed basis $Wb$. Note that, in general, this representation will not be unique, and there are some issues with the coherence of the basis that are beyond the scope of this manuscript (see \cite{luBS09}). However, the projection (hallucination) \index{Hallucination} of the representation onto the space of images can be used to compute the residual, so any ambiguity in $\alpha$ is annihilated by the corresponding choice of basis vectors $Wb$.

A consequence of this fact is that one can just augment the dictionary with transformed versions of the bases, thus obtaining an enlarged dictionary, and then use exactly the same algorithm for encoding $\rho$ and $I$. The only difference is that one would have to organize the dictionary by ``linking'' bases that are transformed versions of each other, or ``deformation hypercolumns'' 
\begin{equation}
B\alpha = [Wb_1, \dots, Wb_n]\alpha
\end{equation}
somewhat akin to ``orientation hypercolumns'' in visual cortex. In practice, one would sample the group $w$ according to its prior (which is a combination of $dP(\nu)$ and $dP(g)$), yielding a set of samples $\{W_j\}$, from which the enlarged basis can be constructed, with elements $W_j b_i$, and the understanding that any non-zero element of $\alpha$ multiplying a set of bases $W_j b_i$ will have a non-zero coefficient $\alpha_i$. 

Unfortunately, the coefficients in this representation are not {\em identifiable}: Indeed, not only are they not unique, but they are not continuously dependent on the data, so it is possible for infinitesimally close images to have wildly different set of coefficients. This is not an issue if the goal is {\em data transmission and storage}, where the only role of the representation is to generate a faithful copy of the original (un-encoded) signal. However, in order to use a representation for decision or control, this program does not carry through.

\chapter{Marginalizing time}
\label{ch-time}

\begin{figure}[htb]
\begin{center}
\includegraphics[width=.7\textwidth]{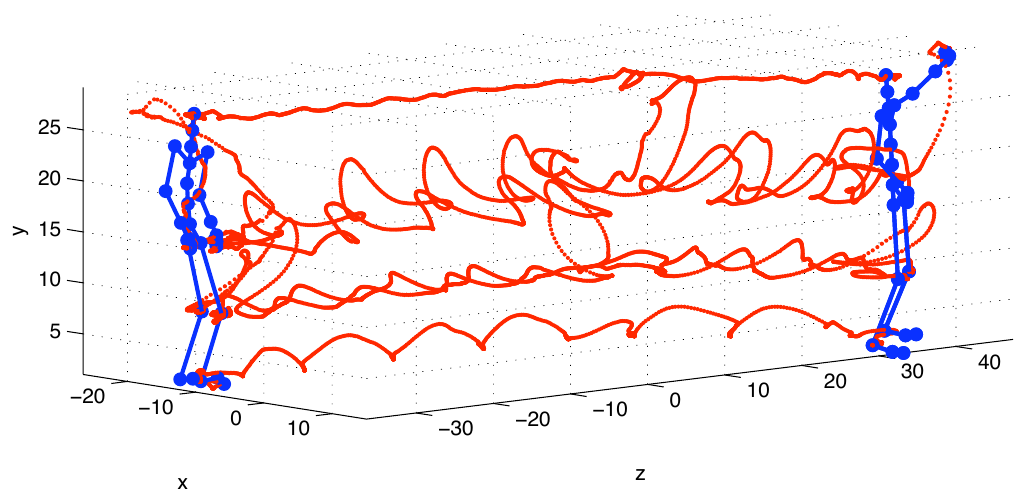}
\end{center}
\caption{\sl With his pioneering light-dot displays, Johansson \cite{johansson73} showed that a great deal of ``information'' is encoded in the temporal evolution of visual signals. For instance, by the motion of a collection of points positioned at the joints, one can easily infer the type of action, the age group of the actor, his or her mood, etc. despite the fact that all pictorial cues have been eliminated from the image. }
\label{fig-johansson}
\end{figure}

As we have discussed in the previous chapter, templates average the data with respect to the distribution of the nuisances, regardless of temporal ordering. Temporal continuity provides a mechanism for tracking, as discussed in Section \ref{sect-correspondence}, but is otherwise irrelevant for the purpose of describing static scenes or rigid objects. Temporal averaging in the template is, in general, suboptimal, so one could consider statistics other than the average, for instance the entire temporal distribution, as we have done in Section \ref{sect-time-hog}. While that approach improves the descriptor in the sense of retaining the distributional properties, it still does not respect temporal ordering. These are just two approaches to {\em canonizing} time by either averaging or binning. Other approaches include a variety of ``spatio-temporal interest point detectors'' \cite{willems2008efficient,laptev2005space} or other aggregated spatio-temporal statistics \cite{davis1997representation}. 

If we are interested in visual decisions involving classes of ``objects\footnote{We call such objects ``actions'' or ``events.''}'' \index{Action} \index{Event} \marginpar{\tiny \sc actions, events} that have a temporal component, then all approaches that ``canonize'' time necessarily mod-out also the temporal variation of interest, in the same way in which canonizing viewpoint eliminates the dependency of the invariant descriptor on the shape of the scene (Section \ref{sect-disentangle}). At the opposite end of the spectrum one could retain the entire time series (the temporal evolution of feature descriptors) (\ref{eq-phi-t}), and compare them as functions of time using any functional norm in the calculation of the classifier (Section \ref{sect-visual-decisions}). This would not yield a viable result because the same action, performed slower or faster, or observed from a different initial time, would result in completely different time series (\ref{eq-phi-t}), if not properly managed. Proper management here means that time should not just be canonized (as in a template or Time HOG) or ignored (by comparing time series as functions), but instead should be {\em marginalized} or {\em max-outed}. The only difference between these approaches is the prior on time, which is typically either uniform or exponentially decaying, so we will focus on a uniform prior, corresponding to a max-out process for time. 

The simplest mechanism to max-out time is known as {\em dynamic time warping} (DTW). \index{Dynamic time warping}\marginpar{\tiny \sc dynamic time warping} \marginpar{\tiny \sc dtw} {\color{pea} Dynamic time warping consists in a reparametrization of the temporal axis, via a function $\tau \leftarrow h(t)$, that is continuous and monotonic (in other words, a {\em contrast transformation} as defined it in Section \ref{sect-im-form}). \index{Contrast transformation} Accordingly, the {\em dynamic time warping distance} \index{Distance!Dynamic time warping} \index{Dynamic time warping distance} is the distance obtained by max-outing the time warping between two time series.} The problem with DTW is that it preserves temporal ordering but little else. The moment one alters the time variable, all the velocities (and therefore accelerations, and therefore forces that generated the motion) are changed, and therefore the outcome of DTW does not preserve any of the dynamic characteristics of the original process that generated the data. In other words, if the time series was generated by a dynamical model, DTW does not respect the constraints imposed by the dynamics. This means that a large number of different ``actions'' are lumped together under DTW. The pioneering work of the psychologist Johansson, however, showed that a great deal of information can be encoded in the temporal signal (Figure \ref{fig-johansson}). This information is destroyed by DTW.

The next two sections describe DTW and point the way to generalizing it to take into account dynamic constraints. The reader who is uninterested in the subtleties of recognizing events or actions that have similar appearance but different temporal signatures can skip the rest of this chapter.

\section{Dynamic time warping revisited}

{\color{pea}
If we consider two time series of images, $I_1$ and $I_2$, where $I_j \doteq \{I_j(t)\}_{t=1}^T$, for simplicity assumed to have the same length,\footnote{The case of different lengths can be also considered at the cost of a more complex optimization.} then the simplest distance we could define is the ${L}^2$ norm of the difference, $d_0(I_1, I_2) = \int_0^T \| I_1(t) - I_2(t) \|^2dt$, which corresponds to a generative model where both sequences come from an (unknown) underlying process\footnote{The notation we use in this chapter abuses the symbols defined previously, but is chosen on purpose so that the final model will resemble those of previous chapters.} $\{h(t)\}$, corrupted by two different
realizations of additive white zero-mean Gaussian ``noise'' (here the word noise lumps all unmodeled phenomena, not necessarily associated to sensor errors)
\begin{equation}
I_j(t) = h(t) + n_j(t) ~~~ j = 1, 2; \ t \in [0, \ T]
\label{mod1}
\end{equation}
The ${L}^2$ distance is then the (maximum-likelihood) solution
for $h$ that minimizes
\begin{equation}
d_0(I_1, I_2)  = \min_h \phi_{data}(I_1, I_2  | h) \doteq \sum_{j=1}^2 \int_0^T \| n_j(t)\|^2dt
\label{eq-data}
\end{equation}
subject to (\ref{mod1}).} Here $h$ can be interpreted as the {\em   average} of the two time series, and although in principle $h$ lives in an infinite-dimensional space, no regularization is necessary at this stage, because the above has a trivial closed-form solution. However, later we will need to introduce regularizers, for instance of the form $\phi_{reg}(h) = \int_0^T \| \nabla h \|dt$. This admittedly unusual way of writing the ${L}^2$ distance makes the extension to more general models simpler, as we discuss in the next sections.

{\color{pea}
Consider now an arbitrary contrast transformation\footnote{Note that this is not the contrast transformation operating on the image values, introduced in Section \ref{sect-im-form}, but it is a contrast transformation applied to the temporal domain.} $m$ of the interval $[0, T]$, called a {\em time warping}, so that (\ref{mod1}) becomes
\begin{equation}
I_j(t) = h(m_j(t)) + n_j(t) ~~~ j = 1, 2.
\label{mod2}
\end{equation}
The data term of the cost functional we wish to optimize is still $\sum_{i=1}^2 \int_0^T \| n_j(t)\|^2dt$, but now subject to (\ref{mod2}), so that minimization is with respect to the unknown functions $m_1$ and $m_2$ as well as $h$. Since the model is under-determined, we must impose regularization \cite{kirsch96} to compute the {\em time-warping distance}
\begin{equation}
d_1(I_1, I_2) = \min_{h\in {\bf H}, m_j \in {\cal M}} \phi_{data}(I_1, I_2 | h, m_1, m_2) + \phi_{reg}(h).
\label{dist2}
\end{equation}
Here $\bf H$ is a suitable space where $h$ is assumed to live and $\cal M$ is the space of monotonic continuous (contrast) transformations. 
In order for $\tau \doteq m(t)$ to be a viable temporal index, $m$ must satisfy a number of properties. The first is continuity (time, alas, does not jump); in fact, it is common to assume a certain degree of smoothness, and for the sake of simplicity we will assume that $m_i$ is infinitely differentiable. The second is causality: The ordering of time instants has to be preserved by the time warping, which can be formalized by imposing that $m_i$ be monotonic. We can re-write the distance above as
\begin{equation}
\min_{h\in {\bf H}, m_i \in {\cal U}} \sum_{j=1}^2 \int_{0}^T \| I_j(t) - h(m_j(t)) \|^2 + \lambda \| \nabla h(t) \| dt
\label{dtw0}
\end{equation}
where $\lambda$ is a tuning parameter that can be set equal to zero, for instance by choosing $h(t) = I_1(m_1^{-1}(t))$, and the assumptions on the warpings $m_i$ are implicit in the definition of the set $\cal M$. This is an optimal control problem, that is solved globally using dynamic programming in a procedure called ``dynamic time warping'' (DTW). 
}

It is important to note that {\em there is nothing ``dynamic'' about dynamic time warping,} other than its name.  There is no requirement that the warping function $x$ be subject to dynamic constraints, such as those arising from forces, inertia etc. However, some notion of dynamics can be coerced into the formulation by characterizing the set $\cal M$ in terms of the solution of a differential equation. {\color{pea} Following \cite{ramsayS}, as shown by \cite{martinSE99}, one can represent allowable $x\in {\cal M}$ in terms of a small, but otherwise unconstrained, scalar function $u$: ${\cal M} = \{ m\in {H}^2([0, \ T]) \ | \ddot x = u \dot x; \ u\in {L}^2([0, T])\}$ where ${H}^2$ denotes a Sobolev space. 
If we define\footnote{$s_i$ is not to be confused with the scale parameter.} $s_i \doteq \dot m_i$ then $\dot s = u s$; we can then stack the two into\footnote{$g$ in this section is not to be confused with viewpoint; the notation $g$ is used because, consistent with the notation adopted earlier, it is an invertible nuisance.} $g \doteq [m, \  s]^T$, indicative of a group (invertible nuisance) and $C = [1, \ 0]$, and write the data generation model as
\begin{equation}
\begin{cases}
\dot g_j(t) = f(g_j(t)) + l(g_j(t))u_i(t) \\
I_j(t) = h(C g_j(t)) + n_i(t)
\label{mod5}
\end{cases}
\end{equation}
as done by \cite{martinSE99}, where $u_i \in {L}^2([0, T])$. Here $f, l$ and $C$ are given, and $h, m_j(0), u_i$ are nuisance parameters that are eliminated by minimization of the same old data term $\sum_{j=1}^2 \int_0^T \| n_j(t) \|^2dt$, now subject to (\ref{mod5}), with the addition of a regularizer $\lambda
\phi_{reg}(h)$ and an energy cost for $u_i$, for instance $\phi_{energy}(u_i) \doteq \int_{0}^T \| u_i \|^2dt$. Writing explicitly all the terms, the problem of dynamic time warping can be written as
\begin{equation}
d_3(I_1, I_2) = \min_{h, u_j, m_j} \sum_{j=1}^2 \int_0^T \| I_i(t) - h(C g_j(t)) \| + \lambda \| \nabla h(t) \| + \mu \| u_i(t) \| dt
\end{equation}
subject to $\dot g_j = f(g_i) + l(g_j) u_i$. Note, however, that this differential equation is only an expedient to (softly) enforce causality by imposing a small ``time curvature'' $u_i$. }

\section{Time warping under dynamic constraints}
\label{sect-twdc}

The strategy to enforce {dynamic constraints in dynamic time warping} is illustrated in Figure \ref{fig2}:
\begin{figure}[htb]
\begin{center}
\includegraphics[width=0.5\textwidth,height=0.3\textwidth]{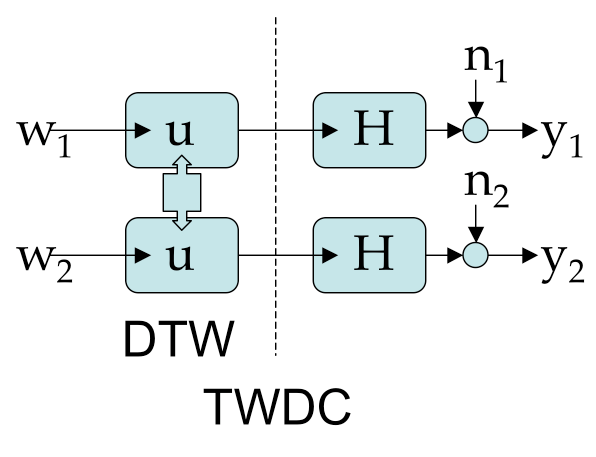}
\end{center}
\caption{\sl Traditional dynamic time warping (DTW) assumes that the data come from a common function that is warped in different ways to yield different time series. In time warping under dynamic constraints (TWDC), the assumption is that the data are the output of a dynamic model, whose inputs are warped versions of a common input function.}
\label{fig2}
\end{figure}
Rather than the data being warped versions of some common function, as in (\ref{mod2}), we assume that {\em the data are   outputs of dynamical models driven by inputs that are warped   versions of some common function.} {\color{pea} In other words, given two time series $\{I_i\}, i = 1, 2$, we will assume that there exist suitable matrices $A, B, C$, state functions $m_i$ of suitable dimensions, with their initial conditions, and a {\em common input} $u$ such that the data are generated by the following model, {\em for some warping  functions $w_i \in {\cal M}$:}
\begin{equation}
\begin{cases}
\dot m_j(t) = A m_j(t) + B u(w_j(t))  \\
I_j(t) = C m_j(t) + n_j(t).
\end{cases}
\label{mod6}
\end{equation}
Our goal is to find the distance between the time series by minimizing with respect to the nuisance parameters the usual data discrepancy $\sum_{j=1}^2 \int_0^T \| n_j(t) \|^2dt$ subject to (\ref{mod6}), together with regularizing terms $\bar \phi_{reg}(u)$ and with $w_j \in {\cal M}$. Notice that this model is considerably different from the one discussed in the previous section, as the state $g$ earlier was used to model the temporal warping, whereas now it is used to model the data, and the warping occurs at the level of the input.  It is also easy to see that the model (\ref{mod6}), despite being linear in the state, includes (\ref{mod5}) as a special case, because we can still model the warping functions $w_i$ using the differential equation in (\ref{mod5}). In order to write this {\em time warping   under dynamic constraint} problem more explicitly, we will use the
following notation: 
\begin{equation}
I(t) = Ce^{A t}I(0) + \int_0^t C e^{A(t-\tau)}Bu(w(\tau))d\tau
\doteq L_0(x(0)) + L_t(u(w))
\label{def-L}
\end{equation}
in particular, notice that $L_t$ is a convolution operator, $L_t(u) = F*u$ where $F$ is the transfer function. We first address the problem where $A, B, C$ (and therefore $L_t$) are given.  For simplicity we will neglect the initial condition, although it is easy to take it into account if so desired. In this case, we define the distance between the two time series
\begin{equation}
d_4(I_1, I_2) = \min \sum_{j=1}^2 \int_0^T \| I_j(t) - L_t(u_j(t)) \|
+ \lambda \| u_j(t) - u_0(w_j(t)) \|dt
\end{equation}
subject to $u_0 \in {\bf H}$ and $w_j \in {\cal M}$. Note that we have introduced an auxiliary variable $u_0$, which implies a possible discrepancy between the actual input and the warped version of the common template. This problem can be solved in two steps: A deconvolution, where $u_i$ are chosen to minimize the first term, and a standard dynamic time warping, where $w_i$ and $u_0$ are chosen to minimize the second term. Naturally the two can be solved
simultaneously.

When the model parameters $A, B, C$ are common to the two models, but otherwise unknown, minimization of the first term corresponds to {\em blind system identification,} which in general is ill-posed barring some assumption on the class of inputs $u_i$.  These can be imposed in the form of generic regularizers, as common in the literature of blind deconvolution \cite{giannakisM89}, or by restricting the classes of inputs to a suitable class, for instance sparse ``spikes'' \cite{raptisWS10}.  This is a general and broad problem, beyond our scope here, so we will forgo it in favor of an approach where the input is treated as the output of an auxiliary dynamical model, also known as {\em exo-system} \cite{isidori89}. This combines standard DTW, where the monotonicity constraint is expressed in terms of a double integrator, with TWDC, where the actual stationary component of the temporal dynamics is estimated as part of the inference. The generic warping $w$, the output of the exo-system  satisfies
\begin{equation}
\begin{cases}
\dot w_j(t) = g_j(t), ~~~ j = 1, 2 \\
\dot g_j(t) = v_j(t) g_j(t)
\end{cases}
\label{twdi}
\end{equation}
and\footnote{$w_j$ in this section is not to be confused with a diffeomorphism of the image domain. It is, however, a diffeomorphism of the time domain, hence the choice of notation.} $w_j(0) = 0, \ w_j(T) = T$. This is a multiplicative double integrator; one could conceivably add layers of random walks, by representing $v_i$ as Brownian motion. Combining this with the time-invariant component of the realization yields the generative model for the time series $I_i$:
\begin{equation}
\begin{cases}
\dot w_j(t) = g_j(t), ~~~ i = 1, 2 \\
\dot g_j(t) = v_i(t) g_j(t) \\
\dot m_j(t) = A m_j(t) + B u(w_j(t)) \\
I_j(t) = C m_j(t) + n_j(t).
\end{cases}
\label{modelfinal}
\end{equation}
Note that the actual input function $u$, as well as the model parameters $A, B, C$, are common to the two time series. A slightly relaxed model, following the previous subsection, consists of defining $u_i(t) \doteq u(w_i(t))$, and allowing some slack between the two; correspondingly, to compute the distance one would have to minimize the data term
\begin{equation}
\phi_{data}(I_1, I_2  | u, w_i, A, B, C) \doteq \sum_{j=1}^2 \int_0^T \| n_j(t) \|^2dt
\end{equation}
subject to (\ref{modelfinal}), in addition to the regularizers
\begin{equation}
\bar \phi_{reg}(v_i, u) = \sum_{i=1}^2 \int_0^T \| v_i(t) \|^2 + \| \nabla u (t) \|^2 dt
\end{equation}
which yields a combined optimization problem
\begin{equation}
d_5(I_1, I_2) = \min_{u, v_j\in {L}^2, A, B, C}  \sum_{j=1}^2 \int_0^T (\| I_j(t)-C m_j(t) \|^2 + \| v_j(t) \|^2 + \| \nabla u(t) \|^2) dt
\label{eq-twdc}
\end{equation}
subject to (\ref{modelfinal}). This distance can be either computed in a globally optimal fashion on a discretized time domain using dynamic programming, or we can run a gradient descent algorithm based on the first-order optimality conditions.} The bottom line is that one could use any of the distances introduced in this section, $d_0, \dots, d_5$, to compare time series, corresponding to different forms of marginalization or max-out. 

This concludes the treatment of descriptors. The next chapters focus on how to construct a representation from data, and how to aggregate different objects under the same category as part of the learning process.

\chapter{Visual exploration}
\label{sect-exploration}
\index{Actionable information}
\index{Information!actionable}
\index{Actionable Information Gap}
\index{Visual exploration}
\index{Exploration}

In this chapter we study the inverse problem of hallucination, introduced in Section \ref{sect-hallucination}, that is the problem of exploration. Whereas hallucination produces images {\em given} a representation; exploration produces a representation {\em given} images. The goal of exploration is to actively control the data acquisition process in such a way that aggregating Actionable Information eventually yields a complete representation. As we acquire more and more data, the hope is that the set of representations that are {\em compatible} with the data shrinks, although not necessarily to a singleton (a complete representation is not necessarily unique). In other words, we hope that the exploration process will reduce the uncertainty on the representation. When and if the inferred representation is complete, it can synthesize the light-field of the original scene up to the uncertainty of the sensors, and we say we have performed {\em sufficient exploration}. \index{Sufficient exploration} Exploration can be more or less {\em efficient}, in the sense that sufficiency can be achieved with a varying cost of resources (\eg time, energy).

Exploration is the process that links maximal invariants, $\phi^\wedge(I)$, whose complexity we called Actionable Information (AI), to minimal sufficient statistics of a complete representation, whose complexity we called complete information (CI). Because of non-invertible nuisances, the {\em gap} between AI and CI can be filled by exercising some form of control on the sensing process. Such control could be exercised in {\em data space}, for instance by choosing the most informative features, or in {\em sensor space}, for instance by selecting sensing assets in a sensor network, or in {\em physical space}, for instance by moving around an occlusion or moving closer to an unresolved region of the scene.

We start by designing a {\em myopic explorer}, driven simply by the maximization of Actionable Information. Ideally, by accumulating AI, one would hope to converge to the CI. Unfortunately, this is in general not the case for such a myopic explorer. We therefore consider a slightly less primitive explorer seeking to maximize its reward over a {\em receding horizon}. Again, in general, this strategy is not guaranteed to achieve complete efficient exploration. Thus, the modeling exercise points to the need to endow the explorer with {\em memory} that can summarize in finite complexity the results of past explorations. Such a memory will be precisely the representation we are after, and we discuss inference criteria that can drive the building of a representation. In some simple instances, such criteria yield viable computational procedures.

In all cases, occlusions and scaling/quantization play a critical role in exploration, a process that can be thought of as the {\em inversion} of such nuisances. Therefore, at the outset, we will need efficient methods to perform {\em occlusion detection}, to be used either instantaneously -- by comparing two temporally adjacent images, as in myopic memoryless exploration -- or incrementally, by comparing each new image with the one hallucinated from the current representation.

\section{Occlusion detection}
\label{sect-occlusion}
\index{Occlusion}

\begin{figure*}[bth!]
  \centerline 
      {
        \hbox
            {
              \begin{tabular}{ccc}
               {\includegraphics[width=0.25\textwidth]{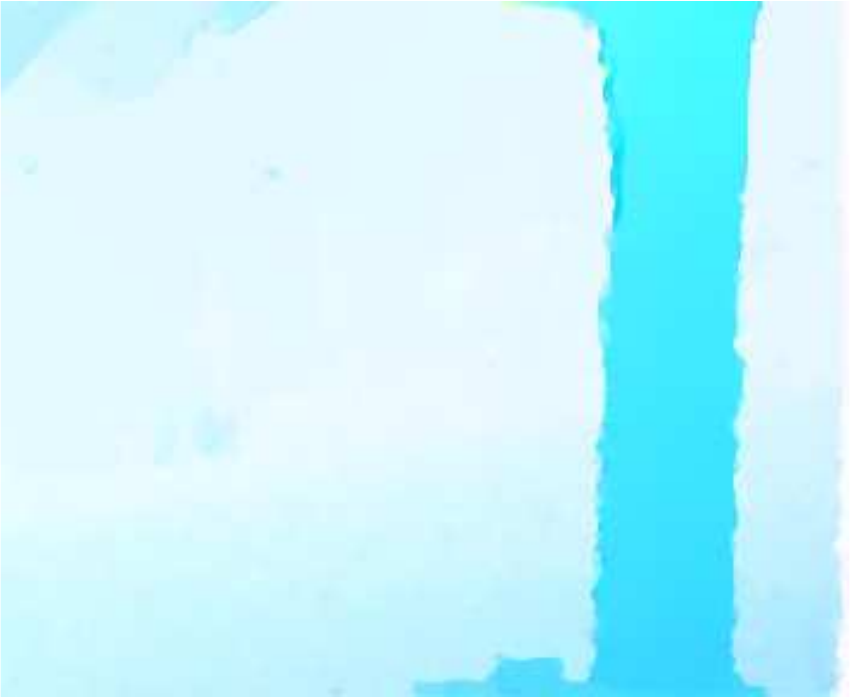}}      
               {\includegraphics[width=0.25\textwidth]{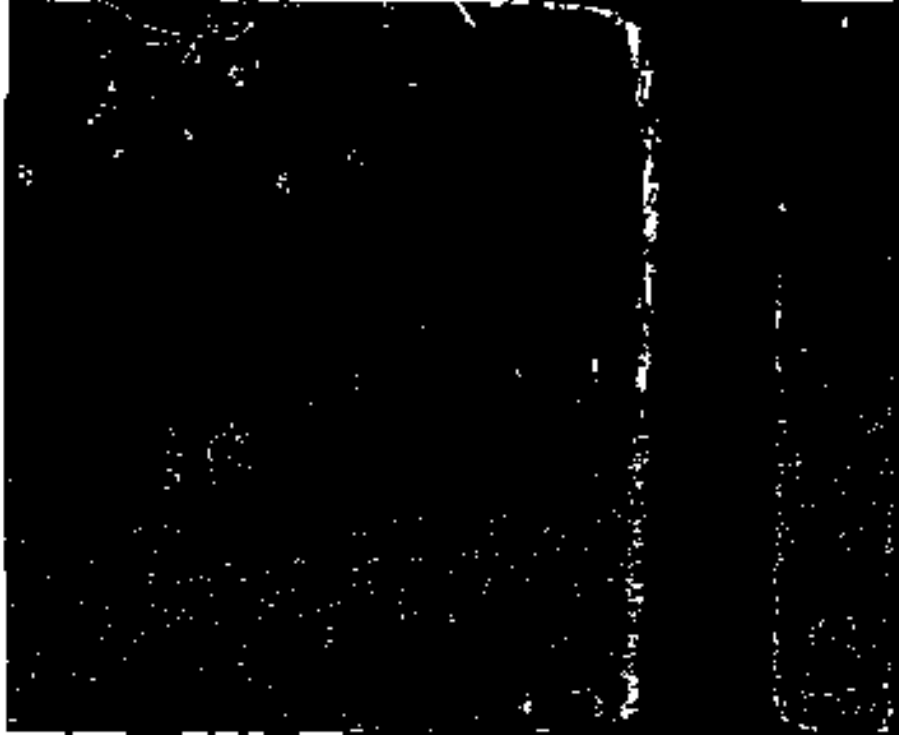}}      
               {\includegraphics[width=0.25\textwidth]{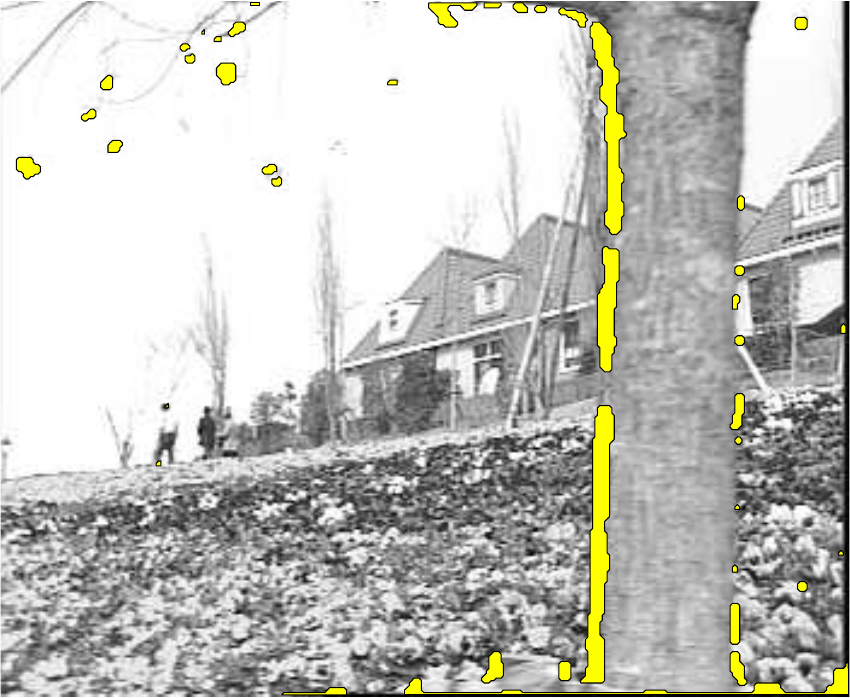}}      \\ 
               {\includegraphics[width=0.25\textwidth]{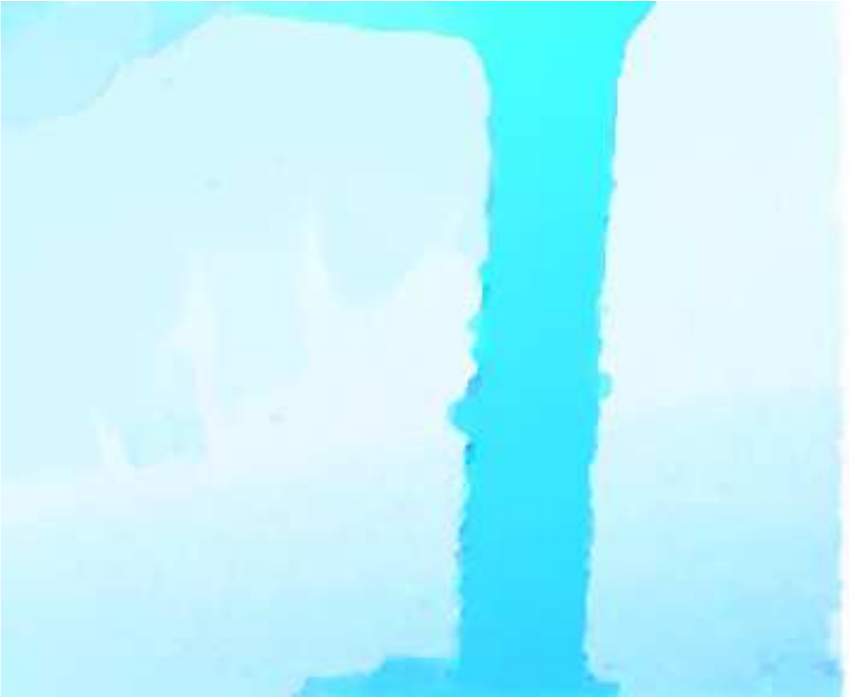}}      
               {\includegraphics[width=0.25\textwidth]{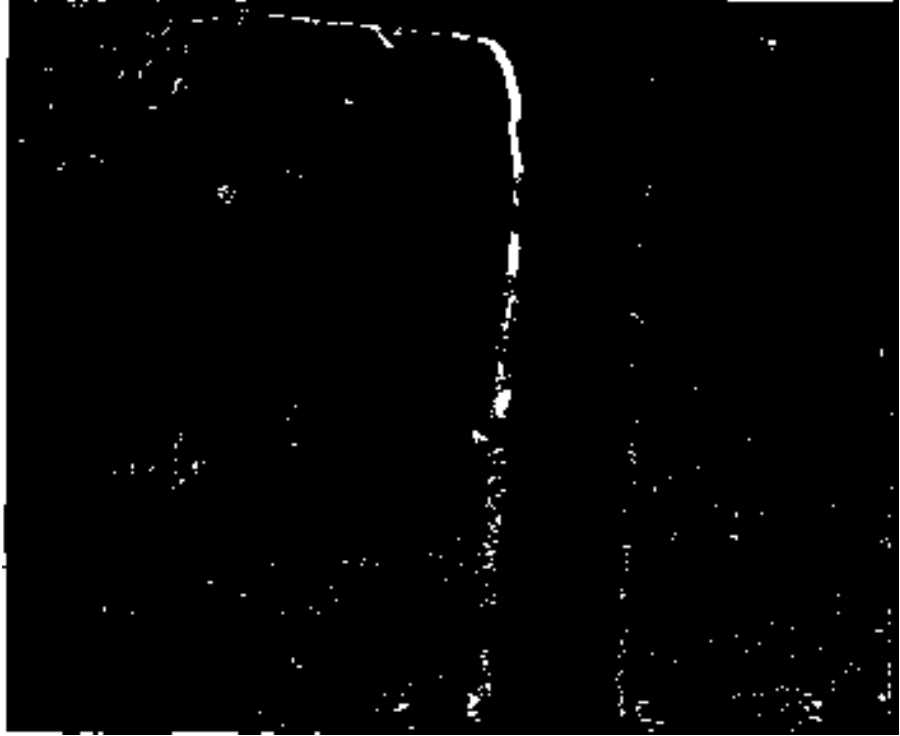}}      
               {\includegraphics[width=0.25\textwidth]{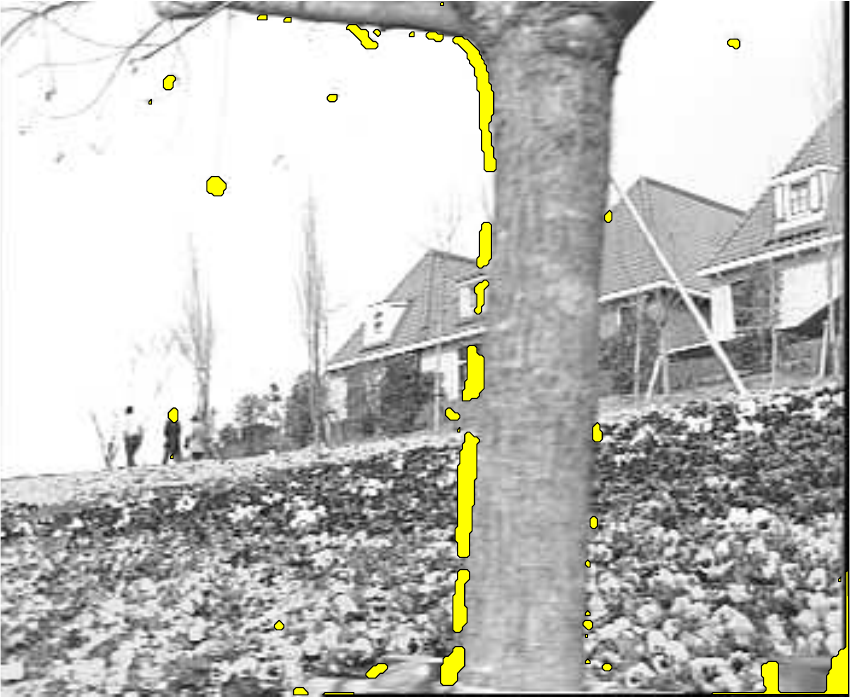}}      \\ 
               {\includegraphics[width=0.25\textwidth]{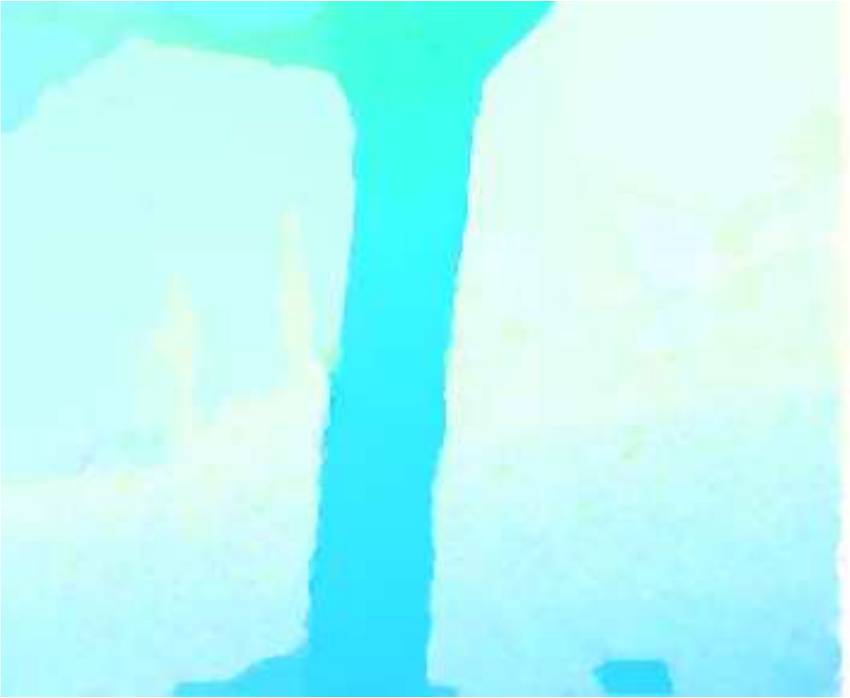}}      
               {\includegraphics[width=0.25\textwidth]{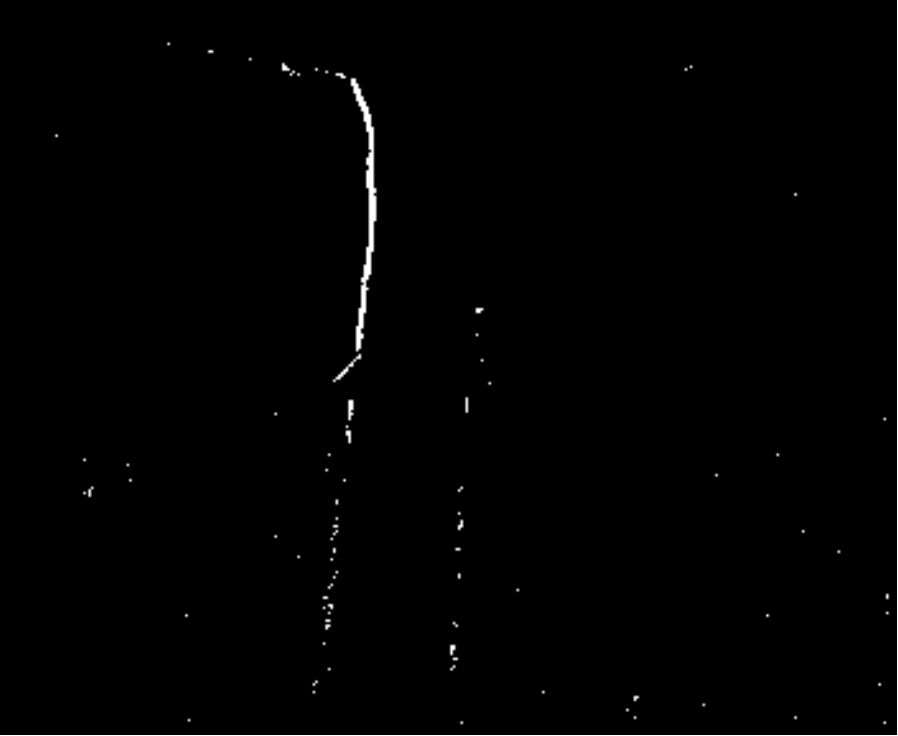}}      
               {\includegraphics[width=0.25\textwidth]{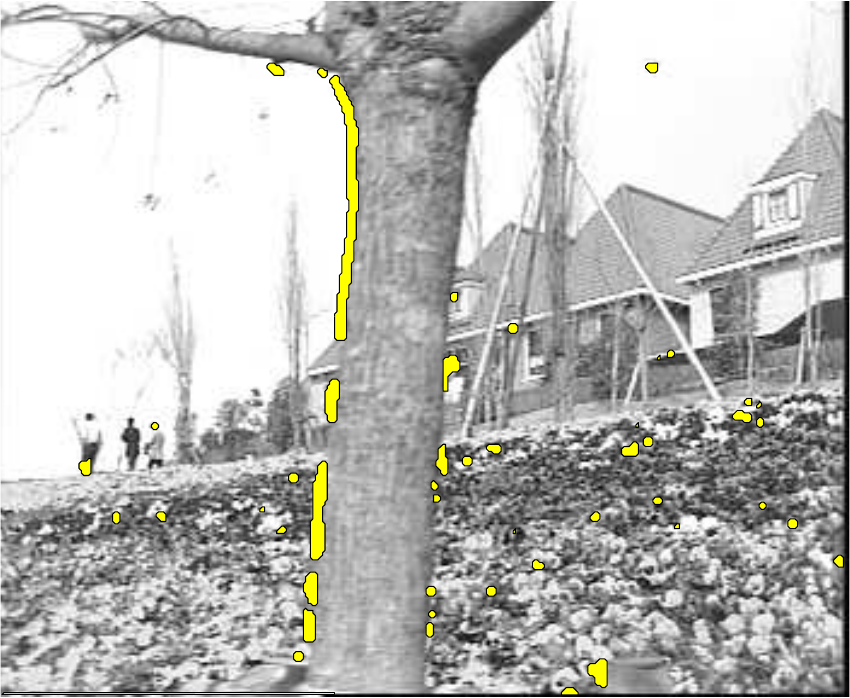}}      \\ 
               {\includegraphics[width=0.25\textwidth]{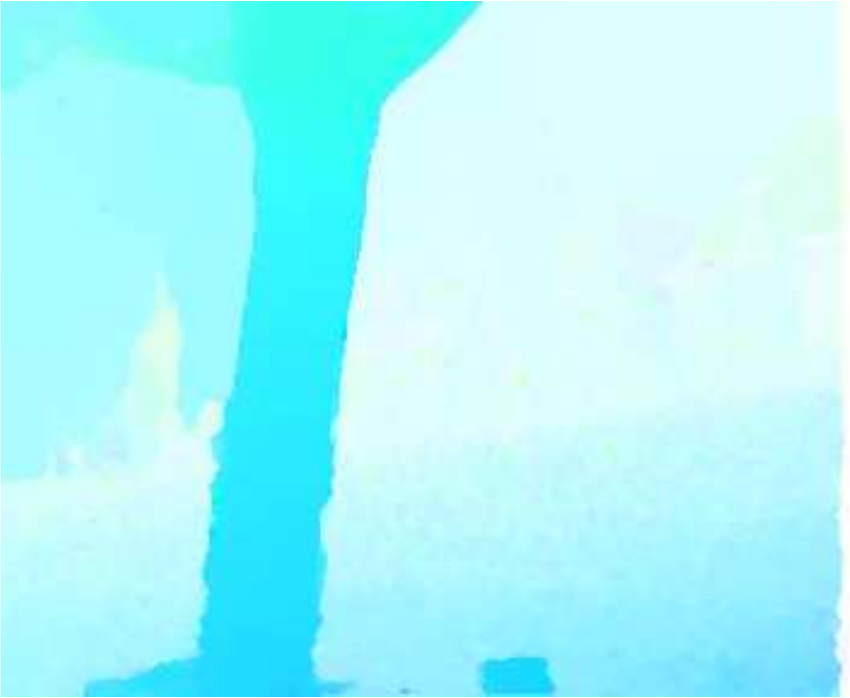}}      
               {\includegraphics[width=0.25\textwidth]{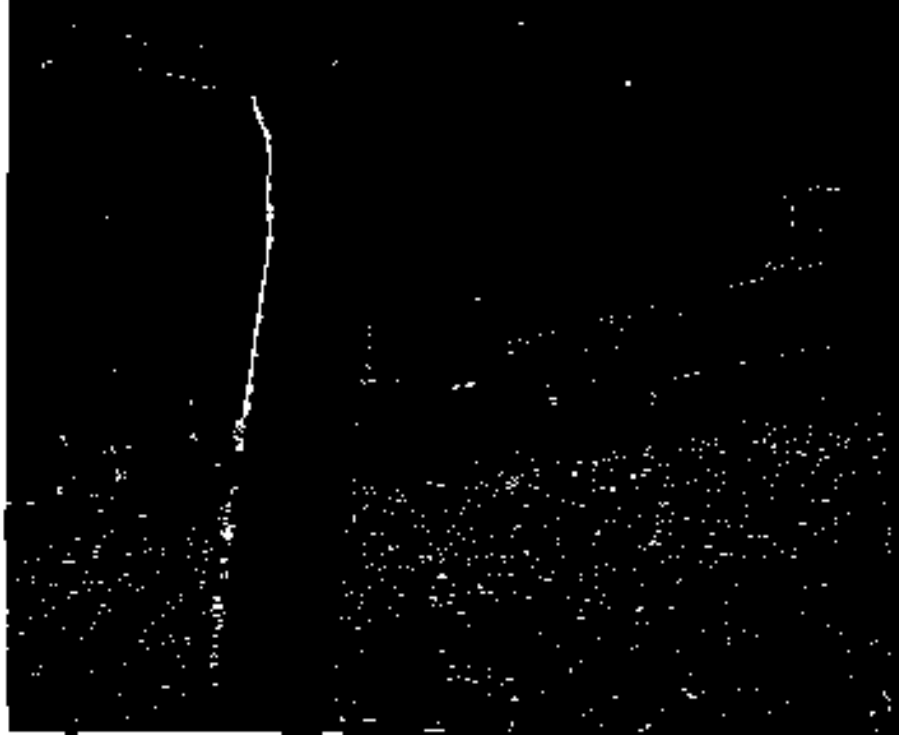}}      
               {\includegraphics[width=0.25\textwidth]{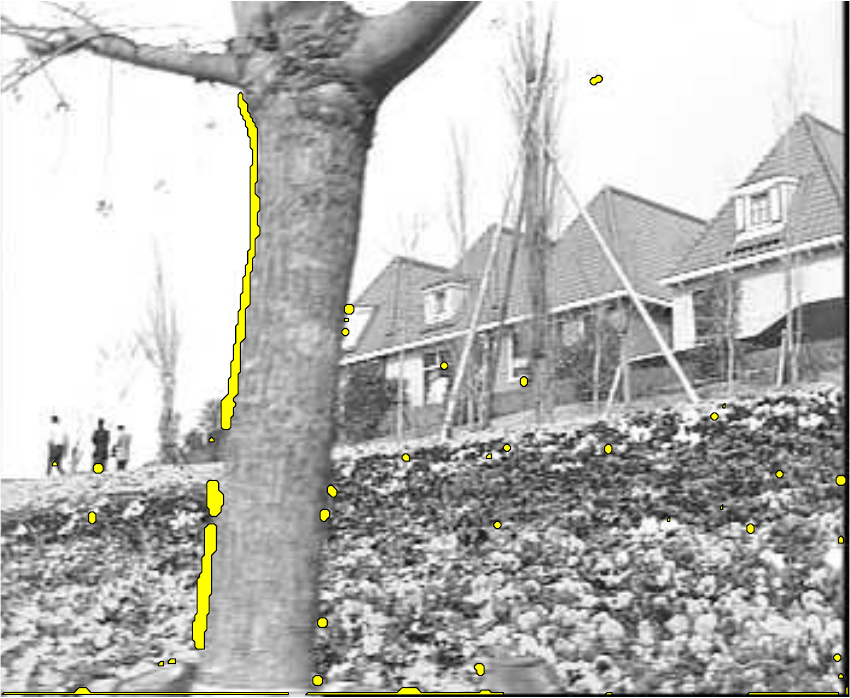}}      \\ 
              \end{tabular}
            }
      }
      \caption{\sl Motion estimates for the Flower Garden sequence (left), residual $e$ (center), and occluded region (right) (courtesy of \cite{ayvaciRS10}).}
     \label{figure:flower:garden}
\end{figure*}

In this section we summarize the results of \cite{ayvaciRS10}, where it is shown that occlusion detection can be formulated as a variational optimization problem and relaxed to a convex optimization, that yields a globally optimal solution. We do not delve on the implementation of the solution, for which the reader is referred to \cite{ayvaciRS10}, but we describe the formalization of the problem. The reader uninterested in the specifics can skip the rest of this section and assume that the region $\Omega(t) \subset D \subset \real^2$ that is not {\em co-visible} in two adjacent images has been inferred. Note that this region is not necessarily simply connected or regular, as Figure \ref{fig-flower-occlusion} illustrates.

\begin{rem}[Optical flow and motion field]
{\em Optical flow} \index{Optical flow} \marginpar{\tiny \sc optical flow}  refers to the deformation of the domain of an image that results from ego- or scene motion. It is, in general, different from the {\em motion field}, (\ref{eq-motion-field}), that is the projection onto the image plane of the spatial velocity of the scene \cite{verri}, unless three conditions are satisfied: Lambertian reflection, constant illumination, and co-visibility. We have already adopted the Lambertian assumption in Section \ref{sect-im-form}, and it is true that most surfaces with benign reflectance properties (diffuse/specular) can be approximated as Lambertian almost everywhere under diffuse or sparse illuminants (\eg the sun). In any case, widespread violation of Lambertian reflection does not enable correspondence\cutTwo{ \cite{soattoY02cvpr}}, so we will embrace it like the majority of existing approaches\cutTwo{ to motion estimation}. Similarly, constant illumination is a reasonable assumption for ego-motion (the scene is not moving relative to the light source), and even for objects moving (slowly) relative to the light source. Co-visibility is the third and crucial assumption that has to do with occlusions and until recently has been widely neglected in the computation of optical flow. If an image contains portions of its domain that are {\em not visible} in another image, these can \cutThree{patently} {\em not} be mapped onto it by optical flow vectors. Constant visibility is often assumed because optical flow is defined in the limit where two images are sampled infinitesimally close in time, in which case {\em there are no occluded regions,} and one can focus solely on discontinuities of the motion field. But the problem is {\em not} that optical flow is discontinuous in the occluded regions; it is simply not defined; {\em it does not exist.} By definition, an occluded region cannot be explained by a displaced portion of a different image, since it is not visible in the latter. Motion in occluded regions can be {\em hallucinated} ({\em extrapolated}, or {\em ``inpainted}'') but not {\em validated on data.}\cut{ Thus, the great majority of variational motion estimation approaches provide an estimate of a dense flow field, defined at each location on the image domain, {\em including occluded regions.} In their defense, it can be argued that\cutTwo{ even if we do not take the limit,} for small parallax (slow-enough motion, or far-enough objects, or fast-enough temporal sampling) occluded areas are {\em small.} However, small does not mean unimportant, as occlusions are critical to perception \cite{gibson84} and a key for developing representations for recognition.}
\end{rem}
\begin{rem}[The Aperture problem]
\label{rem-aperture}
The aperture problem refers to the fact that the motion field at a point can only be determined in the direction of the gradient of the image at that point (Figure \ref{fig-aperture}). Therefore, occlusions can only be determined if the gradient of the radiance intersects transversally the boundary of the occluded region. When this does not happen (for instance when a region with constant radiance occludes another one with constant radiance), an occlusion cannot be positively discerned from a material boundary or an illumination boundary. For instance, in the Flower Garden sequence (Figure \ref{figure:flower:garden}), the (uniform) sky could be attached to the branches of the tree, or it could be occluded by them. Often, priors or regularizers are imposed to resolve this ambiguity, but again such a choice is not validated from the data.  We prefer to maintain the ambiguities in the representation, deferring the decision to the exploration process, when data becomes available to disambiguate the solution (\eg the tree passing in front of a cloud). This naturally occurs over time as the initial representation converges towards the complete representation.
\end{rem}
{\color{pea} For clarity, we denote a time-varying image via $I:D \subset \real^2\times \real^+ \rightarrow \real^+; \ (x,t) \mapsto I(x,t)$. Under the Lambertian assumption, as we have seen in Section \ref{sect-correspondence},  the relation between two consecutive images in a video $\{I(x,t)\}_{t=0}^T$ is given by
\begin{equation}
I(x,t+dt) = \begin{cases} I(w(x,t),t) + n(x,t), ~~~ x\in D\backslash \Omega(t;dt) \\ 
\nu(x,t),  ~~~ x\in \Omega(t;dt)
\end{cases} 
\end{equation}
where $w:D\times \real^+\rightarrow \real^2$ is the {\em optical flow}\footnote{Sometimes the deformation field $w$ is represented in the form $w(x) = x + v(x)$ and the term ``optical flow'' refers to $v(x)$. The two representations are equivalent, for one can obtain $w$ from $v$ as above, and $v$ from $w$ via $v(x) = w(x)  - x$. Therefore, we will not make a distinction between the two here, and simply call $w$ the optical flow.} (\ref{eq-opt-fl}) which approximates the motion field (\ref{eq-motion-field}) and is defined everywhere {\em except at occluded regions} $\Omega$. These can change over time and depend on the temporal sampling interval $dt$; $\Omega$ is in general not simply-connected, so even if we call $\Omega$ the occluded region (singular), it is understood that it is made of multiple connected components.\footnote{Multiple connected component means that each component is a compact simply connected set, that is disjoint from other components, it does not mean that the individual components of the occluded domain $\Omega$ are connected {\em to each other}.} \index{Multiply-connected components} \index{Simply-connected components} In the occluded region, the image can take any value $\nu:\Omega \times \real^+ \rightarrow \real^+$ that is in general unrelated to $I(x,t)_{|_{x\in \Omega}}$. Because of (almost-everywhere) continuity of the scene and its motion (i), and because the additive term $n(x,t)$ compounds the effects of a large number of independent phenomena and therefore we can invoke the Law of Large Numbers (ii), in general we have that
\begin{equation}
{\rm (i)}~~  \lim_{dt\rightarrow 0} \Omega(t;dt) = \emptyset, ~~~ {\rm and} ~~~
{\rm (ii)} ~~ n\stackrel{IID}{\sim} {\cal N}(0, \lambda)
\label{eq-i}
\end{equation}
\ie, the additive uncertainty is normally distributed in space and time with an isotropic and small variance $\lambda>0$. Note that we are using the (overloaded\footnote{We had already used $\lambda$ to indicate the loss function in Section \ref{sect-visual-decisions}, but that should cause no confusion here.}) symbol $\lambda$ for the covariance of the noise, whereas usually we have employed the symbol $\sigma$. The reason for this choice will become clear in Equation (\ref{eq-cost-occl}), where $\lambda$ will be interpreted as a Lagrange multiplier. \marginpar{\tiny \sc lagrange multiplier} \index{Lagrange multiplier} We define the residual $$e:D\rightarrow \real^+\cutTwo{; (x,t) \mapsto e(x,t;dt) \doteq I(x,t+dt)-I(w(x,t),t)}$$ on the entire image domain $x\in D$, via (for simplicity we omit the arguments of the functions when clear from the context)
\begin{equation}
e(x,t;dt) \doteq I(x,t+dt)-I(w(x,t),t) =  \begin{cases}
n(x,t), ~~~ x\in D\backslash \Omega\cut{(t;dt)} \\
\nu(x,t) - I(w(x,t),t) ~~~~ x\in \Omega\cut{(t;dt)}
\end{cases}
\end{equation}
which we can write as the sum of two terms, $e_1: D \rightarrow \real^+$ and $e_2:D\rightarrow \real^+$, also defined on the entire domain $D$ in such a way that 
\begin{equation}
\begin{cases}
e_1(x,t;dt) = \nu(x,t) - I(w(x,t),t), ~~~ x\in \Omega\cut{(t;dt)} \\
e_2(x,t;dt) = n(x,t), ~~~~~~ x\in D\backslash \Omega\cut{(t;dt)}.
\end{cases}
\end{equation}
Note that $e_2$ is undefined in $\Omega$, and $e_1$ is undefined in $D\backslash \Omega$, in the sense that they can take any value there, including zero, which we will assume henceforth. We can then write, for any $x\in D$, 
\begin{equation}
I(x,t+dt) = I(w(x,t),t) + e_1(x,t;dt) + e_2(x,t;dt)
\label{eq-mod+}
\end{equation}
and notice that, because of (i), $e_1$ is {\em large but sparse},\footnote{In the limit $dt\rightarrow 0$, ``sparse'' stands for almost everywhere zero; in practice, for a finite temporal sampling $dt > 0$, this means that the area of $\Omega$ is small relative to $D$. Similarly, ``dense'' stands for almost everywhere non-zero.} while because of (ii) $e_2$ is {\em small but dense}\footnote{In the limit $dt\rightarrow 0$, ``sparse'' stands for almost everywhere zero; in practice, for a finite temporal sampling $dt > 0$, this means that the area of $\Omega$ is small relative to $D$. Similarly, ``dense'' stands for almost everywhere non-zero.}. We will use this as an inference criterion for $w$, seeking to optimize a data fidelity \index{Data fidelity} criterion that minimizes the $L^0$ norm of $e_1$ (a proxy of the area of $\Omega$), and the log-likelihood \index{Likelihood!log} \index{Log-likelihood} of $n$ (the Mahalanobis \index{Mahalanobis norm} norm relative to the variance parameter $\lambda$) 
\begin{eqnarray}
\psi_{\rm data}(w,e_1) \doteq \| e_1 \|_{{L}^0(D)} + \frac{1}{\lambda} \| e_2 \|_{{L}^2(D)} ~~~ {\rm subject \ to \ } (\ref{eq-mod+}) \label{eq-psi-data} \\
= \frac{1}{\lambda}\| I(x,t+dt) - I(w(x,t),t) - e_1 \|_{{L}^2(D)} +  \|e_1 \|_{{L}^1(D)} \nonumber
\end{eqnarray}
where  $\| f\|_{L^0(D)} = \int \chi_{f(x) \neq 0}(x)dx$ is relaxed as usual to $\| f \|_{{L}^1(D)} \doteq \int_D |f(x)|dx$ and $\| f \|_{{L}^2(D)} \doteq \int_D |f(x)|^2dx$. Since we wish to make the cost of an occlusion independent of the brightness of the (un-)occluded region, we can replace the $L^1$ norm with a re-weighted $\ell^1$ norm that provides a better approximation of the $L^0$ norm \cite{ayvaciRS11IJCV}.

 Unfortunately, we do not know anything about $e_1$ other than the fact that it is sparse, and that what we are looking for is $\chi_\Omega \propto e_1$, where $\chi:D\rightarrow \real^+$ is the characteristic function that is non-zero when $x\in \Omega$, \ie, where the occlusion residual is non-zero. So, the data fidelity term depends on $w$ {\em but also on the characteristic function of the occlusion domain} $e_1$. Using the formal (differential) notation for first-order differences  (note that this is just a symbolic notation, as we do not let $dt \rightarrow 0$, lest we would have $\Omega \rightarrow \emptyset$)

\begin{eqnarray}
\nabla I(x,t) &\doteq& \ba{c} I\left(x+\ba{c}1\\0\ea,t\right) - I(x,t) \\
 I\left(x+\ba{c}0\\1\ea,t\right) - I(x,t) \ea^T \\
I_t(x,t) &\doteq& I(x,t+dt)-I(x,t) \\
v(x,t) &\doteq& w(x,t) - x \label{eq-v-deform}
\end{eqnarray}

\noindent we can approximate, for any $x\in D\backslash \Omega$, 
\begin{equation}
I(x,t+dt) = I(x,t) + \nabla I(x,t) v(x,t) + n(x,t)
\end{equation}
where the linearization error has been incorporated into the uncertainty term $n(x,t)$. Therefore, following the same previous steps, we have
\begin{eqnarray}
\boxed{\psi_{\rm data}(v,e_1) = \int_D |\nabla I v  + I_t - e_1|^2 dx +  \lambda \int_D |e_1 |dx}
\label{eq-cost-occl}
\end{eqnarray}
Since we typically do not know the variance $\lambda$ of the process $n$, we will treat it as a tuning parameter, and because $\psi_{\rm data}$ or $\lambda \psi_{\rm data}$ yield the same minimizer, we have attributed the multiplier $\lambda$ to the second term. In addition to the data term, because the unknown $v$ is infinite-dimensional, we need to impose regularization, \index{Regularization} for instance by requiring that the total variation (TV) of $v$ be small \index{Total variation}
\begin{equation}
\psi_{\rm reg}(v) = \mu \int_D \| \nabla v \|dx
\end{equation}
where $\mu$ is a multiplier to weigh the strength of the regularizer. TV is desirable in the context of occlusion detection because it does not excessively penalize motion discontinuities, that can occur where motion is parallel to the occluding boundary (so there is an occluding boundary without an occluded region, for instance where the direction of motion is parallel to the occluding boundary, see Remark \ref{rem-aperture}). The overall problem can then be written as the minimization of the cost functional $\psi = \psi_{\rm data} + \psi_{\rm reg}$, which in a discretized domain (the lattice $D \cap {\mathbb Z}^2$) becomes 
\begin{equation}
{\hat v, \hat e_1 = \arg\min_{v, e}  \underbrace{\| \nabla I v + I_t - e \|^2_{\ell^2} + \lambda \| e \|_{\ell_{rw}^1} + \mu \|\nabla v \|_{{\ell}^1}}_{\psi(v,e)}}
\label{eq-optim}
\end{equation}
where $\ell^1_{rw}$ is the re-weighted $\ell^1$ norm \cite{ayvaciRS11IJCV} and $\ell^1, \ell^2$ are the finite-dimensional version of the functional norms $L^1(D), L^2(D)$.
}
The remarkable thing about this minimization problem is that $\psi$ is {\em convex} in the unknowns $v$ and $e$. Therefore, convergence to a global solution can be guaranteed for an iterative algorithm regardless of initial conditions \cite{boydV04}. {This follows immediately after noticing that the ${\ell}^2$ term  
\begin{equation}
\| \underbrace{[\nabla I , -Id]}_{A}\ba{c}v\\ e\ea + \underbrace{I_t}_{b} \|_{{\ell}^2}
\label{eq-def-Ab}
\end{equation} 
is linear in both $v$ and $e$, the gradient (first-difference) $\nabla v$ is a linear operator, and the ${\ell}^1$ norm is a convex function.} 

Not only is this result appealing on analytical grounds, but is also enables the deployment of a vast arsenal of efficient optimization schemes, as proposed in \cite{ayvaciRS10}. 
\cut{\begin{theorem}
There exists a global minimizer of $\psi(v,e)$.
\end{theorem}
}
In practice, although the global minimizer does not depend on the initial conditions, it does depend on the multipliers $\lambda$ and $\mu$, for which no ``right choice'' exists, so an empirical evaluation of the scheme (\ref{eq-optim}) is necessary \cite{ayvaciRS10}.

It is tempting to impose some type of regularization on the occluded region $\Omega$ (for instance that its boundary be ``short''), but we have refrained from doing so, for good reasons. In the presence of smooth motions, the occluded region is guaranteed to be small, but under no realistic circumstances is it guaranteed to be {\em simple}. For instance, walking in front of a barren tree in the winter, the occluded region is a very complex, multiply-connected, highly irregular region that would be very poorly approximated by some compact regularized blob (Figure \ref{fig-flower-occlusion}).
\begin{figure}[htb]
\begin{center}
\includegraphics[width=.3\textwidth]{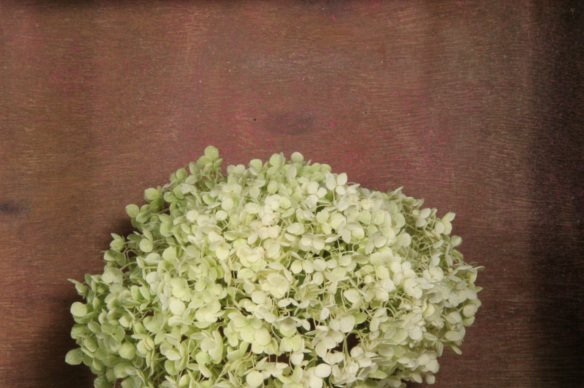}
\includegraphics[width=.3\textwidth]{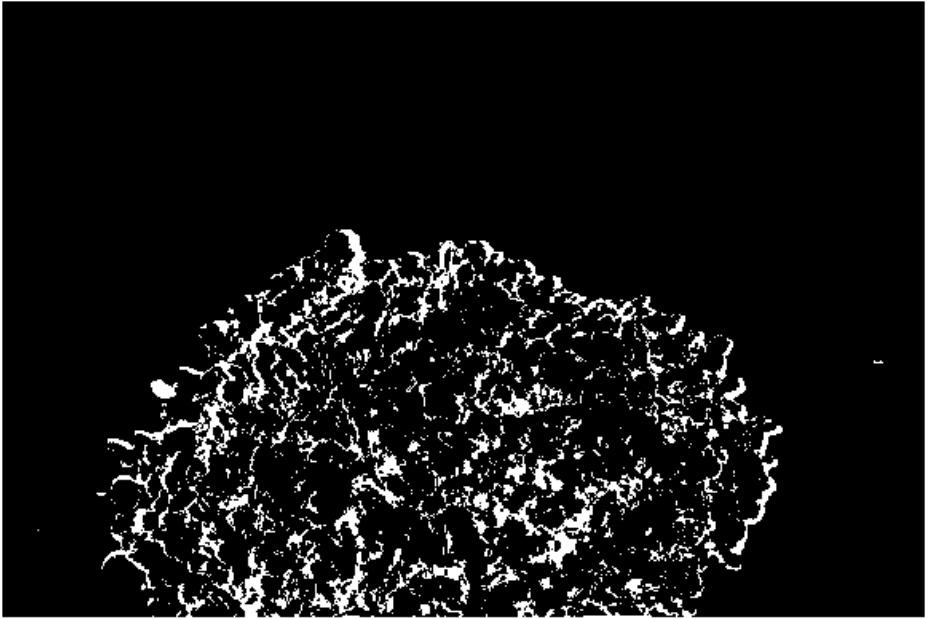}
\includegraphics[width=.3\textwidth]{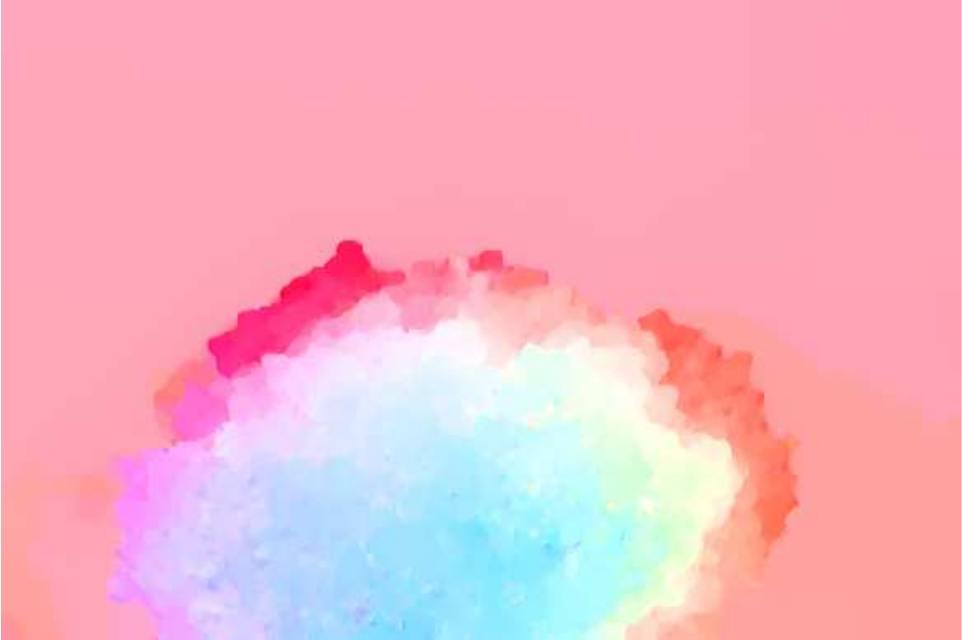}
\end{center}
\caption{\sl Occlusion regions for natural scenes (left) can have rather complex structure (middle) that would be destroyed if one were to use generic regularizers for the occluded region $\Omega$. The disparity maps (right) should only be evaluated in the co-visible regions (black regions in the middle figure), since disparity cannot be ascertained in the occluded region.}
\label{fig-flower-occlusion}
\end{figure}

Source code to perform simultaneous motion estimation and occlusion detection is available on-line from \cite{ayvaciRS10}. It can be used as a seed to detect {\em detached objects}, or better {\em detachable objects} \cite{ayvaciS11}, or to temporally integrate occlusion information so as to infer {\em depth layers} as done in \cite{jacksonYS05,jacksonYS06}. We will, however, defer the temporal integration of occlusion information to the next sections, where we discuss the exploration process. For now, what matters is that it is possible, given either two adjacent images, $I_t, I_{t+1}$, or an image predicted via hallucination, for instance $\hat I_{t+1}$ and a real image $I_{t+1}$, to determine both the deformation taking one image onto the other in the {\em co-visible} region, and the characteristic function of such a region, that is the {\em occluded domain}.

\begin{figure}[htb]
\begin{center}
\includegraphics[width=.6\textwidth]{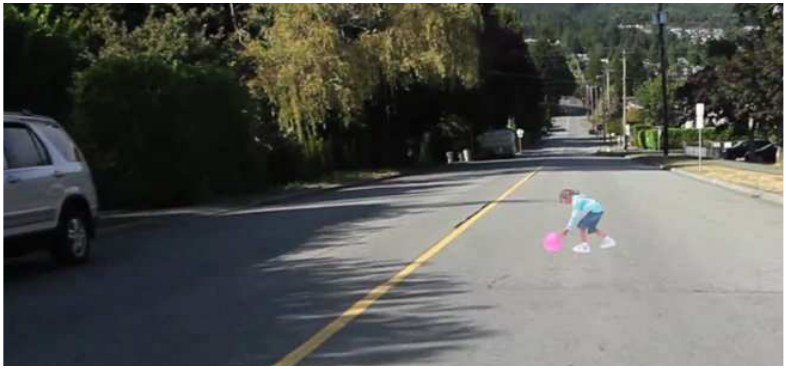}
\includegraphics[width=.6\textwidth]{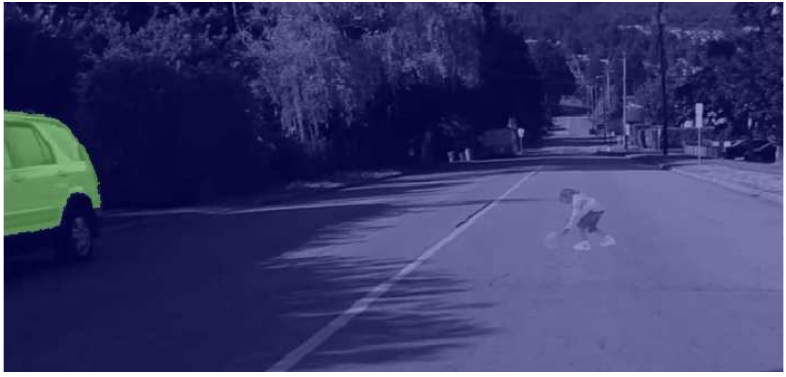}
\includegraphics[width=.6\textwidth]{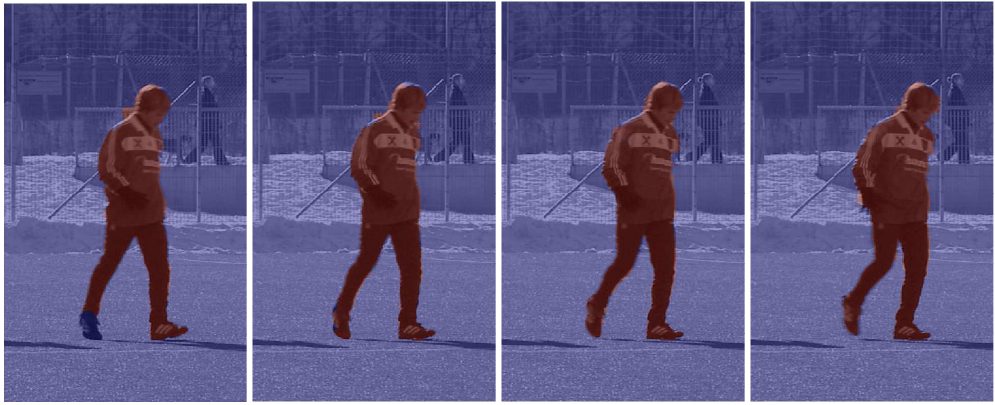}
\end{center}
\caption{\sl {\bf Detachable object detection} (courtesy of \cite{ayvaciS11}). The ``girl'' in Figure \ref{fig-illusions} and the soccer player would be similarly classified as {\em bona-fide} ``objects'' by a passive classification/detection/segmentation scheme. However, an extended temporal observation during either ego-motion (top) or object motion (bottom) correctly classifies the car on top and the soccer player on the bottom as ``detached objects'' but not the girl painted on the road pavement (middle).}
\label{fig-detachable}
\end{figure}

In the next section we begin discussing the implication of occlusion detection for exploration.

\section{Myopic exploration}

Consider an image $I_t$, taken at time $t$, and imagine using it to predict  the next image, $I_{t+1}$. In the next section we describe this process in the absence of non-invertible nuisances. We show that in this case one image is sufficient to determine a complete representation, and there is no need for exploration at all. The story is, of course, different when there are invertible nuisances. 

\subsection{In the absence of non-invertible nuisances} 

In the absence of occlusions (for instance, if the images are taken by an omni-directional camera inside an empty room) and quantization errors or noise (if the images are samples of the scene radiance that satisfy Nyquist's condition -- if that was possible, that is), \index{Nyquist conditions} and if the scene obeys the Lambert-Ambient-Static model (\ref{eq-lambert-ambient}), the new image can be generated by the old image via a suitable contrast transformation of its range and diffeomorphic deformation of its domain. In other words, the first image $I_t$ can be used to construct a representation as we have discussed in Section \ref{sect-representation2}, which in turn can be used to hallucinate any other image of the same scene, for instance $I_{t+1}$. {\color{pea} In fact, from (\ref{eq-lambert-ambient}), we can solve for $\rho$ in the first image via $\rho(p(x)) = I_t(x)$, where $p(x) = \bar x/ \| \bar x\|$ and substitute in the second, so we have 
\begin{equation}
I_{t+1}(w(x)) = k \circ I_t(x)
\label{eq-bccc}
\end{equation}
where $w(x) = \pi g \pi^{-1}(\bar x / \| \bar x \|)$ and where we have aggregated the two contrast transformations $k_{t+dt}$ and $k_t$ into one.} This is exactly the hallucination process described in Section \ref{sect-hallucination}, whereby a single image $I_t$, supported on a plane, is interpreted as a representation and used to hallucinate the next image $I_{t+1}$. 

In the absence of non-invertible nuisances, the only purpose of the second image is to enable the estimation of the domain deformation $w$  and the contrast transformation $k$ through equation (\ref{eq-bccc}) \cite{jinSY05IJCV}. So, the first image captures the radiance $\rho$, and in combination with the second image they capture the shape $S$, entangled with the nuisance $g$ in the domain deformation $w$, as discussed in Section \ref{sect-disentangle} and further elaborated in \ref{sect-ambiguity}. Any additional image $I_\tau$, not necessarily taken at an adjacent time instant, can be explained by these two images. In fact, any additional image can be used to test the hypothesis that it is generated by the same scene, by performing the same exercise with either of the two (training) images, to test whether they are compatible with (\ie can be generated by) the same scene, except for the residual error $n$ that is temporally and spatially white, and isotropic (homoscedastic).\footnote{This kind of geometric validation is used routinely to test sparse features for compatibility with an underlying epipolar geometry \cite{maSKS}.} This test consists of an extremization procedure (Section \ref{sect-ml} and Figure \ref{fig-dalmatian}), whereby nuisance factors, as well as the scene, are inferred as part of the decision process, and the residual has high probability under the noise model. This can be thought of as a ``recognition by (implicit) reconstruction'' process, and was common in the early days of visual recognition via template matching. \index{Homoscedastic}

Notice that, as discussed in Section \ref{sect-disentangle},  contrast (a nuisance) interacts with the radiance, so one can only determine the radiance up to a contrast transformation. Similarly, motion (a nuisance) interacts with the shape/deformation, so one can only determine shape up to a diffeomorphism. If one were to canonize both contrast and domain diffeomorphisms, then some discriminative power would be lost as any two scenes that are equivalent (modulo a domain diffeomorphism, whether or not they are generated by a scene with the same shape) would be indistinguishable from a maximal invariant feature (or from any invariant feature for that matter) \cite{vedaldiS05}. 

In summary, {\em in the absence of non-invertible nuisances, a single} (omni-directional, infinite-resolution, noise-free) {\em image can be used to construct a} representation from which the entire light field of the scene can be hallucinated. This is what we called a {\em complete representation.} Its minimal sufficient statistic is the Complete Information, and there is {\em no need for exploration}. The need for exploration does, of course, arise as soon as non-invertible nuisances are present. The case discussed above is a wild idealization since, even in the absence of occlusions, quantization and noise introduce uncertainty in the process, and therefore there is always a benefit from acquiring additional data, even in the absence of any active control. 

\subsection{Dealing with non-invertible nuisances}
\index{Feature!non-invertible}

In general, equation (\ref{eq-bccc}) is valid only in the {\em co-visible} domain (Section \ref{sect-im-form}). {\color{pea} Neglecting contrast transformations for simplicity, we have that
\begin{equation}
I_{t+1}(w(x)) = I_t(x),  ~~~ x \in D \cap w^{-1}(w(D)) \subset D.
\end{equation}
The complement of the co-visible domain is the {\em occluded domain} \index{Occluded domain} \index{Co-visible domain}
\begin{equation}
\Omega \doteq D \backslash w^{-1}(w(D)) \subset D.
\end{equation}
Since we will assume that there is a temporal ordering (causality), we will consider only {\em forward occlusions} \index{Forward occlusion} \index{Occlusion!forward} \index{Un-occlusion} (sometimes called ``un-occlusion'' or ``discovery''); that is, portions of the domain of $I_{t+1}$, $\Omega \subset D$, whose pre-image $w^{-1}(\Omega)$ was not visible in $I_t$. The restriction of the image to this subset, $\{I_{t+1}(x), \ x \in \Omega \}$ cannot be explained using $I_t$.}

Consider now an image $I_t$, and imagine using it to predict the next image $I_{t+1}$, as we have done in the previous section. No matter how we choose $w$, we cannot predict exactly what the new image is going to look like in the occluded domain $\Omega$, even if we had an omni-directional, infinite-resolution, noise-free camera. Therefore, in the occluded domain we have a distribution of possible ``next images,'' based on the priors. The entropy of this distribution measures the uncertainty in the next image, and therefore the potential ``information gain'' from measuring it (before we actually measure it). Once we discount other nuisance factors, by considering a maximal invariant of the next image in the occluded domain, we have the {\em innovation}
\index{Occlusion}
\begin{equation}
\epsilon(I, t+1 ) \doteq \phi^\wedge({I_{t+1}}_{|_{\Omega}})
\label{eq-innovation}
\end{equation}
\index{Innovation} \marginpar{\tiny \sc innovation}
whose complexity represents the {\em Actionable Information Increment} (AIN): \index{AIN}\index{Actionable Information Increment (AIN)}
\begin{equation}
AIN(I,t) \doteq H(\epsilon(I, t+1)) = {\cal H}({I_{t+1}}_{|_{\Omega}}).
\end{equation}
The AIN is the {\em uncertainty} in the innovation, or the ``degree of unpredictability'' of future data. The above is the contribution to the AIN due to occlusions, that can be determined as in Section \ref{sect-occlusion}. However, we also have uncertainty due to quantization and noise.\footnote{Quantization can also be thought of as a form of occlusion mechanism, since details of the radiance that exist at a scale smaller than the back-projection of the area of a pixel onto the scene are ``hidden'' from the measurement. Rather than moving {\em around} an occluder, one can ``undo'' quantization by moving {\em closer} to the scene. } This is simpler to model, as it is usually independent of the scene: Uncertainty due to scaling and quantization increases linearly with distance, and uncertainty due to noise is independent of both the scene and the viewer. In all cases, however, there exists a control action that can reduce uncertainty. For the case of scale, it is zooming, or translating along the optical axis, or increasing the sensor's resolution. For the case of noise, it is taking repeated (registered) measurements. 

In all cases, what yields a non-zero innovation is the presence of non-invertible nuisances, $\nu_t$, that include occlusion, quantization, and other unmodeled uncertainty (noise). We then have that the AIN can be computed as the conditional entropy of 
\begin{equation}
p(\phi^\wedge(I_{t+1}) | I_{t}) = \int p(\phi^\wedge(I_{t+1}) | I_{t}, \nu_t)dP(\nu_t)
\end{equation}
where we have marginalized over all non-invertible nuisances. We therefore have, for the specific case of just two images,  
\begin{equation}
AIN(I,t) \doteq {\cal H}({I_{t+1}} | I_t).
\label{eq-def-AIN}
\end{equation}
This construction can be extended to the case where we have measured multiple images up to time $t$, as we discuss in the next section.

\begin{rem}[Background subtraction] \index{Background subtraction}
The innovation can be thought of as a form of generalized background subtraction. In the most trivial case, the representation is derived from one image only (the background image) and the only motion in the image is due to a ``foreground object.'' Other background subtraction schemes, including those based on multiple layers or on an aggregate model of the background \cite{koES10}, can be understood in the framework of occlusion detection and innovation under different prediction models \cite{jacksonYS05,jacksonYS06}.
\end{rem}

\subsection{Enter the control}

Clearly, the $AIN$ depends on the motion between $t$ and $t+dt$, as well as on the (unknown) scene $\xi$. {\color{pea} For a sufficiently small $dt$, $g_{t+dt} = \exp{(\widehat u dt)} g_t $, where $u \in {\mathfrak{se}}(3)$ \index{Rigid body velocity} is the rigid body velocity of the viewer \cite{maSKS} (assumed constant between $t$ and $t+dt$), the operator $\widehat{~~}$ is the ``hat'' operator that maps a $6$-dimensional vector of rotational and translational velocities into a ``twist'' \cite{maSKS} and ${\mathfrak{se}}(3)$ denotes the Lie algebra of ``twists'' associated with $SE(3)$ \cite{maSKS}. \index{Lie Algebra} \index{Twist} Thus, the motion between $t$ and $t+dt$ is given by $g_{t+dt}g^{-1}_t \simeq Id + \widehat u dt$, which one can control by moving the sensor with a velocity $u$. By assuming velocity to be sufficiently small, we can take $dt$ to be the unit time interval, $dt = 1$.} Therefore, the $AIN$ can, to a certain extent, be controlled. When emphasizing the dependency on the control input, we write $AIN(I, t; u) \doteq {\cal H}(I_{t+1}| I_t, u_t)$. 
\index{AIN}
\index{Actionable Information Increment (AIN)}
A myopic controller would simply try, at each instant, to perform a control action (\eg a rigid motion, zooming, or capturing and averaging multiple images) so as to maximize the $AIN$:
\begin{equation}
\hat u_t = \arg\max_u AIN(I, t; u).
\label{eq-ent-inn}
\end{equation}
Note that the AIN is computed {\em before} the control $u_t$ is applied, and therefore before the ``next image'' $I_{t+1}$ is computed. Thus the AIN involves a distribution of next images, marginalized with respect to all non-invertible nuisances. Once we actually measure the next image, we sample an instance of the innovation process. If we have discovered nothing in the next image, the sample of the innovation will be zero.

Therefore, a myopic agent could easily be stuck in a situation where {\em none of the allowable control actions} $u_t \in {\cal U}$ {\em yield any information gain}. The observer would then stop despite having failed to attain the Complete Information. The observer could also be trapped in a loop where it keeps discovering portions of the scene it has seen before, albeit not in the immediate past, for instance as it moves around a column. This is because the controller (\ref{eq-ent-inn}) does not have {\em memory}.

To endow the controller with memory we can simply consider the innovation relative to the {\em history} $I^t  \doteq I_0^t \doteq \{I_\tau\}_{\tau = 0}^t$, rather than relative to just the current image $I_t$. 
\begin{equation}
\hat u_t = \arg\max_u {\cal H}(I_{t+1}|I^t, u )
\label{eq-ent-inn2}
\end{equation}
There are two problems with this approach: One is that it is still myopic. The other is that the history $I^t$ grows unbounded. 

\subsection{Receding-horizon explorer}

A slight improvement can be had by planning a controller that maximizes the AIN over a finite horizon, for instance of length $T>1$: 
\begin{equation}
\hat u_t = \arg\max_u {\cal H}(I^{t+T}_{t+1}|I^t, u ).
\label{eq-ent-inn3}
\end{equation}
Of course, as soon as the agent makes one step, and acquires $I_{t+1}$, the history changes to $I^{t+1}$; therefore, the agent can use the control planned for the $T$-long interval to perform the first step, and then re-plan the control for the new horizon from $t+1$ to $t+T+1$. One can also consider an infinite horizon $T = \infty$, which presents obvious computational challenges. Regardless of the horizon, however, all controllers above have to cope with the fact that the history keeps growing. Even setting aside the fact that no provable guarantees of {\em sufficient exploration} exist for this strategy, it is clear that the agent needs to have a finite-complexity {\em memory} that summarizes all prior history. 

\subsection{Memory and representation}

A memory is a function of past data, $\phi(I^t)$. Of all functions, we are interested in those that are {\em parsimonious} (minimal), and ``informative'' in the sense that they can generate copies {\em not} of the data, but of the {\em maximal invariants} of the data. These are the characteristics of what we have defined as a {\em representation}. At any given time $t$, we can infer a representation $\hat \xi_t$ that is compatible with the history $I^t$ and that, eventually, may converge to a complete representation $\hat \xi$. Building such a representation (memory) is an inference problem that can be framed in the context of exploration as a System Identification problem under a suitable prediction-error criterion \cite{ljung99}. Given a collection of data $I^t$, we are interested in determining a ``model'' $\phi$ such that the statistic $\phi(I^t)$ best summarizes the entire past $I^t$ for the purpose of predicting the entire future $I_{t+1}^\infty$. In Remark \ref{rem-IB} we will shows that this is equivalent to determining  
\begin{equation}
\hat \xi_t = \arg\min_\xi H(\xi | I^t)  \  {\rm subject \ to} \ I^t \in {\cal L}(\hat \xi_t).
\end{equation}
In prediction-error methods, usually the complexity constraint is enforced by choosing the bound on the order of the model (the number of free parameters), and the entropy is approximated with its empirical estimate, assuming stationarity. For the case where all the densities at play are Gaussian, minimizing the entropy above reduces to minimizing the sum of squared one-step prediction errors.
This problem has also been addressed in the literature of information-based filtering \cite{mitter05}, but can only be realistically solved for low-dimensional state-spaces \cite{ryan08} using particle filtering, or for linear-Gaussian models (although see \cite{guoW06}).

{\color{pea} Note that the condition $I^t \in {\cal L}(\hat \xi_t)$ is to be understood up to the noise in the measurement device $n_t$, according to (\ref{eq-compatibility}), in the sense that the residual $I_t - h(g, \hat \xi_t, \nu)$ has high probability under the noise model $p_n$ for all times $t$. Such a ``noise'' lumps the residual due to all unmodeled phenomena and can be model as spatially and temporally white, and isotropic (homoscedastic). \index{Homoscedastic} If not, the spatial or temporal structure can be included as part of the representation, until the residual is indeed white. The above equation can then be interpreted as a reduction of uncertainty, specified by the mutual information ${\mathbb I}(\xi; I^t)$, as we have anticipated in Section \ref{sect-actinf}.}

In a causal exploration process, where data is gathered incrementally, we would like to update the estimate of the representation; by minimizing the conditional entropy above. At the same time, we would like to control the exploration process so as to maximize the AIN. This is discussed in the next section where we describe {\em dynamic explorers}.


\section{Dynamic explorers}
\index{Exploration}
\index{Visual exploration}
\index{Dynamic explorer}

In this section we consider an entity including a sensor (a video camera) capable of exercising a {\em control}, \marginpar{\tiny \sc control} \index{Control} for instance by changing the vantage point or the characteristics of the sensor (zooming). We call such an entity an {\em explorer}. \marginpar{\tiny \sc explorer} \index{Explorer} In the absence of any restriction on the control, so long as there are non-invertible nuisances, the $AIN$ provides guidance to perform exploration, and memory provides a way to incrementally build a representation $\hat \xi$. 
\index{Memory}
\index{Explorer}
Ideally, such a representation would eventually converge to a minimal sufficient statistic of the light field, ${\cal L}(\xi)$, thus accomplishing {\em sufficient exploration}.\index{Sufficient exploration} Sufficient exploration, if possible, would enable the {\em absence of evidence} to be taken as {\em evidence of absence} \cite{prettoCS11}.

We start with one image, say $I_0$, and its maximal invariant $\phi^\wedge(I_0)$, that is compatible with a (typically infinite) number of scenes, $\hat \xi_0$, that are distributed according to some prior $p(\hat \xi_0)$ that has high entropy (uncertainty). Compatibility means that the image $I_0$ can be synthesized from $\hat \xi_0$ up to the modeling residual $n$. {\color{pea} As we have seen in Section \ref{sect-hallucination}, we can easily construct a sample scene by choosing $\hat S_0$ to be a unit sphere, $S_0(x) = p(x) = \frac{\bar x}{\| \bar x \|}$, and choosing $\rho_0(p) = I(x)$ where $p = p(x)$ in the entire sphere but for a set of measure zero. Therefore, we call $\hat \xi_0 = \{\rho_0, S_0\} \doteq h^{-1}(I_0)$.} Clearly, if we had only one image, as we did in Section \ref{sect-hallucination}, we would not need a representation to explain it, and indeed we do not even need a notion of scene. So, in our case, this is just the initialization of the explorer, which corresponds to one of many possible representations that are compatible with the first image (see Figure \ref{fig-kanisza}). 

Given the initial distribution, $p(\hat \xi_0)$ and the next measurement, $I_1$, we can compute an innovation $\epsilon_1 = I_1 - h(g_1,  \hat \xi_0, \nu_1)$, that is a stochastic process that depends on the (unknown) nuisances $g_1, \nu_1$ as well as on the distribution of $\hat \xi_0$. We can then update such a distribution by requiring that it minimizes the uncertainty of the innovation. The uncertainty in the innovation is the same as the uncertainty of the next measurement, and therefore we can simply minimize the conditional entropy of the next datum, once we have marginalized or max-outed the nuisances: $\hat \xi_1 = \arg\min_{p(\hat \xi)} {\cal H}(I_1 | \hat \xi_0)$ where $p(I_1 | \hat \xi_0) = \int {\cal N}(I_1 - h(g,  \hat \xi_0, \nu))dP(g)dP(\nu)$. In practice, carrying around the entire distribution of representations is a tall order, for the space of shape and reflectance functions do not even admit a natural metric, let alone a probabilistic structure that is computable. So, one may be interested in a {\em point-estimate}, for instance one of the (multiple) modes of the distribution, the mean, median, or a set of samples. In any case, we indicate this via $\hat \xi_1 = \arg\min_{p(\hat \xi_1)} {\cal H}(I_1 | \hat \xi_0)$. Now, instead of marginalizing over all (invertible and non-invertible) nuisances, we can canonize the invertible ones, and therefore, correspondingly, minimize the {\em actionable information gap}, instead of the conditional entropy of the raw data: At a generic time $t$, assuming we are given $p(\hat \xi_{t-1})$, we can perform an {\em update} of the representation via
\begin{equation}
\boxed{\hat \xi_{t} = \arg\min_{p(\hat \xi_{t})} {\cal H}(I_{t} | \hat \xi_{t-1})} = 
\arg\min H\left(\int {\cal N}(\phi^\wedge(I_{t}) - h(\hat \xi_{t-1}, \nu); \sigma)dP(\nu)\right)
\label{eq-min-ent0}
\end{equation}
where ${\cal N}(\mu; \sigma)$ denotes a Gaussian density with mean $\mu$ and isotropic standard deviation $\sigma$. To the equation above we must add a complexity cost, for instance $\lambda H(\hat \xi_{t-1})$ where $\lambda$ is a positive multiplier.


Now, at time $t$, the updated representation $\hat \xi_t$ can be used to extrapolate the next image $I_{t+1}$. The hallucination process carries some uncertainty because of the non-invertible nuisances, and this uncertainty is precisely the information gain to come from the next image. Since the next image depends on where we have moved (or, more in general, on what control action we have exercised), we can choose the control $u_t$ so that the next image $I_{t+1}$ will be most informative, \ie
\begin{equation}
\boxed{\hat u_{t} = \arg\max_{u \in {\cal U}} {\cal H}(I_{t+1} | \hat \xi_{t}, u)} = 
\arg\max H\left(\int {\cal N}(\phi^\wedge(I_{t+1}) - h(\hat \xi_{t}, \nu); \sigma) dP(\nu)\right)
\label{eq-max-ent0}
\end{equation}


where $\cal U$ includes complexity or energy costs associated with the control action $u$.
Thus, inference and control are working together, one to maximize the uncertainty of the next data, the other to minimize it. 
\index{Control}
\index{Actuation}

{\color{pea} Technically, the most informative next data would be the ones that produce the largest reduction in uncertainty, \ie $\arg\max {\mathbb I}(\xi; I^{t+1}) = H(\xi) - H(\xi| I^{t+1})$. Unfortunately, as we have already discussed, even defining a base measure in the space of scenes is difficult, so computing $H(\xi)$ is problematic. However, we can rewrite the mutual information above in terms of the entropy of the data, $H(I_{t+1}|I^t , u)- H(I_{t+1} | \xi, u).$ Now, given the scene $\xi$, even the non-invertible nuisances $\nu$ become invertible, so $p(I_{t+1}|\xi, u)$ only depends on the residual uncertainty, that comes from all unmodeled phenomena and can realistically be considered white, independent, and isotropic (otherwise, if it has considerable structure, this should be modeled explicitly as a nuisance $\nu$). Therefore, the conditional entropy $H(I_{t+1} | \xi, u)$ is independent of $u$, and we can focus on the first term $H(I_{t+1}|I^t , u)$, which is what we have done in (\ref{eq-max-ent0}) after handling the invertible nuisance via pre-processing with $\phi^\wedge$.}

The construction of a representation from a collection of data treated as a {\em batch} has been described in \cite{jacksonYS05,jacksonYS06} for the case of multiple occlusion layers portraying arbitrarily deforming objects, and in \cite{jinSY05IJCV} for the case of rigid objects generating self-occlusions. This requires the dynamic update of the visible portion of the domain as a result of the update of the representation. In the case of \cite{jinSY05IJCV}, the representation consisted of an explicit model of the geometry of the scene (a collection of piecewise smooth surfaces) as well as of the photometry of the scene, consisting of the radiant tensor field, that could be used to generate ``super-resolution images'' (\ie images hallucinated at a resolution higher than that of the native sensor that collected the original data), as well as to generate views from different vantage points despite non-Lambertian reflection. \index{Super-resolution} In all these cases, the uncertainty was assumed Gaussian, so minimum-entropy estimation reduces to wide-sense filtering (minimum-variance).

The design of a control action, given the current representation, for the case of uncertainty due to visibility has been described in \cite{valenteTS12CISS,valenteTS12JMIV} for compact spaces, and extended in \cite{hernandezKS12} for unbounded domains. The case where uncertainty due to scale and noise is also present has been described in \cite{karasevCS12}.

Note that, in general, there is no guarantee that $\hat \xi_t \rightarrow \xi$ in any meaningful sense. The most we can hope for is that ${\cal L}(\hat \xi_t) \rightarrow {\cal L}(\xi)$, as we have pointed out. In other words, what we can hope is that our representation can at most generate data that is indistinguishable from the real data, up to the characteristics of the sensor. Since the inference of a representation is guided by a control, and the hallucination process requires the simulation of the invertible nuisance $g$ (which includes vantage point), Koenderink's famous characterization of images as ``controlled hallucinations'' is particularly fitting. Following the analysis above we can say that the representation $\hat \xi$ is obtained through a controlled exploration (perception) process, and from the representation we can then hallucinate images in a controlled fashion.

In general, however, it is not possible to guarantee that the exploration process will converge, even in the sense of (\ref{eq-compatibility}) ${\cal L}(\hat \xi_t) \rightarrow {\cal L}(\xi)$. However, it is trivial to design exhaustive control policies that, under suitably benign assumptions on the environment {\color{pea} (that the scene is bounded, that the topology is trivial, and that the radiance obeys some sparsity or band-limited assumption)} will achieve sufficient exploration, at least asymptotically: 
\index{Coverage}
\begin{equation}
\hat \xi_t \rightarrow \hat \xi  ~~~~~ {\rm s. \ t.} ~~~~~ {\cal L}(\hat \xi) = {\cal L}(\xi).
\end{equation}
For instance, a Brownian motion restricted to the traversable space will, eventually, achieve complete exploration of a static environment.

The goal of exploration is, therefore, to trade off the {\em efficiency} of the exploration process, including the cost of computing an approximation to the policy (\ref{eq-min-ent0})-(\ref{eq-max-ent0}), with the probability of achieving sufficient exploration. This is beyond the scope of this manuscript and we refer the reader to the vast literature on Optimal Control, Path Planning, Robotic Exploration, and partially-observable Markov Decision Processes (POMDP) in Artificial Intelligence. For the case of uncertainty due to visibility (occlusions) both \cite{hernandezKS12,valenteTS12CISS} provide bounds on the expected path length as a function of the complexity of the environment.

The discussion above suffices to our purpose of closing the circle on the issue of {\em representation} introduced in Section \ref{sect-representation2} and discussed in Section \ref{sect-hallucination}, by providing means to approximate it asymptotically from measured data. At any given instant of time, our representation $\hat \xi_t$ is incomplete, and any discrepancy between the observed images and the images hallucinated by the representation (the innovation) can be used to update the representation and reduce the uncertainty in $\hat \xi$.

The fact that, to ensure inference of a complete representation, a control $u$ must be exercised, links the notion of representation (and therefore of information) inextricably to a control action. More precisely, the exploration process links the control to the representation, and actionable information to the complete information. 
This discussion, of course, only pertains to the limiting case where we have arbitrary control authority to move in free space, into every nook and cranny (to invert occlusions), to zoom-in or move arbitrarily close and have arbitrarily high sampling rate (image resolution, to invert quantization), and to stay arbitrarily long in front of a static scene (or to sample in time arbitrarily fast relative to the time constant of the temporal changes in the scene), to invert noise. The opposite extremum is when we have no control authority whatsoever, in which case all we can compute is a maximal invariant and its actionable information, as discussed in Section \ref{sect-actinf}. In this case the Actionable Information Gap cannot be closed.\footnote{In addition to mobility, another active sensing modality can be employed, for instance by controlling accommodation\cut{ (Figure \ref{fig-accommodation}}) or by flooding the space with a controlled signal and measuring the return, as in Radar.} \index{Actionable Information Gap}
\index{Invertibility}
\index{Nuisance!invertible}

In between, we can have scenarios whereby a limited control authority can afford us an increase in actionable information by aggregating the $AIN$ over time, thus taking us closer to the complete information. Therefore, we can think of the ``degree of invertibility'' of the nuisance, properly defined, as a proxy of recognition performance. This we do in Section \ref{controlled-recognition}. 

This chapter concludes the treatment of active exploration. In the next chapter we turn our attention to constructing models of {\em not} individual objects, but object classes with some intra-class variability.

\begin{rem}[Information Bottleneck]
\label{rem-IB}
The functional optimization problem (\ref{eq-min-ent0}) is closely related to the Information Bottleneck principle \cite{tishbyPB00}, that prescribes finding the representation $\hat \xi$ that best trades off complexity and task fidelity by solving
\begin{equation}
\arg\min_{p(\hat \xi | I^t)} {\mathbb I}(I^t; \hat \xi) - \beta {\mathbb I}(\hat \xi; {\cal L}(\xi)).
\end{equation}
Note that the last mutual information is not with respect to the ``true'' scene $\xi$, but to its lightfield ${\cal L}(\xi)$, which we can think of the collection of all possible images we can take from time $t+1$ until $t = \infty$, that is, ${\mathbb I}(\hat \xi; {\cal L}(\xi)) = {\mathbb I}(\hat \xi; I_{t+1}^\infty).$ Note that the minimization is with respect to the (degenerate) density $p(\hat \xi | I^t) = p(\phi(I^t) | I^t) = \delta(\hat \xi - \phi(I^t))$, and therefore  with respect to the unknown ``model'' $\phi(\cdot)$, from which the ``state\footnote{The ``state'' is the statistic the best summarizes the past for the purpose of predicting the future. In other words, it ``separates'' the past from the future. In the linear Gaussian case these sentences have precise meanings in terms of Markov splitting subspaces, whereby the state is the projection of the future $I_{t+1}^\infty$ onto the past $I_{-\infty}^t$, for stationary Markovian processes.}'' $\hat \xi_t$ can be computed via $\hat \xi_t = \phi(I^t)$.
Using these facts and the properties of Mutual Information we have that the above is equal to
\begin{eqnarray}
&&\arg\min_{\phi} H(\hat \xi) - H(\hat \xi | I^t) - \beta H(I_{t+1}^\infty) + \beta H(I_{t+1}^\infty | \hat \xi) = \\ 
&&= \arg\min_{\phi} H(I_{t+1}^\infty | \hat \xi) + \lambda H(\hat \xi)
\end{eqnarray}
once we substitute $\lambda = 1/\beta$ and recall that $H(\hat \xi | I^t) = 0$
since $\hat \xi = \phi(I^t)$, and $H(I_{t+1}^\infty)$ does not depend on $\hat \xi$. When the underlying processes are stationary and Markovian, minimizing $H(I_{t+1}^\infty | \hat \xi)$ is equivalent to minimizing $H(I_{t+1} | \hat \xi)$. 
\end{rem}

\section{Controlled recognition bounds}
\label{controlled-recognition}
\index{Recognition bound}


In Communication Theory, given sufficient resources (bits), one can make the performance in the task (transmission of data) arbitrarily good, at least in principle, and for a given limit on the resources, one can quantify a bound on performance that does not depend on the particular signal being transmitted, but only on its distributional properties \cite{shannon}. It would be desirable to have a similar tool for the case of visual decision problems, whereby one could quantify performance in a visual decision (detection, localization, recognition, categorization etc.), rather than in a transmission task. The critical question is: what represents the {\em resource}, that plays the role of the bit rate in communications? What do we need to have ``enough of'' in order to guarantee a given level of performance in a visual decision task? Enough pixels? Enough visibility? Enough views? Enough computing power? In this section, we will see that the critical {\em resource} for visual decisions is the {\em control authority the viewer has on the sensing process}. We will argue that, for a {\em passive} observer \index{Passive observer} with no control authority whatsoever, there is no amount of pixels or computing power that will suffice to {\em guarantee} an arbitrarily high level of performance in a visual decision task (Section \ref{sect-passive-bounds}). In the opposite limiting case, we will show that an {\em omnipotent} explorer, capable of going everywhere and staying there indefinitely, not only can guarantee a bound on the decision error, but can make this error arbitrarily small in the limit (Section \ref{sect-active-bounds}). In between, we will attempt to quantify the {\em amount of control authority} and characterize its tradeoff with the decision error (Section \ref{sect-control-recognition}).

Some of the results in this chapter may appear obvious to some (of course we cannot recognize something we cannot see!), misleading to others, and confusing to others yet. Some examples are admittedly straw-men, meant to illustrate the importance of mobility for cognition, but we will try to state our assumptions as clearly and unequivocally as possible, and hopefully what matters will emerge, which is the fact that {\em control} plays a key role in {\em perception}, and that the notion of {\em visual information}, which is the topic of this manuscript, is the knot that ties them. Of course it is always possible to construct specific cases and counter-examples that violate the statements, but our point is that these statements are valid on average, once one considers all possible objects and all possible scenes.
So, {\em if we want to get visual decisions under control,} in the sense of being able to provide guaranteed bounds on the decision performance, {\em we have to put control in visual decisions}, in the sense of being able to exercise some kind of control authority over the sensing process.

\chapter{Learning Priors and Categories}
\label{sect-learning}
\index{Prior distribution}
\index{Learning}
\index{Category}
\index{Categorization}


Previous chapters have shown how to handle canonizable nuisances (Section \ref{sect-canonization}), and non-invertible nuisances  via either marginalization  (\ref{eq-marg}), extremization (\ref{eq-ML}), or -- if we are willing to sacrifice optimality for decision-time efficiency -- by designing invariant descriptors (Chapters \ref{sect-detectors} and \ref{sect-descriptors}). In all cases, the design of a visual classification algorithm requires knowledge of priors on the nuisances as well as on the scene. 

The procedure we have outlined for building a template (\ref{eq-blurred-template}), or a Time HOG descriptor (Section \ref{sect-time-hog}) using a training sample $\{I_t\}_{t=1}^T \sim p(I|c)$, assumes that a sequence of frames $\hat g_t$ is available (Section \ref{sect-correspondence}). Thus we have implicitly assumed that the {\em scene} $\xi$ {\em is the same}, not just the {\em class} $c$. Indeed, before Chapter \ref{ch-time} we even assumed that the underlying scene was {\em static}, which enabled us to attribute the variability in the data to nuisances, rather than to intrinsic factors. Even in Chapter \ref{ch-time} we assumed that local variability was due to nuisance factors, and in both cases this enabled us to determine the {\em object-specific} nuisance distribution.

What we have deferred until now is the possibility for intrinsic (intra-class) variability. It cannot be realistically assumed that an explicit model be available for all classes of interest, and therefore such variability should be learned, although it is likely that some basic components of models can be shared among classes. In this chapter we describe an approach to build category models starting from the local representations described in previous chapters.

The starting point are occlusions, detected as described in Section \ref{sect-occlusion}. These can be used to bootstrap a partitioning of images into {\em detachable objects} as we will show in Section \ref{sect-detachable}. Such a ``segmentation'' process is different than traditional (single-image) segmentation, and can be accomplished through relatively simple computations (linear programming). Such detachable objects can then be tracked over time (Section \ref{sect-tracking}), providing the support where the local descriptors of Section \ref{sect-time-hog} can be aggregated.

However, knowledge that a certain descriptor belongs to a certain object -- while an improvement on than the so-called ``bag-of-feature'' approach that considers the distribution of descriptors on the entire image -- still fails to capture important geometric relationships. In some cases, there may be an advantage in further subdividing objects into {\em ``parts''}, or subsets of descriptors based on spatial relations (Section \ref{sect-parts}).

This process produces a model of a specific object, removing nuisance variability from the data. In order to represent intrinsic variability and arrive at a categorical model it is necessary to aggregate different instances of the same class. We will assume that the class label is provided as part of the training process. This is because sometimes categories can be defined based on non-visual characteristics. For instance, an object can be called a chair if someone can sit on it, which is a functional property that may or may not have visual correlates. Thus the class distribution of chairs may include largely disparate objects with different shapes and materials. 

This approach is somewhat different from the conventional approach to object categorization, that aims to detect or recognize  the category {\em before} the specific object. This is motivated by studies of primate perception, where there is an evolutionary advantage in being able to assess coarse categories (\eg animal or not) pre-attentively. The category model that is implicit in many of these approaches is not very different from an object model, and an effort is under way to make these models more specific and more discriminative to perform fine-scale categorization, or in other words getting closer to models of individual objects. In our case, objects come before categories. When we learn a model of a chair, we first learn a model of {\em this particular chair}. The fact that someone may sit on it makes it a chair just like another one regardless of its shape and appearance. This also allows one particular object to be easily attributed to multiple categories: A toy elephant can be a chair if someone can sit on it. Of course, there may be particular categories that are defined by visual similarity, in which case one can expect relatively simple categorical distribution that is well summarized by few samples.

In summary, a (detachable) object is one that triggers occlusions, and that supports a collection of descriptors that are organized into parts. A collection of objects induces a distribution of parts and their spatial configuration. Such a distribution can be multi-modal and highly complex, and often requires external supervision to be learned. Of course, descriptors and parts can be shared among objects and also among categories, but this is beyond our scope here and is the subject of current research.

\chapter{Discussion}

Physical processes contributing to image formation occur at a level of granularity that is infinitesimal relative to the sampling capacity of optical sensors. And yet, we seem to take for granted that the epistemological process requires breaking down the data into elementary ``atoms.''\cut{ Why does this need to be so? Can we not acquire knowledge without breaking down the data into pieces only to reassemble it later? And if we need to break down the data, where do we stop? Is there an ``information atom'' that is independent of the particular sensing mechanism or device? From our analysis it appears that if knowledge involves decisions, and if decisions are to be taken in a way that is insensitive to ``accidents'' (nuisances) that are not invertible, then the answer is yes. The ``breaking down'' of the data into constituents is not only manifest in neuronal encoding (concentrations of ions are for all practical purposes continuous, yet spikes are discrete events), but also in molecular biology (\eg chemotaxis, morphogenesis).} Information theory suggests that such a symbolization process would occur at a loss, and that integrated systems capable of sensing and action would be best designed in an end-to-end fashion.

However, we have seen that nuisance factors such as illumination and viewpoint changes account for almost all the variability in the data, and therefore the process of eliminating their influence in the data can lead to a {\em lossless symbolization}. In other words, under certain circumstances, it is possible to throw away almost all the data, and yet none of the information.

Furthermore, nuisances that are not invertible, such as occlusions, can be factored out through a controlled sensing action. Indeed, control is the ``currency'' that trades off performance in a perception process. The more control an agent can exercise on the sensing process, the tighter the bound that can be guaranteed on the reduction of the Actionable Information Gap. 

The models we describe in this manuscript are idealized abstractions.  Nevertheless, an abstraction is useful to guide investigation and to evaluate existing schemes. In fact, one of the many possible objections to our program is that the models we used in our analysis (for illumination, deformation, occlusion etc.) are so
simplistic as to make the formalization exercise futile. Better is to
devise new algorithms and test them on empirical benchmarks. 
That is generally true, although an abstraction allows us to ascertain how different algorithms are related, and determine not just whether one works better than another, but why. Shannon's sampling theorem is valid only for strictly
band-limited signals, a wildly idealized abstraction of real signals. And yet,
it provides useful guidelines for the design of certain algorithms, for instance for audio processing. 
\cut{\begin{rem}[Limitations of the symbolic notation]
The notation $h(g,\xi,\nu)$ is meant to separate the role of invertible (group) nuisances $g$ and non-invertible nuisances $\nu$. However, non-invertible nuisances such as occlusions can arise from the interplay of group transformations $g$ and acting on the scene $\xi$. For instance, a non-convex scene can generate self-occlusions via the action of a planar translation, which is a group. This is why we call the entity $h(g,\xi, 0)$ a ``hypothetical'' image, because in practice it is not possible to act on the scene $\xi$ through a group $g$ without generating non-invertible nuisances. Instead, in general we have $\nu = \nu(\xi, g)$, so non-invertible nuisances can appear through the action of a group acting on the scene. Thus also the notation $I\circ g \circ \nu$ and the composition of group actions and nuisances is inconsistent unless we take into account this dependency through a more elaborate notation, referring to an explicit model such as one of those described in Appendix \ref{sect-image-formation}. As a consequence, the notion of commutativity has also to be specified with respect to a specific image-formation model, rather than the generic symbolic model.
This can be confusing at first reading, but the fact that non-invertible nuisances cannot be dealt with in pre-processing, and that reducing the resulting information gap requires control of the sensing process, remains, and is already illustrated by the simplified formal notation.
\end{rem}}

Another valid objection is that we have reduced visual perception to pattern classification. We have not tackled high-level vision, perceptual organization, and all the high-level models that lie at the foundations of our understanding of higher visual processing. This does not imply that these problems are not important. We have just chosen to focus on the lowest level, which is the conversion from analog signals to discrete entities, on which high-level models can be built.

So far we have been deliberately vague about the difference between information and knowledge. We believe knowledge acts on information, but also needs tools to manipulate representations, including counterfactuals and causal analysis \cite{pearl00causality}. Nevertheless, most investigations that we are aware of take a discrete, atomic representation as a starting point, and fail to bridge the ``gap'' required to explain why we need such a representation in the first place. We hope to have addressed this in a number of ways.  First, by showing that even for invertible nuisances, invariants can be a set of measure zero \cite{sundaramoorthiPS09}. Second, non-invertible nuisances call for breaking down the image into pieces (segments); \cite{vedaldiS05} show that to recognize an object {\em with a viewpoint invariant feature} you have to discard its shape. Interestingly, this was already understood by Gibson, who expressed it in words: (\cite{gibson84}, page 271): 
\begin{quote}
``Despite the argument that because a still picture presents no transformation it can display no invariants under transformation, [in Gibson, 1973] I ventured to suggest that it did display invariants''.
\end{quote}
We have shown that\cut{ what is left after discounting viewpoint variability is a set of measure zero relative to the image: Almost all the data is gone; what is left is information. {From there, knowledge can be built by induction or deduction, by logic or probabilistic inference, or a combination thereof as in Bayesian inference.} But the crucial step is how precisely data are discarded to arrive at information.
We have also shown that}, in the absence of specific modeling of the nuisances, one would have to marginalize them as part of the matching process, which entails infinite-dimensional optimization. 
This does not mean that one cannot obtain positive, even impressive, results on a {\em specific domain}, for instance in semi-structured or fully structured environments where the variability due to nuisances is kept at bay. However, one should not extrapolate the optimism stemming from success on a specific domain with the solution of the general problem.
\section{James Gibson}
\label{sect-gibson}

The notion of Actionable Information presented in this manuscript is closely related to the notion of Information proposed by Gibson, although he never formalized these notions; in words, however,  page 245 of \cite{gibson84} recites
\begin{quote}
 ``The hypothesis that invariance under optical transformation constitutes information for the perception of a rigid persisting object goes back to the moving-shadow experiment (Gibson and Gibson, 1957)''
\end{quote}
And again on page 310: 
\begin{quote}
``Four kinds of invariants have been postulated: those that underlie change of illumination, those that underlie change of the point of observation, those that underlie overlapping samples, and those that underlie a local disturbance of structure. [...] Invariants of optical structure under changing illumination [...] are not yet known, but they almost certainly involve ratios of intensity and color among parts of the array. [...] Invariants [...] under change of the point of observation [...] some of the changes [...] are transformations of its nested forms, but the major changes are gain and loss of form, that is, increments and decrement of structures, as surfaces undergo occlusion. [...] The theory of the extracting of invariants by a visual system takes the place of theories of ``constancy'' in perception, that is, explanations of how an observer might perceive the true color, size, shape, motion and direction-from-here of objects despite the wildly fluctuating sensory impressions on which the perceptions are based.''
\end{quote}
The line of the program sketched by Gibson in his theory of ``information pickup'' where ``the occluded becomes unoccluded'' is very closely related to the notion of invertibility of the nuisance and controlled recognition that we discuss in Chapters \ref{sect-exploration} and \ref{controlled-recognition}.


\section{Alan Turing}

Alan Turing is perhaps the researcher that showed the deepest insight into the questions raised in this manuscript. On one hand, he specifically addressed the rise of  ``discontinuous behavior'' in continuous chemical systems through reaction-diffusion processes \cite{turingMorphogenesis}. On the other hand, he addressed the issue of intelligent behavior in machines. While Turing's theory of morphogenesis, despite its limits, can be taken as sufficient evidence that biological systems may evolve towards discrete structures, it does not provide evidence of the need to organize {\em measured data} into discrete ``information entities.''  

Since in building machine vision systems we are not constrained by the chemistry of reaction-diffusion, Turing's theory remains incomplete. In particular, it does not address why a data-processing system (biological or otherwise) built from optimality principles should exhibit discrete internal representations rather than a collection of continuous input-output maps. This ``gap'' thus remains open in Turing's work. Indeed, it is summarily dismissed: 
\begin{quote}
``the confusion between [analog and digital machines] is to be ignored. Strictly speaking, there are no 
[digital] machines. Everything really moves continuously. But there are 
many kinds of machine which can profitably be thought of as being 
discrete-state machines. For instance in considering the switches for a 
lighting system it is a convenient fiction that each switch must be 
definitely on or definitely off. There must be intermediate positions, 
but for most purposes we can forget about them.'' 
\end{quote}
He therefore moved on to characterize ``intelligent behavior'' in terms of symbols, as convenient to sustain the discourse, without regard to how such symbols might come to be in the first place. Digitization is indeed an abstraction. It is, however, an abstraction that ``destroys'' information, in the classical sense (Section \ref{sect-dpi}). And yet, it seems to be a necessary step to knowledge or ``intelligence''.\footnote{Note that Godel's assertion of the limitations of logic (1931) is irrelevant in this context, as our argument is not in favor of logic, but about the necessity of a discrete/finite internal representation. This could be used as an approximation tool, and continuous techniques be used for inference rather than logic.}

\section{Norbert Wiener}

As we have remarked earlier, despite anticipating the role of information in making a decision, Wiener remains anchored to the notion of information as entropy. To be fair, this was revolutionary at the time, and indeed Wiener is credited as one of the proposers of using notions from statistical mechanics, including entropy, in the processing of signals. Part of the problem is that the task Wiener was implicitly considering is the reconstruction of a signal under noise. This is not surprising since so much of Wiener's work was devoted to Brownian motions and to the characterization of stochastic processes driven by ``noise''.

It is interesting, however, to note that Wiener had the intuition that transformations play an important role, including the notion of the invariant under a group. On page 135 of \cite{wiener49}, he remarks about the ability of the human visual system to recognize line drawings, pointing the attention to discontinuities. He also introduces the notion of ``group scanning'' (what we have called max-out, or {\em registration}), hypothesizing computational hardware in the brain that could be implementing such an operation (page 137). Indeed, he suggests that McCulloch's apparatus could perform such a group-scanning. He even introduces the first moment as an invariant statistic to a group in equation (6.01) on page 138, and called it a {\em gestalt}! It is unfortunate that Wiener did not further elaborate on these points.

Wiener also implies some of the seeds of ecological vision, mixed by the notion of invariant statistics, suggesting that 
\begin{quote}
``we tend to bring any object that attracts our attention into a standard position and orientation, so that the visual image which we form of it varies within as small a range as possible.'' 
\end{quote}
This would be equivalent to ``physical canonization'' which, as Gibson suggested, can always be performed even when the nuisance is not invertible from a single image. Wiener does not explain, however, how his notion of information is compatible with his hypothesis that  
\begin{quote}
``processes occur [...] in a considerable number of stages each step in this process diminishes the number of neuron channels involved in the transmission of visual information''
\end{quote}
a sentence that reveals both the attachment to transmission of information as the underlying task, and the notion of ``compression'' as information that is, however, not developed further.

\section{David Marr}

This manuscript could be interpreted as an attempt to frame the ideas of Marr into an analytical framework, although Marr did not frame the questions he asked in the context of a task.
\begin{quote}
``Our view is that vision goes symbolic almost immediately, right at the level of zero-crossings, and the beauty of this is that the transition [...] is probably accomplished without loss of information'' 
\end{quote}
This statement, from \cite{marrPU79}, is incorrect if one means ``information'' in the sense of Shannon \cite{shannon}. However, it is precisely correct if one is to take the approach described in this manuscript. The use of zero-crossings was ultimately rejected because the decoding process was unstable; however, as we have argued, an internal representation is not needed for reconstruction; instead, for the task of recognition, an internal representation is necessary, and zero-crossings (a special case of feature detector $F$ in our parlance) might not have been a bad idea after all.\footnote{Marr's theories pertains not to vision in general, but to visual recognition. There is no need for an internal representation (such as that afforded by the primal sketch, the 2-1/2D sketch and the full sketch) for navigation, 3-D reconstruction, rendering or control. It is interesting that Marr uses stereo, and in particular Julesz' random dot stereograms, to validate his ideas, where a fully continuous algorithm that acts directly on the data would do as well or better.}

\section{Other related investigations} 

Donald L. Snyder used to often purport his mother's advice to ``never throw away information.'' Our discussion reveals that any sensible notion of information that relates to the recognition task (as opposed to transmission) requires {\em data} (not information) {\em to be thrown away}. In other words, to gather information you must first throw away data, the more the better. 
So, what is ``lossy'' for image compression or data transmission is not lossy for recognition, and in general for understanding images. This seems at first to defy any notion of traditional information theory and decision theory, for it involves multiple intermediate decisions. We hope that our arguments have convinced the reader that this is not the case. 

Some of the statements made in this manuscript may seem controversial. Certainly I do not wish to imply that individual organisms that do not exhibit visual recognition (\eg blind people) do not have intelligence. Nor do I imply that every organism that has a visual system exhibits intelligent behavior. What I have argued is that organisms and species that need efficient visual recognition (limitations on the classifier to maximize computational efficiency at decision time) benefit from signal analysis (and therefore symbolic manipulation, internal representation etc.)

 It is interesting to speculate whether there are organisms that have vision, and that use vision for regression or control tasks (\eg visual navigation) but not to perform decisions (\eg visual recognition). For instance, it is interesting to speculate whether the fly navigates optically, but decides olfactorily.\footnote{In a personal communication, Steve Zucker remarked that the fly does not exhibit long-range lateral interactions, and yet it is known to use vision for navigation, and have strong olfactory system to guide actions.} To this end, there is evidence that the fly lacks the wide-ranging lateral connections exhibited in higher mammals. {Marr's account on the fly's visual system (and the so-called ``representation'' that it implies, page 34 of \cite{marr}) really describes a collection of analog input-output maps, or collection of sensing-action maps, with a simple decision switch, with no need for an internal representation. Wiener points out that 
\begin{quote}
``complicated as the behavior patterns of birds are -- in flying, in courtship, in the care of the young, and in nest building -- they are carried out correctly on the very first time without the need of any large amount of instruction from the mother.'' 
\end{quote}
He uses this argument in support of phylogenic (species, as opposed to ontogenic, or individual) learning, and speaks in favor of endowed input-output maps without an underlying internal representation that is easily manipulated.} A contrarian view on the topic is supported by experiments in the development of an individual organism's vision system in the absence of mobility \cite{heldH63}.

\cut{\begin{rem}
This also answers a nagging question for anyone wishing to extrapolate the conclusion of running recognition experiments on benchmark datasets. How big should a dataset be in order to guarantee, with reasonable confidence, that the result of any given algorithm will extend to any other dataset? It is well-known that several algorithms that were achieving state-of-the-art performance in the Caltech 101 dataset were performing very poorly in the Caltech 256 dataset. So, clearly $N = 101$ is not sufficient in order to guarantee that the performance of a given dataset will extend to another $M > N$. So, what should the ``Caltech N'' dataset be, in order to guarantee that the performance of algorithms on the ``Caltech N'' will be the same for any other dataset? The theorem above shows that there exists no such $N$, and therefore benchmark datasets can only provide suggestive evidence of the performance of a given algorithm -- the larger the dataset the better -- but never a performance guarantee.
\end{rem}
This does not mean that benchmark dataset are useless. It just means that they have to be taken with a grain of salt, and that they cannot be used as performance guarantees. Note that this result is unlike results in classical Information Theory, where given enough resources (large-enough bandwidth, fast-enough sampling), one could guarantee perfect recovery of a signal. Here, even in the presence of infinitely-many samples, one cannot achieve perfect recognition.}

\cut{\begin{rem}[Segmentation and information]
Many practitioners of segmentation argue that the process, although technically non-falsifiable, is useful because it ``extracts information'' from the data. However, segmentation reduces entropy, and therefore if one embraces the traditional notion of entropy the process actually reduces information. The discussion above indicates that segmentation -- if done properly in a way that is robust to noise, quantization and other unmodeled phenomena -- should not entail a loss of (suitably defined notion of) information for the context of recognition under changes of viewpoint and illumination.
\end{rem}}

Because mobility plays such an important role in this manuscript, it naturally relates to visual navigation and robotic localization and planning \cutTwo{\cite{thrun98,batalinS03,hughesL05}}. In particular, \cite{whaiteF90,bourgault2002iba,stachnissGB05} propose ``information-based'' strategies, although by ``information'' they mean localization and mapping uncertainty based on range data.\cutTwo{ Range data are not subject to illumination and viewpoint nuisances, which are suppressed by the active sensing, \ie by flooding the space with a known probing signal (\eg laser light or radio waves) and measuring the return.} There is a significant literature on vision-based navigation \cite{brooks85,zhangO02,newmanH05,peruchVG95,taylorK98,simD03,franzSMB98,davidsonM02,seLL05,andersenJC97} that is relevant to occlusion-driven navigation \cite{kutulakosD95,kutulakosJ95,bajcsyM93}. In most of the literature, stereo or motion are exploited to provide a three-dimensional map of the environment, which is then handed off to a path planner, separating the {\em photometric} from the {\em geometric and topological} aspect of the problem. This separation is unnecessary, as the regions that are most informative are occlusions, where stereo provides no disparity.
Another stream of related work is that on Saliency and Visual Attention \cite{ittiK01}, although there the focus is on navigating the {\em image}, whereas we are interested in navigating the {\em scene}, based on image data.\cutTwo{ In a nutshell, robotic navigation literature is ``all scene and no image,'' the visual attention literature is ``all image, and no scene.'' The gap can be bridged by going ``from image to scene, and vice-versa'' in the process of visual exploration (Chapter \ref{sect-exploration}). The relationship between visual incentives and spatial exploration has been a subject of interest in psychology for a while \cite{butler}.}

This work also relates on visual recognition, by integrating structures of various dimensions into a unified representation that can, in principle, be exploited for recognition. In this sense, it presents an alternative to \cite{guoZW03,todorovicA06}, that could also be used to compute Actionable Information. \cutTwo{However, the rendition of the ``primal sketch'' \cite{marr} in \cite{guoZW03} does not guarantee that the construction is ``lossless'' with respect to any particular task, because there is no underlying task guiding the construction.} This work also relates to the vast literature on segmentation, particularly texture-structure transitions \cite{wuGZ08}.\cutTwo{ Alternative approaches to this task could be specified in terms of sparse coding \cite{olshausen98sparse} and non-local filtering \cite{buadesCM05}.}\cutOne{ I stress the fact that, while no single segmentation is ``right'' or ``wrong,'' the collection of all possible segmentations, with respect to all possible statistics pooled at all possible scales, is ``useful'' in the sense of providing pre-computation of the optimization or marginalization functional implicit in any recognition task.}\cutTwo{ This paper also relates to the literature of ocular motion, and in particular saccadic motion. The human eye has non-uniform resolution, which affects motion strategies in ways that are not tailored to engineering systems with uniform resolution. One could design systems with non-uniform resolution, but mimicking the human visual system is not our goal.}

Our work also relates to other attempts to formalize ``information'' including\cut{ the so-called Epitome \cite{jojicFK03}, that could be used as an alternative to our Representational Structure if one could compute it fast enough. Furthermore, the Epitome does not capture compactness and locality, that are importantly related to the structure of the scene (occlusions) and its affordances (relationship to the viewer). For instance, if one has same texture patch in different locations in space, these are lumped together, regardless of compactness. Another alternative is} the concept of Information Bottleneck, \cite{tishbyPB00}, and our approach can be understood as a special case tailored to the statistics and invariance classes of interest, that are task-specific, sensor-specific, and control authority-specific. These ideas can be seen as seeds of a theory of {\em ``Controlled Sensing''} that generalizes Active Vision to different modalities whereby the purpose of the control is to counteract the effect of nuisances. This is different than Active Sensing, that usually entails broadcasting a known or structured probing signal into the environment. Our work also relates to attempts to define a notion of information in statistics \cite{lindley,bernardo}, economics \cite{marschak60,arrow} and in other areas of image analysis \cite{keeler} and signal processing \cite{goodO}. Our particular approach to defining the underlying representational structure relates to the work of Guillemin and Golubitsky \cite{guilleminG}. 
\cut{Our work also relates to video coding/compression: As I have pointed out, poor man's versions of some of our constructions could be computed using standard operations from the video coding standards. However, I advocated structures that are adapted to the image data (superpixels, TAG, representational graph) rather than on fixed blocks. We can do this because, to achieve invariance to viewpoint, we have no need to encode deformation of these regions, just their correspondence. }

Last, but not least, our work relates to Active Vision \cite{aloimonosW88,blakeY93,bajcsy88}\cutTwo{, and to the ``value of information'' \cite{marschak60,fogelH82,gould74,claxtonNAW01,castronKNQRZ08}}.\cutTwo{ The specific illustration of the experiment to the sub-literature on next-best-view selection \cite{pito99,bajcsyM93}. Although this area was popular in the eighties and nineties, it has so far not yielded usable notions of information that can be transposed to other visual inference problems, such as recognition and 3D reconstruction.} A notable exception is the application of active vision to the automotive environment, pioneered by Dickmanns and co-workers \cite{dickmannsC89,dickmannsG88,dickmanns94}.

Similarly to previously cited work \cite{whaiteF90,bourgault2002iba}, \cite{denzlerB02} propose using the decrease of uncertainty as a criterion to select camera parameters, and \cite{arbelF95} uses information-theoretic notions to evaluate the ``informative content'' of laser range measurements depending on their viewpoint. Other influential literature on the relation between sensing and action include \cite{oreganN01,goodaleH98}.

This manuscript of course also relates to data compression, in particular video compression, but not in the traditional sense, where the goal is to reconstruct a copy of the signal on the other side of the channel, but where the goal is to transmit a compressed representation that is to be used for decision or control tasks.


\appendix

\chapter{Background material}

\def\lmbd{{l}}
\def\radn{{R_L}}
\def\rad2{{R_S}}
\def\xsi{{{{x}}^{(\i)}}}
\def\ysi{{{{y}}_e^{(\i)}}}
\def\hsi{{{{h}}^{(\i)}}}
\def\x{{x}}
\def\X{{X}}
\section{Basics of differential geometry}
\label{sect-differential-geometry}

Some of the concepts discussed in this manuscript use concepts from differential geometry, including the notions of {\em group, orbit space, equivalence classes, quotients and homogeneous spaces}. A good introduction to this material is \cite{boothby86}. While an expository review of differential geometry is beyond the scope of this appendix, most of the concepts treated in this manuscript can be understood after going through Appendix A of \cite{murrayLS94}, on pages 403--433 (skipping Section 1.3). Some of that material, specifically relating to the Lie groups $SE(3)$ and $SO(3)$, and their corresponding Lie algebra ${\mathfrak{se}}(3)$, can also be found in Chapter 3 of \cite{maSKS}. Our notation in this manuscript is consistent with, and in fact derived from, both \cite{maSKS} and \cite{murrayLS94}.

\section{Basic topology (by Ganesh Sundaramoorthi)}
\label{sect-topology}
{\color{orange} 

\newcommand{\ud}{\,\mathrm{d}}
\newcommand{\sh}{\hat{s}}
\newcommand{\ph}{\hat{p}}
\newcommand{\st}{\tilde{s}}
\newcommand{\Nc}{\mathcal{N}}
\newcommand{\Tc}{\mathcal{T}}
\newcommand{\Fa}{F_{\alpha}}
\newcommand{\R}{\mathbb{R}}
\newcommand{\D}{\mathbf{D}}
\newcommand{\Z}{\mathbb{Z}}
\newcommand{\C}{M}
\newcommand{\ip}[3]{\left< {#1}, {#2} \right>_{#3}}
\newcommand{\inner}[3]{\left\langle #2, #3\right\rangle_{_{\!\!H^#1}}}
\newcommand{\mean}[1]{\overline{#1}}
\newcommand{\grad}[1]{\nabla_{\!\!\mathbb{L}^{2}}}
\newcommand{\gradt}[1]{\nabla_{_{\!\!\textrm{Sobolev}} }}
\newcommand{\der}[2]{\frac{\ud {#1}}{\ud {#2}}}
\newcommand{\pder}[2]{\frac{\partial {#1}}{\partial {#2}}}
\newcommand{\pdersq}[2]{\frac{\partial^2 {#1}}{\partial {#2}}}
\newcommand{\comment}[1]{ }

In this appendix we describe some basic notions of topology that are useful to follow the orange-colored sections of the manuscript. We have privileged simplicity to rigor, so some of the statements made are imprecise and others are not fully developed. The interested reader can consult a differential topology book for details and clarifications.

\subsection{Morse Functions}

We consider functions $f : \R^2 \to \R^+$ as models for images. Define
the {\bf gradient} as 
\[
\nabla f(x) \doteq \left( \pder{f}{x_1}(x), \pder{f}{x_2}(x) \right) 
\mbox{ where } x=(x_1,x_2),
\]
and define the {\bf Hessian} as
\[
\nabla^2 f(x) \doteq 
\left(
\begin{array}{cc}
  \pdersq{f}{x_1^2}(x) & \pdersq{f}{x_1x_2}(x) \\
  \pdersq{f}{x_2x_1}(x) & \pdersq{f}{x_2^2}(x)
\end{array}
\right).
\]
Define a {\bf level set} of a function at level $a\in \R^+$ as 
\[
f^{-1}(a) \doteq \{ x\in \R^2 : \, f(x) = a \}.
\]

\begin{defn}[Critical Point]
For a $C^1$ function, a {\bf critical point} $p \in \R^2$ is a point
    such that $\nabla f(p) = 0$.
  \begin{itemize}
  \item A critical point $p$ is a {\bf local minimum} if $\exists \,
    \delta > 0$ such that $f(x)\geq f(p)$ for $x$ such that
    $|x-p|<\delta$.
  \item A critical point $p$ is a {\bf local maximum} if $\exists \,
    \delta > 0$ such that $f(x)\leq f(p)$ for $x$ such that
    $|x-p|<\delta$. 
  \item A critical point $p$ is a {\bf saddle} if it is neither
    a local minimum or local maximum. 
  \end{itemize}
\end{defn}

\begin{defn}[Morse Function]
  A {\bf Morse function} $f$ is a $C^2$ function such that all
  critical points are non-degenerate, i.e., if $p$ is a critical
  point, then $\det{\nabla^2f(p)} \neq 0$.
\end{defn}
\begin{rem}
 By Taylor's Theorem, we see that Morse functions are well
    approximated by quadratic forms around critical points:
    \begin{align*}
      f(x) &= f(p) + \nabla f(p) \cdot (x-p) + (x-p)^T \nabla^2f(p) (x-p)
      + o(|x-p|^2) \\
            &= (x-p)^T \nabla^2f(p) (x-p) + o(|x-p|^2)
    \end{align*}
    provided that $f(p)=0$ (if not, set $f$ to $f-f(p)$).
In particular, this means that Morse functions have {\bf
      isolated} critical points.
\end{rem}

The previous Remark, leads to the following observation:
\begin{theorem}[Morse Lemma]
  If $f$ is a Morse function, then for a critical
  point $p$ of $f$, there is a neighborhood $U \ni p $ and a chart
  (coordinate change) $\psi : \tilde{U} \subset \R^2 \to U \subset
  \R^2$ so that 
  \[
  f(\hat x) = f(p_i) +
  \begin{cases}
    -(\hat x_1^2 + \hat x_2^2)  & \mbox{if $p$ is a maximum} \\
    \hat x_1^2 + \hat x_2^2   & \mbox{if $p$ is a minimum} \\
    \hat x_1^2 - \hat x_2^2   & \mbox{if $p$ is a saddle}
  \end{cases}
  \]
  where $(\hat x_1, \hat x_2) = \psi_i(x_1,x_2)$ and $(x_1,x_2) \in \R^2$ are the
  natural arguments of $f$.
  \begin{figure}[h]
    \centering
    \includegraphics[totalheight=2in]{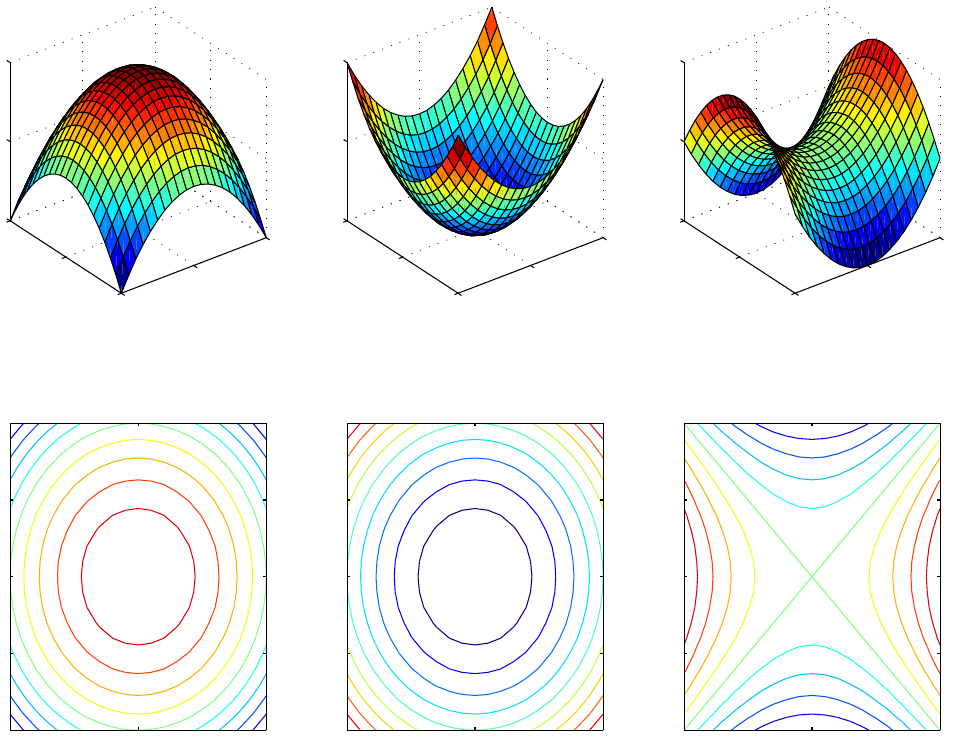}
    \caption{\sl Morse's Lemma states that in a neighborhood of a critical
      point of a Morse function, the level sets are topologically
      equivalent to one of the three forms (left to right: maximum,
      minimum, and saddle critical point neighborhoods).}
    \label{fig:morse_lemma}
  \end{figure}
\end{theorem}
\begin{rem}
  Morse originally used the previous theorem as the definition of
  Morse functions. This way, a Morse
  function does not need to be differentiable.
\end{rem}

\subsection{Examples of Morse/non-Morse Functions}
\noindent Examples of Morse/Non-Morse functions are the following:
\begin{itemize}
\item Obviously, $f(x_1,x_2)=x_1^2+x_2^2$ and
  $f(x_1,x_2)=x_1^2-x_2^2$ are Morse Functions.
\item The \emph{height function} $f : S \to \R^2$ of the embedding of
  a compact surface $S \subset \R^3$ without boundary is a Morse
  function. For example, the height function of the torus:
  \begin{center}
  \end{center}
\item $f(x_1,x_2) = x_1^4 + x_2^4$ is not a Morse function (degenerate
  critical point).
\item A monkey saddle, i.e., $f(x_1,x_2) = x_1^3 - 3x_1x_2^2$:
  \begin{center}
  \end{center}
  is not a Morse function (degenerate; level sets of saddle must cross
  at an `X').
\item All non-smooth functions are not Morse functions, \eg images
  that have edges!
\item Functions that have co-dimension one critical sets (\eg ridges
  and valleys) are not Morse functions. Such critical sets are
  commonplace in images!
\end{itemize}

\subsection{Morse Functions are (Almost) all Functions}
\noindent Morse functions seems to be a very restricted class of functions from
the previous examples, however, a basic result from Morse Theory says
otherwise. 
\begin{defn}[Dense Subset of a Normed Space]
  Let $\|\cdot\|$ denote a norm on a topological space $X$. A set $S\subset X$ is
  {\bf dense} if $\mbox{closure}(S) = X$, i.e., for every $x\in X$ and
  $\delta > 0$, there exists $s\in S$ such that $\|s-x\| < \delta$.
\end{defn}
\begin{theorem}[Morse Functions are Dense]
  Let $\|\cdot\|$ denote the $C^2$ norm on the space of $C^2$
  functions, i.e.,
  \[
  \|f\| = \sup_{x \in \R^2} |f(x)| + |\nabla f(x)| + |\nabla^2 f(x)|.
  \]
  Then Morse functions form an open, dense subset of $C^2$ functions.
\end{theorem}
\begin{rem}
By the previous theorem, any smooth ($C^2$) function can be well approximated
  by a Morse function up to \emph{arbitrary precision} (defined by
  $\|\cdot\|$).
For example, ridges and valleys can be approximated with a Morse
  function, \eg consider the following circular ridge that can be
  made Morse by a slight tilt. Let $f(x_1,x_2) =
  \exp{(-(\sqrt{x_1^2+x_2^2}-1)^2)}$ which is a ridge and non-Morse:
  \begin{center}
  \end{center}
  Now consider the function $g(x_1,x_2) = f(x_1,x_2) + \varepsilon
  x_1$, which is arbitrarily close to $f$ (in $C^2$ norm) and is a
  Morse function:
  \begin{center}
  \end{center}  
It is a basic fact that $C^2$ functions under the norm
  above are \emph{dense} in all square integrable functions ($L^2$)
  functions. Therefore, even non-smooth functions can be approximated
  to arbitrary precision by Morse functions.
\end{rem}

\subsection{Reeb Graph}


\begin{defn}[Equivalence Relation]
  Let $X$ be a set. An {\bf equivalence relation} on $X$ denoted
  $\sim$ is a binary relation with the following properties: for all
  $x,y, z \in X$:
  \begin{itemize}
  \item (reflexivity) $x \sim x$
  \item (symmetry) if $x \sim y$ then $y \sim x$
  \item (transitivity) if $x \sim y$ and $y \sim z$, then $x \sim z$
  \end{itemize}
  We denote by $[x]$ all elements of $X$ related to $x$, \eg $[x] = \{
  y \in X : \, x \sim y \}$.
\end{defn}
\begin{defn}[Topological Space]
  A {\bf topology} denoted $\mathcal{T}$ on a set $X$ is a collection
  of subsets of $X$ (called \emph{open sets}) such that the following
  properties hold:
  \begin{itemize}
  \item $\emptyset, X \in \mathcal{T}$
  \item for $U_{\alpha} \in \mathcal{T}$ where $\alpha \in
    \mathcal{J}$ is an index set (perhaps uncountable), we have
    $\bigcup_{\alpha \in \mathcal J} U_{\alpha} \in \mathcal T$.
  \item for $U_i \in \mathcal T$ where $i \in \mathcal I$ is a finite
    index set, we have $\bigcap_{i \in \mathcal I} U_i \in \mathcal T$.
  \end{itemize}
\end{defn}
\begin{defn}[Quotient Space]
  Let $X$ be a topological space. Let $\sim$ denote an equivalence
  relation on $X$. The {\bf quotient space} of $X$ under the
  equivalence relation $\sim$, denoted $X/\sim$ is the topological
  space whose elements are
  \[
  X/\sim \, \doteq \{ [x] : \, x\in X \},
  \]
  and whose topology is induced from $X$. The {\bf quotient map} is
  the (continuous) function $\pi : X \to X/\sim$ defined by $\pi(x) = [x]$.
\end{defn}

\begin{defn}[Reeb Graph]
  Let $f : \R^2 \to \R$ be a function. Define a equivalence relation
  $\sim$ on the space $\mbox{Graph}(f) \doteq \{ (x,f(x)) : \, x\in \R^2
  \}$ by 
  \[
  (x,f(x)) \sim (y,f(y))\, \mbox{iff}\, f(x)=f(y)\, \mbox{and there
    is a continuous path from $x$ to $y$ in $f^{-1}(f(x))$}.
  \]
  The {\bf Reeb graph} of the function $f$, denoted $Reeb(f)$, is the
  topological space $\mbox{Graph}(f)/\sim$.
\end{defn}
\begin{rem}
  The Reeb graph of a function $f$ is the set of connected components
  of level sets of $f$ (with the additional information of the
  function value of each level set).
\end{rem}

\subsection{Examples}

We will depict the Reeb graph in the following way: an element
$[(x,f(x))] \in \mbox{Reeb}(f)$ will be represented by a point $p_{[(x,f(x))]}$ in the
$x-y$ plane, and if $f(z_1)>f(z_2)$ then the $y$-coordinate of
$p_{[(z_1,f(z_1))]}$ will be larger than $p_{[(z_2,f(z_2))]}$.
\begin{itemize}
\item $f(x_1,x_2) = x_1^2 + x_2^2$
  \begin{center}
  \end{center}
\item $f(x_1,x_2) = \exp{[-(x_1^2 + x_2^2)]} + \exp{[-((x_1-1)^2 + x_2^2)]}$
  \begin{center}
  \end{center}
\item $f(x_1,x_2) = \exp{[-(x_1^2 + x_2^2)]} - 0.1\exp{[-10((x_1-0.2)^2 + x_2^2)]}$
  \begin{center}
  \end{center}
\item $f(x,y) = \exp{\left[ -(x_1^2+x_2^2) \right]} +
\exp{\left[ -((x_1-3)^2+x_2^2) \right]} +
\exp{\left[ -((x_1+3)^2+x_2^2) \right]}$
  \begin{center}
  \end{center}
\end{itemize}

\subsection{Properties of Reeb Graphs}

\begin{lemma}[Reeb graph is connected]
  If $f : \R^2 \to \R$ is a function, then $Reeb(f)$ is connected.
\end{lemma}

\begin{lemma}[Reeb Tree]
  If $f : \R^2 \to \R$ is a function, then $Reeb(f)$ does not contain cycles.
\end{lemma}

\begin{rem}
  Both of these results follow from basic results in topology, namely,
  that connectedness and contractibility of loops are preserved under
  quotienting. That is, since $\mbox{Graph}(f)$ is connected and loops
  in  $\mbox{Graph}(f)$  are contractible (so long as $f$ is
  continuous), we have that $\mbox{Reeb}(f) = \mbox{Graph}(f) / \sim$
  must also have these properties.
\end{rem}

Assume now that $f : \R^2 \to \R$ is a Morse function whose critical
points have  {\bf distinct values}, then we may associate an
\emph{attributed graph} to $\mbox{Reeb}(f)$.
\begin{defn}[Attributed Graph]
  Let $G = (V,E)$ be a graph ($V$ is the vertex set and $E$ is the
  edge set), and $L$ be a set (called the \emph{label set}). Let $a :
  V \to L$ be a function (called the \emph{attribute} function). We
  define the \emph{attributed graph} as $AG = (V,E,L,a)$.
\end{defn}

\begin{defn}[Attributed Reeb Tree of a Function]
Let $V$ be the set of critical points of $f$. Define $E$ to be
\begin{multline}
E = \{ 
(v_i, v_j) : i \neq j, \, \exists \mbox{ a continuous map }
\gamma : [0,1] \to Reeb(f) \mbox{ such that } \\
\gamma(0) = [(v_i,f(v_i))], \,
\gamma(1) = [(v_j,f(v_j))]  \mbox{ and } \gamma(t) \neq [(v,f(v))] 
\mbox{ for all } v \in V \mbox{ and all }
t\in (0,1)  
\}.
\end{multline}
Let $L = \R^+$, and 
\[
a(v) = f(v).
\]
\end{defn}

\begin{defn}[Degree of a Vertex]
  Let $G=(V,E)$ be a graph, and $v\in V$, then the degree of a vertex,
  $deg(v)$, is the number of edges that contain $v$.
\end{defn}

\begin{theorem}
  Let $f : \R^2 \to \R^+$ be a Morse function with distinct critical
  values. Let $(V,E,\R,f)$ be its Attributed Reeb Tree. Then
  \begin{enumerate}
  \item $(V,E)$ is a connected tree
  \item $n_0 - n_1 + n_2 = 2$ where $n_0$ is the number of maxima,
    $n_1$ the number of saddles and $n_2$ the number of minima
  \item If $v \in V$ and $v$ is a local minimum/maximum, then $deg(v)=1$
  \item If $v \in V$ and $v$ is a saddle, then $deg(v) = 3$
  \end{enumerate}
\end{theorem}
\begin{rem}
  Property 2 above is a remarkable fact from Morse Theory, which is
  more general than it is shown above. Indeed, given a compact surface
  $S \subset \R^3$, the number $n_0 - n_1 + n_2$ is the same for any Morse
  function $f : S \to \R^+$, i.e., $n_0 - n_1 + n_2$ (although
  seemingly a property of the function) is an {\bf
    invariant} of the surface $S$.
\end{rem}
\begin{rem}
  Using the fact that for any tree $(V,E)$, we have that $|V|-|E| =
  1$ and Property 2, we can conclude by simple algebraic manipulation
  that $deg(v)=3$ for a saddle.
\end{rem}

\begin{theorem}[Stability of Attributed Reeb Tree Under Noise]
  Let $f : \R^2 \to \R$ be a Morse function and set $g_{\varepsilon} =
  f + \varepsilon h$ where $h : \R^2 \to \R$ is $C^2$. Then for
  all $\varepsilon$ sufficiently small, $g_{\epsilon}$ is Morse and
  $ART(f) = ART(g_{\epsilon})$.
\end{theorem}

\subsection{Diffeomorphisms and the Attributed Reeb Tree}

\begin{defn}[Diffeomorphism of the Plane]
  A function $\psi : \R^2 \to \R^2$ is a {\bf diffeomorphism} provided
  that $\nabla \psi(x)$ and $\nabla \psi^{-1}(x)$ exists for all $x
  \in \R^2$.
\end{defn}

\begin{theorem}[Invariance of ART Under Diffeomorphisms]
  \label{thrm:art_invar}
  Let $f : \R^2 \to \R$ be a Morse function and $\psi : \R^2 \to \R^2$
  be a diffeomorphism. Then $ART(f) = ART(f\circ \psi)$.
\end{theorem}
\begin{rem}
    Note that if $p$ is a critical point of $f$, then $\psi^{-1}(p)$ is a
    critical point of $f \circ \psi$:
    \[
    \nabla ( f \circ \psi ) ( \psi^{-1}(p) ) = 
    \nabla \psi( \psi^{-1}(p) ) \circ \nabla f(
    \psi( \psi^{-1}(p) ) ) = \nabla \psi(p) \circ \nabla f(p) = 0
    \mbox{ if } \nabla f(p) =0.
    \]
    Therefore the vertex set in the ART of both $f$ and $f \circ \psi$
    are equivalent.
Moreover, if $\gamma$ is a continuous path in $f^{-1}(f(x))$
    then $\psi \circ \gamma$ is a continuous path in
    $(f\circ\psi)^{-1}(f\circ\psi(x))$, as diffeomorphisms do not
    break continuous paths. Therefore, the edge sets in the ART of $f$
    and $f\circ \psi$ are equivalent.
\end{rem}

\begin{theorem}
  \label{thrm:ART_injective}
  If $f, g : \R^2 \to \R$ are Morse functions with distinct critical
  values and if $ART(f)=ART(g)$, then there exists a monotone function
  $h : \R \to \R$ and a diffeomorphism $\psi : \R^2 \to \R^2$ such
  that $f = h \circ g \circ \psi$.
\end{theorem}
\begin{rem}
  By Morse Lemma, we can construct diffeomorphisms $\psi_i$ around
  critical points, the idea is then to ``stitch'' these diffeomorphisms
  up with ``patches''  to form the diffeomorphism $\psi$ of interest.
\end{rem}

\begin{theorem}[Reconstruction of Function from ART]
  \label{thrm:ART_surjective}
  If $(V,E)$ is a tree such that each vertex $v\in V$ is of degree 1
  or 3, then there exists a Morse function $f : \R^2 \to \R$ such that
  $ART(f) = (V,E)$.
\end{theorem}

\begin{defn}[Orbit Space]
  Let $X$ be a set, and $G$ be a group.
  \begin{itemize}
  \item $G$ {\bf acts on} $X$ if each $g \in G$ is also $g: X\to X$
    such that 
    \begin{enumerate}
    \item For each $g,h \in G$ and $x\in X$, $(gh)x=g(hx)$.
    \item For the identity element $e\in G$, we have $ex=x$ for all
      $x\in X$.
    \end{enumerate}
  \item If $G$ acts on $X$, then the {\bf orbit} of a point $x\in X$ is $Gx
    = \{gx : g\in G\}$.
  \item Define an equivalence relation in $X$ by $x \sim y$ if there
    exists $g\in G$ such that $gx=y$. The {\bf orbit space} (or the
    quotient of the action $G$) is the set $X/G = \{ [x] : \, x\in X \}$.
  \end{itemize}
\end{defn}

\begin{theorem}
  Let $\mathcal F$ be the set of Morse functions with distinct
  critical values, $\mathcal H$ denote the set of monotone functions
  $h : \R \to \R$, and $\mathcal W$ denote the set of diffeomorphisms
  of the plane. Then
  \begin{itemize}
  \item $\mathcal H \times \mathcal W$ acts on $\mathcal F$ through
    the action : $(h,w)f \doteq h \circ f \circ w$ for $h\in \mathcal H$,
    $w \in \mathcal W$, and $f \in \mathcal F$.
  \item The orbit space $\mathcal F / (\mathcal H \times \mathcal W) =
    \mathscr T$ where $\mathscr T$ is the set of trees whose vertices
    have degree 1 or 3.
  \end{itemize}
\end{theorem}
\begin{rem}
  The second result above is simply a restatement of the Theorems
  above. Indeed, we can define the mapping $ ART : \mathcal F / (\mathcal H
  \times W) \to \mathscr T$ by
  \[
  ART([f]) \doteq  ART(f), \mbox{ where } [f] = \{ (h,w)f \in \mathcal F
  :\, (h,w) \in \mathcal H \times \mathcal W \}
  \]
  The function above is well-defined since by
  Theorem~\ref{thrm:art_invar}, any representative $g\in [f]$ will
  have the same Attributed Reeb Tree. Note
  \begin{itemize}
  \item Theorem~\ref{thrm:ART_injective} states that $ ART : \mathcal F / (\mathcal H
    \times W) \to \mathscr T$ is injective.
  \item Theorem~\ref{thrm:ART_surjective} states that $ ART : \mathcal F / (\mathcal H
    \times W) \to \mathscr T$ is surjective.
  \item Therefore, $ ART : \mathcal F / (\mathcal H
    \times W) \to \mathscr T$ is a bijection and therefore, $\mathcal F / (\mathcal H
    \times W) = \mathscr T$.
  \end{itemize}
\end{rem}



}

\section{Radiometry primer (by Paolo Favaro)}
\label{app-radiometry}

{\color{pink}
We describe the light source using its {\em radiance},
$\radn(q,\lmbd)$, which indicates the power density per unit area and
unit solid angle emitted at a point $q \in L$ in a given direction
$\lmbd\in {\mathbb H}^2$, and is measured in [W/sterad/$m^2$]. This is
a property of the light source. When we consider the particular
direction $\lmbd$ from a point $q\in L$ on the light source towards a
point $p\in S$ on the scene, this is given by ${g_q}_*(p-q) = g_q p -
0 = g_q p$. Therefore, given a solid angle $d\Omega_L$ and an area
element $dL$ on the light source, the power per solid angle and unit
foreshortened\footnote{If the area element on the light source is
  $dL$, the portion of the area seen from $p$ is given by $\langle
  \nu_q,g_qp\rangle dL$; this is called the {\em foreshortened area}.} area
radiated from a point $q$ towards $p$ is given by
\begin{equation}
\radn(q,g_qp)d\Omega_L\langle \nu_q, g_qp\rangle dL
\label{eq-radnd-1}
\end{equation}
where $g_qp \in {\mathbb H}^2$ is intended as a unit vector.
Now, how big a patch $dL$ of the light we see standing at a point $p$
on the scene depends on the solid angle $d\Omega_S$ we are looking
through. Following Figure \ref{fig-balance-equation} we have that
\begin{equation}
dL = d\Omega_S \| p-q \|^2/\langle \nu_q, l_{qp} \rangle
\end{equation}
where we have defined $l_{qp} \doteq q-p/\|q-p\|$ and the
inner product at the denominator is called {\em
foreshortening}. Similarly, the solid angle $d\Omega_L$ shines a patch
of the surface $dS$. The two are related by
\begin{equation}
d\Omega_L = \frac{dS}{\|p-q\|^2} \langle \nu_q, l_{pq} \rangle
\label{eq-dom-l}
\end{equation}
where $l_{pq} = -l_{qp} = p-q/\|p-q\|$. Substituting the expressions
of $d\Omega_L$ and $dL$ in the previous two equations into
(\ref{eq-radnd-1}), one obtains the infinitesimal power received at
the point $p$.
\begin{figure}
\begin{center}
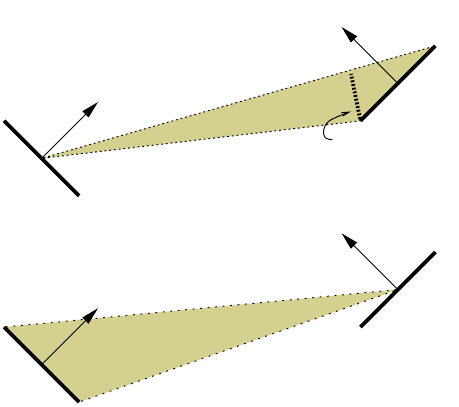
\end{center}
\caption{\sl Energy balance: a light source patch $dL$ radiates energy
  towards a surface patch $dS$. Therefore, the power injected in the
  solid angle $d\Omega_{L}$ by $dL$ equals the power received by $dS$
  in the solid angle $d\Omega_{S}$. Equation (\ref{eq-dom-l})
  expresses this balance in symbols.}
\label{fig-balance-equation}
\end{figure}

Now, we want to write the portion of power exiting the surface at $p$
in the direction of a pixel $\x$ through an area element $dS$. First,
we need to write the direction of $\x$ in the local reference frame at
$p$. We assume that $\x$ is a unit vector, obtained for instance via
central perspective projection
\begin{equation}
\pi:\real^3 \longrightarrow {\mathbb S}^2; \ p\mapsto \pi(p)\doteq \x.
\label{eq-projection}
\end{equation}
However, the point $p$ is written in the inertial frame, while $\x$ is
written in the frame of the camera at time $t$. We need to first
transform $\x$ to the inertial frame, via $g_*(t)^{-1}\x$, and then
express this in the local frame at $p$, which yields ${g_p}^{-1}_*
g_*(t)^{-1}\x$. We call the normalized version of this vector
$l_{p\x}(t)$. Then, we need to integrate the infinitesimal power
radiated from all points on the light source through their solid angle
$d\Omega_L$ against the BRDF\footnote{The following equation, which
  specifies that the scene radiance is a linear transformation of the
  scene radiance via the BRDF is merely a model, and not something
  that can be proven. Indeed this equation is often used to define the
  BRDF.}, which specifies what portion of the incoming power is
reflected towards $\x$. This yields the infinitesimal energy that $p$
radiates in the direction of $\x$ through an area element $dS$:
\begin{equation}
\rad2(p,\x)dS(p) = \int_{L}
\beta(l_{p\x}(t),g_pq)\radn(q,g_qp)d\Omega_L(q)\langle \nu_q, g_q p\rangle
dL(q)
\end{equation}
where the arguments in the infinitesimal forms $dS, dL, d\Omega_L$
indicate their dependency.  Now, we can substitute\footnote{Most often
  in radiometry one performs the integral above with respect to the
  solid angle $d\Omega_S$, rather than with respect to the light
  source. For those that want to compare the expression of the
  radiance $\rad2$ with that derived in radiometry, it is sufficient
  to substitute the expressions of $dL$ and $d\Omega_L$ above, to
  obtain $\rad2(p,\x) = \int_{{\mathbb H}^2}
  \beta(l_{p\x}(t),g_pq)\radn(q,g_qp)\langle N_p, g_qp\rangle
  d\Omega_S(p).$ In our context, however, we are interested in
  separating the contribution of the light and the scene, and
  therefore performing the integral on $L$ is more appropriate.} the
expression of $d\Omega_L$ from (\ref{eq-dom-l}) and simplify the area
element $dS$, to obtain the {\em radiance} of the surface at $p$
\begin{equation}
\rad2(p,\x) = \int_{L}
\beta(l_{p\x}(t),g_pq)\radn(q,g_qp)\frac{\langle \nu_q, g_qp\rangle}{\|p-q\|^2}
\big\langle N_p, l_{pq} \big\rangle dL(q)
\end{equation}
Since the norm $\| p-q \|$ is invariant to Euclidean
transformations, we can write it as $\| g_q p \|$. Now, if the size of
the scene is small compared to its distance to the light, this term is
almost constant, and therefore the measure
\begin{equation}
dE(q,g_qp) \doteq \radn(q,g_qp)\frac{\langle \nu_q, g_qp\rangle}{\|g_q p
  \|^2} dL(q)
\end{equation}
can be thought of as a property of the light source. Since
we cannot untangle the contribution of $\radn$ from that of $dL$, we just
choose $dE$ to describe the power distribution radiated by the light
source. Therefore, we have
\begin{equation}
\rad2(p,\x) = \int_{L}
\beta(l_{p\x}(t),g_pq)\big\langle N_p, l_{pq} \big\rangle dE(q,g_qp).
\label{eq-def-rad}
\end{equation}
This is the portion of power per unit area and unit solid angle
radiated from a point $p$ on a reflective surface towards a point $\x$
on the image at time $t$. The next step consists of quantifying what
portion of this energy gets absorbed by the pixel at location
$\x$. This follows a similar calculation, which we do not report here,
and instead refer the reader to \cite{horn-book} (page 208). There, it is
argued that the irradiance at the pixel $\x$ is equal to the radiance
at the corresponding point $p$ on the scene, up to an approximately
constant factor, which we lump into $\rad2$. The point $p$ and its
projection $\x$ onto the image plane at time $t$ are related by the
equations
\begin{equation}
\x = \pi(g(t)p) \quad p = g(t)^{-1}\pi_S^{-1}(\x)
\label{eq-inv-projection}
\end{equation}
where $\pi_S^{-1}:{\mathbb S}^2 \rightarrow \real^3$ denotes the
inverse projection, which consists in scaling $\x$ by its depth
$Z(\x)$ in the current reference frame, which naturally depends on
$S$.  Therefore, the equation below, known as the {\em irradiance
  equation}, takes the form
\begin{equation}
I(\x,t) = \rad2(p, \pi(g(t)p)) = \rad2(g(t)^{-1}\pi_S^{-1}(\x), \x).
\label{eq-irradiance}
\end{equation}
After we substitute the expression of the radiance (\ref{eq-def-rad}),
we have the {\em imaging equation}, which we describe in Section
\ref{sect-image-formation}, Equation (\ref{eq-imaging}).

}
\chapter{}
\section{Basic image formation}
\label{sect-image-formation}

This section spells out the conditions under which the Ambient-Lambert-Static model is a reasonable approximation of the image formation process.

\subsection{What is the ``image'' ...}

An ``image'' is just an array of positive numbers that measure the
intensity (irradiance) of light (electromagnetic radiation) incident a
number of small regions (``pixels'') located on a surface. We will
deal with gray-scale images on flat, regular arrays, but one can easily
extend the reasoning to color or multi-spectral images on curved
surface, for instance omni-directional mirrors.  In formulas, a digital
image is a function $I:[0,N_x-1]\times[0,N_y-1] \rightarrow [0,N_g-1];
\ (x,y) \mapsto I(x,y)$ for some number of horizontal and vertical
pixels $N_x, N_y$ and grey levels $N_g$. For simplicity, we neglect
quantization in both pixels and gray levels, and assume that the image
is given on a continuum $D \subset \real^2$, with values in the
positive reals:
\begin{equation}
I:D \subset \real^2 \rightarrow \real_+; \ \x \mapsto I(\x)
\end{equation}
where $\x \doteq [x, \ y]^T \in \real^2$. When we consider more than
one image, we index them with $t$, which may indicate time, or generically an index when images are not captured at adjacent time instants:
$I(\x,t)$. We often use time as a subscript, for convenience of
notation: $I_t(\x)$. This abstraction in representing images is all we
need for the purpose of this manuscript.

\subsection{What is the ``scene''...}

A simple description of the ``scene'', or the ``object'', is less
straightforward. This is a {\em modeling} task, for which there is no
right or wrong choice, and finding a suitable model is as much an art
as it is a science; one has to exercise discretion to strike a
compromise between simplicity and realism.  We consider the scene as a
collection of ``objects'' that are volumes bounded by closed,
piecewise smooth surfaces embedded in $\real^3$. We call the generic
surface $S$, which may for convenience be separated into a number $i =
1, \dots, N_o$ of simply connected components, or ``objects'' $S =
\cup_{i=1}^{N_o} S_i$. The surface is described relative to a
(Euclidean) reference frame, which we call $g \in SE(3)$. The two
entities
\begin{equation}
S \subset \real^3; \ g \in SE(3) 
\end{equation}
describe the {\bf geometry} of the scene, and in particular we call
$g$ the {\em pose} relative to a fixed (or ``inertial'') reference
frame\footnote{If a point $p$ is represented in coordinates via $\X
  \in \real^3$, then the transformed point $g p$ is represented in
  coordinates via $R \X + T$, where $R \in SO(3)$ is a rotation matrix
  and $T \in \real^3$ is a translation vector. The action of $SE(3)$
  on a vector is denoted by $g_* v$, so that if the vector $v$ has
  coordinates $V\in \real^3$, then $g_*v$ has coordinates $RV$. See
  \cite{maSKS}, Chapter 2 and Appendix A, for more details.} and $S$
the {\em shape} of objects, although a more proper definition of shape
would be the quotient $S/g$ \cite{kendall84}. This is, however,
inconsequential as far as our discussion is concerned.

Objects interact with light in ways that depend upon their {\em
  material} properties. Describing the interaction of light with
matter can be rather complicate if one seeks physical realism: one
would have to start from Maxwell's equations and describe the
scattering properties of the volume contained in each object. That is
well beyond our scope.  Besides, we do not seek physical realism, but
only to capture the phenomenology of the material to the extent in
which it can be used for detection, recognition or other visual
decision tasks. We will therefore start from a much simpler model, one
that is popular in computer graphics, because it can describe with
sufficient accuracy a sufficient number of real-world objects: each
point $p$ on an object $S$ has associated with it a function $\beta:
{\mathbb H}^2\times {\mathbb H}^2 \rightarrow \real_+; (v, \lmbd)
\mapsto \beta(v, \lmbd)$ that determines the portion of
energy\footnote{The term ``energy'' is used colloquially here to
  indicate radiance, irradiance, radiant density, power etc.}  coming
from a direction $\lmbd$ that is reflected in the direction $v$, each
represented as a point on the half-sphere ${\mathbb H}^2$ centered at
the point $p$. This is called the {\em bi-directional reflectance
  distribution function} (BRDF) \index{BRDF}\index{Bi-directional reflectance distribution function (BRDF)} and is measured in [1/sterad]. This
model neglects diffraction, absorption, subsurface scattering; it only
describes the reflective properties of materials ({\em reflectance}).

To make the notation more precise, we define a {\em local} (Euclidean) {\em reference
frame}, centered at the point $p$ with
the third axis along the normal to the surface, $e_3 = N_p \perp T_p
S$ and first two axes parallel to the tangent plane. 
\begin{figure}[htb]
\begin{center}
\includegraphics[height=3cm]{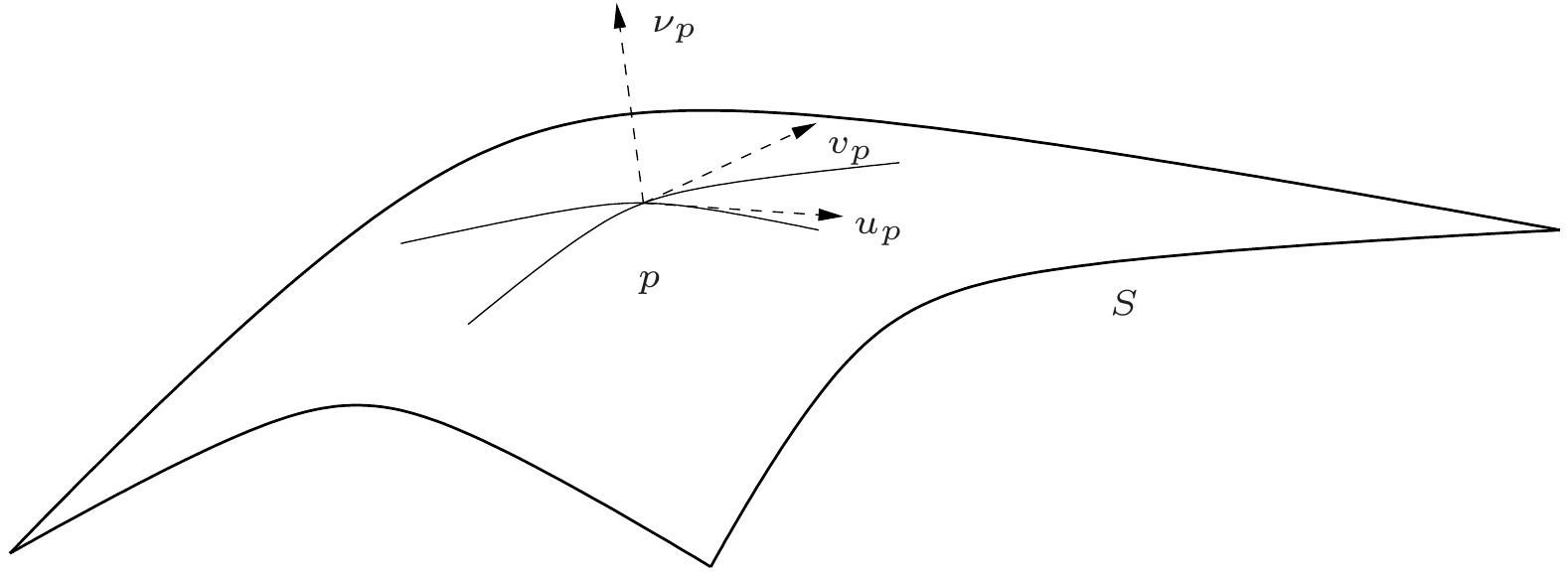}
\end{center}
\caption{\sl Local reference frame at the point $p$.}
\label{fig-local-frame}
\end{figure}
We call such a local reference
frame $g_p$, which is described in homogeneous coordinates by
\begin{equation}
g_p = \ba{cc} \ba{ccc} u_p & v_p & N_p \ea & p \\ 0 & 1 \ea
\label{eq-def-gp}
\end{equation}
where $u_p$, $v_p$ and $N_p$ are unit vectors. Therefore, a point
$q$ in the inertial reference frame will transform to $g_p q$ in the
local frame at $p$. Similarly, a vector $v$ in the inertial frame will
transform to ${g_p}_*v$ in the local frame where
\begin{equation}
{g_p}_* = \ba{cc} \ba{ccc} u_p & v_p & N_p \ea & 0 \\ 0 & 0 \ea.
\label{eq-def-gps}
\end{equation}
The total energy radiated by the point $p$ in a direction $v$ is
obtained by integrating, of all the energy coming from the {\em light
  source}, the portion that is reflected towards $v$, according to the
BRDF. The light source is the collection of objects that can radiate
energy. In principle, every object in the scene can radiate energy
(either by reflection or by direct radiation), so the light source is
just the scene itself, $L = S$, and the energy distribution can be
described by a distribution of directional measures on $L$, which we
call $dE \in {L}_{\rm loc}(L\times \mathbb{H}^2)$, the set of
locally integrable distributions on $L$ and the set of directions.
These include ordinary functions as well as ideal delta measures. The
distribution $dE$ depends on the properties of the light source, which
is described by a function $\radn:L\times \mathbb{H}^2\rightarrow
\real$ of the point $q$ on the light source and a direction (see
Appendix \ref{app-radiometry} for the relationship between $dE$ and
$\radn$). The collection
\begin{equation}
\beta(\cdot, \cdot):{\mathbb H}^2\times {\mathbb H}^2\rightarrow \real_+; \ L \ {\rm and} \  dE: L \times \mathbb{H}^2 \rightarrow \real_+
\end{equation}
describes the {\bf photometry} of the scene (reflectance and
illumination).  Note that $\beta$ depends on the point $p$ on the
surface, and we are imposing no restrictions on such a dependency. For
instance, we do {\em not} assume that $\beta$ is constant with
respect to $p$ (homogeneous material). When emphasizing such a
dependency we write $\beta(v, \lmbd; p)$.

In addition, reflectance (BRDF) and geometry (shape and pose) are
properties of each object that can change over time. So, in principle,
we would want to allow $\beta, \ S, \ g$ to be functions of
time. In practice, we assume that the material of each object
does not change, but only its shape, pose and of course
illumination. Therefore, we will use
\begin{equation}
S = S(t); \ g = g(t), \  \ t \in [0, T]
\end{equation}
to describe the {\bf dynamics} of the scene.  The index $t$ can be
thought of as {\em time}, in case a sequence of measurements is taken
at adjacent instants or continuously in time, or it can be thought of
as an {\em index} if disparate measurements are taken under varying
conditions (shape and pose). We often indicate the index $t$ as a
subscript, for convenience: $S_t \doteq S(t); \ g_t \doteq g(t)$. Note
that, as we mentioned, the light source $(L, dE)$ can also change over
time.  When emphasizing such a dependency we write $L(t)$ and $dE(q,
\lmbd; t)$, or $L_t, dE_t(\lmbd)$.

\begin{example}
The simplest surface $S$ one can conceive of is a plane: $S = \{ p
\in \real^3 \ | \ \langle N, p\rangle = d\}$ where $N$ is the
unit normal to the plane, and $d$ is its distance to the origin. For a
plane not intersecting the origin, $1/d$ can be lumped into $N$, and
therefore three numbers are sufficient to completely
describe the surface in the inertial reference frame. In that
case we simply have $S$ a constant, and $g = e$, the identity. A
simple light source is an ideal point source, which can be modeled as
$L \in \real^3$ with infinite power density $dE = E_l
\delta(q-L)dL(q)$. Another common model is a constant ambient illumination,
which can be modeled as a sphere $L = {\mathbb S}^2$ with $dE =
E_0dL$. We will discuss examples of various models for the BRDF shortly.
\end{example}

\begin{rem}[Choosing a level of granularity in the representation]
  Note that by assuming that the world is made of surfaces we are
  already imposing significant restrictions, and we are implicitly
  choosing a level of granularity for our representation. Consider for
  instance the fabric shown in Figure \ref{fig-maglione}. There is no
  surface there.  The fabric is made of thin one-dimensional threads,
  just woven tightly enough to give the impression of spatial
  continuity.  Therefore, we choose to represent them as a smooth
  surface. Of course, the variation in the appearance due to the
  fine-scale structure of the threads has to be captured somehow, and
  we delegate this task to the reflectance model. Naturally, one could
  even describe each individual thread as a cylindrical surface
  modeled as an object $S$, but this is well beyond the level of
  detail that we want to capture. This example illustrates the fact
  that describing objects entails a notion of {\em scale}. Something
  (\eg a thread) is an object at one scale, but is merely part of a
  texture at a coarser scale. Figure \ref{fig-maglione} highlights the
  modeling tradeoff between shape and reflectance: one could model the
  fabric as a very complex object (woven thread) made of homogeneous
  material (wool), or as a relatively simple object (a smooth surface) made
  of textured material. Although physically different, these two scenarios are phenomenologically indistinguishable, which relates to the discussion earlier in the manuscript of the role between the light field and the complete representation. 
\end{rem}
\begin{figure}[htb]
\begin{center}
\includegraphics[height=5cm]{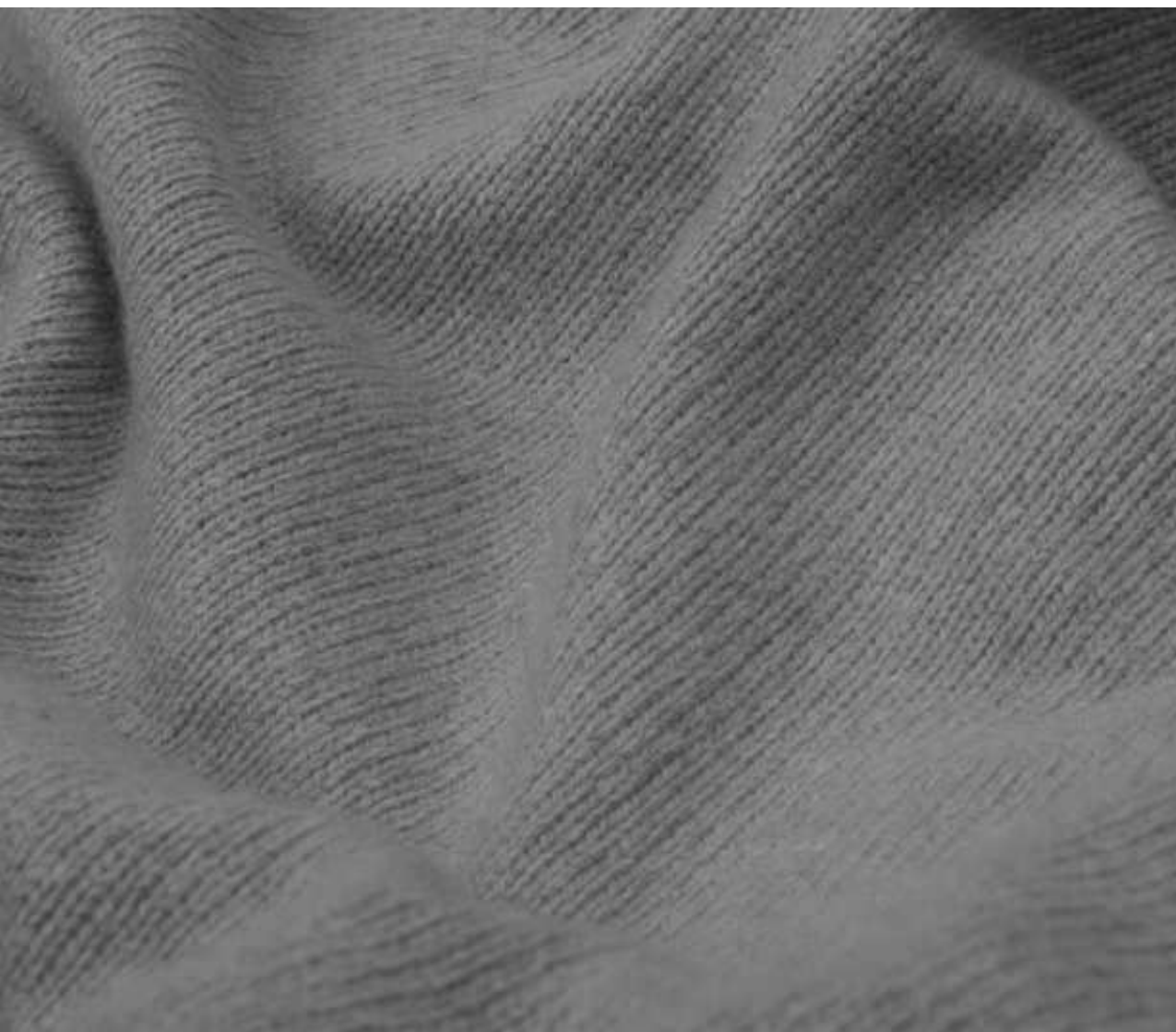}
\end{center}
\caption{\sl A complex shape (woven thread) with simple reflectance (homogeneous material), or a simple shape (a smooth surface) with complex reflectance (texture)?}
\label{fig-maglione}
\end{figure}
\begin{rem}[Tradeoff between shape and motion]
\label{rem-shape-motion}
As we have already noted, instead of allowing the surface $S$ to deform
arbitrarily in time via $S(t)$, {\em and} moving rigidly in space
via $g(t) \in SE(3)$, we can lump the motion and deformation into
$g(t)$ by allowing it to belong to a more general class of
deformations $G$, for instance diffeomorphisms, and let $S$ be
constant. Alternatively, we can lump the deformation $g(t)$ into
$S$ and just describe the surface in the inertial reference frame
via $S(t)$. This can be done with no loss of generality, and it
reflects a fundamental tradeoff in modeling the interplay between shape
and motion \cite{soattoY02}.
\end{rem}
Now, if we agree that a scene can be described by its {\em geometry,
photometry} and {\em dynamics}, we must decide how these relate to the
measured images.

\subsection{And how are the two related?}
\label{sect-image-formation2}

Given a description of the geometry, photometry and dynamics of a
scene, a model of the image is obtained through a description of the
{\em imaging device}. An imaging device is a series of elements
designed to control light propagation. This is typically modeled
through diffraction, reflection, and refraction. We ignore the first
two, and only consider the effects of refraction. For simplicity, we
can also assume that the set of objects that act as light sources and
those that act as light sinks are disjoint, so that $S \cap L =
\emptyset$, i.e. we ignore inter-reflections. Note that $S$ needs not
be simply connected, so we can divide simply connected regions of the
scene into ``light'' $L$ and ``objects'' $S_i, \ i = 1, \dots, N_o$.

Now, using the notation introduced in the previous section, we want
to determine the energy that impinges on a given pixel as a function
of the shape of the scene $S$, its BRDF $\beta$, the light source $L$
and its energy distribution $dE$, and the position and orientation of
the camera.  For simplicity, given the tradeoff between shape and
motion discussed in Remark \ref{rem-shape-motion}, we describe the
(possibly time-varying) shape of the scene in the inertial frame and
drop the explicit description of its pose. In fact, to further
simplify the notation, we can choose the inertial frame to coincide
with the position and orientation of the viewer at time $t=0$, so that
if $I_0(\x_0)$ is the first image, then the scene can be described as
a surface parameterized by $\x_0$: $S_t(\x_0)$. We then describe the
position and orientation of the camera at time $t$ relative to the
camera at time $0$ using a moving Euclidean reference
frame $g_t \in SE(3)$.  Following the derivation in Appendix
\ref{app-radiometry}, the intensity (irradiance) measured at a pixel
$\x$ on the image indexed by $t$ is given by
\begin{equation}
\boxed{
\begin{cases} I_t(\x) = 
\int_{L}
\beta(l_{p\x}(t) ,g_pq)\big\langle N_p, l_{pq} \big\rangle dE(q,g_qp); \\ 
\x = \pi(g_tp); \ p\in S 
\end{cases}
} ~~~~~ {\rm The \ Imaging \ Equation}
\label{eq-imaging}
\end{equation}
where the symbols above are defined as follows:
\begin{description}
\item[Directions:] In the equation above, we have defined $l_{p\x}
  \doteq {g_p}^{-1}_* g_*(t)^{-1}\x$, $g_p$ and ${g_p}_*$ are defined by
  Equation (\ref{eq-def-gp}) and (\ref{eq-def-gps}) respectively,
  $l_{pq} \doteq p-q/\|p-q\|$ and $g_pq$ indicates the (normalized)
  direction from $p$ to $q$, and similarly for $g_q p$; 
\item[Light source:] $L\subset \real^3$ is the (possibly time-varying)
  collection of light sources emitting energy with a distribution
  $dE:L\times {\mathbb H}^2\rightarrow \real_+$ at every point $q\in
  L$ towards the direction of a point $p$ on the 
\item[Scene:] a collection of (possibly time-varying) piecewise smooth
  surfaces $S\subset \real^3$; $\beta:{\mathbb H}^2\times {\mathbb
    H}^2\times S\rightarrow \real$ is the bi-directional reflectance
  distribution function (BRDF) that depends on the incident direction,
  the reflected direction and the point $p\in S$ on the scene $S$ and
  is a property of its material. 
\item[Motion:] relative motion between the scene and the camera is
  described by the motion of the camera $g_t\in SE(3)$ and possibly
  the action of a more complex group $G$, or simply by allowing the
  surface $S_t$ to change over time.
\item[Projection:] $\pi:\real^3\mapsto {\mathbb S}^2$ denotes ideal
  (pinhole) perspective projection, modeled here as projection onto
  the unit sphere, although the same model applies if $\pi: \real^3
  \rightarrow {\mathbb P}^2$, in which case $l_{p\x}$ has to be
  normalized accordingly. 
\item[Visibility and cast shadows:] One should also add to the
  equation two characteristic function terms: $\chi_v(\x,t)$ outside
  the integral, which models the visibility of the scene from the
  pixel $\x$, and $\chi_s(p,q)$ inside the integral to model the
  visibility of the light source from a scene point (cast shadows). We
  are omitting these terms here for simplicity. However, in some cases
  that we discuss in the next section, discontinuities due to
  visibility or cast shadows can be the only source of visual
  information.
\end{description}

\begin{rem}[A philosophical aside on scene modeling]
  One could argue that the {\em real} world cannot be captured by
  simple mathematical models of the type just described, and even
  classical physics is largely inadequate for the task. However, we
  are not looking for an {\em absolute} model. Instead, we are looking
  to describe the scene at the level of granularity that is suitable
  for us to be able to perform inference and accomplish certain tasks.
  So, what is the ``right'' granularity? For us a suitable model of
  the scene is one that can be validated with other existing sensing
  modalities, for instance touch. This is well illustrated by the
  fabric of Figure \ref{fig-maglione}, where at the level of
  granularity required the scene can be safely described as a smooth
  surface. Notice that this is similar to what other researchers have
  suggested by describing the scene as a {\em functional} that cannot
  be directly measured. However, such a functional can be {\em
    evaluated} with various test-functions. Physical instruments
  provide a set of test functions, and imaging device provide yet
  another set of test functions. The goal of the imaging model,
  therefore, can be thought of as relating the value of the scene
  functional obtained by probing with physical instruments to the
  value obtained by probing with images.
\end{rem}
The imaging equation is relevant because most of computer vision is
about inverting it; that is, inferring properties of the scene (shape,
material, motion) regardless of pose, illumination and other nuisances
(the visual reconstruction problem). However, in the general
formulation above, one cannot infer photometry, geometry and dynamics
from images alone. Therefore, we are interested in deriving a model
that strikes a balance between tractability (i.e. it should contain
only parameters that can be identified) and realism (i.e. it should
capture the phenomenology of image formation). We
will use simple models that are widely used
in computer graphics to generate realistic, albeit non-perfect,
images: Phong (corrected) \cite{phong}, Ward \cite{ward92} and
Torrance-Sparrow (simplified) \cite{torranceS}.  All these models
include a function $\rho_d(p)$ called {\em (diffuse) albedo}, and a
function $\rho_s(p)$ called {\em specular albedo}. Diffuse albedo is
often called just albedo, or, improperly, {\em texture}. In 
the next section we discuss various special cases of the imaging equation and
the role they play in visual reconstruction. Here we limit ourselves
to deriving the model under a generic\footnote{See Theorem
  \ref{thm-gaussian-light}.}  illumination consisting of an ambient
term and a number of concentrated point light sources at infinity: $L
= {\mathbb S}^2 \cup \{L_1, L_2, \dots, L_k\}, \ L_i \in {\mathbb
  R}^3$, $dE(q) = E_0dL(q) + \sum_{i=1}^k E_i \delta(q-L_i)$.  In this
case the imaging equation reduces to
\begin{equation}
\boxed{
\begin{cases}
I_t(\x) = \rho_d(p)\left(E_0 + \sum_{i=1}^k E_i \langle N_p, L_i\rangle\right) + \rho_s(p) \sum_i E_i \frac{\big\langle g_t^{-1}\x + L_i/\|L_i\|, N_p \big\rangle^c}{\langle g_t^{-1}\x,N_p\rangle}\\
\x = \pi(g_tp);  \quad p = S(\x_0).
\end{cases}
}
\label{eq-general-light}
\end{equation}
\begin{rem}
  Note that this model does not explicitly include occlusions and
  shadows. Also, note that the first (diffuse) term does not depend on
  the viewpoint $g$, whereas the second term (specular) does.
  However, note that, depending on the coefficient $c$, the second
  term is only relevant when $\x$ is close to the specular direction
  and, therefore, if one assumes that the light sources are
  concentrated, the second term is relevant in a small subset of the
  scene. If we threshold the effects of the second term based on the
  angle between the viewing and the specular direction, then we can
  write the above model as
\begin{equation}
I_t(\x) = \begin{cases}
\rho_d(p)\left(E_0 + \sum_{i=1}^k E_i \langle N_p, L_i\rangle\right) ~~~{\rm if} ~~ \big\langle g_t^{-1}\x + L_i/\|L_i\|, N_p \big\rangle< \gamma(c) \ \forall \ i\\
\rho_s(p) E_{\hat i}~~~ {\rm otherwise}
\end{cases}
\label{eq-rank2}
\end{equation}
where $\hat i = \arg\min_{i}  \frac{\big\langle g_t^{-1}\x + L_i/\|L_i\|, N_p \big\rangle^c}{\langle g_t^{-1}\x,N_p\rangle}$
which justifies the rank-based model of \cite{jinSY02cvpr}.
Empirical evaluation of the validity of this model, and the
  resulting ``brightness constancy constraint,'' discussed in the next
  subsection, has not been thoroughly addressed in the literature.
\end{rem}
The ``identity'' of a scene or an object is specified by its shape $S$
and its reflectance properties $\beta$. The illumination $L(t),
dE(\cdot, t)$, visibility $\chi(\x,t)$ and pose/deformation $g(t)$ are
{``nuisance factors''} that affect the
measurements but {\em not} the identity of the
scene/object\footnote{Depending on the problem at hand, some unknowns
  may play either role: motion, for instance, could be a quantity of
  interest in tracking, but it is a nuisance in recognition.
  Illumination will almost always be a nuisance.}. They change with
the view, whereas the identity of the object does not. 
In the imaging equation (\ref{eq-general-light})
we measure $I_t(\x)$ for all $\x\in D$ and $t = t_1, t_2, \dots,
t_m$, and the unknowns are $L(t_j), dE(\cdot, t_j), g(t_j)$, which for
simplicity we indicate as $L_j, dE_j(\cdot), g_j$ respectively, for
all $j = 1, \dots, m$. For simplicity, we indicate all the unknowns of
interest with the symbol $\xi$ (note that some unknowns are
infinite-dimensional), and all the nuisance variables with $\nu$.
Equation (\ref{eq-general-light}), once we write the coordinates of
the point $p$ relative to the pixel in the moving frame, $p =
g(t)^{-1}\pi^{-1}(\x,t)$, can then be written as a functional $h$,
formally, as follows:\footnote{Note that the symbol $\nu$ for
  ``nuisance'' in the symbolic equation may be confused with $N_p$,
  the normal to the surface in the physical model. Since the two
  symbols will be used exclusively in different contexts, it should be
  clear which one we are referring to.}
\begin{equation}
\boxed{ I = h(\xi, \nu) + n.}~~~ {\rm The \ Imaging \ Equation \ Lite}
\label{eq-imaging-symbolic}
\end{equation}∑
where, to summarize the equivalence of (\ref{eq-imaging-symbolic})
with (\ref{eq-general-light}), we have
\begin{equation}
\begin{cases}
I:D\subset \real^2 \rightarrow \real^3 \\
\xi \in C(\real^2 \backslash {\cal D} \rightarrow \real^3) \times BV({\mathbb S}^2\times {\mathbb S}^2 \rightarrow \real^+) \doteq {\cal S}\\
\nu \in SE(3) \times BV(\real^3\rightarrow \real^+)\times {\mathbb P}(\real^3\rightarrow \real^2) \doteq {\cal \nu} \\
h:\real^3\times BV(\real^3\rightarrow \real^+)\times SE(3) \times \real^+ \times \real^{3k}\times \real^k \rightarrow \real^+ \\
~~~~~~~ (p, \beta, g, E_0, \{L_1, \dots, L_k\}, \{E_1, \dots, E_k\}) \mapsto I \\
n \sim {\cal N}(0,Q)
\end{cases}
\end{equation}
where $\cal D$ is a subset of measure zero (the set of
discontinuities), $BV$ denotes functions of bounded variation.  We
will use the symbolic notation of (\ref{eq-imaging-symbolic}) and the
explicit notation of (\ref{eq-general-light}) interchangeably,
depending on convenience. In some cases we may indicate the arguments
of the functions $I, \xi, \nu, n$ explicitly.
\begin{rem}[Occlusions and cast shadows]
  Occlusions are an accident of image formation that significantly
  complicates our modeling efforts. In fact, while they are
  ``nuisances'' in the sense that they do not depend solely on the
  scene, the {\em do} depend on both the scene and the viewpoint (for
  occlusions) and illumination (for cast shadows). That is why,
  despite depending on the nuisance, under suitable conditions they
  can be exploited to infer the shape of the scene (see
  \cite{yezziS01IJCV} for occlusions and \cite{bouguetP99} for cast
  shadows). For the case of illumination, as we will show, there is no
  loss of generality in assuming ambient + point-light illumination, at
  which point cast shadows are simple to model as a selection process
  of what sources are visible from each point. Nevertheless, it is a
  global inference problem that requires a global solution.
\end{rem}

\section{Special cases of the imaging equation and their role in visual reconstruction (taxonomy)}
\label{app-reconstruction}

In its general
formulation above, the imaging equation cannot be inverted. Therefore,
it is common to make assumptions on some of the unknowns in order to recover
the others. In this section we aim at enumerating a collection of
special cases that compounded characterize most of what can be done in
visual inference.  We start with models of reflection.

Many common materials can be fruitfully described by a BRDF. Exceptions include
translucent materials (\eg skin), anisotropic material (\eg brushed
aluminum), micro-structured material (\eg hair) etc. However, since
our goal is not realism in a physical simulation, we are content with
some common BRDF that are well established in computer graphics: 
Phong (corrected) \cite{phong}, Ward \cite{ward92} and Torrance-Sparrow
(simplified) \cite{torranceS}.
\begin{description}
\item[Phong (corrected)] $\beta(v,\lmbd) = \rho_d(p) + \rho_s(p)\cos^c
\delta/\cos\theta_i\cos\theta_o$. \\ Here $\cos \delta = \langle
g(t)^{-1}\x + q/\|q\|, N_p \rangle$ where each term in the inner
product is normalized, and $\theta_i\doteq \arccos \langle \lmbd,
N_p \rangle$, and $\arccos(\theta_o)\doteq\langle v, N_p\rangle$;
$c\in \real$ is a coefficient that depends on the material.
\item[Ward] $\beta(v,\lmbd) = \rho_d(p) + \rho_s(p) \frac{\exp(-\tan^2(\delta)/\alpha^2)}{\sqrt{\cos\theta_i\cos\theta_o}}$.\\
Here $\alpha \in \real$ is a coefficient that depends on the material
and is determined empirically.
\item[Torrance-Sparrow (simplified)] $\beta(v,\lmbd) = \rho_d(p) + \rho_s(p) \frac{\exp(-\delta^2/\alpha^2)}{\cos\theta_i\cos\theta_o}$. 
\item[Separable radiance] As Nayar and coworkers point out \cite{nayarIK91}, the radiance for the latter model can be written as the sum of products, where the first factor depends solely on material (diffuse and specular albedo), whereas the second factor compounds shape, pose and illumination.
\end{description}
In all these cases, $\rho_d(p)$ is an unknown function called {\em
  (diffuse) albedo}, and $\rho_s(p)$ is an unknown function called
{\em specular albedo}. Diffuse albedo is often called just albedo, or,
improperly, {\em texture}. 

Note that the first term (diffuse reflectance) is the same in all
three models. The second term (specular reflectance) is different.
Surfaces whose reflectance is captured by the first term are called
Lambertian, and are by far the most studied in computer vision.

\subsection{Lambertian reflection}

Lambertian surfaces essentially look the same regardless of the
viewpoint: $\beta(v,\lmbd) = \beta(w, \lmbd) \ \forall w\in
{\mathbb H}^2$. This yields major simplifications of the image
formation model. Moreover, in the case of constant illumination, it
allows relating different views of the same scene to one another
directly, bypassing the image formation model. This is known as the
{\em local correspondence problem}, which relies crucially on the Lambertian
assumption and the resulting brightness constancy
constraint.\footnote{Although the constraint is often used {\em locally} to
  approximate surfaces that are {\em not} Lambertian.} We address this
case first.

\subsubsection{Constant illumination}

In this case we have $L(t) = L$ and $dE(q, \lmbd; t) =
dE(q,\lmbd)$. We consider two simple light source models first.

\subsubsection*{Self-luminous}
Ambient light is due to inter-reflection between different surfaces in
the scene. Since modeling such inter-reflections is quite
complicated,\footnote{There is some admittedly sketchy evidence that
  inter-reflections are not perceptually salient \cite{ennsRensink90}.} we
will approximate it by assuming that there is a constant amount of
energy that ``floods'' the ambient space. This can be approximated by
a (half) sphere radiating constant energy: $L = {\mathbb S}^2$ and $dE = E_0
dL$. In this case, the imaging equation reduces to
\begin{equation}
I(\x,t) = \rho_d(p) E_0 \int_{{\mathbb S}^2}\langle N_p, \lmbd \rangle d\Omega(\lmbd) \doteq E(p).
\label{eq-vignetting}
\end{equation}
Due to the symmetry of the light source, assuming there are no shadows
and having a full sphere, we can always change the global reference
frame so that $N_p = e_3$. However, it is only to first
approximation (for convex objects) that the integral does not depend
on $p$, i.e. we can neglect {\em vignetting}. In this case, $E_0$ can
be lumped into $\rho_d$, yielding the simplest possible model that,
when written with respect to a moving camera, gives
\begin{equation}
\boxed{
\begin{cases}
I(\x,t) = \rho(p) \\
\x = \pi(g(t)p);  \quad p = S(\x_0).
\end{cases}
}
\label{eq-diffuse-ambient}
\end{equation}
Note that this model effectively neglects illumination, for one can
think of a scene $S$ that is self-luminous, and radiates an equal
amount of energy $\rho(p)$ in all directions. Even for such a simple
model, however, performing visual inference is non-trivial. It has
been done for a number of special cases:
\begin{description}
\item[Constant albedo: silhouettes] When $\rho(p)$ is constant, the
  only information in Equation (\ref{eq-diffuse-ambient}) is at the
  discontinuities between $\x = \pi(g(t)p), p\in S$ and $p \notin S$,
  i.e. at the occluding boundaries. Given suitable conditions, that
  have been first studied by Astr\"om et al. \cite{astromCG95}, motion
  $g(t)$ and shape $S$ can be recovered. The reconstruction of shape
  $S$ and albedo $\rho$ has been addressed in an infinite-dimensional
  optimization framework by Yezzi and Soatto
  \cite{yezziS01,yezziS02cvpr} in their work on stereoscopic
  segmentation.
\item[Smooth albedo] The stereoscopic segmentation framework has been
  extended to allow the albedo to be smooth, rather than constant. The
  algorithm in \cite{jinYTCS02} provides an estimate of the shape of
  the scene $S$ as well as its albedo $\rho(p)$ given its motion
  relative to the viewer, $g(t)$.
\item[Piecewise constant/piecewise smooth albedo] The same framework
  has been recently extended to allow the albedo to be piecewise
  constant in \cite{jinCYS04}. This amount to performing region-based
  segmentation a' la Mumford-Shah \cite{mumfordShah89} on the scene
  surface $S$. Although it has not been done yet, the same ideas could be
  extended to piecewise smooth albedo.
\item[Nowhere constant albedo] When $\nabla \rho(p) \neq 0$ everywhere
  in $p$, the image formation model can be bypassed altogether,
  leading to the so-called correspondence problem which we will see
  shortly. This is at the base of most traditional stereo
  reconstruction algorithms and structure from motion. Since these
  techniques apply without regard to the illumination, we will address
  this after having relaxed our assumptions on illumination.
\end{description}

\subsubsection*{Point light(s)}

A countable number of stationary point light sources can be modeled as
$L = \{L_1, L_2, \dots, L_k\}, \ L_i \in \real^3$, $dE = \sum_{i=1}^k
E_i \delta(q-L_i)$. In this case the imaging equation reduces to 
\begin{equation}
I(\x,t) = \sum_{i=1}^k E_i \rho_d(p)\langle N_p, p-L_i/\|p-L_i\|\rangle.
\end{equation}
Note that, if we neglect occlusions and cast shadows,\footnote{Cast
  shadows for the case of point light sources is simply modeled as a
  selection process to determine which source is visible from which
  point.} the sum can be taken inside the inner product and therefore
there is no loss of generality in assuming that there is only one
light source. If the light sources are at infinity, $p$ can be dropped
from the inner product; furthermore, the intensity of the source $E$
multiplies the light direction, so the two can be lumped into the
vector $L$. We can therefore further simplify the above model to
yield, taking into account camera motion,
\begin{equation}
\boxed{
\begin{cases}
I(\x,t) = \rho(p)\langle N_p, L\rangle \\
\x = \pi(g(t)p);  \quad p = S(\x_0)
\end{cases}
}
\label{eq-diffuse-point}
\end{equation}
Inference from this model has been addressed for the following cases.
\begin{description}
\item[Constant albedo] Yuille et al. \cite{yuille02KGBR} have shown that
  given enough viewpoints and lighting positions one can reconstruct
  the shape of the scene. Jin et al. \cite{jinCYS04} have proposed an
  algorithm for doing so, which estimates shape, albedo and position
  of the light source in a variational optimization framework. If the
  position of the light source is known and there is no camera motion,
  this problem reduces to classical shape from shading
  \cite{hornB89}.
\item[Smooth/piecewise smooth albedo] In this case, one can easily
  show that albedo and light source cannot be recovered since there
  are always combinations of the two that generate the same
  images. However, under suitable conditions shape can still be
  estimated, as we discuss next.
\item[Nowhere constant radiance] If the combination of albedo and the
  cosine term (the inner product in (\ref{eq-diffuse-point})) result
  in a radiance function that has non-zero gradient, we can think of
  the radiance as an albedo under ambient illumination, and therefore
  this case reduces to multi-view stereo, which we will discuss
  shortly. Naturally, in this case we cannot disentangle reflectance
  from illumination, but under suitable conditions we can still
  reconstruct the shape of the scene, as we discuss shortly in the
  context of the correspondence problem.
\item[Cast shadows] If the visibility terms are included, under
  suitable conditions about the shape of the object and the number and
  nature of light sources, one can reconstruct an approximation of the
  shape of the scene.
\end{description}

\subsubsection{General light distribution: 
the reflectance/illumination ambiguity}
\label{sect-light-gauss}

As we have already discussed, in the absence of mutual illumination the light source $L$ and the
scene $S$ are disjoint. We make the assumption that the light source
is ``far,'' i.e. the minimum distance between $L$ and $S$ is much
smaller than the maximum distance between two points in $S$,
$\min_{p\in S, q\in L} d(p,q) >> \max_{p,r \in S} d(p,r)$.  Under these
assumptions, we can approximate the light source with a half-sphere
with infinite radius, $L={\mathbb H}^2(\infty)$. The radiance of the
light source therefore only depends on the position $q\in L$, but not
on the direction, since the latter is always normal to $L$; therefore,
$dE:L\rightarrow \real_+; \ q \mapsto dE(q)\ge 0$. An image
is obtained by integrating the light source against the BRDF:
\begin{equation}
I(x) = \int_L \beta_p(x, \lambda)dE(\lambda) = \int_{{\mathbb H}^2}
\beta_p(\nu_{px},\lambda)e(\lambda)dS
\label{eq-ill-refl1}
\end{equation}
where $dS$ is the area form of the sphere, and we neglect edge
effects. For notational simplicity we write $\beta_p(x,\lambda)$ as a
short-hand for $\beta_p(\nu_{px}, \nu_{p\lambda})$. Clearly, if we call
$\tilde \beta_p(x,\lambda) \doteq \beta_p(x,
\lambda)h(\lambda)$ and $\tilde e(\lambda) \doteq
h^{-1}(\lambda)e(\lambda)$, by substituting $\tilde \beta$ and $\tilde
e$ in (\ref{eq-ill-refl1}), we obtain the same image, and therefore
illumination and reflectance can only be determined up to an
invertible function $h$ that preserves the positivity, reciprocity and
losslessness properties of $\tilde \beta$.

To reduce this ambiguity, we show that, under the assumptions outlined
above, there is no loss of generality in assuming that the
illumination is a constant (ambient) term and a collection of ideal
point light sources. In fact, Wiener (``closure theorem,''
\cite{wiener33} page 100) showed that one can approximate a positive
function in $L^1$ or $L^2$ on the plane arbitrarily well with a sum of
isotropic Gaussians: $\lim_{N\rightarrow \infty} \sum_{i=0}^N E_i G(x-
\mu_i; \sigma I)$, where $G(x-\mu; \Sigma) \sim
\exp(-(x-\mu)^T\Sigma(x-\mu))$.  Wilson \cite{wilson00} has shown that
this can be done with {\em positive coefficients}, using Gaussians
with arbitrary covariance, i.e.  $\sum_{i=0}^N E_i G(x- \mu_i;
\Sigma_i); \ E_i \ge 0$. One could combine the two results by showing
that for any $\epsilon$ there exists a $\sigma = \sigma(\epsilon) > 0$
and an integer $N = N(\epsilon)$ such that an anisotropic Gaussian
with covariance $\Sigma$ can be approximated arbitrarily well with a
sum of $N$ isotropic Gaussians with covariance $\sigma I$, i.e.  $\|
G(x-\mu; \Sigma) - \sum_{i=0}^N E_i G(x-\mu_i; \sigma I) \| \le
\epsilon$. Then one could adapt this result to the sphere by showing
that the so-called angular Gaussian density 
approximates arbitrarily well the Langevin density (minimum entropy
density on the sphere, also known as VonMises-Fisher (VMF), or Gibbs
density on the sphere) $G_s(\lambda- \mu; \Sigma) = \exp({\rm
  trace}(\Sigma \mu^T \lambda))$ where $\mu, \lambda \in {\mathbb
  S}^2$ and $\Sigma = \Sigma^T > 0$ is the dispersion. The notation
$\lambda - \mu$ in the argument of $G_s$ should be intended as the
angle between $\lambda$ and $\mu$ on the sphere. Finally, one can
attribute the kernel $G_s$ to the BRDF $\beta$ instead of the light
source. Neglecting the second argument in $G_s(\mu; \Sigma)$ and
neglecting visibility effects, we have: 
\begin{eqnarray}
I(x) &=& \int_{{\mathbb S}^2} \beta_p({x}, \lambda)\sum_{i=0}^N
E_i G_s(\lambda-\mu_i) dS = \\ &=& \sum_{i=0}^N E_i \int_{{\mathbb S}^2} \beta_p({x}, \lambda)\int_{{\mathbb S}^2} G_s(\lambda')\delta(\lambda - \mu_i - \lambda') dS(\lambda') dS(\lambda) = \nonumber \\ 
&=& \sum_{i=0}^N E_i \int_{{\mathbb S}^2} 
\beta_p({x},\lambda'+\lambda'') G_s (\lambda') \delta(\lambda''-\mu_i) dS(\lambda')dS(\lambda'') = \nonumber \\ &=& \int_{{\mathbb S}^2} \tilde \beta_p(x, \lambda'')
\sum_{i=0}^N E_i \delta(\lambda''- \mu_i)dS
\end{eqnarray}
where $\tilde \beta_p(x, \lambda) \doteq \int_{{\mathbb S}^2} \beta(x,
\lambda+\lambda')G(\lambda')dS(\lambda')$ and therefore one cannot
distinguish $\tilde \beta$ viewed under point-light sources located at
$\mu_1, \dots, \mu_N$ from $\beta$ viewed under the general
illumination $dE$. With some effort, following the trace above, one
could prove the following result:


\begin{theorem}[Gaussian light]
\label{thm-gaussian-light}
  Given a scene viewed under distant illumination, with no
  self-reflections, there is no loss of generality in assuming that
  the light source consists in an ambient (constant) illumination term
  $E_0$ and a number $N$ of point-light sources located at $\mu_1,
  \dots, \mu_N$ each emitting energy with intensity $E_i \ge 0$.
\end{theorem}

Given these considerations, we restrict our attentions to illumination
models that consist of the sum of a constant ambient term and a
countable number of point light sources. The general case, therefore,
reduces to the special cases seen above: 
\begin{equation}
L = {\mathbb S}^2; \quad dE(q) = E_0 dS + \sum_{i=1}^k E_i
\delta(q-L_i)dS.
\end{equation}
Note that the energy does not depend on the direction, since for
distant lights (sphere of infinite radius) all directions pointing
towards the scene are normal to $L$.

\begin{rem}[Spherical harmonics]
\label{rem-spherical}
Note that current work on general representations of illumination
uses a series expansion of the distribution $dE$ on $L={\mathbb S}^2$
into spherical harmonics \cite{ramamoorthiH01}. While this is appropriate
for simulation, in the context of inference this is problematic
for two reasons: first, spherical harmonics are {\em global}, so the
introduction of another term in the series affects the entire image.
Second, while any smooth function on the sphere can be approximated with
spherical harmonics, there is no guarantee that such a function be
{\em positive}, hence physically plausible. Indeed, the harmonic terms in the series are
themselves not positive, and therefore each individual component does
not lend itself to be interpreted as a valid illumination, and there
is no guarantee except in the limit where the number of terms goes to
infinity that the truncated series will be a valid illumination. The
advantage of a sum of Gaussian approximation is that one can
approximate any positive function, and given any truncation of the
series one is guaranteed to have a positive distribution $dE$.
\end{rem}
\begin{rem}[Local discrete kernels]
  An alternative to using Gaussians is to use simple functions defined
  on a tiling of the sphere. For such functions to be
  translation-invariants, however, the sphere would have to be tiled
  in regular subdivisions, and this is known to be impossible, as it
  would entail the existence of regular polygons with an arbitrary
  number of faces. The same holds for discrete approximations of
  wavelets on the sphere.
\end{rem}

\begin{rem}[Illumination variability of a Lambertian plane]
\label{rem-lambert-ill}
Consider an image generated by a model (\ref{eq-rank2}). We are
interested in modeling the variability induced in two images of the
same scene under different illumination. We will assume that
illumination can be approximated by an ambient term $E_0$ and a
concentrated point source with intensity $E_1$ located at $L$, so that
each image $I_i(x_i)$ can be approximated by $\rho_d(p)(E_0(t_i) +
E_1(t_i) \langle N_p, L(t_i) \rangle + \beta(i)$ where the latter
term lumps together the effects of non-Lambertian reflection. 
This neglects vignetting (eq. (\ref{eq-vignetting})). The
relationship between two images, the, can be obtained by eliminating
the diffuse reflection $\rho_d$, so as to obtain
\begin{equation}
I(x_1, t_1) = I(x_2, t_2) \frac{E_1(t_1) + E_1(t_1) \langle N_p, L(t_1) \rangle }{E_0(t_2) + E_1(t_2) \langle N_p, L(t_2) \rangle} - \frac{- \beta(t_1) }{E_0(t_2) + E_1(t_2) \langle N_p, L(t_2) \rangle} + \beta(t_2). \nonumber
\end{equation}
Now, if the scene is a plane, then the first fraction on the right
hand side does not depend on $p$, i.e. it is a constant, say $\alpha$.
The second and third term depend on $p$ if the scene is
non-Lambertian.  However, if non-Lambertian effects are negligible
(i.e. away from the specular lobe), or
altogether absent like in our assumptions, then the second term can also be
approximated by a constant, say $\beta$. Furthermore, for the case of
a plane $x_1$ and $x_2$ are related by a homography, $x_1 = H x_2$
where $x_1$ and $x_2$ are intended in homogeneous coordinates.
Therefore, the relationship between the two images can be expressed as
\begin{equation}
I(x_2, t_2) = \alpha I(H x_2, t_2) + \beta.
\end{equation}
One can therefore think of one of the images (\eg $I(\cdot, t_0)
\doteq \rho$) as the scene, and the images are obtained by a warping
$H$ of the domain and a scaling $\alpha$ and offset $\beta$ of the
range. All the nuisances, $H, \alpha, \beta$ are {\em invertible}, and
therefore a planar Lambertian scene one can construct a complete invariant
descriptor. 
\end{rem}

\subsubsection*{Multi-view stereo and the correspondence problem}

If the radiance of the scene $R_S(p)$ is not constant, under suitable
conditions one can do away with the image formation model
altogether. Consider in fact the irradiance equation
(\ref{eq-irradiance}). Under the Lambertian assumption, given (at
least) two viewpoints, indexed by $t_1$ and $t_2$, we have that
\begin{equation}
I(\x_1,t_1) = \rad2(p, \pi(g(t_1)p)) = I(\x_2,t_2)
\end{equation}
without regard to how the radiance $\rad2$ comes to existence. The
relationship between $\x_1$ and $\x_2$ depends solely on the shape of
the scene $S$ and the relative motion of the camera between the two
time instants, $g_{12}\doteq g(t_1)g(t_2)^{-1}$:
\begin{equation}
\x_1 =\pi(g_{12}\pi_S^{-1}(\x_2)) \doteq w(\x_2;S,g_{12}). 
\end{equation}
Therefore, one can forget about how the images are generated, and
simply look for the function $w$ that satisfies (substitute the last
equation into the previous one)
\begin{equation}
\boxed{I(w(\x_2;S,g_{12}),t_1) = I(\x_2, t_2)}.
\end{equation}
Finding the function $w$ from the above equation is known as the {\em
  correspondence problem}, and the equation above is the {\em
  brightness constancy constraint}.

More recently, Faugeras and Keriven have cast the problem of stereo
reconstruction in an infinite-dimensional optimization framework,
where the equation above is integrated over the entire image, rather
than just in a neighborhood of feature points, and the correspondence
function $w$ is estimated implicitly by estimating the shape of the
scene $S$, with a given motion $g$.  This works even if $\rho$ is
constant, but due to a non-uniform light and the presence of the
Lambertian cosine term (the inner product in equation
(\ref{eq-diffuse-point})) the radiance of the surface is nowhere
constant (shading effect, or attached shadow) and even in the case of
cast shadows, if the light does not move. In the presence of regions
of constant radiance, the algorithm interpolates in ways that depend
upon the regularization term used in the infinite-dimensional
optimization (see \cite{faugerasK96} for more details).

\subsubsection{Constant viewpoint: photometric stereo}

When the viewpoint is fixed, but the light changes, inverting the
model above is known as photometric stereo \cite{horn-book}.  If the
light configuration is not known and is allowed to change between
views, Belhumeur and coworkers have shown that this problem cannot be
solved \cite{belhumeurKY99}. In particular, given two images
one can pick a surface $S$ at will, and construct two light
distributions that generate the given images, even if the scene is
known to be Lambertian. However, this result relies on the presence of
a single point light source. We conjecture that if the illumination is
allowed to contain an ambient term, these results do not apply, and
therefore reconstruction could be achieved. Note that psychophysical
experiments suggest that face recognition is extremely hard for humans
under a point light source, whereas a more complex illumination term
greatly facilitates the task.

\subsection{Non-Lambertian reflection}

In this subsection we relax the assumption on reflectance. While,
contrary to intuition, a more complex reflectance model can in some
cases facilitate recognition, in general it is not possible to
disentangle the effects of shape, reflectance and illumination. We
start by making assumptions that follow the taxonomy used for the
Lambertian case in the previous subsection.

\subsubsection{Constant illumination}

\subsubsection*{Ambient light}
In the presence of ambient illumination, the specular term of an empirical
reflection model, for instance Phong's, takes the form
\begin{equation}
\rho_s(p)\int_{-\pi}^\pi \int_0^{\pi/2} \frac{\cos^k\delta}{\cos\theta_o} \sin\theta_id\theta_id\phi_i
\end{equation}
If the exponent $c\rightarrow \infty$, only one point on the light
surface ${\mathbb S}^2$ contributes to the radiance emitted from the
point $p$. Since the distribution $dE$ is uniform on $L$, we conclude
that, if we exclude occlusions and cast shadows, this term is a
constant. This can be considered as a limit argument to the conjecture
that, in the presence of ambient illumination, the specular term is
negligible compared to the diffuse albedo. Naturally, if an object is
perfectly specular, it renders the viewer an image of the light
source, so in this case inter-reflection is the dominant contribution,
and the ambient illumination approximation is no longer justified. See
for instance Figure \ref{fig-bmw}.
\begin{figure}[htb]
\begin{center}
\includegraphics[height=6cm]{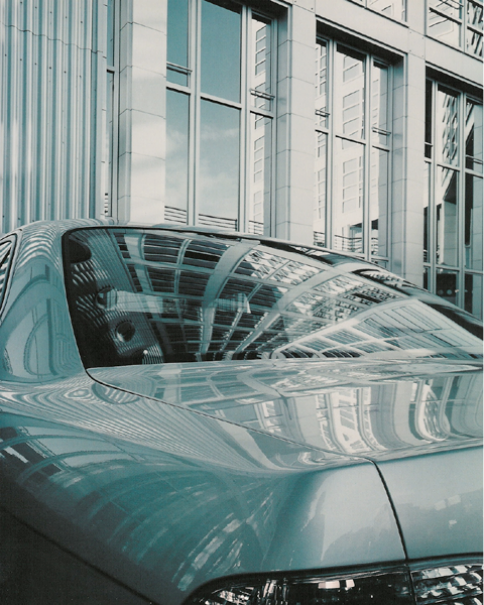}
\end{center}
\caption{\sl In the presence of strongly specular materials, the image is
  essentially a distorted version of the light source. In this case, modeling inter-reflections with an ambient illumination term is inadequate.}
\label{fig-bmw}
\end{figure}

\subsubsection*{Point light(s)}

In the presence of point light sources, the specular component of the
Phong models becomes
\begin{equation}
\sum_i E_i \rho_s(p) \frac{\big\langle g^{-1}(t)\x + L_i/\|L_i\|, N_p \big\rangle^c}{\langle g^{-1}(t)\x,N_p\rangle}
\end{equation}
where the arguments of the inner products are normalized. In this
case, assuming that a portion of the scene is Lambertian and therefore
motion and shape can be recovered, one can invert the equation above
to estimate the position and intensity of the light sources. This is
called ``inverse global illumination'' and was addressed by Yu
and Malik \cite{yuDMH99}. If the scene is dominantly specular, so no
correspondence can be established from image to image, we are not
aware of any general result that describes under what condition shape,
motion and illumination can be recovered. Savarese and Perona
\cite{savareseP01} study the case when assumptions on the position and
density of the light, such as the presence of straight edges at known
position, can be exploited to recover shape.

\subsubsection*{General light}

In general, one cannot separate reflectance properties of the scene
with distribution properties of the light sources. Jin et
al. \cite{jinSY02cvpr} showed that one can recover shape $S$ as well as
the radiance of the scene, which mixes the effects of reflectance and
illumination.

\subsubsection{Constant viewpoint}

In the presence of multiple point light sources, Many have studied the
conditions under which one can recover the position and intensity of
the light sources, see for instance \cite{nayarIK91} and references
therein. Variations of photometric stereo have also been developed for
this case, starting from \cite{ikeuchi81}.

\subsubsection{Reciprocal viewpoint and light source}

Zickler et al. \cite{zicklerBK02} have developed techniques to exploit
a very peculiar imaging setup where a point light source and the
camera are switched in pairs of images, which allows us to eliminate the
BRDF from the imaging equation. 

\section{Analysis of the ambiguities}
\label{sect-ambiguity}

The optimal design of invariant feature requires an analysis of the ambiguities in shape, motion
(deformation), reflectance and illumination. While special cases of
this analysis have been presented in the past, especially in the field
of reconstruction (shape/motion \cite{yezziS03IJCV},
reflectance/illumination \cite{narasimhan03a-class}, shape/reflectance
\cite{soattoY02cvpr}, exemplified below), to this date there is
no comprehensive analysis of the ambiguities in shape, motion,
reflectance and illumination. 

\begin{example}[Tradeoff between shape and motion]
\label{rem-shape-motion1}
We note that, instead of allowing the surface $S$ to deform
arbitrarily in time via $S(t)$, {\em and} moving rigidly in space
via $g(t) \in SE(3)$, we can lump the motion and deformation into
$g(t)$ by allowing it to belong to a more general class of
deformations $G$, for instance diffeomorphisms, and let $S$ be
constant. Alternatively, we can lump the deformation $g(t)$ into
$S$ and just describe the surface in the inertial reference frame
via $S(t)$. This can be done with no loss of generality, and it
reflects a fundamental tradeoff in modeling the interplay between shape
and motion \cite{soattoY02}.
\end{example}

The reflectance/illumination ambiguity has been addressed in Section \ref{sect-light-gauss}. Given the conclusions reached there, we restrict our attention to illumination models that consist of the sum of a constant ambient term and a
countable number of point light sources.

Even under these restrictive modeling assumption, it can be shown that illumination-invariant statistics do not exist.
\begin{theorem}[Non-existence of single-view illumination invariants]
 \label{thm-noDII} There exists no discriminative
  illumination invariant for illumination under the Lambert-Ambient model.
\end{theorem}
This theorem was first proved by Chen et al. in \cite{chenBJ00}
(although also see \cite{shashua97}, as cited by
Zhou, Chellappa and Jacobs in ECCV 2004).  Here we give a simplified
proof that does not involved partial differential equations, but just
simple linear algebra. We refer to the model (\ref{eq-general-light})
where we restrict the scene to be Lambertian, $\rho_s=0$: If
illumination invariants do not exist for Lambertian scenes, they
obviously do not exist for more general reflectance models. Also, we
neglect the effects of occlusions and cast shadows: If an illumination
invariant does not exist without cast shadows, it obviously does not
exist in the presence of cast shadows, since these are generated by
the illumination. We neglect occlusions in the sense that we
consider equivalent all scenes that are equal in their visible range.

\proof{Without loss of generality, following Theorem
  \ref{thm-gaussian-light}, we consider the illumination to be the
  superposition of an ambient term $\{{\mathbb S}^2, E_0\}$ and a
  number of point light sources $\{L_i, E_i\}_{i=1,\dots, l}$. Rather
  than representing each light with a direction $L_i\in {\mathbb S}^2$
  and an intensity $E_i\in \real^+$, we just consider their product
  and call it $L_i \in \real^3$. In the absence of occlusions we can
  lump all the light sources onto one $L = \frac{1}{l}\sum_{i=1}^l L_i
  \in \real^3$ and therefore, again without loss of generality, we can
  assume that the model of the image formation model is given by
  $I_t(x_t) = \rho(p)(E_0 + \langle N_p, L\rangle)$ where $x_t =
  \pi(g_tp)$ and $p\in S(x_0)$, $x_0\in D$. If we consider a
  fixed viewpoint, without loss of generality we can let $g(t) = e$ so
  that $x(t)=x_0 = x$, and the image-formation model is
\begin{equation}
I(x) = \ba{cc}\rho(p) & \rho(p)\nu^T_p \ea \ba{c} E_0 \\ L \ea \doteq A_p X
\end{equation}
where $A_p$ depends on the scene, and $X\in \real^4$ depends on the
nuisance. If we collect measurements at a number of pixels $x_1,
\dots, x_n$, and stack the measurements into a column vector $I \doteq
[I(x_1), \dots, I(x_n)]^T$, and so for the matrix $A$, we can write
the image-formation model in matrix form as $I = A X$. Now, since
$A\in \real^{n\times 4}$ has rank at most $4$, without loss of
generality we will assume that $n=4$. Now, using this notation, the
question of existence of an illumination invariant can be posed as the
existence of a function $\phi$ of the image $I$ that does not depend
on the nuisance $X$. In other words, $\phi$ has to satisfy the
following condition
\begin{equation}
\phi(I) = \phi(AX_1) = \phi(AX_2) ~~ \forall \ X_1, X_2 \ \in \real^4.
\end{equation}
Now, if such a function existed, then, because $AX = AH H^{-1}X \doteq
B \bar X$ for any $H\in {\mathbb GL}(4)$, and the matrix $B\in
\real^{4\times 4}$ can be arbitrary, we would have that $\phi(AX) =
\phi(B\bar X) = \phi(BX)$ for all $X$, and therefore the function
$\phi$ would not depend on $A$, hence it would be non-discriminative.
}

\begin{rem}[Invariance to ambient illumination]
\label{rem-ambient}
Notice that if the vector $X\in \real^4$ in the previous proof was
instead a scalar (for instance, if $X = E_0$ and $L=0$, i.e. there is
only ambient illumination), then it is possible to find a function of
$I$ that does not depend on $E_0$, simply $\phi(I) = I(x_1)/I(x_2)$ or
any other normalizing ratio. See Remarks \ref{rem-lambert-ill} 
for more details on the illumination model implied by a
Lambertian plane and its invariance. Note, however, that the ambient
illumination term depends on the global geometry of the scene, which
determines cast shadows. It may be possible to show that
invariance to illumination can be achieved.
\end{rem}

\subsection{Lighting-shape ambiguity for Lambertian scenes}

In a series of recent papers \cite{yuille02KGBR}, Yuille and coworkers have
shown that for Lambertian objects viewed under point light sources
from an arbitrary viewpoint there is an important class of ambiguities
for 3-D reconstruction. According to \cite{yuille02KGBR}, given a certain
scene, seen from a collection of viewpoints under a certain light
configuration, there exists a different scene, a different collection
of viewpoints and a different light configuration that generates
exactly the same images, and therefore from these it is not possible
to reconstruct the scene geometry (shape), photometry (albedo,
illumination), and dynamics (viewpoints) uniquely. In particular,
equivalent scenes are characterized by a global affine transformation
of the surface normals, the viewpoints, and the lighting
configuration.

Yuille's results pertain to a Lambertian scene viewed under ideal
point light sources from a weak perspective (affine) camera. In
\cite{vedaldiS05}, however, it is argued that a more realistic model
of illumination in real-world scenes is not a collection of point
sources, but an ambient illumination term, which captures the
collective effect of mutual illumination, and a collection of isolated
sources. Here we show that, under this illumination model, the KGBR
ambiguity described in \cite{yuille02KGBR} disappears, and one is left with
the usual projective reconstruction ambiguity well-known in
structure-from-motion \cite{maSKS}. Note that Kriegman and co-workers showed that mutual illumination causes the KGBR to disappear (CVPR 2005).

To introduce the problem, we follow the notation of \cite{vedaldiS05},
where the basic model of image formation for a Lambertian scene is
given by
\begin{equation}
I_t(x_t) = \rho(p)\left(E_0(p) + \sum_{i=1}^N \langle N_p, L_i \rangle \right)
\end{equation}
where $\rho(p):S\rightarrow \real^+$ is the diffuse albedo, $N$ is the
number of light sources, $N_p \in {\mathbb S}^2$ is the outward
normal unit vector to the scene at $p \in S$, $L_i \in {\mathbb S}^2$
is a collection of positions on the sphere at infinity which denotes
the ideal point light sources, and $E_0(p)$ is the ambient
illumination term. Since the ambient light $L$ is assumed to be the
sphere at infinity, in the absence of self-occlusions and cast shadows
(i.e. for convex objects) this term is constant. Otherwise, it is a
constant modulated by a solid angle that determines the visibility of
the ambient sphere from the point $p$:
\begin{equation}
E_0(p) \doteq E_0 \int_{{\mathbb S}^2}\langle N_p, l \rangle d\Omega(l)
\doteq E_0 \Omega(p).
\label{eq-light}
\end{equation}
In the presence of multiple viewpoints the different images are
obtained from the imaging equation by the correspondence equation that
establishes the relationship between $x$ and $p$:
\begin{equation}
x_t = \pi(g_t p), ~~~ p \in S, \ g_t \in SE(3)
\end{equation}
where $\pi:\real^3 \rightarrow \real^2$ is the ideal perspective
projection map. If we rewrite the right hand-side of
the imaging equation as $I_t(x_t) = \rho(p)l(p)$, it is immediate to
see that we cannot distinguish the radiance $\rho(p)$ and $l(p)$ from
$\tilde \rho(p) \doteq \rho(p)\alpha(p)$ and $\tilde l(p) \doteq
\alpha^{-1}(p) l(p)$, where $\alpha:S\rightarrow \real^+$ is a
positive function whose constraints are to be determined. In the
absence of the ambient illumination term, $E_0$, Yuille et al. showed
that $\alpha(p) = | K^T N_p |$. The following theorem establishes
that $\alpha(p) = p$, the identity map, and therefore the ``ambient
albedo'' $\rho(p)E_0(p)$ is an illumination invariant, up to affine transformations of the radiance, which can be factored out using level lines or other contrast-invariant statistics.
\begin{theorem}[KGBR disappears with ambient illumination]
\label{thm-KGBR}
  Let a collection of image of a scene with surface $S$, albedo
  $\rho$, viewed under an ambient illumination $E_0$ and a number $N$
  of point light sources $L_1, \ldots, L_N$, be given: $I_t(x), \ t
  =1, \ldots, T$. The only other scene that generates the same images
  consists of a projective transformation of $S$ and the corresponding
  cameras $\pi$. In the presence of calibrated cameras, the only
  ambiguity is a scale factor affecting $S$ and the translational
  component of $g_t$.
\end{theorem}
The sketch of the proof 
by contradiction is as follows. Under the
assumption of weak perspective projection (affine camera), changes in
the viewpoint $g_t$ can be represented by an affine transformation of
the scene, $K_tp, \ p\in S$. Dropping the subscript $t$, we can write
the image of the scene from an arbitrary viewpoint as
\begin{eqnarray}
I(x) &=& \rho(Kp)\left( E_0(Kp) + \sum_{i=1}^N \frac{\langle K^{-1}N_p, E_i\rangle}{|K^{-1}N_p|} \right) = \\ &=& \frac{\rho(Kp)}{|K^{-1}N_p|}
\left( E_0(Kp)|K^{-1}N_p| + \sum_{i=1}^N \langle K^{-1}N_p, E_i\rangle \right)
\end{eqnarray}
from which it can be seen that $\rho, E_0$ and $E_i$ are
indistinguishable from 
\begin{eqnarray}
\tilde \rho(p) & \doteq & \frac{\rho(Kp)}{|K^{-1}N_p|}  \\
\tilde E_0(p) & \doteq &  E_0(Kp)|K^{-1}N_p|\\
\tilde E_i & \doteq & KE_i.
\end{eqnarray}
However, the function $\tilde E_0(p)$ cannot be arbitrary since, from
(\ref{eq-light})
\begin{equation}
\tilde E_0(p) = \tilde E_0 \Omega(p) = E_0 \Omega(Kp)  |K^{-1}N_p|.
\end{equation}
From this equation one concludes that the ratio of the solid angles
$\frac{\Omega(p)}{\Omega(Kp)} = |K^{-1}N_p|$ is a function of the
tangent plane at $p$, $N_p$, which is not the case, hence the
contradiction.

\chapter{Legenda}
\label{sect-notation}
Symbols are defined in the order in which they are introduced in the text. 
\begin{itemize}
\item $I$: an image, either intended as an array of positive numbers $\{I_{ij} \in \real^+ \}_{i,j = 1}^{M,N}$, or a map defined on the lattice ${\mathbb Z}^2$ with values in the positive reals, $I: {\mathbb Z}^2 \rightarrow \real^+; \ (i, j) \mapsto I(i,j)$, or in the continuum as a map from the real plane, or a subset $D$ of the real plane, $I: D\subset \real^2 \rightarrow \real^+; \ x \mapsto I(x)$. For a time-varying image, or a video, this is interpreted as a map $I:D\subset \real \rightarrow \real^+; \ (x,t) \mapsto I(x,t)$ or sometimes $I_t(x)$. The temporal domain can be continuous $t\in \real$, or discrete, $t \in {\mathbb Z}$. For color images, the map $I$ takes values in the sphere ${\mathbb S}^2 \subset \real^3$, represented with three normalized coordinates (RGB, or YUV, or HSV), and more in general, $I$ can take values in $\real^k$ for multi-spectral sensors.
\item $D$: the domain of the image, usually a compact subset of the real plane or of the lattice, $D\subset \real^2$ or $D\subset {\mathbb Z}^2$.
\item $h$: Symbolic representation of the image formation process.
\item $\xi$: symbolic representation of the {\em ``scene''}. $\Xi$, the space of all possible scenes.
\item $p$, $S$: symbolic representation of the geometry of the scene. $S\subset \real^3$ is a piecewise smooth, multiply-connected surface, and $p\in S$ the generic point on the surface.
\item $\rho$: a symbolic representation of the reflectance of the scene. $\rho: S \rightarrow \real^+$ represents the diffuse albedo of the surface $S$, or the radiance if no explicit illumination model is provided.
\item $g\in G$: motion group. If $G = SE(3)$, $g$ represents a Euclidean (rigid body) transformation, composed of a rotation $R\in SO(3)$ (an orthogonal matrix with positive determinant) and a translation $T\in \real^3$. When $g$ is indexed by time, we write $g(t)$ or $g_t$.
\item $\nu$: Symbolic representation of the nuisances in the image formation process. They could represent viewpoint, contrast transformations, occlusions, quantization, sensor noise.
\item $m$: Contrast transformation, $m: \real \rightarrow \real$ is continuous and monotonic. Usually the arguments are normalized so that $m:[0, \ 1] \rightarrow [0, \ 1]$.
\item $\cal M$ the space of contrast transformations.
\item $n$: Noise process, including all unmodeled phenomena in the image formation process.
\item $SE(3)$: Special Euclidean group of rigid body motions in three-dimensional Euclidean space.
\item $\pi$: perspective projection: $\pi: \real^3 \rightarrow {\mathbb P}^2 \simeq \real^2; \ X \mapsto \pi(X) = x$ where $x_1 = X_1/X_3$ and $x_2 = X_2/X_3$.
\item Homogeneous coordinates $\bar x = \ba{c} x \\ 1 \ea$.
\item $w$: Domain diffeomorphism, $w:D\subset \real^2 \rightarrow \real^2$ that is differentiable, with differentiable inverse.
\item $Z$: Depth map, $Z:D\subset \real^+; x \mapsto Z(x)$.
\item $\pi^{-1}_S(x) = \{p \in S \ | \ \pi(p) = x \}$: Pre-image of the pixel $x$, the intersection of the projection ray $\bar x$ with every surface on the scene.
\item $\pi^{-1}(x)$: Pre-image of the pixel $x$, the intersection of the projection ray $\bar x$ with the closest point on the scene.
\item $R$, $T$: Rotational and translational component of the motion group $g\in G$, usually indicated with $g = (R, T)$, or in homogeneous coordinates as a $4\times 4$ matrix $\bar G = \ba{cc} R & T \\ 0 & 1 \ea$.
\item $\Omega$: the occluded domain, a subset of $D$, that back-projects onto a portion of the scene that is visible from the current image, $I(x,t)$, but not from neighboring images $I(x,t+1)$ or $I(x,t-1)$. 
\item $c$: Class label, without loss of generality assumed to be a positive integer, or simply $c\in \{0, 1\}$ for binary classification.
\item $R(\cdot)$: a risk functional. Not to be confused with $R\in SO(3)$, a rotation matrix.
\item $\cal N$: Normal (Gaussian) density function.
\item $d\mu$: Base (Haar) measure on a space.
\item $dP$, $dQ$: Probability measures. When these are (Radon-Nikodym) differentiable with respect to the base measure, we have $dP = p d\mu$ and $dQ = q d\mu$ where $p, q$ are probability density functions.
\item $\phi$: A feature, i.e. any statistic, or deterministic function of the data.
\item $\hat \xi$: A representation.
\item $H$: a complexity measure, for instance coding length, or algorithmic complexity, or entropy if the image is thought of as a distribution of pixels.
\item $\cal H$: Actionable information
\item ${\cal H}_\xi$: Complete information
\item $\psi$: A feature detector, a particular case of feature that serves to fix the value of a particular group element $g\in G$.
\item $d(\cdot, \cdot)$: A distance.
\item $\| \cdot \|$: A norm.
\item $/$: Quotient.
\item $\backslash$: Set difference.
\item $E[\cdot ]$, $E_p[\cdot]$: Expectation, expectation with respect to a probability measure, $E_p[f] \doteq \int f dP$.
\item $\sim$: Similarity operator, $x \sim y$, denoting the existence of an equivalence relation, and $x, y$ belonging to the same equivalence class; also used to denote that an object is ``sampled'' from a probability distribution, $x \sim dP$.
\item $[\cdot]$: An equivalence class.
\item $\circ$: Composition operator
\item $\forall$ $\exists$: For all, exists.
\item $\widehat{~~}$: The ``hat'' operator mapping $\real^3$ to $se(3)$, the set of $3\times 3$ skew-symmetric matrices.
\item $\hat{~}$: The estimate of an object, for instance $\hat x$ is an estimate of $x$.
\item $| \cdot|$: Determinant of a matrix $|A|$, or volume of a set $|\Omega |$, or absolute value of a real number $| x |$, or the Euclidean norm of a vector in $\real^N$, $| {\bf x} |$.
\item $\chi$: A characteristic function. For a set $\Omega$, $\chi_\Omega(x) = 1$ if $x\in \Omega$, and $0$ otherwise. Sometimes it is indicated by $\chi(\Omega)$ when the independent variable is clear from the context.
\item ${\cal B}$, ${\cal B}_\sigma(x)$: An open set (a ``ball'') of radius $\sigma$ centered at $x$.
\item $\cal G$: A set, or a convolution kernel.
\item $J$: The Jacobian determinant (the determinant of the derivative of a transformation).
\item ${\mathbb{GL}}(3)$ the general linear group of $3\times 3$ invertible matrices. Similarly, ${\mathbb{GL}}(4)$.
\item $u$: The input to a dynamical system, usually denoting a variable that is actively controlled and therefore known with high precision.
\item $\nabla I$: The gradient of the image, consisting of two components $\nabla_x I$ and $\nabla_y I$.
\item $\omega$: Either a set, or a rotational velocity vector, depending on the context.
\item $I_{|_\omega}$: The restriction of an image to a set.
\item $\{ \cdot \}$: A set.
\item $\cal I$: The set of images.
\item $\delta$: Either Dirac's delta distribution, defined implicitly by $\int \delta(x-y)f(y)dy = f(x)$ and $\int \delta(x)dx = 1$; or Kronecker's delta, $\delta(i,j) = 1$ if $i=j$, and $0$ otherwise.
\item $*$ convolution operator. For two functions $f, g$, $f*g = \int f(x-y)g(y)dy$.
\item ${\mathbb H}^2$ the half-sphere in $\real^3$.
\item $\ell^p$: finite-dimensional spaces of $p$-convergent sequences, for instance absolutely convergent ($p=1$) or square-summable ($p=2$) sequences.
\item $L^p$: infinite-dimensional spaces of $p$-integrable functions, for instance Lebesgue-measurable functions $L^1$ or square-integrable functions $L^2$.
\item $\bf H$: The space where the average temporal signal lives, for use in Dynamic Time Warping.
\item $H^2$: A Sobolev space.
\item $\oplus$: The composition operator to update a representation with the innovation. It would be a sum of all the elements involved were linear. 
\item $\mathbb I$: Mutual Information.
\end{itemize}

\printindex

\end{document}